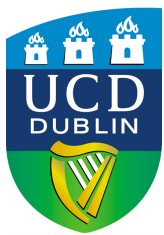

# Ontology-Enhanced Decision-Making for Autonomous Agents in Dynamic and Partially Observable Environments

by

## Saeedeh Ghanadbashi

Student Number: 19204510

This thesis is submitted to University College Dublin in fulfillment of the requirements for the degree of Doctor of Philosophy

School of Computer Science
Head of School:  Assoc. Prof. Neil Hurley

Principal Supervisor: Assist. Prof. Fatemeh Golpayegani

Doctoral Studies Panel Members:
Assist. Prof. Rob Brennan
Assoc. Prof. Gavin McArdle

May 2024

# CONTENTS

# LIST OF TABLES

# LIST OF FIGURES

# ABSTRACT

Agents are intelligent software or hardware entities that can perceive their environment through sensors and act on it through actuators. Such agents usually operate in dynamic and partially observable environments and make decisions on what actions to take to achieve their goals. In such environments, they often encounter unforeseen situations and their observations may differ from the actual state of the environment. Also, in such challenging environments, not all the goals are known or can be predefined, and therefore agents might need to identify the changes and define new goals or adapt their predefined goals on the fly. To do so, agents face the following challenges; (1) Agents' observation is often based on incomplete, ambiguous, and noisy sensed data. As a result, distinct states of the environment might appear the same to the agents, and consequently, they fail to take suitable actions. (2) Facing unforeseen situations (i.e., unpredictable or rare events) is unavoidable and agents need to make decisions and take actions on the fly while only accessing incomplete information which can cause uncertainty in the agents' action selection process.

To address these challenges, collaboration, and practical reasoning techniques have been used to handle agents' limited knowledge and restricted capabilities, enabling them to adapt to new circumstances or emerging requirements by choosing an action from a predefined set. However, the uncertainty caused by partially observable environments makes reasoning more complex and leads to inconsistencies in many traditional reasoning systems. Also, unforeseen situations may require generating new goals because the agent's initial goals may no longer be relevant or achievable. Machine Learning (ML) and in particular Reinforcement Learning (RL) algorithms are used to address these challenges, however, depending on the availability of large amounts of data, the presence of predefined goals, or the possibility of inferring goals from expert human demonstrations and no consideration for a huge number of actions and the long exploration periods make their application limited. other works have employed ontologies, serving as formal knowledge representations, to enable efficient information fusion from diverse sources and provide a comprehensive understanding of the dynamic environment, thus enhancing the agents' decision-making capabilities in complex and uncertain scenarios. Using ontologies, agents are enabled to use the integrated conceptual features extracted from domain ontologies, which allows them to find optimal or near-optimal actions faster.

This thesis proposes a novel ontology-enhanced decision-making model for autonomous agents, enhancing their performance in dynamic and partially observable environments. This model proposes novel techniques for employing ontologies and reasoning mechanisms to enrich the agents' domain knowledge, enabling them to interpret unforeseen events, generate new or evolve the current goals accordingly, and ultimately make suitable decisions and improve their performance in the real world. Specifically, the contributions of this thesis are as follows: (1) An ontology-based observation modeling method is proposed where agents' partial observations are improved on a real-time basis by using prior ontological knowledge. (2) A novel ontology-enhanced decision-making model (OntoDeM) is proposed that allows agents to successfully handle unforeseen situations on the fly in dynamic and partially observable environments by enabling them to evolve their goals or generate new ones. (3) Implementation and evaluation of the proposed model in four different real-world application areas and addressing their challenging problems by employing the relevant methods from the proposed model.

The proposed model is compared to traditional learning algorithms including Q-learning, SARSA, and Deep Q Network, and state-of-the-art methods including Proximal Policy Optimization (PPO), Trust Region Policy Optimization (TRPO), and Deep Deterministic Policy Gradient (DDPG). The results show that OntoDeM can improve the agents' observation and decision-making, which consequently improves their performance in dynamic and partially observable environments.

Statement of Authorship

I hereby certify that the submitted work is my own work, was completed while registered as a candidate for the degree stated on the Title Page, and I have not obtained a degree elsewhere on the basis of the research presented in this submitted work.

Signature     Saeedeh Ghanadbashi     *saeedeh ghanadbashi*

# LIST OF PUBLICATIONS

- (**P1:** Published) S. Ghanadbashi and F. Golpayegani. **Using Ontology to Guide Reinforcement Learning Agents in Unseen Situations**. Applied Intelligence Journal (APIN) 52.2, 5.019 Impact Factor, (2022): 1808-1824.

- (**P2:** Published) S. Ghanadbashi and F. Golpayegani. **An Ontology-based Intelligent Traffic Signal Control Model**. 24th IEEE International Intelligent Transportation Systems Conference (ITSC), pp. 2554-2561. Indianapolis, 2021.

- (**P3:** Published) F. Golpayegani, S. Ghanadbashi, and M. Riad. **Urban Emergency Management Using Intelligent Traffic Systems: Challenges and Future Directions**. 7th IEEE International Smart Cities Conference (ISC2), pp. 1-4. Manchester, 2021.

- (**P4:** Published) Z. Safavifar, S. Ghanadbashi, and F. Golpayegani. **Adaptive Workload Orchestration in Pure Edge Computing: A Reinforcement-Learning Model**. 33rd IEEE International Conference on Tools with Artificial Intelligence (ICTAI), pp. 856-860. Washington, 2021.

- (**P5:** Published) F. Golpayegani, M. Guériau, P. A. Laharotte, S. Ghanadbashi, J. Guo, J. Geraghty, and S. Wang. **Intelligent Shared Mobility Systems: A Survey on Whole System Design Requirements, Challenges and Future Direction**. IEEE Access 10, 3.9 Impact Factor, (2022): 35302-35320.

- (**P6:** Published) R. Collier, S. Russell, S. Ghanadbashi, and F. Golpayegani. **Towards the Use of Hypermedia MAS and Microservices for Web Scale Agent-Based Simulation**. SN Computer Science 3.6, 3.78 Impact Factor, (2022): 1-19.

- (**P7:** Published) S. Ghanadbashi, A. Zarchini, and F. Golpayegani. **An Ontology-based Augmented Observation for Decision-making in Multi-agent Systems**. [Poster]. 30th Irish Conference on Artificial Intelligence and Cognitive Science (AICS), Cork, 2022.

- (**P8:** Published) M. Riad, S. Ghanadbashi, and F. Golpayegani. **Run-time Norms Synthesis in Dynamic Environments with Changing Objectives**. 30th Irish Conference on Artificial Intelligence and Cognitive Science (AICS), pp. 462-474. Cork, 2022.

- (**P9:** Published) S. Ghanadbashi, Z. Safavifar, F. Taebi, and F. Golpayegani. **Handling Uncertainty in Self-adaptive Systems: An Ontology-based Reinforcement Learning Model**. Journal of Reliable Intelligent Environments (JRIE), 5.26 Impact Factor, (2023): 1-26.

- (**P10:** Published) S. Ghanadbashi, A. Zarchini, and F. Golpayegani. **An Ontology-based Augmented Observation for Decision-making in Partially Observable Environments**. 15th International Conference on Agents and Artificial Intelligence (ICAART), pp. 343-354. Lisbon, 2023.

- (**P11:** Submitted) F. Ghasemi, H. Tabatabaei, S. Ghanadbashi, R. Kucharski, and F. Golpayegani. **Using Reinforcement Learning to Optimize Platform Revenue in Ride-sourcing Market**. Submitted to European Conference on Artificial Intelligence (ECAI) (2023).

- (**P12:** Accepted) J. Guo, S. Ghanadbashi, S. Wang, and F. Golpayegani. **Edge-based Ontology DRL for Urban Traffic Signal Control**. 26th IEEE International Intelligent Transportation Systems Conference (ITSC), Bilbao, 2023.

- (**P13:** Accepted) S. Ghanadbashi, A. Zarchini, and F. Golpayegani. **Ontology-based Adaptive Reward Functions**. Modeling and Representing Context Workshop (MRC) at 26th European Conference on Artificial Intelligence (ECAI), Kraków, 2023.

- (**P14:** Submitted) S. Ghanadbashi, A. Zarchini, and F. Golpayegani. **Reinforcement Learning with Adaptive Reward Machine: An Ontology-Based Approach**. Submitted to Sustainability Journal, 3.9 Impact Factor, (2024).

- (**P15:** Submitted) S. Ghanadbashi, A. Zarchini, and F. Golpayegani. **Enhancing Autonomous Agent Decision-Making in Complex Environments with Ontology-Driven Models**. Submitted to ACM Transactions on Intelligent Systems and Technology (TIST), 10.489 Impact Factor, (2024).

- (**P16:** Accepted) I. Mikolasek, S. Ghanadbashi, and F. Golpayegani. **Data Sharing at the Edge of the Network: A Disturbance Resilient Multi-modal ITS**. Submitted to Transport Research Arena (TRA). Dublin, 2024.

# ACKNOWLEDGEMENTS

This thesis would have been impossible without the help of many. First and foremost, I would like to express my sincere gratitude to my supervisor, Dr. Fatemeh Golpayegani, for her constant encouragement and advice throughout my Ph.D. having very enjoyable debates about my research and challenging questions that sparked a lot of new ideas or refined older ones and deepened my understanding of the matter. Her optimistic approach and words of motivation during the tough times of my Ph.D. were of great help. I want to express my gratitude to Dr. Gavin McArdle and Dr. Rob Brennan, my committee members, for their invaluable guidance throughout my Ph.D. journey. Additionally, I extend my thanks to Prof Evangelos Pournaras and Prof Brian MacNamee for their insightful feedback and comments on my thesis. I am also sincerely thankful to everyone in the entire MAS3 lab for their support and help; I am honored to have been a member of this lab. Finally, thank you to all of my family and friends for all their love and constant encouragement. I could not have accomplished as much without the moral support I received especially during stressful, sleepless nights. This memorable, unique journey has only been possible because of these people. Thanks to my mother and father, who always encouraged me to continue my studies. Very special thanks go to my husband Farshad Taebi who helped support me through hard times and had a big part in my decision to pursue a Ph.D. degree in the first place but also encouraged me with his experiences when things did not work as I wished, during the Ph.D. time but as well throughout my whole life. I would like to thank my children, Arian and Sara, who have accompanied me to make this work possible.

# CHAPTER ONE

# INTRODUCTION

This chapter provides an introduction to the problem domain, the research gaps, and this thesis's research objectives and questions. In the Motivation section, the complexities arising from dynamic environments are explored, where unforeseen events can disrupt agents' decision-making, and the additional challenge of partial observability is emphasized. Moving to the Research Context, the focus is directed towards the categorizations of agents and environments that form the core of this Ph.D. investigation. Additionally, four distinct real-world applications are introduced to serve as experimental areas for evaluating the proposed model. The primary challenges are outlined in the section addressing the Research Challenges, namely, decision-making intricacies under limited observations and the necessity for generating new goals when unforeseen events occur. The key objectives and questions of this thesis are detailed in the Research Objectives and Research Questions sections. The interrelations between research questions, challenges, and objectives are then elucidated. Finally, within the Thesis Contributions, the core contributions of the thesis are presented as the innovative Ontology-enhanced Decision-Making model (OntoDeM). This model employs novel methods to enhance agents' decision-making in dynamic and partially observable environments through structured knowledge representations.

## 1.1 Motivation

Agents are entities that can perceive their environment through sensors and act on it through actuators, the aim of which is to achieve a desired goal [11]. They operate in environments comprising observable physical and virtual entities, where their decision-making process is affected by the properties of the environment including its dynamism and partial observability [17].

In **dynamic** environments, encountering unforeseen events is inevitable [5], and when such an environment is **partially observable**, the agent's sensed data can be noisy, ambiguous, inaccurate, or incomplete [181]. In such challenging real-world environments, agents need to adapt their current goals or generate new goals and make decisions on what actions to take accordingly. The agent's decision-making process can include three stages: (S1) Observation modeling, (S2) Goal selection, and (S3) Action selection. *Partial observability of the environment* and *possibility of unforeseen events occurrence* (i.e., unpredictable, unexpected events) are two main challenges that must be addressed in these stages.

Literature has proposed several techniques to handle these challenges. **Goal generation approaches** allow agents to choose a new goal from a pre-defined goal set when the current goal should change [1, 113]. Using Goal Reasoning (GR) techniques, agents continuously reason about the goals they are pursuing, and if necessary, they adjust them according to their preferences [99, 6]. Although GR agents can operate in complex environments and adapt intelligently to changing conditions [126], they do not perform as well when they encounter an unforeseen event that requires generating new goals. **Practical reasoning** is used to enable agents to infer knowledge from their environment and interaction with other agents [6, 100]. Although reasoning allows agents to understand their environment, update their perceptions, and make decisions in real-time, they cannot help with generating new goals. Also, the uncertainty resulting from operating in partially observable environments makes reasoning more complex and leads to inconsistencies in many traditional reasoning systems [15]. For example, in the traffic signal control domain, partial observability due to weather conditions (rainy weather) results in uncertainty about the real-time traffic situation. This uncertainty poses a challenge to the system's ability to reason and make precise decisions.

Some of these challenges have been also studied in the context of **Self-Adaptive System (SAS)**. These systems adapt their behavior autonomously by monitoring the environment and selecting the best-predefined adaptation action to ensure that the goals are consistently achieved under run-time changing conditions [32, 259]. However, anticipating all potential changes in dynamic environments at design time is in most cases infeasible. Therefore, the gap between the SAS's design and run-time models causes uncertainty in the adaptation process [116] and consequently system's behavior deviates from the predefined expectations [171]. To address these challenges, **Learning techniques** are used to enrich agents' knowledge and learn new actions using big data or long exploration periods [240, 188]. However, these techniques are not useful when real-time decision-making is desirable and not enough data is available [167]. Particularly, **Reinforcement Learning (RL)** is a common technique used in such environ-ments [74, 272], which uses a trial-and-error learning technique enabling agents to find suitable actions by maximizing the total cumulative reward through interactions with the environment. However, in partially observable environments, due to lack of full observability, the same observation might be obtained from two distinct states, and result in agents taking the same action, whereas in reality if the full observation was provided two different actions would have been taken in each state. Also, in the presence of unforeseen changes in an environment, random action exploration may exhibit slow learning if there are many possible actions, and it will be more challenging when on-the-fly decision-making is desirable [177, 32].

In 1977, Feigenbaum pointed out that Artificial Intelligence (AI) systems' power lies in their ability to encode and exploit domain-specific knowledge, leading to the paradigm that "in the knowledge lies the power" [72]. Domain-specific knowledge can be acquired from human experience within a specific domain. Ontology is a formal representation of knowledge, facilitating knowledge sharing and reasoning [78, 25]. Experts can represent their knowledge and expertise within the domain through ontology. This process involves identifying the relevant concepts, specifying their properties and relationships, and organizing the knowledge in a structured manner. Ontologies explicitly represent human knowledge about domains of interest [244] and also the hierarchical structure of ontologies, with subclasses inheriting the properties from their ascendant classes, facilitates inferring new concepts and properties. In the literature, ontological knowledge is used to model systems and their environment [234, 10, 159], and it can be used to improve the accuracy of learning algorithms [185, 227]. However, the concerns related to handling challenges of agent's decision-making process in real-world environments, including augmenting agents' partial observations, detecting sudden changes, and reasoning about unforeseen events are not addressed.

**Motivating Example:** Take into account Intelligent Traffic Signal Control (ITSC) systems, where traffic signals are controlled by autonomous and adaptive agents, and their goal is to minimize the waiting time of all vehicles. In such an environment, dynamic changes can arise from dynamic and stochastic traffic flow due to adverse weather conditions and emergency events such as traffic accidents and breakdowns. So, it is not possible to predict all the events, and consequently encountering unforeseen situations is inevitable. Also, the signal controller agents can only observe and communicate with a portion of vehicles that are equipped with communication technologies [294] and the data collected from sensors may be noisy [277]. When an incident happens, it is expected that the traffic signal phase changes from red to green to make clearance for the ambulance's paths. In this situation, the controller agents should identify the change in the environment, create a new goal, and take suitable actions accordingly. The controller agents should make such a decision immediately so the ambulance can deliver emergency services as fast as possible. In such a dynamic and partially observable environment, domain-specific knowledge which encompasses the specialized information relevant to the particular domain can help in understanding the intricacies of the domain and providing insights into the hidden aspects of the environment. For example, recognizing the importance of emergency vehicles in the context of traffic control is vital for saving lives, providing timely medical care, managing traffic congestion, and maintaining public safety.

In summary, operating agents in dynamic and partially observable environments, where unforeseen events are inevitable, face the challenge of identifying changes and responding promptly. Limited observation capabilities and noisy sensor data further complicate decision-making. In such complex scenarios, domain-specific or expert knowledge becomes invaluable, providing specialized insights into the intricacies of the domain.

## 1.2 Research Context

This section discusses the types of agents and environments that are the focus of this Ph.D. research.

**Types of agents:** Agents are grouped into various classes, and are not necessarily disjoint classes, based on their degree of perceived intelligence and capability. Intelligent, adaptive, and autonomous agents are the main focus of this research:

- An intelligent agent can make decisions or perform a service based on its environment, user input, and experiences [65].

- An adaptive agent learns by observing and perceiving its environment and can adapt to the changing environment [28].

- An autonomous agent operates independently and makes decisions on its own, based on its perception of the environment, knowledge, and goals. Autonomy can vary in levels, where the highest level represents full autonomy without human intervention, while other levels allow for varying degrees of human involvement [76]. An adaptive intelligent agent is an autonomous agent that learns and adapts its behavior over time. The RL agent is an adaptive intelligent agent. The RL agent's adaptability arises from its continuous learning and adaptation to the environment through interactions, adjusting its decision-making process based on rewards or punishments. Moreover, its intelligence is evident in utilizing ac-quired knowledge and experience to optimize actions and make informed decisions within the given environment.

**Distinction between Goal and Reward:**

- A goal refers to the objective or desired outcome that the agent aims to achieve within its environment. It represents what the agent is trying to accomplish or the state it wants to reach. Goals provide direction and purpose to the actions of an agent, guiding its decision-making process toward achieving desired results.

- Reward serves as a feedback signal that indicates how well an agent is performing concerning its goals. In RL, the reward function quantifies the desirability of different states or actions, providing the basis for the agent to learn and optimize its behavior. Rewards can be positive or negative values assigned to specific states or actions, reflecting the consequences of the agent's decisions.

**Environment's properties:** This research is concerned with dynamic and partially observable environments:

- Dynamic environment: An environment is static if only the actions of an agent change its state (e.g., playing a game of chess). However, dynamic environments can be affected by the actions of the agents and other factors that are not in control of the agents (e.g., driving a car in traffic) [17].

- Partially observable environment: In a fully observable environment, agents have complete and accurate information about the environment's state (e.g., game of tic-tac-toe). In a partially observable environment, agents do not have access to complete information about the environment (e.g., game of poker) [158].

**Application areas:** The following application areas are selected in this thesis to evaluate the proposed solutions as the target environment and agent types can be simulated to resemble near real-world scenarios.

- **Intelligent Traffic Signal Control (ITSC)**, addresses the challenges posed by the rapid growth of urban populations and increasing vehicular traffic. Conventional traffic signal systems are no longer sufficient in addressing congestion and optimizing transportation efficiency in this context. ITSC systems utilize real-time data from various sources, such as vehicle detectors, cameras, and sensors, to dynamically adjust signal timings based on actual traffic conditions. The environment in ITSC is dynamic due to constantly changing traffic conditions influenced by factors like weather, accidents, and construction, making prediction difficult [123]. Moreover, some traffic data collection sensors have limitations, such as relying on vehicles equipped with communication technologies and having a limited range, making the environment partially observable [295]. In ITSC, adaptive agents, using real-time data dynamically adjust traffic signal timings according to the current traffic conditions. Also, AI techniques are incorporated into intelligent agents to enhance decision-making capabilities.

- **Edge Computing (EC)** brings computation and data storage closer to the source of data generation, such as sensors, devices, and Internet of Things (IoT) endpoints, rather than relying solely on centralized data centers. By processing and analyzing data at the edge of the network, edge computing minimizes latency, and enhances real-time responsiveness. **Task offloading** in edge computing involves deciding whether to process a task locally on the edge device or offload it to an edge server for processing. Since the requirements of applications and mobile networks, such as delay requirements, length of tasks, bandwidth, and utilization of edge servers, may change over time, the environment is dynamic [40]. There is also the possibility that edge servers will fail, thus introducing uncertainty into the system. Moreover, it may be difficult to maintain an up-to-date view of the environment when devices have limited computational power and battery life, as well as intermittent network connectivity [120, 39], so the environment is partially observable. In edge computing, especially in the context of task offloading, there are adaptive agents. These agents continuously adapt to changing conditions (i.e., current conditions of the edge device, the network, and the workload) to optimize task execution and resource utilization. Furthermore, agents can attain intelligence through the integration of AI techniques, enhancing their capacity for informed decision-making.

- **Job Shop Scheduling (JSS)** is a crucial application area in operations management and production planning, where the goal is to optimize the sequencing and allocation of orders or jobs on various machines. The objective is to minimize production time, maximize resource utilization, and meet delivery deadlines efficiently. In a JSS environment, unforeseen events such as machine failure, longer than usual processing times, or orders that must be introduced into production urgently can hugely impact the scheduling process [29, 213], so the environment is dynamic. Also, observations may be unavailable for certain machines or orders because they are delayed, corrupted, or even missing, so the environment is partially observable [270]. In JSS, intelligent agents can analyze job requirements, machine capacities, and other constraints to find optimal schedul-ing solutions that minimize the makespan, reduce idle time, and improve overall efficiency. Adaptive agents, on the other hand, have the ability to adjust their scheduling strategies based on changes in the environment or dynamic job requirements. They can learn from past scheduling experiences and adapt their decision-making process to optimize schedules in real-time as new jobs are added or existing ones are modified.

- A **Heating System Control (HSC)** is a vital component of heating systems responsible for regulating and managing the heating process. Its primary purpose is to maintain a comfortable and controlled indoor environment by dynamically adjusting the temperature and other relevant parameters. The HSC utilizes sensors and measurements to collect data on the present indoor temperature, but it may lack complete access to all pertinent information concerning the indoor environment and external factors. This may occur due to sensor corruption or the absence of sensors for certain parameters, such as sudden changes in weather conditions. As a result of these data collection limitations and the presence of unobservable external influences, the HSC operates within a partially observable environment. The HSC operates in a dynamic environment because the indoor conditions and external factors it needs to consider are subject to constant changes over time. Moreover, unforeseen events could arise, such as malfunctions in heating system components, disruptions from power outages, indoor temperature fluctuations due to occupant activities, changes in outdoor weather, or the operation of other heating or cooling systems [87, 284]. Adaptive intelligent agents in an HSC continuously learn from their interactions with the environment and adapt to varying heating requirements based on occupancy patterns, time of day, and changing weather conditions [296].

## 1.3 Research Challenges

In partially observable and dynamic environments agents face two main challenges: (**CH1**) Decision making when only having access to partial observation and/or (**CH2**) New goal generation when encountering unforeseen events and non-of the predefined goals can be suitable. Each of these challenges can impact each of the three stages of the agent's decision-making process (S1, S2, S3), and are described as follows (see Figure 1.1):

- **S1 - Observation Modeling:** One of the main issues that agents face when having access to *partial observations* is modeling the environment accurately. When an agent has access to noisy or incomplete information about the state of the environment, it can result in uncertainty and ambiguity in the agent's perception of the environment. Additionally, the complexity of the environment and the interactions between different entities in the environment can make it difficult to model the environment accurately. Furthermore, the environment may change over time, which requires the agent to be able to adapt its model to the changes in the environment.

- **S2 - Goal Selection:** The other issue when only having access to *partial observations* is efficiently searching through the space of possible goals and selecting those that are *most relevant* [6]. *Preferences* between goals may change over time, and the agent must update them when this happens [115]. Also, *unforeseen situations* may require *generating new goals* because the agent's original goals may not be relevant or achievable in the new situation.

- **S3 - Action Selection:** Agents operating in *partially observable* environments lack complete knowledge of their states and the outcomes of their actions. Because of this uncertainty, choosing the optimal action for achieving goals may be difficult. [229]. Also, for handling *unforeseen events*, the agent must balance the need to explore the environment to find new, potentially better actions with the need to exploit its current knowledge to maximize its rewards [127]. This trade-off can be especially difficult with *large action spaces* when there is a need to make decisions on-the-fly [122].

## 1.4 Research Objectives

The main research objectives of this thesis are as follows:

- **RO1 - Augmented Observation Modelling:** The aim is to enable agents to augment their observations by providing additional information about the state of the environment that may not be directly observable.

- **RO2 - Adaptive Goal Selection:** The aim is to enable agents to adapt their behavior by evolving their goals or even generating new goals to address the emerging requirements of the dynamic environment.

- **RO3 - Enhanced Action Selection:** The aim is to enable agents to explore - exploit their action space efficiently by tuning their reward functions and reasoning about unforeseen situations autonomously and on-the-fly.

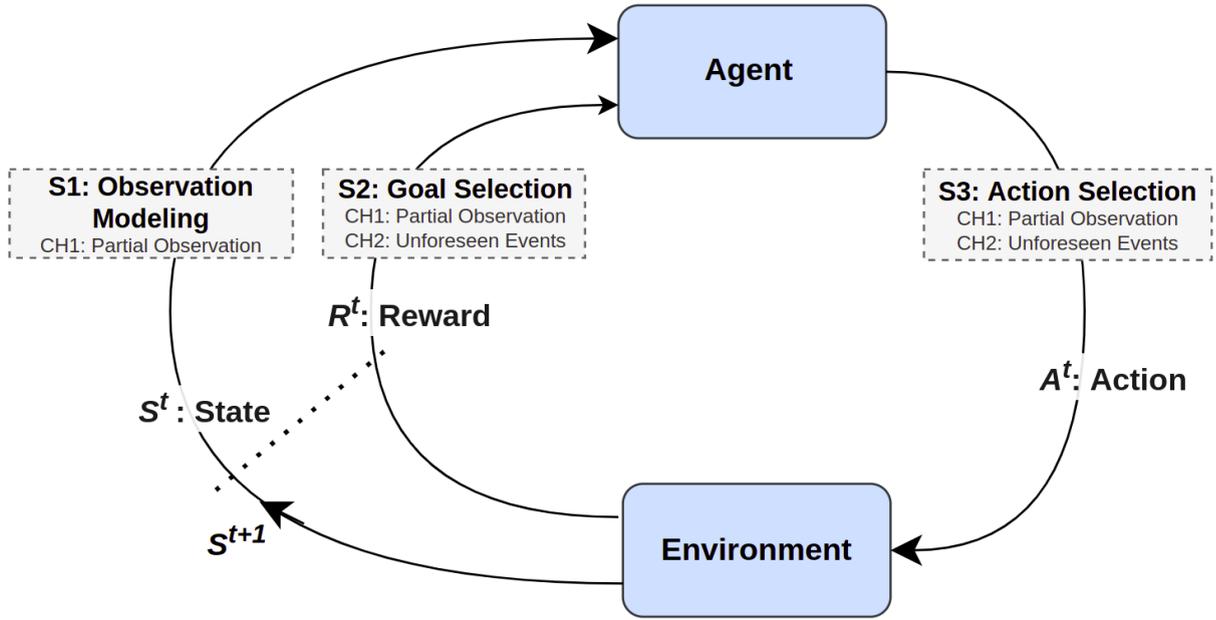

Figure 1.1: Research challenges in the reinforcement learning agent's decision-making process.

## 1.5 Research Questions

The following are the research questions this Ph.D. has addressed. Table 1.1 shows relations between challenges, research objectives, and research questions.

- **RQ1 -** To what extent an ontology-enhanced observation modeling can improve agents' performance in dynamic and partially observable environments?

  Augmenting the observations can help provide additional information that may not be directly observable, but can still be useful for decision-making [151]. Examples of augmented observations can include adding sensor data, generating simulated samples, or incorporating expert knowledge. The hypothesis is that agents can take more effective actions when they are equipped with an augmented model of the environment, for example, in ITSC, vehicles' waiting time is decreased when agents' decision-making process is improved by the provided augmented information for undetectable vehicles.

- **RQ2 -** To what extent ontology-driven knowledge can enable agents to generate new goals that are compatible with the emerging requirements in dynamic and partially observable environments?

  Agents can generate new goals that are compatible with emerging requirements by assessing the changes in the environment and determining which goals are still relevant and achievable. The process may involve re-prioritizing existing goals or setting entirely new ones based on the changing environment. Expert knowledge can also be used to identify new goals that are in line with the evolving requirements. Here, the hypothesis is that agents' action selection and consequently their performance will improve when a new goal is generated to address an emerging requirement. Overall, *generating new goals* requires a process of ongoing assessment and adaptation to ensure that the agent can continue to make progress toward its objectives in dynamic environments [202]. For example, a JSS agent, when observing low due date orders, can deduce a new goal to reduce waiting times for such orders, leading to an increase in the total processed orders and a decrease in the total number of failed orders.

- **RQ3** - To what extent ontology-driven action selection can improve an agent's performance in dynamic and partially observable environments?

  Leveraging expert knowledge, contextual conditions encompass information about the agent's environment (including other agents or entities) along with external factors such as time, location, and user preferences [258]. In [51], the authors have mentioned that agents operating in complex environments with a wide range of varying contextual conditions can use high-level policies to achieve their goals whilst adhering to varying environmental constraints. The Hypothesis here is that having access to contextual information (provided by domain experts) allows agents to make more informed decisions and to better understand their environment, leading to more effective and efficient behavior in new or *unforeseen situations*. For example, to minimize total failed tasks, the same task offloading policy used for Voice over Internet Protocol (VoIP) tasks can also be applied to healthcare tasks since both have the same low due date requirement.

Table 1.1: Relations between challenges, research objectives, and research questions at different stages of the agent's decision-making process.

| Research Questions | Research Objectives | Challenges | | Stages | | |
|---|---|---|---|---|---|---|
| | | CH1 | CH2 | S1 | S2 | S3 |
| RQ1 | RO1 | ✓ | | ✓ | | |
| RQ2 | RO2 | ✓ | ✓ | | ✓ | |
| RQ3 | RO3 | ✓ | ✓ | | | ✓ |

## 1.6 Thesis Contributions

The contributions of this thesis is the comprehensive Ontology-enhanced Decision-Making model (OntoDeM) that contributes in the following two categories: (1) a set of novel algorithms or methods that advance agents' decision-making by providing a structured/augmented representation of knowledge, facilitating reasoning and adaptability to dynamic environments (2) advancement achieved in different real-world application areas by applying the proposed novel algorithms (see Table 1.2).

Table 1.2: Contributions to knowledge.

| Paper | Method | RQ1 | RQ2 | RQ3 | Case study | Evaluation metric | Baseline algorithm |
|---|---|---|---|---|---|---|---|
| P1 | Ontology-based modeling Automatic goal selection/generation | - | ✓ | - | ITSC | Waiting time | Q-learning SARSA DQN |
| P2 P12 | Observation augmentation Observation sampling Reward augmentation Action exploration | ✓ | ✓ | ✓ | ITSC | Waiting time Travel time Travel delay Number of finished trips | Q-learning SARSA DQN |
| P9 | Observation expansion Observation masking Action masking Action prioritization Execution prioritization | ✓ | - | ✓ | EC | Task success rate Task failure rate | DDPG [38] |
| P10 | Observation augmentation | ✓ | - | - | JSS | Average utilization rate Total processed orders | TRPO [142] |
| P13 P14 | Adaptive reward definition | - | ✓ | - | JSS | Average utilization rate Average waiting time Total processed orders Total failed orders | TRPO [142] |
| P15 | Observation abstraction | ✓ | - | - | HSC | Average time spent out of range Weighted MSE error Total reward | DDPG PPO [254] |

#### 1.6.0.1 OntoDeM Scientific Contributions

Figure 1.2 depicts OntoDeM's contribution that relates to different stages of the RL agent's decision-making process. These contributions are as follows:

1. In **ontology-based environment modeling** method, an RL agent models its observation as an ontology-based schema that maps low-level sensor data to high-level concepts, relationships, and properties. This schema allows the agent to access the structure of ontological knowledge of the observation and to conduct logical reasoning based on this knowledge.

2. **Ontology-enhanced observation modeling** to enhance the agent's observations:

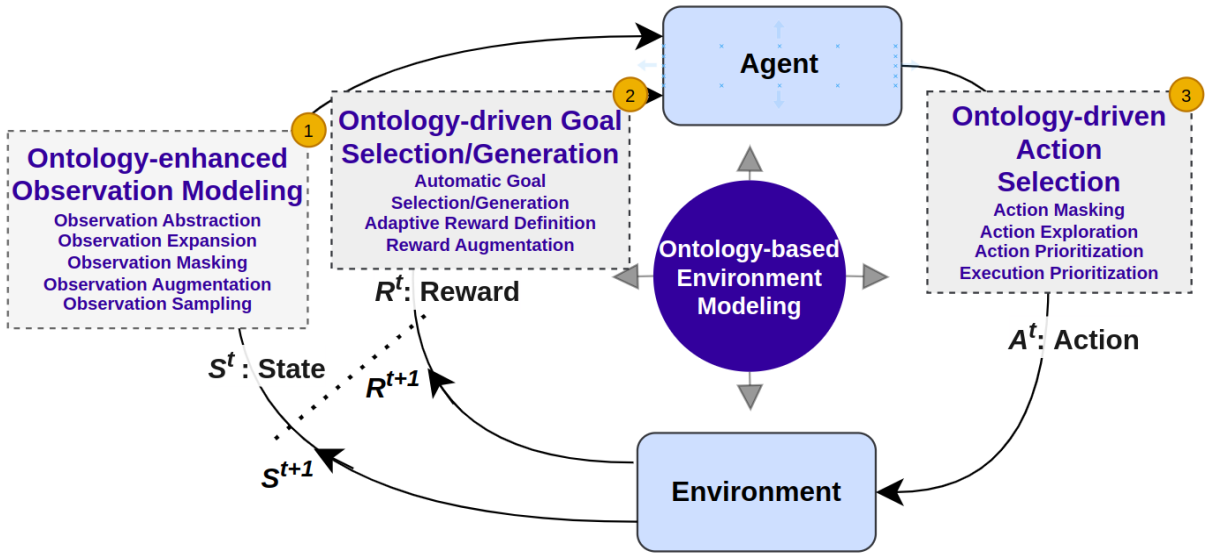

Figure 1.2: The reinforcement learning algorithm is complemented by the Ontology-enhanced decision-making model (OntoDeM) to enhance the decision-making process.

- A novel **observation abstraction** method has been proposed where an agent employs ontological structure and knowledge graph to abstract observations into higher-level representations. Observation abstraction allows the agent to reduce the dimensionality of their observations by focusing on the most relevant information, enhancing their ability to discriminate or identify commonalities between different states and make more accurate predictions about the consequences of their actions. In *partially observable* environments, observation abstraction allows the agent to generalize from observed instances, enabling them to make informed decisions even in situations where specific observations are not directly available or when unforeseen events happen.

- A new **observation expansion** method has been proposed in which the agent extends observations with new concepts, properties, or relationships using the domain-specific ontology. By incorporating these new relevant features into the agent's observation modeling, the agent can better distinguish between similar observations and help the agent generalize its experiences to *unforeseen situations*. In *partially observable* environments, the observation expansion method can incorporate contextual information into the observations, enabling the agent to gain a better understanding of the current situation and make decisions that take into account the relevant context.

- A new **observation masking** method has been proposed in which domain-specific ontology and action ontology help the agent to identify and mask out irrelevant or unnecessary parts of the observation for a specific action selection process. *Partially observable* environments often contain a large amount of irrelevant or noisy information. Observation masking allows the agent to filter out or mask irrelevant observations, focusing only on the relevant information. By reducing the amount of distracting or misleading data, the agent can direct its attention to the most informative and significant aspects of the environment, leading to more accurate decision-making.

- A new **observation augmentation** method has been proposed in which the agent uses an inference engine and forward reasoning technique to extract implicit observation data from explicit ones. In *partially observable* environments, observation augmentation can enrich the observed features with additional information that is not directly observable. By augmenting the missing or incomplete observations with derived features, the agent can obtain a more informative representation of the environment, enabling the agent to make more accurate and informed decisions.

- A new **observation sampling** method has been proposed in which the agent adapts the sampling rate proportional to the concepts' importance, so, observation sampling can be performed adaptively, focusing on areas or instances where more accurate estimations are required. The proposed observation sampling method allows the agent to select a representative subset of observations that captures the variability and characteristics of the environment. By ensuring a diverse and representative sample, the agent can make decisions based on a more comprehensive understanding of the environment, even with limited observations. Moreover, observation sampling can ensure that rare events are captured and included in the agent's observations, enabling the agent to make more effective decisions in situations that occur less frequently but have significant consequences.

3. A set of novel **ontology-driven goal selection/generation** methods that allow agents to adapt their behavior according to the changes in the environment:

    - A novel **automatic goal selection/generation** method has been proposed allowing the agent to continuously search for significant changes in the environment and if necessary, to select a goal from the predefined goal set (based on the proposed goal prioritization mechanism through concept weighting technique) or generate a new goal for itself through a backward reasoning technique. In *dynamic* environments, when agents' current goals are not useful or relevant, new goal generation can significantly improve an agent's decision-making by allowing it to adapt to changing circumstances more

effectively. Moreover, new goal generation can facilitate the agent's exploration by introducing novel objectives or variations of existing goals resulting in gathering more information about the environment.

- Reward Machines (RMs) provide different rewards to agents at different times based on their observations. A new **adaptive reward definition** method has been proposed that dynamically/adaptively creates or modifies RMs, by allowing the agents to extract new transitions and reward functions based on the related ontological concepts and properties they observe. In *dynamic* environments, it may be challenging to define reward functions that accurately capture the desired behavior. Adaptive reward functions can facilitate goal adaptation by dynamically adjusting the rewards associated with different objectives.

- A novel **reward augmentation** method is proposed enabling agents to augment their reward function using the actions' efficiency rate (i.e., the amount of improvement that the action has for the most important concepts of the environment) to improve its action selection. In *dynamic* environments, it can be challenging for the agent to receive timely and informative rewards due to the sparsity of positive feedback. Reward augmentation can address this issue by providing additional intermediate rewards that are easier to obtain or more frequent.

4. **Ontology-driven action selection**, especially in dynamic environment with unforeseen events:

    - A novel **action masking** method has been proposed enabling agents to mask their actions that violate semantic constraints imposed by domain-specific ontology. In *dynamic* environments, the constraints or limitations on the agents' actions can change over time. Action masking allows agents to adapt to these changing constraints by dynamically masking out actions that become infeasible or undesirable or less relevant due to new conditions. Moreover, action masking can help improve decision-making by reducing the action space to a subset of more promising or feasible actions.

    - A new **action exploration** method has been proposed in which agents dedicate a certain percentage of exploration time to the actions that favor the most important concepts of the environment. In *dynamic* environments, action exploration helps the agent maintain flexibility by continuously exploring alternative actions. This flexibility allows the agent to detect and understand unexpected changes (e.g., new obstacles, opportunities, or goals) and find

new solutions to emerging challenges. Moreover, the agent's ability to explore actions based on the weights assigned to ontological concepts allows it to prioritize actions that align with its current objectives.

- New **action prioritization** and **execution prioritization** methods have been proposed in which the agent prioritizes the actions and their execution sequence using criteria extracted from logical rules. In *dynamic* environments, by prioritizing actions based on their urgency or time sensitivity, the agent can ensure that critical tasks are addressed promptly. Also, by focusing on actions that directly contribute to achieving desired objectives, the agent can allocate its resources efficiently. When agents need to make decisions in a sequence of actions, execution prioritization can help the agent determine the optimal order or sequence of actions by considering the value of different actions.

5. The final contribution is the formulation of a comprehensive **guideline**, a flowchart, enabling agents to navigate effectively in various environments when they face different circumstances. This guideline unifies all the proposed methods under a single framework, empowering agents to choose the most suitable approach based on their specific situation.

#### 1.6.0.2 Real-World Application Contributions

1. **ITSC - Improving traffic management systems' performance in emergency events:** In this scenario the proposed OntoDeM's performance is compared to traditional and widely used learning algorithms such as Q-learning, SARSA (State-Action-Reward-State-Action), and Deep Q-Network (DQN). Vehicles' average waiting time, average travel delay, and average number of finished trips have been used to measure the efficiency of the proposed model and baseline algorithms. The results show that the OntoDeM outperforms the baselines under partially observable and dynamic traffic flow, especially when handling unforeseen situations such as emergency and congestion cases.

2. **EC - Improving edge computing systems' performance during unforeseen events and limited data availability:** In this scenario the OntoDeM is compared with the baseline Deep Deterministic Policy Gradient (DDPG) algorithm. The results show that the proposed model increases the total number of processed tasks, decreases the total number of failed tasks, increases the total number of processed tasks satisfying security requirements, and decreases the average standard deviation of usage history compared to the baseline algorithm.

3. **JSS - Enhancing job shop scheduling adaptability during unforeseen events while addressing challenges of limited observations for machines/orders:** In this scenario a JSS environment with a high level of partial observability is chosen as a case study where the OntoDeM's performance is compared to the Trust Region Policy Optimization (TRPO) learning algorithm. Machines' average utilization rate, orders' average waiting time, total processed orders, and total failed orders have been used to measure the efficiency of the proposed model and baseline algorithms. The results show that OntoDeM can improve the agent's observation and outperform the baseline algorithm.

4. **HSC - Enhancing heating system control's robustness amidst unforeseen conditions, while addressing limited observations:** In this scenario the focus is on managing noisy outdoor temperature data, and the performance of the OntoDeM is compared against the Proximal Policy Optimization (PPO) and Deep Deterministic Policy Gradient (DDPG) algorithms. Evaluative metrics encompass average duration spent outside the optimal temperature range, weighted minimum square error, and the extent to which temperature values remain within the desired interval. The findings prominently illustrate that OntoDeM enhances the system's observation modeling and showcases superior performance when compared with the baseline algorithms.

## 1.7  Thesis Outline

Chapter 2 presents a review of the relevant literature. Chapter 3 demonstrates the process of mapping raw sensor data into high-level concepts through environment modeling and Chapter 4 describes proposed OntoDeM in detail. In Chapter 5, the case studies are described. In Chapter, 6 the experimental scenarios are defined and the results are analyzed, with the conclusion and future plan being discussed in Chapter 7.

# CHAPTER

# TWO

# RELATED WORK

In this chapter, I present a comprehensive review of existing literature and contemporary approaches employed to address the issues surrounding decision-making by agents in environments that are partially observable and dynamic. This discussion is organized into three key stages: observation modeling, goal selection, and action selection (refer to Table 2.1 for a concise summary of the approaches under review). Also, there is a fourth subsection that will illustrate the ontology-based approaches.

| Sec. | Subsection | Key points | Page number |
|---|---|---|---|
| | Partially Observable Markov Decision Processes | - Challenges of sequential decision-making in partially observable environments. - Introduction to MDPs and POMDPs. - Role of belief states. | 21 |
| | Hidden Markov Models | - Probabilistic modeling with hidden states. - Transition and emission probabilities. - Limitations of Markov assumptions. | 22 |
| | Monte Carlo Methods | - Particle filters. - Estimating hidden states and expected utilities. - Handling noisy or high-dimensional observations. | 23 |
| Observation Modeling Approaches | Neural Network-based Models | - RNNs for capturing temporal dependencies. - Processing sequential data in partially observable environments. - Handling noisy or incomplete observations. | 24 |
| | Scene Graphs | - Capturing spatial and semantic relationships. - Enhancing environment understanding and decision-making. - Limitations and challenges in handling complex semantic relationships. | 26 |
| | Bayesian Learning | - Combining prior knowledge with observed data. - Making probabilistic inferences. - Handling data limitations and real-time constraints. | 27 |
| | Predictive State Representation | - Representing the environment using predictive state features. - Sensitivity to observation inaccuracies and dynamic environments. | 28 |

(a) Summarized overview of literature review (observation modeling approaches).

| Sec. | Subsection | Key points | Page number |
|---|---|---|---|
| Goal Selection Approaches | Goal Formation, Goal Generation, and Goal Reasoning Techniques | - Transformation of unachievable goals into concrete ones. - Goal generation factors. - Goal reasoning as a case-based system. | 29 |
| | Practical Reasoning | - Incorporating desires and preferences. - Decision-making based on available information. - Social reasoning. - Predefined reasoning rules. | 30 |
| | Heuristic-based Approaches | - Use of heuristics for goal prioritization. - Factors guiding heuristic-based decisions. - Computational efficiency. - Adaptability. | 31 |
| | Learning Techniques | - Adaptation to unforeseen situations. - Data availability challenges. | 32 |
| | Adaptive Reward Signals | - Continuous evaluation of goal progress. - Shaping rewards. - Challenges with dynamic environments. - Reward function design. | 33 |

(b) Summarized overview of literature review (goal selection approaches).

| Sec. | Subsection | Key points | Page number |
|---|---|---|---|
| Action Selection Approaches | Reinforcement Learning | - RL agents learn from trial and error in dynamic environments. - RL agents adapt their strategies and optimize action sequences. - Value-based and policy gradient RL algorithms. - Challenges in high-dimensional state spaces and dynamic environments. | 36 |
| | Optimization Techniques | - Minimize/maximize an objective function, subject to certain constraints. - Flexible scheduling. - Local optima, choice of algorithm and hyperparameters, computational resources, noisy objective functions, and struggles with uncertainty and ambiguity. - Role of ontology. | 37 |
| | Self-adaptive Systems | - SASs employ adaptive algorithms based on real-time feedback. - Handling uncertainties with probabilistic reasoning and risk assessment. - Use of AI and ML techniques in SAS. - Challenges in adapting to rare events and balancing exploration. | 39 |
| | Online Planning | - Continuous update of action selection based on real-time feedback. - Challenges in real-time decision-making and adapting to changing environments. | 42 |
| | Exploration Algorithms | - Active exploration to reduce agent uncertainty. - Count-based, intrinsic motivation, and probability-based exploration. - Challenges in balancing exploration and exploitation. | 43 |
| | Semi-stochastic Action Selection | - Combining deterministic and stochastic components. - Adjusting the balance based on learning progress and uncertainty. - Challenges in adapting to dynamic environments. | 44 |

Table 2.1 (Continued): Summarized overview of literature review (action selection approaches).

## 2.1 Observation Modeling Approaches

In the exploration of observation modeling approaches, I begin with Partially Observable Markov Decision Processes (POMDPs), a fundamental framework for handling sequential decision-making in partially observable environments. In the Hidden Markov Models (HMMs), I explore probabilistic modeling, where observations are used to estimate hidden states. Monte Carlo Methods then offer a probabilistic alternative for estimating states while incorporating uncertainty. In the context of deep learning, Neural Network-based Models utilize sequential data to capture complex dependencies. Scene Graphs combine spatial and semantic information, enabling richer environment understanding. Bayesian Learning subsequently merges prior knowledge with observed data to make informed inferences. Finally, Predictive State Representation (PSR) employs predictive state features to model and predict the environment's state.

## 2.1.1 Partially Observable Markov Decision Processes

Sequential decision-making problems refer to problems in which an agent (decision-maker) must make a series of decisions over time, each decision influencing the outcome of subsequent decisions [70]. *Markov Decision Process (MDP)* [247] is a mathematical framework used for modeling sequential decision-making problems. In the formulation of MDP $\mathcal{M} = (\mathcal{S}, \mathcal{A}, r, p, \gamma)$, at time step $t \geq 0$, an agent is in state $s^t \in \mathcal{S}$, takes an action $a^t \in \mathcal{A}$, receives an instant reward $r^t = R(s^t, a^t) \in \mathbb{R}$ and transitions to a next state $s^{t+1} \sim p(\cdot \mid s^t, a^t) \in \mathcal{S}$. $R^t$ denotes the specific reward function tailored to the problem at time step $t$, and it is employed to compute the value of the reward $r^t$. Let $\pi : \mathcal{S} \mapsto P(\mathcal{A})$ be a policy, where $P(\mathcal{A})$ is the set of distributions over the action space $\mathcal{A}$. The discounted cumulative reward under policy $\pi$ is $J(\pi) = \mathbb{E}_\pi \left[ \sum_{t=0}^{\infty} \gamma^t r \right]$, where $\gamma \in [0, 1)$ is a discount factor. The objective of MDP is to search for a policy $\pi$ that achieves the maximum cumulative reward $\pi^* = \arg\max_\pi J(\pi)$. Under policy $\pi$, the value function can be defined as the expected cumulative reward of the agent in a certain state $s$, $V^\pi(s) = \mathbb{E}_\pi \left[ J(\pi) \mid s_0 = s, a_0 \sim \pi(\cdot \mid s_0) \right]$. The action value function returns the expected cumulative reward for using action $a$ in a certain state $s$, $Q^\pi(s, a) = \mathbb{E}_\pi \left[ J(\pi) \mid s_0 = s, a_0 = a \right]$. The advantage function measures how much a certain action $a$ is a good or bad decision given a certain state $s$, $A^\pi(s, a) = Q^\pi(s, a) - V^\pi(s)$.

Sequential decision-making problems in partially observable and dynamic environments refer to situations where an agent needs to make a series of decisions over time while having limited or incomplete information about the underlying state of the environment (e.g., due to sensor noise or occlusions), and where the environment itself (e.g., characteristic of different entities, availability of different resources) can change or evolve over time. These types of problems, commonly encountered in real-world scenarios such as traffic signal control systems, cannot be completely captured by MDPs and are typically formulated as a *Partially Observable Markov Decision Process (POMDP)* [147]. A POMDP $(\Omega, \mathcal{S}, \mathcal{A}, r, p, \mathcal{W}, \gamma)$ [247] is an MDP with two additional components: the possibly infinite set $\Omega$ of observations, and the $\mathcal{W} : \mathcal{S} \to \Omega$ function that generates observations $v^t$ based on the unobservable state $s^t$ of the process through a set of conditional observation probabilities (i.e., probability distribution over possible states based on past observations and actions). At each time, the agent receives an observation $v^t \in \Omega$ which depends on the new state of the environment, $s^t$, and on the most recently taken action, $a^t$, with probability $\mathcal{W}(v^t, a^t, s^t)$. Learning is difficult in these set-tings due to partial observability, as the same observation may be obtained from two different states, each requiring a different optimal action.

*Example in ITSC.* Consider a busy intersection with four lanes, each equipped with sensors to monitor the traffic flow. The system maintains a belief state, which is a probability distribution over the possible states of the traffic system, such as the number of vehicles in each lane or the traffic flow rate. Initially, the belief state assigns equal probabilities to all possible traffic conditions. As time progresses, the sensors provide partial observations about the traffic, such as the number of vehicles detected in each lane. Using these partial observations, the POMDP model updates the belief state by incorporating the observed information. For example, if the sensor in lane 1 detects a high number of vehicles, while the sensor in lane 2 detects a low number of vehicles, the POMDP model may update the belief state to assign a higher probability to the possibility of lane 1 having congested traffic and lane 2 having smooth traffic.

**Limitations:** The complexity of finding an optimal policy in a POMDP grows exponentially with the size of the state and observation spaces, making it challenging for real-time decision-making in complex environments [83, 98]. Also, in practice, sensors can be noisy, resulting in imperfect observations which can affect the accuracy of the belief state estimation [176, 286]. For instance, imagine a large city with numerous intersections, each equipped with sensors to gather information about traffic conditions. This can make it challenging to find an optimal policy in real-time, especially in complex environments where there are numerous intersections and a large number of possible traffic states. Additionally, a sensor might occasionally fail to detect vehicles impacting the accuracy of the belief state estimation.

### 2.1.2 Hidden Markov Models

Hidden Markov Models (HMMs) [193] are probabilistic models used to analyze sequential data. They consist of a set of hidden states representing the underlying system and a set of observable symbols associated with each state. HMMs assume that the observed symbols are generated by the hidden states, but the states themselves are not directly observable. The model is characterized by transition probabilities, which define the likelihood of transitioning from one hidden state to another, and emission probabilities, which determine the probability of observing a particular symbol given a hidden state. HMMs employ the Forward-Backward algorithm [18] or the Viterbi algorithm [211] to efficiently compute the most likely sequence of hidden states given the observed symbols or vice versa. In a partially observable environment, HMMs allow for estimating the underlying hidden states based on observable data.

*Example in ITSC.* The HMM model considers two hidden states: "high traffic" and "low traffic". The observed states are the traffic flow rates measured by sensors at the intersection. When the traffic flow rates are high, indicating congestion, the HMM will transition to the "high traffic" state. On the other hand, when the traffic flow rates are low, indicating smooth traffic flow, the HMM will transition to the "low traffic" state.

**Limitations:** HMMs assume that the underlying system follows the Markov property, meaning that the probability of transitioning to a new state depends only on the current state and not on the past states. This assumption may not always hold in real-world scenarios, particularly in dynamic environments where the system's behavior may be influenced by a long history of states and observations [223, 207]. In other words, HMMs assume that observations directly depend only on the current hidden state and do not account for more complex dependencies or context [218]. This can be a limitation in scenarios where richer contextual information or non-linear relationships exist. Also, HMMs heavily rely on sufficient training data to estimate model parameters accurately [182, 13]. In situations where data is scarce, or noisy, the performance of HMMs may be compromised. Additionally, HMMs may struggle when faced with rare events that are not well-represented in the training data [136]. As an example of limitations in the ITSC domain, during peak hours, traffic congestion can build up gradually, and the behavior of the system may depend on the traffic conditions experienced over an extended period. Additionally, contextual factors like weather conditions, special events, or road infrastructure changes may affect traffic behavior. Moreover, when an ITSC system is deployed in a newly developed neighborhood with limited historical traffic data, there may be a scarcity of representative training data for the HMM to accurately estimate the underlying traffic patterns and model parameters. Furthermore, when a rare event such as a major sporting event or a large-scale construction project significantly impacts traffic patterns in the area, and it is not adequately represented in the training data, the HMM may struggle to handle this rare event effectively.

### 2.1.3 Monte Carlo Methods

Monte Carlo methods [34], such as particle filters or sequential Monte Carlo methods, can estimate the hidden state of a system in a partially observable environment. By propagating a set of particles through the system's state space and updating their weights based on observations, Monte Carlo methods provide a probabilistic estimate of the system's true state. This enables effective state estimation even in the absence of complete observability. Also, by sampling possible actions and simulating their outcomes, Monte Carlo methods can estimate the expected utility or reward associated with different decisions. This information aids decision-making by identifying actions that are likely to lead to favorable outcomes under uncertain conditions.

*Example in ITSC.* During the search process, Monte Carlo Tree Search (MCTS) simulates multiple possible future traffic scenarios by sampling actions and observations from the underlying probability distribution. Each simulation represents a potential trajectory of the traffic system. By repeatedly simulating and updating the search tree, MCTS builds knowledge about the uncertain traffic conditions and identifies the most promising actions to take at each decision point.

**Limitations:** If the observations are noisy or incomplete, it can be challenging to accurately estimate the underlying system states by Monte Carlo methods [165]. Also, when the dimensionality of the observations is high, the number of samples required to achieve accurate estimates grows exponentially. This can limit Monte Carlo's effectiveness in real-time scenarios [189]. Monte Carlo methods typically rely on pre-defined models or assumptions about the underlying system dynamics. However, in the presence of unforeseen events, these models may not be accurate. If the system encounters novel or rare events that were not adequately sampled or represented in the training data, the Monte Carlo estimates may not generalize well [203, 257]. For example in ITSC domain, a vehicle detection sensor might occasionally misclassify a stationary object as a vehicle. This inaccuracy can lead to incorrect estimations of the traffic flow, which can impact the effectiveness of the Monte Carlo method in determining optimal signal timings. If the Monte Carlo method heavily relies on noisy observations without considering methods to compensate for the noise, the estimated traffic flow may not accurately represent the actual traffic conditions. Additionally, if the system needs to consider multiple lanes, each with varying traffic volumes, vehicle types, and speeds, the number of potential combinations becomes exponentially large. This high-dimensional observation space makes it difficult for the Monte Carlo method to explore and sample enough scenarios to achieve accurate estimations of the traffic state within real-time constraints. Moreover, if the system encounters novel or rare events (e.g., public protests, major sporting events, unplanned road construction, and natural disasters) that were not adequately sampled or represented in the training data, the Monte Carlo estimates may not generalize well, resulting in suboptimal signal control decisions.

### 2.1.4 Neural Network-based Models

A neural network is a computational model composed of interconnected artificial neurons that can be trained to learn and make predictions from data. In traditional feedfor-ward neural networks, each input is processed independently, and there is no inherent notion of order or sequence. However, in many real-world applications, the order of data points is crucial for understanding and making predictions. Recurrent Neural Networks (RNNs) address this limitation by introducing a feedback loop that allows information to flow from one time step to another [129]. The hidden state in an RNN acts as a memory that captures the information from previous time steps. This way, the network can capture temporal dependencies present in the sequential data. In the context of partially observable environments, the agent receives a series of observations over time, where each observation provides partial information about the current state. By leveraging the sequential nature of the data, an RNN can use the hidden state to encode information from past observations, enabling the agent to understand the context and make more informed predictions or decisions [79].

*Example in ITSC.* Suppose an autonomous traffic signal control system aims to optimize traffic signal timings to reduce congestion and improve traffic flow. The system receives sequential sensor observations from various sources, such as traffic cameras, vehicle detectors, and pedestrian sensors. Each observation provides partial information about the current state of the intersection, including the number and types of vehicles, pedestrian activity, and current traffic conditions. As the RNN processes the sequential sensor observations, it learns to capture the relationships between different time steps. For example, it may learn that during peak hours, increased traffic volume in one direction leads to longer queues and delays in the opposite direction. By encoding this temporal knowledge in the hidden state, the RNN can understand the context of the traffic flow and make more informed predictions about future traffic patterns.

**Limitations.** Although RNNs can capture short-term dependencies, they may struggle to model long-term dependencies effectively [135, 292]. Also, neural networks can be sensitive to noisy or incomplete observations. Even small errors in the input data can propagate through the network and affect the predictions or decisions [81, 237]. Additionally, neural networks learn patterns and relationships from training data, and their performance heavily relies on the data they have been exposed to. If the network encounters unforeseen events or situations that were not adequately represented in the training data, its predictions or decisions may not generalize well [156]. Moreover, their computational demands may restrict the feasibility of deploying these models in systems with strict latency requirements [26]. Additionally, neural networks are often considered black-box models due to their complex internal representations. Understanding how the network arrives at its predictions or decisions can be challenging [200]. For instance, developing accurate neural network models for ITSC systems often requires extensive data that includes information about traffic flow, historical patterns, and corresponding optimal signal timings. Traffic patterns and congestion levels can vary significantly across different intersections and time periods, making it challenging to obtain representative and diverse data. Also, traffic patterns can change based on various factors such as time of day, day of the week, special events, and road construction. Neural network models trained on historical data may struggle to adapt to these dynamic changes and may require frequent retraining or updating to maintain their accuracy. In ITSC systems, decisions need to be made quickly and efficiently to optimize traffic flow. Neural networks, especially deep architectures, can be computationally intensive and may not always meet the real-time requirements of ITSC systems.

### 2.1.5 Scene Graphs

A scene graph [281] generates a graph that captures spatial and semantic relationships between objects in the environment. Agents can utilize this information to reason about the dependencies and interactions among objects, infer the state of unobserved objects, anticipate the effects of their actions on the entire environment, rather than just the observed parts, and make contextually appropriate decisions. Also, scene graphs facilitate the generalization of knowledge across similar environments and enhance the agent's decision-making in unfamiliar situations. In [15], the authors proposed an algorithm called Scene Analysis for Robot Planning (SARP) that enables robots to reason with visual contextual information toward achieving long-term goals under uncertainty. SARP uses local scene graphs of single images to build and augment global scene graphs.

*Example in ITSC.* Suppose we have a road intersection with multiple lanes, traffic signals, and vehicles. The scene graph captures the connections and attributes of these elements. The intersection node is the root of the graph, connected to nodes representing each lane, traffic signal, and vehicle. Each lane node contains information such as lane direction, speed limit, and lane occupancy. The traffic signal nodes represent the state of the traffic signal phases, indicating whether they are red, green, or yellow. The vehicle nodes contain details about each vehicle, including its position, speed, and destination.

**Limitations:** Scene graphs lack the expressive power to capture complex semantic relationships and domain-specific knowledge that ontologies can provide [269, 84, 121, 16]. Ontologies allow for a more nuanced representation of concepts, attributes, and relationships, enabling agents to reason at a higher level of abstraction. Scene graphs can struggle with handling unanticipated changes in the environment or the introduction of new objects or relationships [15, 97]. Ontologies, with their flexibility and extensibility, can better accommodate changes and allow for more adaptive decision-making.

For instance, in the traffic domain, using a scene graph representation, the system may have a structured representation of the physical objects at an intersection, such as: "Traffic signal", "Vehicle detector", "Pedestrian crossing", and "Traffic camera". The scene graph can capture basic relationships like "Traffic signal is connected to vehicle detectors" or "Pedestrian crossing is located near the traffic signal". However, when it comes to capturing complex semantic relationships and domain-specific knowledge, scene graphs can be limited. For instance, when there is a need to represent and reason about the impact of air quality and noise levels on signal timings, this would involve considering factors such as pollutant levels, noise thresholds, and their influence on the health and well-being of pedestrians and nearby residents. Scene graphs may struggle to capture and represent these complex semantic relationships between environmental factors, traffic signals, and human well-being. Furthermore, when new types of environmental sensors are added or changes are made to environmental regulations, the scene graph may lack the flexibility to accommodate these changes without modifications to its structure and relationships.

### 2.1.6 Bayesian Learning

Bayesian learning [265] is a statistical learning approach that combines prior knowledge with observed data to make inferences and update beliefs. It follows the principles of Bayes' theorem, which calculates the posterior probability of a hypothesis given the observed data. Bayesian learning provides a framework for incorporating prior knowledge, handling uncertainty, and making probabilistic inferences based on observed data.

*Example in ITSC.* For instance, the traffic engineer has a prior belief about the expected traffic patterns at different times of the day. This prior knowledge could be based on historical data or an expert judgment. As vehicles pass through the intersection, sensors collect data on the number of vehicles, their speeds, and the duration of traffic congestion. Bayesian learning is then used to update the beliefs about the traffic patterns based on the observed data. By considering both the prior knowledge and the observed data, Bayesian learning allows for a more accurate and adaptive adjustment of signal timings to match the actual traffic conditions.

**Limitations:** As the number of dimensions increases, the amount of data required to accurately estimate the probability distributions grows exponentially [300]. Insufficient training data may lead to unreliable posterior estimates [52]. Also, Bayesian learning heavily relies on the choice of a suitable probabilistic model that represents the dynamics of the system. Selecting an appropriate model can be challenging, especially in dynamic environments where the underlying processes may change over time [55]. Bayesian learning relies on historical data to estimate probability distributions and update beliefs. However, for example, in ITSC, historical data might be limited or insufficient, especially in scenarios where traffic patterns change frequently or when the intersection is in a newly developed area. Moreover, traffic conditions can be highly dynamic, influenced by factors such as time of day, weather, special events, and road incidents, making it challenging for Bayesian learning to adapt quickly to these dynamic conditions. Traffic signal control requires real-time decision-making to dynamically adjust signal timings. However, Bayesian learning often involves computationally intensive tasks such as posterior estimation, which may not be feasible in real-time scenarios.

### 2.1.7 Predictive State Representation

In Predictive State Representation (PSR) [154], the state of the environment is represented by a vector of predictive state features. These features capture the relevant information from the history of observations and actions that allow the agent to make predictions about future observations. In standard PSR, the number of predictive state features grows linearly with the length of the history, which can lead to scalability issues in complex environments. Sparse PSR addresses this problem by using a subset of predictive state features that capture the most relevant information for making predictions. While the standard PSR captures the history of observations and actions, extended PSR includes extra information (i.e., domain-specific rules, expert knowledge) beyond the raw input data.

*Example in ITSC.* Consider a specific intersection that experiences heavy traffic congestion during peak hours. A PSR model can be built to predict the traffic conditions at this intersection. The model incorporates observed data from sensors, such as vehicle counts and flow rates, and contextual information like time of day and historical traffic patterns. By learning the state transition and observation models, the PSR model can predict future traffic conditions, such as the expected queue length or the likelihood of congestion, based on the current state and control actions. This information can then be used to optimize the traffic signal timings, ensuring that green phases are assigned to lanes with high traffic volume and congestion, thereby improving traffic flow efficiency and reducing delays at the intersection.

**Limitations:** While PSR is designed to handle partially observable environments, its effectiveness can be limited in situations where observations provide incomplete or noisy information about the underlying state [103, 20]. Also, PSR does not inherently handle changes or non-stationarity in the environment. If the dynamics of the system evolve over time, the PSR model needs to be adapted to maintain accurate predictions [289]. For example, one limitation of PSR in the ITSC context is its sensitivity to inaccuracies or delays in the observations used to estimate the underlying traffic state. For instance, if there is a delay in receiving traffic data from sensors or if the sensors provide noisy measurements, the PSR model may struggle to accurately predict future traffic conditions. Additionally, if there are sudden changes in traffic patterns, such as due to accidents or road closures, the PSR model may require time to adapt and update its predictions, which can further impact the effectiveness of the ITSC system.

## 2.2 Goal Selection Approaches

Exploring various goal selection approaches, I commence with Goal Formation, Goal Generation, and Goal Reasoning techniques. These techniques show how agents transform objectives into actionable goals, generate relevant goals based on various factors, and employ a case-based system for goal reasoning. However, it reveals limitations in the face of unpredictability. In Practical Reasoning, I explore how agents incorporate preferences and navigate social contexts (social norms, expectations, preferences) to prioritize meaningful goals. However, practical reasoning struggles with predefined rules that may not cover all scenarios. Heuristic-based approaches introduce rule-of-thumb strategies, allowing agents to efficiently prioritize goals, but these strategies might lack adaptability. Learning techniques emerge as a solution, empowering agents to adapt and learn from experiences. Still, they face challenges with data availability and unforeseen events. Lastly, Adaptive Reward Signals offer a continuous evaluation mechanism, shaping rewards based on progress, but they require careful design in dynamic environments, often posing a challenge.

### 2.2.1 Goal Formation, Goal Generation, and Goal Reasoning Techniques

Goal formation is a vital aspect of intelligent agents' decision-making processes. It involves the transformation of initially unachievable goals into concrete ones, breaking down complex objectives into manageable sub-goals that can be pursued step by step, ultimately leading to the desired outcome [59]. During goal generation process, agents consider a combination of factors, including conditional beliefs, obligations, intentions, desires, motivations, and their own preferences. These elements guide the selection of relevant goals that align with the agent's internal state and objectives. Goal generation occurs when the agent detects inconsistencies between sensory inputs and expectations, prompting the creation of goals aimed at resolving these discrepancies [140]. Additionally, goal reasoning is a case-based system that employs active and interactive learning to automatically select goals from a predefined set. This process enables agents to reason about available goals, taking into account their beliefs, the state of the environment, and current priorities, to make informed decisions regarding which goals to pursue in a given context [276].

*Example in ITSC.* Goal formation involves defining objectives such as optimizing traffic flow and reducing congestion. Goal generation techniques can include analyzing traffic patterns to prioritize congested areas, considering environmental factors like air quality to minimize emissions, and accommodating emergency responses or special events. Goal reasoning is used through real-time traffic monitoring, allowing the system to dynamically adjust goals based on changing conditions.

**Limitations.** In partially observable environments, agents may not have complete information about the current state of the system, which can hinder the accurate formation of goals [235, 282]. Unforeseen events may introduce new information or change the environment in unexpected ways, causing the pre-existing goals to no longer align with the new situation. This can make it challenging for agents to formulate appropriate goals promptly without a comprehensive understanding of the situation [49, 126, 276]. For example, consider a situation where an ITSC system has generated goals based on historical traffic patterns and known road conditions. However, a sudden accident occurs, causing road closures and diverting traffic in unexpected ways. The agent responsible for signal control may lack complete information about the accident and the resulting traffic congestion, hindering their ability to accurately formulate goals. The pre-existing goals, optimized for the previous context, may no longer align with the changed situation, making it challenging for the agents to promptly adapt and generate new relevant goals.

### 2.2.2 Practical Reasoning

Practical reasoning incorporates agents' desires and preferences, enabling them to make informed decisions based on the available information and ensuring that the selected goals align with their individual objectives and motivations [140]. This helps agents prioritize goals that are most relevant and meaningful to them in a given context. Additionally, social reasoning plays a crucial role in practical reasoning by enhancing agents' understanding of other agents' goals and dependencies [100]. Practical reasoning relies on predefined reasoning rules to handle different situations. These rules are typically developed based on prior knowledge and experience.

*Example in ITSC.* When the system detects high traffic congestion, a predefined rule can trigger the goal of optimizing traffic flow by adjusting signal timings to reduce delays and queues at congested intersections. In the event of an emergency, a predefined rule can prioritize the goal of providing efficient and safe passage to emergency vehicles by adjusting signal timings and granting them the right-of-way.

**Limitations.** Reasoning rules may not cover all possible scenarios, making it challenging to generate appropriate goals when faced with unforeseen events [139, 140]. For example, consider a scenario where a sudden large-scale public event, such as a protest, takes place near an intersection that is managed by the ITSC system. Since this event was not accounted for in the predefined rules, the system may struggle to adapt its goal selection process to handle the large number of pedestrians and vehicles associated with the event.

### 2.2.3 Heuristic-based Approaches

These approaches leverage heuristics, which are rule-of-thumb techniques or strategies that guide decision-making based on simplified problem-solving methods. Agents can employ heuristics to prioritize goals based on specific factors. Factors guiding heuristic-based decisions refer to the criteria or considerations that heuristics use to prioritize actions or strategies when making decisions. These factors help heuristics determine the most suitable course of action by taking into account elements such as urgency, importance, resource availability, or expected outcomes within a problem-solving context. These heuristics help agents make efficient decisions by focusing on the most relevant goals given the limited observability of the environment. Also, agents can quickly assess the situation and adjust their goal selection strategies to address the changing dynamics of the environment. Heuristic-based approaches offer practical advantages in terms of computational efficiency, simplicity, and adaptability. These approaches allow agents to make reasonable decisions in real-time, even when faced with partial observability and dynamic changes in the environment [199].

*Example in ITSC.* A heuristic-based approach such as the Max Pressure algorithm [157] prioritizes reducing congestion by dynamically adjusting signal timings based on the observed "pressure" at each approach of an intersection. Approaches with higher vehicle queues are allocated longer green times to allow more vehicles to clear the intersection, while approaches with lower queues receive shorter green times. By using this heuristic, the Max Pressure algorithm aims to optimize traffic flow and minimize delays. Another example of a heuristic-based approach is the Time of Day (ToD) scheduling algorithm [150]. The idea behind the ToD scheduling algorithm is to assign different signal plans for different time periods of the day, taking into account expected traffic patterns and demands during each period.

**Limitations.** Heuristics-based approaches are not guaranteed to always yield optimal solutions and may be context-specific. Fine-tuning and refining the heuristics based on the specific problem domain and environmental conditions are necessary to ensure their effectiveness in guiding the agent's goal selection process. Additionally, heuristics are typically designed based on predefined strategies. They may not be capable of adapting to unforeseen events or significant changes in the environment [186, 251, 86]. For example in the traffic domain, the Max Pressure algorithm's focus on minimizing pressure may not adequately address specific traffic patterns that vary throughout the day, leading to suboptimal signal timings during different time periods. Similarly, the ToD scheduling algorithm, while accounting for different time periods, may not fully account for variations in traffic patterns. Without the ability to adapt and respond to such variations, algorithms may fail to optimize signal timings in unforeseen situations, leading to increased congestion or delays.

### 2.2.4 Learning Techniques

To overcome the limitations of previous approaches (e.g., goal formation/generation, practical reasoning, and heuristic-based), researchers have emphasized the importance of incorporating learning mechanisms, enabling agents to acquire new knowledge and update their reasoning rules based on their experiences. By integrating learning into the system, agents can adapt their goal selection strategies to effectively handle unforeseen situations, enhancing their decision-making capabilities in the face of uncertainty and changing circumstances. In [77], a reverse learning approach was proposed, allowing agents to gradually learn from a set of initial states near the goal. [106] presented a framework where agents learn both a world model and a self model concurrently. The world model aids in predicting the dynamic consequences of the agent's actions, while the self model estimates the errors in the world model, guiding future exploration.

*Example in ITSC.* Unexpected events such as a major sporting event, result in a significant increase in traffic volume and altered traffic patterns that were not previously encountered. By observing the state of the traffic, the agent can update its reasoning rules by learning that, during such events, it is beneficial to prioritize longer green signal duration on routes leading away from the stadium to facilitate smoother traffic flow.

**Limitations.** Learning techniques typically rely on historical data to train models and inform goal selection. However, in the presence of unforeseen events, there may be a lack of relevant training data to adequately capture the dynamics of the new situation. This can limit the system's ability to accurately select goals in real-time. Additionally, unforeseen events may introduce novel scenarios that may deviate significantly from the training data, making it difficult for the system to generalize its learned knowledge and adapt its goal selection strategies accordingly [77, 68, 106]. For instance, consider an ITSC system trained on historical data representing regular traffic patterns. If an unexpected event, such as a natural disaster or a large-scale accident, occurs, the system may encounter traffic conditions that it has not encountered before. The system may struggle to generalize its learned knowledge to effectively respond to the unique traffic demands and optimize signal timings in real time.

### 2.2.5 Adaptive Reward Signals

Learning strategies can guide agents toward their goals using reward or fitness functions. In partially observable and dynamic environments, goals may need to be adapted based on the changing conditions. Adaptive Reward Signals can be used to continuously evaluate the progress toward a goal and adjust the rewards accordingly. Additionally, adaptive reward signals can be used to shape the rewards in a way that encourages the agent to focus on specific sub-goals or intermediate objectives. By providing rewards that reflect progress towards these sub-goals, the agent can break down complex tasks into more manageable components and learn to achieve them step by step [152].

Using learning techniques, agents face a difficult challenge when they do not know the reward function [194]. Imitation Learning (IL) enables agents to reach any goal without any need for reward by learning from human demonstrations [60]. Also, using Inverse Reinforcement Learning (IRL) and Inverse Reward Design (IRD) agents can infer a reward function from the expert demonstrations assuming that the expert policy is optimal regarding the unknown reward function [107].

**Reward Machines** (RMs) from [255] directly represent the reward functions as Finite-State Automata (FSA), dividing a complex task into sub-tasks with distinct rewards for each. There have been efforts to learn a so-called perfect RM in [256] from the experience of an RL agent in a partially observable environment. They formulated finding a perfect RM problem as an optimization problem. Their method focuses on the RM that predicts what is possible and impossible in the environment at the abstract level. Their method still needs to know a set of high-level propositional symbols and a labeling function that can detect them prior and does not work when there is an unforeseen observation for which there is no experience in the training set. Learning Finite State Controllers (FSC) [178] suggested looking for the smallest RM that correctly mimics the external reward signal given by the environment and whose optimal policy receives the most reward. It is a desirable property for RMs to have, but it requires computing optimal policies to compare their relative quality, which seems prohibitively expensive. Deterministic Markov Models (DMMs) [172] proposed learning the RM that remembers sufficient information about history to make accurate predictions about the next observation. However, their approach needs to keep track of more information that will not necessarily result in better predictions, especially in dynamic and noisy environments. In [80], the authors defined Hierarchies of RMs (HRMs) in which RM and its associated policies are reusable across several RMs. [299] proposed Symbolic RMs (SRMs) for incorporating high-level task knowledge when specifying the reward functions. It facilitates the expression of complex task scenarios and the specification of symbolic constraints. [299] developed a hierarchical Bayesian approach that can concretize an SRM by inferring appropriate reward assignments from expert demonstrations. However, these approaches do not work when the agent does not have access to expert demonstrations.

**Reward shaping** proposed by [192] adds state-based potentials to the reward in each state. Exploration-driven approaches such as [206] incentivize agents with intrinsic rewards. [173] applied reward shaping approaches to multi-objective problems. However, it is not always apparent how general concepts describing intended behaviors can be translated into meaningful numeric rewards [146]. [280] proposed learning-based reward shaping to enable the agent to generate inner rewards to guide itself. However, their learning approach has challenges with a limited number of training steps and uncertain environments. [195] used genetic programming to find novel reward functions that improve learning system performance. [119] considered how to adaptively utilize a given parameterized shaping reward function, utilizing the beneficial part of the given shaping reward function as much as possible and meanwhile ignoring the unbeneficial shaping rewards. [57] described the use of context-aware potential functions in which different shaping functions should be used for different contexts to address different subgoals. However, the agent must autonomously learn the mapping from context to shaping function. [253] introduced a dynamic reward shaping approach in which a human provides verbal feedback translated into additional rewards. How-ever, humans sometimes give a wrong reward, give feedback out of time, or the agent's speech recognition system misinterprets the command.

*Example in ITSC.* The agent receives a basic reward based on minimizing the total waiting time of vehicles at the intersection. Additional rewards can be shaped by considering factors such as queue length, average vehicle speed, and priority for emergency vehicles. These additional rewards incentivize the agent to reduce congestion, maintain shorter queues, optimize traffic flow, and provide efficient passage for emergency vehicles.

**Limitations.** Adaptive reward signals often require the design and engineering of reward functions. Creating appropriate reward functions can be challenging and time-consuming, especially in complex and dynamic environments [179, 245, 138]. Sudden or significant changes in the environment can make the learned reward functions obsolete or require rapid adjustment, which can be challenging and may introduce instability into the learning process [127, 47]. For instance in the traffic domain, in the case of a sudden accident or road closure, the predefined reward function (i.e., reducing waiting times) may not adequately capture the impact of such events. Since these events are unpredictable, the agent may not have been trained to handle them explicitly. Consequently, the reward function may not provide the agent with the necessary guidance to respond appropriately. This limitation arises from the inherent difficulty of anticipating and accounting for all possible unforeseen events in the design of the reward function, making it challenging for the agent to effectively adapt its behavior in real time to address such situations.

## 2.3 Action Selection Approaches

In this section discussing approaches for selecting actions in dynamic and partially observable environments, I commence with Reinforcement Learning (RL), in which agents learn from trial and error in dynamic environments, adapt their strategies, and optimize action sequences over time. Then I focus on Self-adaptive Systems (SASs), which employ adaptive algorithms to dynamically adjust their decision-making strategies based on real-time feedback and handle uncertainties. Also, I explore Online Planning approaches that continuously update action selection based on real-time feedback, followed by a discussion of exploration algorithms that aid in reducing agents' uncertainty. Finally, I delve into semi-stochastic action selection, a method that combines deterministic and probabilistic elements to guide an agent's decision-making process.

## 2.3.1 Reinforcement Learning

RL agents can learn to extract relevant information from their observations, effectively dealing with partial observability and making informed decisions based on the available information. Also, RL agents continuously update their policy based on the observed rewards and environmental feedback. This adaptability enables RL agents to adjust their action selection strategy to handle changing conditions. Moreover, RL agents learn to optimize action sequences over time, taking into account delayed rewards and potential uncertainties to make optimal decisions in a sequence of actions. Value-based RL algorithms, such as Q-learning, SARSA, Deep Q-Network (DQN), or policy gradient methods such as Trust Region Policy Optimization (TRPO), Proximal Policy Optimization (PPO), Advantage Actor-Critic (A2C), and Actor-Critic with Kronecker-Factored Trust-Region (ACKTR) or a hybrid algorithm that combines elements of both policy gradients and value-based methods such as Deep Deterministic Policy Gradient (DDPG), can be adapted to work in partially observable and dynamic environments [7].

RL techniques used in partially observable environments estimate unobservable state variables. In [148], the authors proposed a hierarchical deep RL approach for learning, where the learning agents need an internal state to memorize important events in partially observable environments. In [196, 204], the authors built a memory system to learn to store arbitrary information about the environment over numerous steps to generalize such information for the new environments. RNNs can compensate for missing information by developing their internal dynamics and memory [62, 114, 180]. Therefore, in many of the current works, the RL agents require some form of memory to learn optimal behaviors over numerous steps, challenging when on-the-fly decision-making is required [63]. To improve RL's performance, agents require mechanisms to augment their observation on the fly. As an example, to tackle the partial detection of vehicles problem in ITSC systems, various RL algorithms are used such as Q-learning, PPO, A2C, ACKTR [295], DQN [294], and Connectionist RL [89]. RL algorithms have high deviation under partial detection of vehicles, especially in off-policy learning algorithms such as Q-learning and DQN, in which rapid Q-value updates in a noisy environment sometimes lead to a "bad update". However, policy gradient methods such as the A2C algorithm try to improve their policy based on their approximation of policy gradient and have better performance at each update [295]. Moreover, the suggested methods for assessing traffic conditions through the utilization of traffic volume and the average speed of vehicles demonstrate limitations when confronted with unanticipated traffic scenarios, and they struggle to handle associated uncertainties [128]. In a different study [88], the authors proposed RL signal controllers designed for dynamic and unpredictable traffic micro-simulations. Nonetheless, this approach relies on fixed traffic signal phase duration, and any alteration to the action space or state space necessitates the retraining of the model.

**Limitations.** RL algorithms often struggle to efficiently explore and learn in high-dimensional state spaces. The lack of complete information about the environment can lead to suboptimal exploration strategies or prolonged exploration periods [210, 160]. Moreover, RL algorithms may struggle to adapt quickly to rapidly changing environments. In dynamic environments where conditions shift frequently, RL agents may require substantial exploration and learning time to update their policies and adjust to the new dynamics [3]. For example, consider a highly complex intersection with multiple lanes, diverse traffic patterns, and a large number of possible signal timings. The state space in such a scenario is high-dimensional, including variables like the number of vehicles in each lane, waiting times, traffic flow rates, and historical data. RL algorithms can struggle to efficiently explore and learn in such high-dimensional state spaces. Also, when a major accident occurs nearby, leading to a road closure and redirecting a significant portion of traffic through alternative routes, this rapid change in traffic conditions poses a challenge for the RL agent. Due to the accident, the agent's learned policy, based on historical data and observations, may become outdated and ineffective in handling the increased traffic volume and altered traffic patterns caused by the road closure.

### 2.3.2 Optimization Techniques

Optimization techniques involve algorithms designed to either minimize or maximize an objective function, subject to certain constraints or conditions. The optimization techniques employed for agent decision-making in various contexts, for example, [191] leveraged collective learning to achieve two optimization objectives: balancing IoT workload globally and minimizing service execution costs locally. Through the local generation of candidate assignments and cooperative selection, agents aim to maximize edge utilization while minimizing service execution costs. [69] introduced an appliance-level flexible scheduling framework based on consumers' self-determined flexibility and comfort requirements. The paper proposed a decentralized network of autonomous scheduling agents to address the challenge of coordinating schedules to reduce demand peaks effectively. This approach involves multi-objective optimization and leverages techniques such as the I-EPOS system (Iterative Economic Planning and Optimized Selections) for decentralized, privacy-preserving optimization. [190] presented a methodology for energy-centric flexible JSS, aimed at minimizing energy consumption in manufacturing processes. The methodology combines a plan generation algorithm with a collective learning tool, I-EPOS, to achieve near-optimal solutions for the energy-centric flexible JSS problem. The decentralized nature of collective learning allows autonomous agents to adaptively coordinate their actions, making it suitable for dynamic and partially observable environments characteristic of manufacturing settings [216]. [117] introduced a Combinatorial Optimization Heuristic for Distributed Agents (COHDA) designed to solve Multiple-Choice Combinatorial Optimization Problems (MC-COP) in Multi-Agent Systems (MAS). MC-COP involves selecting exactly one element from each of multiple sets or classes of elements to form a solution, aiming to minimize or maximize a given objective function. The approach operates entirely in a decentralized manner, leveraging individual knowledge and cooperative behavior among agents. Agents form their knowledge bases through local perceptions and beliefs about global knowledge, even if incomplete or outdated. Traditional centralized optimization approaches often struggle in decentralized systems due to privacy concerns, bandwidth limitations, or interdependencies among distributed search spaces. The heuristic leverages self-organizing strategies, where individuals represent partial solutions within local search spaces. The goal is to find local solutions that, when combined, yield the optimal global solution. Operating without global knowledge, COHDA enables agents to adapt to changing conditions by continuously updating local solutions and collaborating with others. [217] presented a novel approach called Decentralized Collective Learning (DCL) as a decentralized process of collective decision-making among self-organizing agents structured in a hierarchical tree topology. By leveraging remote interactions and aggregate-level information exchange, DCL enables efficient and privacy-preserving optimization of complex combinatorial problems, such as resource allocation in energy management and bike sharing initiatives.

*Example in ITSC.* Traffic signals can use distributed constraint optimization techniques to collaboratively solve the optimization problem of minimizing congestion and reducing travel times across the entire traffic network. This objective function typically involves maximizing the throughput of vehicles through intersections while minimizing delays and queues.

**Limitations:** While optimization techniques offer powerful tools for solving a wide range of problems, they also come with limitations and challenges [261, 75, 23, 239, 149]. One common challenge is the presence of local optima, where the algorithm converges to a suboptimal solution instead of the global optimum. Additionally, the choice of optimization algorithm and its hyperparameters can significantly impact the convergence speed and the quality of the solution obtained. Moreover, optimization techniques may require substantial computational resources, especially when dealing with large-scale problems or high-dimensional search spaces [225]. In some cases, the objective function may be noisy, posing additional challenges for optimization algorithms [224]. This refers to a situation where the function values (outputs) contain random or unpredictable fluctuations, often due to external factors or inherent variability in the system being modeled. Furthermore, the effectiveness of optimization techniques can be influenced by the problem's inherent complexity and structure. Optimization techniques may struggle when the underlying problem exhibits uncertainty or ambiguity [131]. For example, in the ITSC context, optimization algorithms may rely on historical traffic data and predefined models to determine optimal signal timings. However, these algorithms may struggle to adapt quickly to sudden changes in traffic patterns, such as accidents, road closures, or unexpected events. Furthermore, optimization algorithms may not adequately address the trade-offs between conflicting objectives, such as minimizing travel times for vehicles while ensuring pedestrian safety and prioritizing public transportation routes. Balancing these objectives requires sophisticated decision-making mechanisms that take into account various factors that may be difficult to incorporate into traditional optimization models.

Ontologies can facilitate more efficient resource utilization by guiding the algorithm to focus on relevant areas of the search space, particularly in large-scale or high-dimensional problems. Moreover, by incorporating domain expertise into the objective function definition, ontologies can help mitigate the effects of noisy data, improving algorithm robustness. Furthermore, ontologies can support adaptive decision-making in dynamic environments by providing real-time updates on changing objective functions, allowing optimization algorithms to react promptly to unforeseen events or uncertainties. Additionally, ontologies can facilitate the balancing of conflicting objectives by providing a structured representation of the relationships and dependencies between various domain concepts and criteria. By explicitly encoding these concepts and their interrelationships, the ontology enables optimization techniques to understand the trade-offs involved in different decisions.

### 2.3.3 Self-adaptive Systems

Self-Adaptive Systems (SASs) employ adaptive algorithms that analyze the current state, goals, and available information to dynamically adjust their decision-making strategies. These systems can automatically tune their parameters, adjust their strategies, or learn from past experiences to enhance future actions. Moreover, SASs can handle uncertainties by incorporating techniques such as probabilistic reasoning, risk assessment, or adaptive planning. This involves using probabilistic reasoning to assess the likelihood of different outcomes or events occurring in uncertain situations. Additionally, risk assessment techniques are employed to evaluate the potential consequences and associated risks of different adaptation strategies, helping SASs make informed decisions that balance the trade-offs between different actions in uncertain environments. To develop a system that is adaptive to an uncertain environment, various engineering approaches, such as eliciting adaptive requirements from the environment [275], analyzing SAS design while considering an uncertain environment [61], testing SAS implementation with environmental inputs [220], and updating environmental knowledge for optimal run-time decision-making [249], have been proposed. In [260], the authors specified loose self-management policies capable of flexible and nondeterministic choice of behaviors through the new Autonomic System Specification Language (ASSL). In [243], the authors proposed a goal model augmented with uncertainty annotations to guide the synthesis of adaptation policies at run-time. However, it is not specified how rare events such as catastrophic scenarios could be taken into account. Probability theory [66] is widely used to deal with different sources of uncertainty by running averages to mitigate uncertainty due to noise in monitoring, explicit annotation of adaptation strategies with probabilities, and estimation of the future environment and system behavior [31]. For instance, Rainbow framework [42] calculates the average of observations to deal with the uncertainty of the environment. Possibility theory has been mainly used in approaches that deal with the uncertainty of the objectives [21]. In [274], the authors designed a formal specification language RELAX to enable analysts to identify requirements that may be relaxed at run-time when the environment changes. The POISED approach was built on the possibility theory to assess the positive and negative consequences of uncertainty. It makes adaptation decisions that result in the best range of potential behavior, however, it can only deal with the defined uncertain situations [67]. Fuzzy control can tolerate the inaccuracy of sensors' data, but it requires the system engineers to extract fuzzy rules from historical data and cannot adapt to significant changes [109].

AI techniques, especially ML, have been adopted to deal with the difficulty in predicting environmental conditions at run-time [238]. ML can support the MAPE-K model (i.e., a well-known reference model for SASs) [132] to build run-time models and adaptation strategies from complex and high-dimensional data obtained from uncertain environments [273]. For example, in [228], the authors used classifiers over the runtime data to detect real-time constraints. Then, these constraints are used to refine the design model of a SAS. In [287], the authors developed an ML-based approach, ACon, that uses MAPE-K feedback for adapting the SAS's contextual requirements affected by uncertainty at run-time. A self-learning fuzzy neural network [109] was proposed to handle dynamic uncertainties from the sensing phase and deal with considerable changes. FUSION [66] adjusted strategies online through a learning model of the environment. However, ML-based approaches need to train the model before the system runs and cannot deal with the unknowns. Learning methods can handle this challenge by identifying the active context and modifying the adaptation space at run-time [137, 96]. Thus, context changes not anticipated at design time are addressed by learning new adaptation rules dynamically or by modifying and improving existing rules [238]. However, the generation of all possible situations, exclusively at run-time, poses a risk to the system's performance, reliability, and real-time constraints. In [177], the authors used the feature model for the system's adaptation space and thereby leveraged its semantics to guide RL's exploration, but their approach only supports discrete adaptation actions. Mao's team [174] proposed an RL method that can solve a part of unknown environment changes through existing strategies, however, they ignored that the existing strategies cannot solve all the novel situations. [278] proposed a planning method to handle uncertainty from the environment by learning knowledge of the relationship between system states and actions. This method can generate new strategies to deal with unknown situations, however, with the increase in system scale, more knowledge is needed to learn.

*Example in ITSC.* SAS continuously monitors real-time data (e.g., traffic volumes, vehicle queues, waiting times) from various sensors. Based on the collected data, the system builds a model of the traffic environment. The model captures the dynamics of traffic patterns, congestion levels, and the relationships between different intersections. Then SAS utilizes adaptive algorithms to dynamically adjust the signal timings at each intersection. These algorithms take into account the current state of the traffic environment, the model predictions, and the desired objectives, such as minimizing delays or maximizing throughput. Also, the system continuously evaluates the performance of the current signal timings and uses feedback mechanisms to assess their effectiveness.

**Limitations.** It is not specified how rare or unknown events should be considered when SAS has not experienced them before [287, 32, 116]. In addition, SAS needs to find a balance between exploration and real-time decision-making. While exploration is crucial for discovering better strategies and adapting to changing conditions, excessive exploration (i.e., the significant number of possible adaptation actions) can compromise the system's ability to make timely decisions [228, 177]. For example in the traffic domain, a sudden major event like a large-scale protest or an unusual traffic accident might significantly impact traffic patterns, and the SAS may struggle to adapt effectively in such scenarios, as it lacks prior experience and appropriate models to guide its decision-making. Moreover, in a bustling urban environment, traffic condi-tions can change rapidly, requiring quick and efficient decision-making. SAS needs to process and analyze incoming data, update its models, and generate optimized signal timings within tight time constraints. Failing to make timely decisions may lead to traffic congestion, increased delays, and compromised intersection efficiency.

### 2.3.4 Online Planning

Instead of learning a fixed policy, online planning approaches continuously update the agent's action selection based on real-time feedback. These approaches use techniques like Monte Carlo Tree Search (MCTS) or rollout algorithms to simulate possible future trajectories and select actions based on the accumulated rewards. As new observations become available, the agent can update its beliefs about the environment and use the planning process to generate a sequence of actions that maximizes its expected utility [250].

*Example in ITSC.* The agent continuously receives real-time data from vehicle detectors and cameras, providing information on traffic volumes, vehicle queues, and waiting times at the intersection. Using online planning techniques, the agent updates its model of the traffic environment based on the received observations. It estimates the current traffic state, including the number of vehicles approaching from different directions and the severity of congestion. Then instead of optimizing signal timings for a single cycle, the agent extends the planning horizon to cover multiple cycles or a predefined time duration. It takes into account historical traffic data, real-time observations, and predictions of future traffic flows. By incorporating these factors, the agent generates an action sequence that optimizes the traffic flow and minimizes delays over the extended planning horizon.

**Limitations.** Planning algorithms may struggle to generate plans within the limited time constraints of real-time decision-making, especially when the state space is large or the planning horizon is extended [50, 262]. In addition, online planning heavily relies on its models to estimate the current state and predict future outcomes. If the models are inaccurate or fail to capture the true dynamics of the environment (i.e., due to partial observability), the generated plans may be ineffective [163, 170]. For instance, traffic signal control systems operate in real time, and there are strict time constraints for decision-making. As the planning horizon is extended, the computational time required to generate optimal signal timings increases. Additionally, traffic patterns and dynamics are inherently uncertain and can vary significantly over time. The extended planning horizon may exacerbate the impact of uncertainties and make accurate predictions more difficult. Also, if the planning horizon is too long, the system may struggle to adapt and respond promptly to sudden unforeseen changes, potentially resulting in delays and congestion.

### 2.3.5 Exploration Algorithms

Exploration algorithms [22, 14] play a crucial role in reducing an agent's uncertainty in dynamic and partially observable environments. These algorithms aim to maximize the agent's knowledge about the environment by actively exploring and gathering information. Exploration algorithms are categorized into two groups: (A) Count-based exploration methods are a class of exploration algorithms that use the concept of counting (i.e., visitation frequency) to encourage the exploration of less-visited or uncertain actions. (B) Intrinsic motivation is an exploration algorithm that encourages agents to explore and learn by providing internal rewards or curiosity-driven incentives (i.e., seeking novel experiences, and learning new skills) [22]. Some of the commonly used exploration algorithms are (1) Epsilon-Greedy selects the action with the highest estimated value most of the time (exploitation) but occasionally selects a random action with a small probability epsilon (exploration) [53], (2) Upper Confidence Bound (UCB) maintains a measure of uncertainty or confidence in the estimated values and selects actions that have both high estimated value and high uncertainty [85], (3) Thompson Sampling maintains a distribution over the true values of actions and samples from this distribution to select actions [35], (4) Monte Carlo Tree Search (MCTS) prioritizes actions that have not been explored extensively [36], and (5) Optimistic Initialization starts with optimistic estimates of the action values, so the agent is encouraged to explore actions that initially seem promising but are uncertain [166].

*Example in ITSC.* Consider an intersection where the agent, responsible for optimizing the traffic signal timings, aims to minimize congestion and reduce waiting times. Initially, the agent has limited knowledge about the optimal signal timings for different traffic patterns. During the exploration phase, the agent randomly selects signal timings, incorporating a level of randomness into its decision-making process. For instance, it may choose to extend the green signal duration for a particular direction or introduce shorter phases to allow more frequent signal changes. Through this exploration, the agent gathers valuable experience by observing the effects of different signal timings on congestion levels, waiting times, and traffic flow. Additionally, the agent may encounter unusual traffic conditions due to a special event or construction work, where its learned policies may not be optimal. The ongoing exploration enables the agent to explore alternative actions and adjust its signal timings accordingly.

**Limitations:** Growing the number of possible actions exponentially makes it difficult for exploration algorithms to thoroughly explore the space and find optimal solutions [54, 71]. Also, exploration algorithms can get stuck in local optima because they prioritize the exploitation of known good actions over exploring uncertain but potentially better actions [161, 297]. Moreover, exploration can be time-consuming and resource-intensive, particularly in dynamic environments where the agent must constantly adapt to changes [290]. For example in a traffic network with multiple intersections and various signal timings for each intersection, the action space becomes vast. The exponential growth makes it difficult for exploration algorithms to thoroughly explore the entire action space within a reasonable timeframe. Moreover, traffic conditions are subject to constant changes due to factors like varying traffic volumes, accidents, or road closures. To adapt to such changes, the agent needs to explore and learn new optimal policies.

### 2.3.6 Semi-stochastic Action Selection

In semi-stochastic action selection, the agent makes use of both deterministic and stochastic components when selecting actions. In the deterministic component, the agent calculates an action selection policy based on its current knowledge through techniques like value iteration, Q-learning, or policy gradient methods. The stochastic component can take different forms, such as adding a small amount of noise to the deterministic action or using a probability distribution to select actions. The balance between the deterministic and stochastic components can be adjusted based on the agent's learning progress or the level of uncertainty in the environment [242].

*Example in ITSC.* Consider an agent responsible for optimizing traffic signal timings at an intersection. The agent can choose between fixed, predetermined signal timings or adapt the timings based on real-time traffic conditions. During each decision-making step, the agent evaluates the current traffic conditions, such as the number of vehicles waiting, the traffic flow from different directions, and the presence of pedestrians or cyclists. Based on this evaluation, the agent generates a set of potential signal timings that are deemed promising or favorable for the current situation. So, instead of selecting a completely random action, the agent uses a semi-stochastic approach by applying a probabilistic mechanism to choose among the promising signal timings that have been deemed effective in similar situations.

**Limitations:** Finding the optimal balance between exploration and exploitation can be challenging and may require careful tuning of parameters [266, 44]. Also, in partially observable environments, the agent has limited or noisy observations of the underly-ing state. This limited information can affect the effectiveness of semi-stochastic action selection. The agent may struggle to estimate the uncertainty accurately and make suboptimal decisions based on incomplete information [242]. Moreover, dynamic environments introduce changes over time, and semi-stochastic action selection may have difficulty adapting to such changes promptly [242]. For instance in the traffic domain, during peak hours, traffic congestion is high, and the signal timings need to be carefully adjusted to minimize delays. The agent may have learned effective policies during less congested periods, but the same strategies may not be optimal during rush hour. Due to the bias towards exploiting known actions, the agent may struggle to explore alternative options effectively and fail to adapt its signal timings adequately to manage the increased traffic volume. Unexpected events like accidents or road closures can significantly impact traffic conditions. However, the semi-stochastic action selection may not promptly respond to such events, as it relies on previous knowledge and biased action selection.

## 2.4 Ontology-based Approaches

An ontology is a formal framework for representing knowledge within a specific domain, comprising concepts (classes) that define categories of objects or ideas, properties (attributes) that describe characteristics of these concepts, and relationships that specify connections between concepts. Additionally, ontologies often incorporate logical rules and semantic constraints to enforce consistency and define how concepts, properties, and relationships interact within the knowledge representation. Ontology provides user-contributed, augmented intelligence, and machine-understandable semantics of data [78], providing a foundation for reasoning about objects, behavior, and knowledge as a formal methodology [164]. Ontologies can be classified based on abstraction level and field of application, upper ontology formalizing (general concepts that can be applied across various domains), domain ontology including concepts relevant to a particular domain, interface ontology to represent the concepts relevant to the intersection of two disciplines and process ontology describing inputs, outputs, constraints, and sequence of information in a particular process [212]. In the literature, various ontology-based templates (i.e., general and domain-specific ontologies) have been defined to represent, formalize, manage, and organize the knowledge about the agent, its behavior within the environment, and its interaction with other entities [241, 197, 164, 45]. By providing unambiguous and machine-accessible interpretations of terms, ontologies act as a middleware between the physical layer and the inform-ation layer enabling interconnectivity, interoperability, and coordination of activities between intelligent heterogeneous systems in an environment.

Some AI techniques enhance agents' learning process by providing semantic models and augmented feedback loops to optimize their overall accuracy [33, 185]. As part of ML tasks, ontologies are used in several ways, including enrichment of features derived from the ontology, calculating similarities or distances between instances based on the ontology's structure, and determining probabilistic dependencies between instances and features based on the entities' dependencies in the ontology [27]. Domain-specific knowledge encoded with ontology can be used to constrain search and find optimal or near-optimal solutions faster or find a solution that is generalized better [144]. [185] proposed a summarization method that reinforces learning functions by using semantic relationships between ontology concepts. The authors of [283] created an ontology base on classification results and queried this ontology instead of querying the decision tree. To improve the accuracy of a Support Vector Machine (SVM) algorithm, [33] used ontology augmentation as a feedback loop. In [133], the authors modeled the TSC context using a fuzzy ontology for reusing knowledge and firing suitable fuzzy rules using a fuzzy inference engine. The approach proposed in [248], enriches the semantics of the raw sensor data and infers contextual description through context-aware attributes.

In SASs, ontology is mainly used as a modeling method to specify the requirements of a system-to-be. For example, in [10], the authors used a goal ontology containing semantic knowledge about the system's goals and the relationship between goals and sub-goals. In [221], the authors used a shared ontology, which includes concepts about the domain and conditions relating to them to support a shared understanding of concepts between the user and the system. The shared ontology enables the system to update the specification when a refinement or addition of new concepts occurs and also understand the service descriptions for solution provisioning. To manage the heterogeneity of real-world systems, in [214], the authors used ontologies, leveraging their ability to provide a shared and unified representation of a complex and heterogeneous domain of interest. In [183], the authors modeled the environment using an ontology that defines the concepts and available actions to the system for test generation. In [222], the authors proposed a framework to enable Continuous Adaptive Requirements Engineering (CARE) that leverages the requirements-aware systems and exploits a reasoning system. A fuzzy theory and semantic distance technology [43] were proposed to handle inaccurate monitoring data. Self-adaptive activity recognition systems [112] aim to recognize complex Activities of Daily Living (ADL) from a series of observations of the individual's actions and the environmental conditions. In [46], the authors proposed a new approach that relies on an ontology to derive a first set of semantic correlation values between activities and sensor events. In [112], the authors introduced an incremental learning method to integrate new data with new features into current activity recognition models. In [236], the authors proposed a new domain adaptation technique that uses a general ontology to share and transfer activity models across individuals, even though the sensor deployments and operating environments are different. In the literature, ontological knowledge is used to enrich the semantics of the raw sensor data and provide reasoning capabilities to improve learning algorithms, however, the concerns related to environments' partial observability and dynamism are not addressed yet.

## 2.5  Summary and Analysis

The existing literature provides various approaches for addressing challenges related to partial observability and unforeseen events in an agent's decision-making process. In the exploration of **observation modeling** approaches, I begin with Partially Observable Markov Decision Processes (POMDPs), a framework for sequential decision-making in partially observable environments where agents make decisions over time with limited information about the environment's state. POMDPs handle partial observation by maintaining a belief state, incorporating observation models, and selecting actions based on maximizing expected rewards given the agent's current belief about the environment's state. Hidden Markov Models (HMMs) are probabilistic models suitable for modeling partially observable environments by estimating hidden states from observable symbols. Monte Carlo Methods provide effective state estimation in such environments by propagating particles through the state space and updating their weights based on observations. Neural Network-based Models, particularly Recurrent Neural Networks (RNNs), capture temporal dependencies in sequential data, enabling agents to incorporate historical context and fill in gaps in their knowledge over time. Scene Graphs offer graph-based representations to capture spatial and semantic relationships, facilitating reasoning and context-aware decisions. Bayesian Learning incorporates prior knowledge and updates beliefs based on observed data. Predictive State Representation (PSR) employs predictive state features to predict future observations. I outline the limitations associated with the observation modeling approaches in Table 2.2.

*Leveraging Ontology for enhanced Observation Modeling in Dynamic and Partially Observable environments.* Ontologies can help in defining a structured representation of the state and observation spaces in partially observable environments. By explicitly spe-cifying the relationships between different entities and their attributes, ontologies can reduce the complexity of state representation. Also, ontology makes it possible to in-corporate domain-specific knowledge and constraints into the model, leading to more informative state representations. Moreover, ontologies can be used to define the structure of the state space, allowing for a more organized and informed exploration of possible states. Ontology-based reasoning can help address the limitation of incomplete information by enabling agents to infer missing details from available data, while also assisting in identifying relevant features for the agent's decision-making. Additionally, reasoning with ontologies can assist in generating informative samples and guiding the sampling process by taking into account the relationships and constraints defined in the ontology.

| Approach | Limitation | Reference |
|---|---|---|
| Partially Observable Markov Decision Processes (POMDPs) | Complexity grows exponentially with state and observation spaces. | [83] |
| | Performance affected by noisy or imperfect observations. | [176, 286] |
| Hidden Markov Models (HMMs) | Assumption of Markov property may not hold in dynamic environments. | [223, 207] |
| | Limited ability to capture complex dependencies or context. | [218] |
| | Heavily reliant on sufficient training data. | [182, 13] |
| | Struggles with rare events not well-represented in training data. | [136] |
| Monte Carlo Methods | Accuracy is impacted by noisy or incomplete observations. | [165] |
| | Computational demands increase with high-dimensional observations. | [189] |
| | Relies on pre-defined models that may not handle unforeseen events. | [203, 257] |
| Neural Network-based Models | Struggle to model long-term dependencies effectively. | [135, 292] |
| | Sensitive to noisy or incomplete observations. | [81, 237] |
| | Limited generalization to unforeseen events. | [156] |
| | Computational demands may restrict real-time feasibility. | [26] |
| Scene Graphs | Limited expressiveness compared to ontologies. | [269, 84, 121, 16] |
| | Struggles with unanticipated changes or new objects/relationships. | [15, 97] |
| Bayesian Learning | Insufficient training data lead to unreliable posterior estimates. | [52] |
| | Challenging to select an appropriate probabilistic model in dynamic environments. | [55] |
| | Computationally intensive tasks may not be feasible in real-time scenarios. | [300] |
| Predictive State Representation (PSR) | Ineffectiveness with incomplete or noisy observations. | [103, 20] |
| | Limited adaptation to changing dynamics in the environment. | [289] |

Table 2.2: Limitations of observation modeling approaches.

For the **goal selection**, several approaches have been proposed for guiding intelligent agents. Goal Formation, Goal Generation, and Goal Reasoning techniques are pivotal in transforming high-level objectives into actionable goals, generating relevant goals based on various contextual factors, and employing a case-based reasoning system to dynamically select goals. Practical Reasoning aids agents in incorporating their preferences and social understanding, allowing them to prioritize goals aligned with individual motivations, while Heuristic-based approaches provide rule-of-thumb strategies for efficient goal prioritization. Learning techniques empower agents to adapt and evolve by acquiring knowledge from experiences. Finally, Adaptive Reward Signals offer continuous evaluation and dynamic shaping of rewards, enhancing goal pursuit, but necessitating careful design in dynamic environments. I enumerate the limitations associated with the goal selection approaches in table 2.3.

*Leveraging Ontology and Reasoning for enhanced Goal Selection in Dynamic and Partially Observable environments.* Ontologies can provide a structured and formal representation of the domain knowledge. By defining concepts, relationships, and rules within the ontology, agents can have a clearer understanding of the environment and potential goals. Reasoning engines can help agents make inferences based on the information available in the ontology. For example, if the ontology contains information about historical traffic patterns and the current traffic conditions, reasoning can be used to detect inconsistencies or deviations from the norm, triggering the need for new goal formation. When unforeseen events occur, the ontology can be updated to include information about the event, allowing agents to reason about the new situation and formulate relevant goals. Moreover, ontologies can include information about contextual factors, social norms, preferences of different entities (e.g., drivers, pedestrians), and the rules governing traffic management. Reasoning mechanisms can help agents adapt predefined rules to new and unforeseen scenarios. Instead of relying solely on static rules, agents can apply reasoning to dynamically generate or modify rules based on the evolving context. Furthermore, reasoning can be used to dynamically adjust re-

| Approach | Limitation | Reference |
|---|---|---|
| Goal Formation, Goal Generation, and Goal Reasoning Techniques | In partially observable environments, agents may struggle to accurately form goals due to incomplete information. | [235, 282] |
| | Unforeseen events can disrupt the context and render pre-existing goals irrelevant, making it difficult for agents to adapt and generate new relevant goals. | [49, 126, 276] |
| Practical Reasoning | Reliance on predefined reasoning rules may limit the ability to generate appropriate goals for unforeseen events. | [139, 140] |
| Heuristic-based Approaches | Heuristics may not always yield optimal solutions and may struggle to adapt to unforeseen events or significant changes in the environment. | [186, 251, 86] |
| Learning Techniques | Lack of relevant training data in the presence of unforeseen events can limit the system's ability to accurately select goals in real-time. | [77, 68, 106] |
| Adaptive Reward Signals | Designing appropriate reward functions can be challenging and time-consuming, especially in complex and dynamic environments. | [179, 245, 138] |
| | Sudden or significant changes in the environment may make learned reward functions obsolete or require rapid adjustment. | [127, 47] |

Table 2.3: Limitations of goal selection approaches.

ward functions based on the evolving environment. For example, if an accident occurs, reasoning can modify the reward function to prioritize safety-related goals.

There are various approaches for **action selection** in dynamic and partially observable environments. Reinforcement Learning (RL) enables agents to adapt and optimize actions through trial and error, making informed decisions based on available information and observed rewards. Optimization techniques encompass algorithms aimed at optimizing objective functions, subject to constraints, which are applied in decentralized decision-making and self-organizing strategies. Self-adaptive Systems (SASs) employ adaptive algorithms, dynamically adjusting strategies based on real-time feedback to handle uncertainties effectively. Online Planning continuously updates action selection using techniques like Monte Carlo Tree Search (MCTS) to maximize expected utility as new observations arrive. Exploration Algorithms actively reduce an agent's uncertainty by encouraging the exploration of less-visited or uncertain actions. Semi-stochastic Action Selection combines deterministic and stochastic components, allow-ing agents to balance knowledge-based policies with probabilistic mechanisms for effective action choices in dynamic and partially observable environments. The limitations of action selection approaches are presented in table 2.4.

| Approach | Limitation | Reference |
|---|---|---|
| Reinforcement Learning (RL) | Struggle to efficiently explore and learn in high-dimensional state spaces | [210, 160] |
| | Slow adaptation to rapidly changing environments | [3] |
| Optimization Techniques | Presence of local optima | [261, 75, 23, 239, 149] |
| | Requirement for substantial computational resources | [225] |
| | Noisy objective functions | [224] |
| | Struggle with uncertain or ambiguous problems | [131] |
| Self-adaptive Systems (SASs) | Uncertainty in handling rare/unknown events | [287, 32, 116] |
| | Balancing exploration and real-time decision-making | [228, 177] |
| Online Planning | Difficulty generating plans within limited real-time constraints | [50, 262] |
| | Reliance on accurate models for effective plan generation | [163, 170] |
| Exploration Algorithms | Difficulty exploring large action spaces | [54, 71] |
| | Getting stuck in local optima | [161, 297] |
| Semi-stochastic Action Selection | Challenging to find the optimal balance between exploration and exploitation | [266, 44] |
| | Limited information and suboptimal decisions in partially observable environments | [242] |
| | Difficulty adapting to changes in dynamic environments | [242] |

Table 2.4: Limitations of action selection approaches.

*Leveraging Ontology for enhanced Action Selection in Dynamic and Partially Observable environments.* Ontology reduces the dimensionality of state spaces for RL agents by providing hierarchical abstractions, allowing them to group similar states and operate at higher levels of abstraction, thus simplifying the exploration and learning process. By encoding relevant domain knowledge, ontology can highlight the most critical features and relationships, making it easier for RL agents to direct their attention to key elements for the exploration. Additionally, ontology can provide agents with know-ledge about rare or unknown events. By encoding such knowledge, agents can reason about how to react to these scenarios, ensuring adaptability in uncertain environments. For example, ontology can include rules for handling unforeseen traffic disruptions or natural disasters. Moreover, ontology-driven reasoning can help RL agents infer the unobservable aspects of the environment. For instance, if an RL agent lacks direct information about traffic conditions in a certain area, the ontology can provide rules and relationships that allow the agent to make informed estimates based on nearby traffic data.

**How OntoDeM Addresses the Limitation.**

*Limitations of Observation Modeling Approaches:*

1. Complexity grows exponentially with state and observation spaces: OntoDeM mitigates this by leveraging ontology to reduce the dimensionality of state spaces. Hierarchical abstractions provided by ontology allow for grouping similar states, operating at higher levels of abstraction, thus simplifying exploration and learning.

2. Performance affected by noisy or imperfect observations: OntoDeM addresses this through observation augmentation. By enriching the agent's observations with additional context from the ontology, it provides a more reliable basis for action selection despite noisy or imperfect observations.

3. Assumption of the Markov property may not hold in dynamic environments: OntoDeM does not address this challenge directly.

4. Limited ability to capture complex dependencies or context: Through hierarchical abstraction, ontologies break down complex dependencies into manageable components, while semantic relationships capture detailed dependencies. Also, logical reasoning allows for the inference of implicit dependencies, and domain-specific modeling tailors the ontology to capture intricacies unique to the domain, collectively enabling ontologies to comprehensively represent complex dependencies and context in observation modeling approaches.

5. Computational demands increase with high-dimensional observations: OntoDeM reduces computational demands by utilizing the concept similarity method to filter out irrelevant observation information thereby decreasing the size of agent's observation.

6. Struggle to model long-term dependencies effectively: OntoDeM does not address this limitation directly.

7. Limited generalization to unforeseen events: OntoDeM's ontology-driven reasoning enables agents to reason about unforeseen events based on encoded domain knowledge, enhancing generalization capabilities.

8. Limited expressiveness compared to ontologies: OntoDeM's use of ontology provides a richer and more expressive knowledge representation compared to scene graphs.

9. Struggles with unanticipated changes or new objects/relationships: When encountering new objects or relationships, OntoDeM can dynamically expand agent's observation to incorporate these additions. This enables agent to maintain relevance and adaptability in dynamic environments by continuously expanding its understanding of the domain. Additionally, OntoDeM's ontology-driven reasoning allows it to infer potential relationships between new objects and existing concepts, facilitating informed decision-making even in the presence of previously unseen elements.

10. Insufficient training data lead to unreliable posterior estimates: OntoDeM does not address this limitation directly.

11. Challenging to select an appropriate probabilistic model in dynamic environments: OntoDeM does not address this limitation directly.

12. Computationally intensive tasks may not be feasible in real-time scenarios: By leveraging the structured knowledge encoded in the ontology, OntoDeM can quickly analyze and interpret data, prioritize actions, and adapt its strategies accordingly. Thus OntoDeM enhances agent's responsiveness in dynamic environments where computational resources are limited and enables making decisions within the constraints of real-time scenarios.

*Limitations of Goal Selection Approaches:*

1. In partially observable environments, agents may struggle to accurately form goals due to incomplete information: OntoDeM mitigates this limitation by improving the agent's perception and decision-making in partially observable environments through observation augmentation, observation expansion, observation abstraction, and ontology-driven reasoning. By enriching the agent's observations with additional context from the ontology, OntoDeM helps the agent form more informed goals based on the available information.

2. Reliance on predefined reasoning rules may limit the ability to generate appropriate goals for unforeseen events: OntoDeM incorporates dynamic adaptation mechanisms that allow the agent to generate new goals and adjust its decision-making strategies based on real-time observations and contextual factors (automatic goal selection/generation and adaptive reward definition methods).

3. Lack of relevant training data in the presence of unforeseen events can limit the system's ability to accurately select goals in real-time: OntoDeM's observation augmentation and expansion methods enrich the agent's observations with additional context from the ontology. This helps the agent generalize its decision-making abilities beyond the training data and make more informed goal selections in real-time, even in the presence of unforeseen events.

4. Designing appropriate reward functions can be challenging and time-consuming, especially in complex and dynamic environments: OntoDeM's ontology-driven reasoning approach can facilitate the identification of relevant properties and relationships in the environment, which may inform the design of more effective reward functions (adaptive reward definition method).

*Limitations of Action Selection Approaches:*

1. Struggle to efficiently explore and learn in high-dimensional state spaces: OntoDeM prioritizes actions based on importance weight and relevance identified through ontology and inference rules. Also, action masking filters impossible actions, reducing the action space and focusing exploration on relevant areas. Therefore, this limitation is mitigated by OntoDeM.

2. Slow adaptation to rapidly changing environments: OntoDeM incorporates contextual factors and dynamically adjusts action selection based on real-time observations, ensuring timely adaptation to changing circumstances. Thus, OntoDeM also mitigates this limitation.

3. Presence of local optima: While OntoDeM does not directly address local optima challenge, its use of ontology-driven reasoning helps in prioritizing actions, potentially guiding the agent away from local optima.

4. Requirement for substantial computational resources: OntoDeM utilizes ontology-based reasoning and logical rules to filter out infeasible actions, thus reducing the computational burden by focusing only on feasible actions. However, the computational resources required for ontology reasoning might still be substantial depending on the complexity of the ontology and the size of the action space.

5. Noisy objective functions: OntoDeM addresses noisy objective functions by employing the observation augmentation method, enriching the agent's observations with additional context from the ontology, thereby providing a more reliable basis for action selection.

6. Struggle with uncertain or ambiguous problems: OntoDeM addresses uncertainty by incorporating contextual factors and dynamically adjusting action selection based on real-time observations, enabling the agent to adapt to changes and handle uncertain or ambiguous situations.

7. Uncertainty in handling rare/unknown events: OntoDeM systematically searches for potentially effective actions through action exploration method, while observation expansion method improves the agent's perception by generating additional observations from the context, addressing the uncertainty in handling rare or unknown events.

8. Balancing exploration and real-time decision-making: OntoDeM employs a hybrid policy that uses ontology-based reasoning, enabling the agent to dynamically balance exploration and exploitation based on the current context. Thus, OntoDeM mitigates this limitation by providing a mechanism for balancing exploration and real-time decision-making.

9. Reliance on accurate models for effective plan generation: OntoDeM reduces reliance on accurate models by incorporating contextual factors into the decision-making process.

# CHAPTER THREE

# ONTOLOGY-BASED ENVIRONMENT MODELING

In this thesis, ontology is used to enable agents to **represent and interpret** their observations by an ontological structure. Ontology can assist in handling partial observability by explicitly inferring the relationships between observable and unobservable variables through incorporating reasoning mechanisms. Moreover, ontology allows for the integration of contextual information including time, location, weather conditions, or user preferences into the environment model. By representing and reasoning about the contextual aspects, ontology enhances the agents' understanding of the environment's dynamics and supports adaptive decision-making and behavior in response to changing contextual conditions.

This chapter begins with illustrating the *ontology development* process to create domain-specific ontology using the NeOn methodology and Protégé editing environment. Then the *sensor data mapping* is covered which aligns low-level sensor data with high-level concepts using SSN ontology and then establishes an ontology-based schema for agent-monitored data. Also, *action ontology modeling* entails the representation of actions using ontology elements. Thus, I define domain-specific ontology, ontology-based schema, and action ontology to establish a structured framework for modeling and conducting reasoning tasks related to domain-specific concepts, observed concepts by agents, and concepts related to actions. Subsequently, to evaluate the significance of concepts observed by the agent in the context of action selection, the section on *measuring concept similarity in different ontologies* outlines the process of quantifying the similarity between concepts found in separate ontologies, specifically the domain-specific and action ontologies. Also, *measuring ontology distance* describes a measure for comparing the agent's ontology-based schemas across distinct time intervals. Following this, *subsumer extraction* explores the process of identifying subsumers, which are concepts encompassing more general elements within the ontology. To enhance reasoning capabilities, the chapter introduces *rule-based inference strategies*: forward chaining and backward chaining. Further enhancing the utility of the ontology, *assessment of observation significance* is explained for assigning weights to concepts based on relationships. Lastly, the chapter discusses *application-specific ontologies*. These ontologies are tailored for distinct domains. Together, these methods form an ontology-based environment modeling, serving as the basis for the agent's reasoning within the proposed OntoDeM model (refer to Figure 3.1 to view the process model I propose for ontology-based environment modeling).

Figure 3.1: A model outlining the process of environment modeling based on ontology.

## 3.1 Ontology Development

Feilmayr and Wöß (2016) define ontology as "A formal, explicit specification of a shared conceptualization that is characterized by high semantic expressiveness required for increased complexity" [73]. The NeOn (Networked Ontologies) development methodology [246] and the ontology editing environment Protégé [187] are used to develop or evolve the required ontologies for this study. NeOn outlines a structured approach to collaboratively and systematically create and evolve ontologies. NeOn encourages reusing existing ontologies to expedite development, enabling ontology engineers to leverage well-established ontologies. This methodology comprises the following phases:

1. *Requirement Analysis*: This initial phase involves gathering ontology requirements from different stakeholders and domain experts. It defines the ontology's purpose, scope, and intended use, ensuring a clear understanding of the development objectives.

2. *Ontology Design*: Based on the requirements, the ontology is designed, to identify crucial concepts, relationships, and properties representing the domain knowledge. The structure is organized into modular components, addressing specific aspects of the domain.

3. *Ontology Development*: In this phase, ontology engineers create the ontology by defining concepts, properties, relationships, and axioms using ontology languages like Web Ontology Language (OWL) or the Resource Description Framework (RDF). Additionally, this phase involves the establishment of semantic constraints using ontology languages like Semantic Web Rule Language (SWRL) and SPARQL Inferencing Notation (SPIN), which govern the valid relationships and interactions among different ontology elements.

4. *Ontology Integration*: Emphasizing the creation of modular ontologies, the NeOn methodology promotes an interconnected ontology network, where different ontology modules are linked to form a coherent whole.

5. *Ontology Evaluation*: The developed ontology undergoes evaluation to ensure it meets the intended requirements and accurately represents the domain knowledge. Evaluation criteria may include accuracy, completeness, coherence, and relevance.

**Protégé**, an open-source ontology development platform, serves as a widely used ontology editing environment. It offers a user-friendly interface for ontology engineers and domain experts to create, edit, and manage ontologies. Protégé supports various ontology languages, including OWL, RDF, SWRL, and SPIN.

I formally define a *domain-specific ontology* as $O_{d_i}$, to capture the concepts, relationships, and constraints within the specific domain $d_i$ (see Equation 3.1). The ontology $O_{d_i}$ consists of a set of concepts $C_{d_i} = \{c_1, c_2, \ldots, c_x\}$ representing domain entities, a set of properties $F_{d_i} = \{f_1, f_2, \ldots, f_x\}$ describing attributes, and a set of relationships $M_{d_i} \subseteq C_{d_i} \times F_{d_i} \times C_{d_i}$ that express associations between concepts with specific properties. In the context of ITSC, a concept could be "TrafficSignalControl", a property could be "hasTrafficSignalPhase" which describes the phase of a specific traffic signal control, and a relationship could associate a specific phase with a traffic signal control. For example: "TrafficSignalControl A -> hasTrafficSignalPhase -> Red". The domain and range of a relationship define the concepts between which the relationship is established. The domain of this relationship is the concept "TrafficSignalControl", indicating that the relationship applies to traffic signal controls. The range of the relationship is the concept "TrafficSignalPhase" representing the different phases of a traffic signal control. These relationships enable inheritance between concepts and automated reasoning. "TrafficSignalControl A" installed at a particular intersection serves as an instance of the broader concept "TrafficSignalControl" within the ITSC ontology (see Figure 3.3).

$$O_{d_i} = (C_{d_i}, F_{d_i}, M_{d_i}, L_{d_i}, W_{d_i}, H_{d_i}) \tag{3.1}$$

$L_{d_i}$ shows *logical rules* defined by ontology engineers in the domain-specific ontology. A logical rule defines relationships, conditions, and constraints between entities or concepts in the ontology. Logical rules are used for automated reasoning and inference. They enable the ontology to deduce new information based on the existing relationships and constraints. A logical rule in the ITSC domain could be: *IF (Traffic signal phase is green) THEN (Vehicles in that direction are allowed to proceed)*. An inference rule, on the other hand, involves reasoning and decision-making based on available information. For example: *IF (Traffic signal phase is green and there are no vehicles waiting in any other direction) THEN (Traffic signal phase in the opposite direction should be red)*. It uses logical rules, such as the relationship between green phases and vehicle movement, to deduce the appropriate action (changing the opposite phase to red).

Ontology engineers may assign time-based *importance weights* $W_{d_i}$ to the relationships, thereby determining their significance across different temporal contexts. For instance, in the traffic context, take into account the concept "Vehicle" (i.e., domain) and its property "hasType", which can be an "Emergency" one (i.e., range). To integrate the temporal dimension, engineers can allocate varying importance weights to the relationships (i.e., relationship and its domain/range combination). For example, a vehicle classified as "Emergency" might receive the "Highest" importance weight during rush hours. This temporal weighting approach accounts for the varying relevance of relationships in different time periods. The relationship importance can be categorized into degrees such as "Lowest", "Low", "Middle" "High", and "Highest" each of which can be quantified with predefined numerical mappings.

$H_{d_i}$ represents *semantic constraints* that must be upheld within the ontology. They define certain limitations or rules that concepts and relationships must adhere to in order to ensure meaningful and valid representations. For example, "No simultaneous green phases at conflicting intersections", means conflicting intersections should not have simultaneous green signal phases.

## 3.2 Sensor Data Mapping

In order to represent the agent's observation, low-level sensor data streams are mapped to high-level concepts. Sensors often produce raw data and unstructured streams, measuring phenomena values. Data collected by sensor resources can be described by the Semantic Sensor Network (SSN) ontology [108], which bridges the gap between low-level data streams coming from sensors in real-time and high-level concepts used by agents to interpret an observation (i.e., high-level semantic representations). As demonstrated in Figure 3.2, the RL agent uses SSN ontology with domain-specific ontology to annotate and present sensor data using the data model proposed in [64]. Sensor data observed by an agent is described by the "ssn:Observation" class. The "ssn:Property" specifies the property of the feature of interest that is described by a domain-specific ontology.

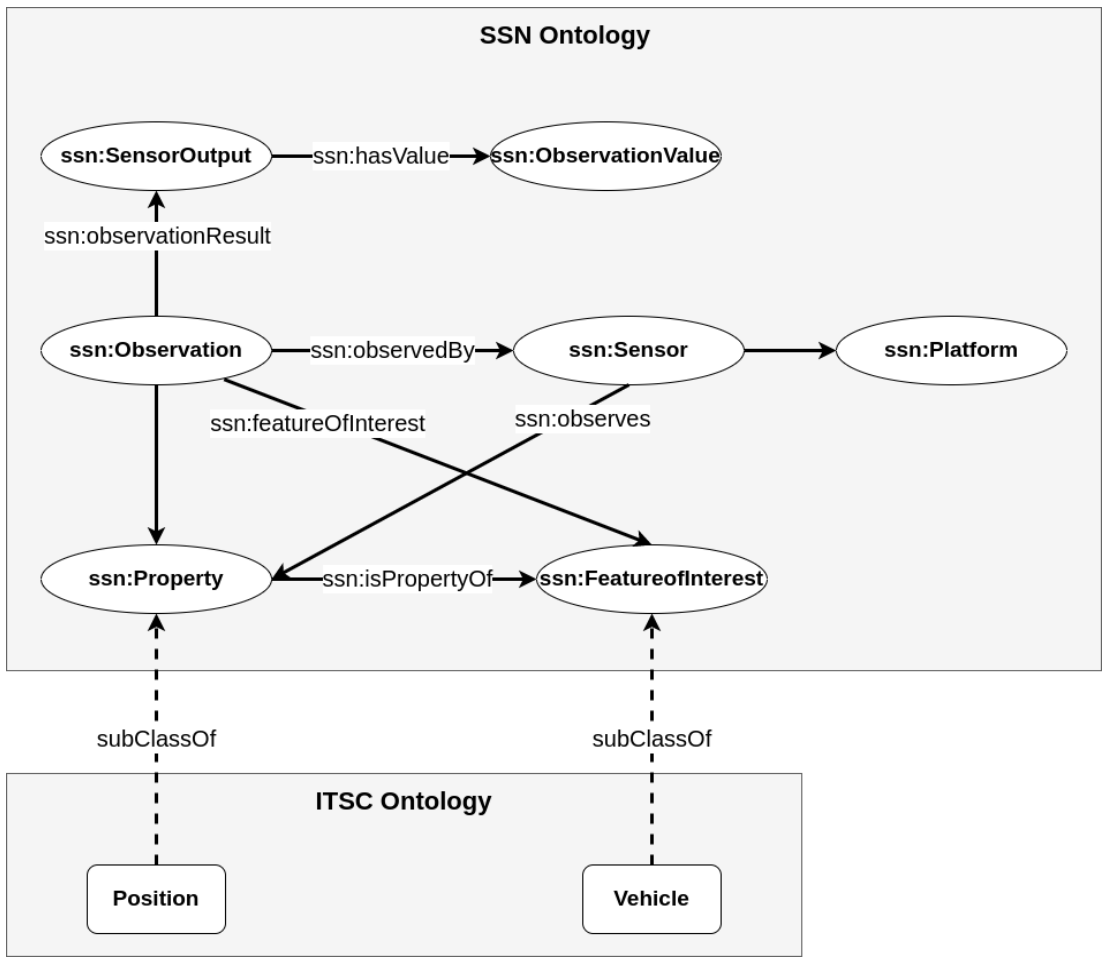

Figure 3.2: The data model to annotate and present sensor data [64].

In this thesis, $O_{g_i}^t$ is defined as the schema describing the data monitored/observed by the agent $g_i$ at time step $t$. $C_{g_i}^t$ represents the set of observed concepts and $F_{g_i}^t$ represents their observed properties (see Equation 3.2).

$$O_{g_i}^t = (C_{g_i}^t, F_{g_i}^t) \tag{3.2}$$

## 3.3 Action Ontology Modeling

Action categories are created to address the need for structured representation and efficient utilization of task-related information [264]. Each action category encapsulates a comprehensive set of information about a specific action, that is mapped to the ontology's elements encompassing concepts, properties, and relationships.

I model the action ontology taxonomy as $O_A$ (see Equation 3.3), where $C_{a_j}$ represents action-bound concepts, which specifies the concepts that are relevant to the action $a_j$, $F_{a_j}$ represents properties, and $M_{a_j}$ represents the relationships between concepts and their properties.

$$O_A = (C_{a_j}, F_{a_j}, M_{a_j}) \tag{3.3}$$

For example, we could have action ontology for the action "ModifyPhaseTime(Intersection, TrafficSignalPhase, Duration)" that represents the ability to adjust the duration of the traffic signal phase for a specific intersection and traffic signal phase. In this action ontology, "TrafficSignalPhase" identifies the specific phase of the traffic signal control for which the timing adjustment is applied. An example of usage is "ModifyPhaseTime(Intersection A, Green, 30 seconds)". By representing actions in the ITSC domain using an ontology, it becomes possible to capture the knowledge and rules related to traffic signal control actions, their parameters, and their effects. This enables intelligent systems to reason about and execute appropriate actions based on the current traffic conditions and control objectives.

## 3.4 Measuring Concept Similarity in Different Ontologies

In the proposed OntoDeM model, I determine which concepts to use for action selection by assessing the similarity between concepts in both the domain-specific ontology and the action ontology. Consequently, this section elaborates on the process of calculating the similarity between concepts across these two ontologies.

The similarity of two concepts $c_x \in O_{d_i}$ and $c_y \in O_{d_j}$ is measured based on their neighborhood concept set and the property set in two different ontologies (see Equation 3.4). The neighborhood concept set in an ontology refers to the collection of directly related concepts to a given concept, and the property set represents the properties associated with that concept.

$$\begin{aligned} \text{Sim}(c_x, c_y) &= \sqrt{\frac{\alpha \, \text{Sim}_N^2 + \beta \, \text{Sim}_F^2}{2}} \\ \alpha &= \frac{|N_{d_i}(c_x)|}{|N_{d_i}(c_x)| + |N_{d_j}(c_y)|} \\ \beta &= \frac{|F_{d_i}(c_x)|}{|F_{d_i}(c_x)| + |F_{d_j}(c_y)|} \end{aligned} \tag{3.4}$$

Where, $Sim(c_x, c_y)$ is considered as the similarity measure between the concepts $c_x$ and $c_y$ and $\alpha$ and $\beta$ specify the relative importance of the two similarity metrics $Sim_N$ and $Sim_F$ respectively. $|N_{d_i}(c_x)|$ and $|F_{d_i}(c_x)|$ represent the number of neighborhood concepts and properties in the domain-specific ontology $d_i$ and $|N_{d_j}(c_y)|$ and $|F_{d_j}(c_y)|$ represent that of the domain-specific ontology $d_j$.

The function $Sim_N$ is used for measuring the similarity of the neighborhood concept set of the selected concepts $c_x$ and $c_y$ in the two ontologies:

$$\text{Sim}_N(c_x, c_y) = \sqrt{\frac{\frac{|N_{d_i}(c_x) \cap N_{d_j}(c_y)|}{|N_{d_i}(c_x)|} + \frac{|N_{d_i}(c_x) \cap N_{d_j}(c_y)|}{|N_{d_j}(c_y)|}}{2}} \tag{3.5}$$

$|N_{d_i}(c_x) \cap N_{d_j}(c_y)|$ represents the number of common neighborhood concepts between the two ontologies. The function $Sim_F$ is used for finding similarity by comparing property sets of the selected concepts $c_x$ and $c_y$ in both ontologies:

$$\text{Sim}_F(c_x, c_y) = \sqrt{\frac{\frac{|F_{d_i}(c_x) \cap F_{d_j}(c_y)|}{|F_{d_i}(c_x)|} + \frac{|F_{d_i}(c_x) \cap F_{d_j}(c_y)|}{|F_{d_j}(c_y)|}}{2}} \tag{3.6}$$

$|F_{d_i}(c_x) \cap F_{d_j}(c_y)|$ represents the number of common properties between the two ontologies.

## 3.5 Measuring Ontology Distance

Ontology distance or ontology similarity is often computed using various metrics that measure the differences or similarities between two ontologies. One common approach is to calculate the "Jaccard Similarity" or "Jaccard Distance" between the sets of concepts, relationships, and properties present in the two ontologies ($A$ and $B$).

Formally, the Jaccard Similarity (JS) between two sets $A$ and $B$ is given by:

$$JS(A, B) = \frac{|A \cap B|}{|A \cup B|}$$

The Jaccard Distance (JD) is then computed as:

$$JD(A, B) = 1 - JS(A, B)$$

A high Jaccard Distance indicates a significant difference between the ontologies, while a low value suggests greater similarity.

In this thesis, I use ontology distance measure to compute the distance between the agent's ontology-based schema in two different time steps $O_{g_i}^{t-1}$ and $O_{g_i}^{t}$. If the distance between them is higher than a predefined threshold, it indicates a substantial change occurred in the environment.

## 3.6 Subsumer Extraction

In the context of ontologies, a subsumer refers to a concept or description that includes all the relevant information and relationships entailed by another concept. It represents a more inclusive description that encompasses the specific concept being considered [12]. For example, "Vehicle" is a subsumer concept, representing the general category of vehicles as a whole. "Emergency" is a subsumee concept, representing specific types of vehicles, such as "FireTruck", "PoliceCar", and "Ambulance". Each vehicle type has its own unique characteristics and additional properties that set it apart from other types. The subsumer-subsumee relationship implies that all instances or concepts categorized as "Emergency" are also instances or concepts of the more general "Vehicle" category. The subsumer concept captures the common aspects and functionalities shared by all types of vehicles, while the subsumee concept represents the specific variations or specializations within the domain [153].

## 3.7 Rule-Based Inference Strategies

The **inference engine** is part of an intelligent system that infers new information based on known facts, using logical rules. Ontology engineers manually assert the logical rules using SWRL or SPIN for the inference engine to compare them with the facts in the knowledge base. When the *IF (Condition)* part of the rule matches a fact, the rule is fired and its *THEN (Action)* part is executed [130].

The Modus Ponens rule is one of the most important rules of inference, and it states that if *(A)* and *(A) THEN (B)* is true, then we can infer that *(B)* will be true. "A" is called the antecedent, and "B" is called the consequent. An inference engine can search for an answer using two basic rule-based inference strategies, such as Forward and Backward chaining, leverage the reasoning capabilities of ontology:

**Forward chaining** infers from logical rules in the knowledge base in the forward direction by applying the Modus Ponens rule to extract more data until a goal is reached. Take into account the following logical rule within the ITSC domain: *IF (Traffic congestion is high) THEN (Extend green phase duration)*. The agent observes the fact that "Traffic congestion is high", using forward chaining, the inference engine applies the rules and extends the green phase duration to alleviate the congestion.

**Backward chaining** starts with a list of goals and works backward from the consequence to the antecedent to see if any known facts support any of the consequences. Suppose the agent uses a knowledge base with the following logical rules to make decisions:

*IF (Congestion is detected at an intersection) THEN (Identify optimal signal timings) and IF (Optimal signal timings are identified) THEN (Update signal timings)*.

The agent's goal is to "Update signal timings" at a congested intersection. Using backward chaining, the agent starts with this goal and works backward through the logical rules to achieve it. First, the agent identifies the need to "Identify optimal signal timings" to alleviate congestion. This requires an analysis of traffic conditions. Backtracking further, the agent acknowledges that it needs to detect "Congestion at an intersection" as a starting point for making any adjustments. Once congestion is detected, the system follows the logical dependencies, identifying optimal signal timings based on traffic analysis, and finally updating the signal timings accordingly.

## 3.8 Assessment of Observation Significance

The agent $g_i$ observes the environment at time step $t$ based on its ontology-based schema $O_{g_i}^t$, and the importance of each observation is determined based on the importance of the concepts $C_{g_i}^t$ involved. I employ the iweighting indicator $iw(c_x)$ intro-duced in [175], which is a concept weighting function. This function helps the agent to quantify how important each concept $c_x \in C_{g_i}^t$ is in an environment (see Equation 3.7).

$$iw(c_x) = 1/|M(c_x)| \sum_{m \in M(c_x)}^{|M(c_x)|} iw(m_{xy}) \tag{3.7}$$

The iweighting indicator is a numerical value calculated from weighting the local context of concept $c_x$ based on its outgoing edges (i.e., relationships to other concepts). The concept's weight $iw(c_x)$ is calculated based on the average importance weights of the relationships $m_{xy} \in M_{g_i}^t$ of domain concept $c_x$ constrained by their particular range $c_y$.

For instance, let's consider the concept "Vehicle" in the ITSC domain. For each property ("hasType", "hasLength", "hasPosition"), there are importance weights assigned to their relationships. These weights reflect the significance of each relationship in the context of the ITSC. Suppose the assigned importance weights for the relationships are delineated as follows: $iw(m_{\text{Vehicle, Emergency}}) = 0.8$, $iw(m_{\text{Vehicle, Medium}}) = 0.6$, and $iw(m_{\text{Vehicle, Stationary}}) = 0.7$. To compute the importance weight for the vehicle denoted as "A", characterized as an ambulance halted at the intersection, the agent takes the average of these importance weights:

$$iw(\text{Vehicle A}) = \frac{iw(m_{\text{Vehicle, Emergency}}) + iw(m_{\text{Vehicle, Medium}}) + iw(m_{\text{Vehicle, Stationary}})}{3}$$

$$iw(\text{Vehicle A}) = \frac{0.8 + 0.6 + 0.7}{3} = 0.7$$

So, in this example, the concept weight for vehicle "A" is calculated to be $0.7$ This process ensures that the importance of relationships contributes to the overall weight of the concept within the ontology.

## 3.9 Application-Specific Ontologies

In this section, I will discuss the ontologies tailored for the application areas covered in this thesis, which will serve as specialized knowledge representations for each of these application areas.

### 3.9.1 Ontology for Intelligent Traffic Signal Control System

For modeling ITSC concepts, I leverage the established ontology from [184] and augment it with new concepts such as "Type", "Position", "Length" and "TrafficSignalAction". The representation of these enhancements is provided in Figure 3.3.

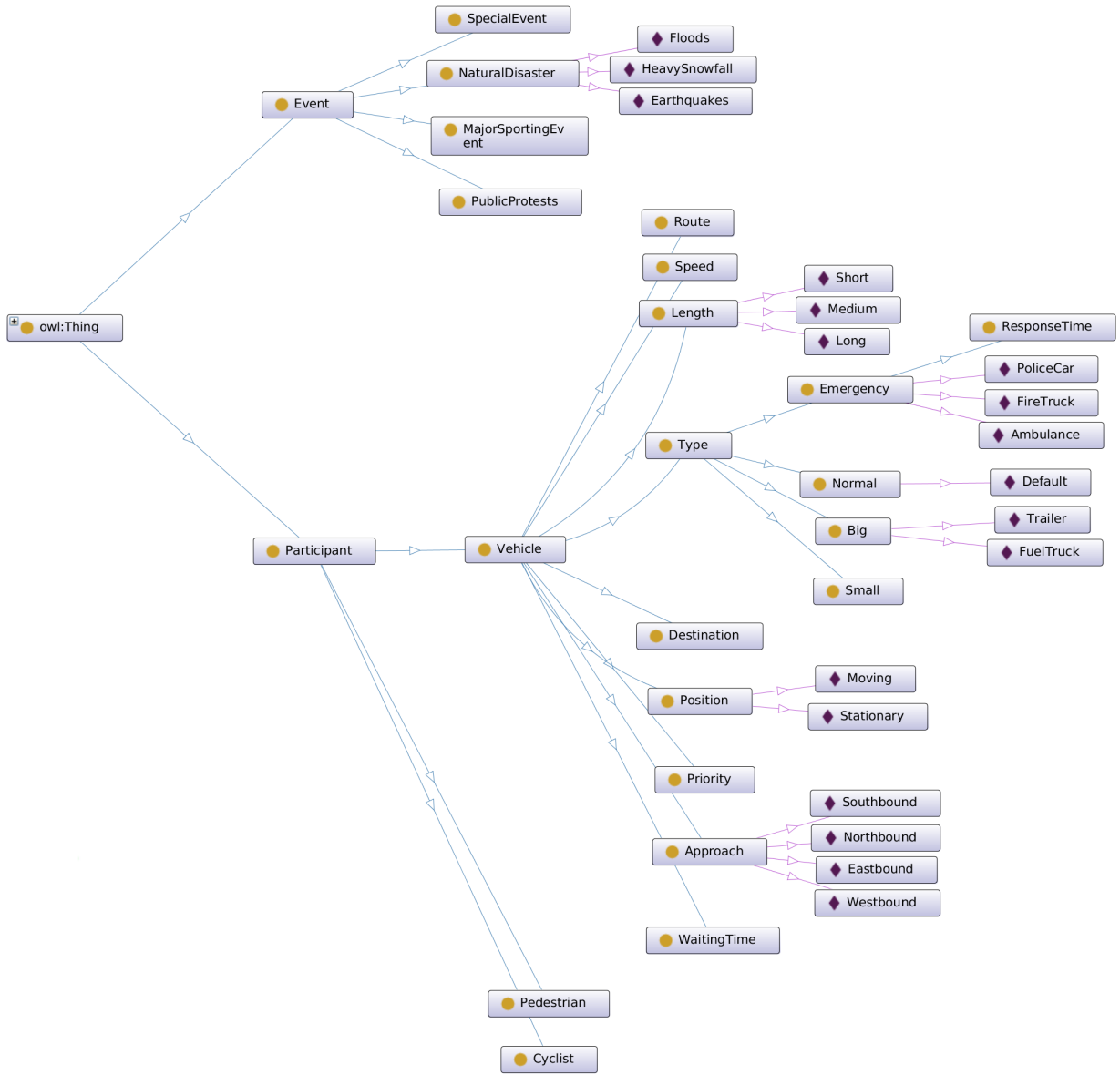

(a) Ontology for intelligent traffic signal control, as represented by OntoGraf.

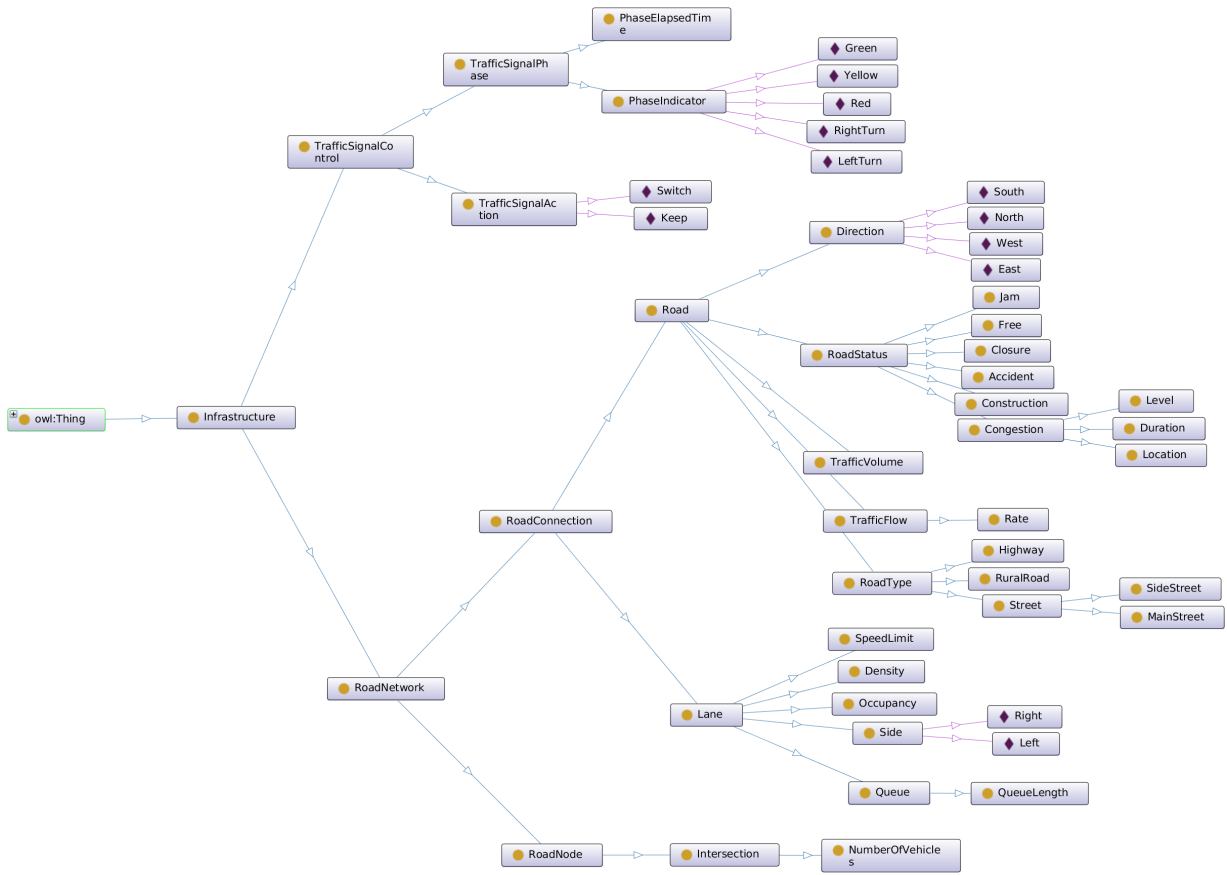

Figure 3.3 (Continued): Ontology for intelligent traffic signal control
, as represented by OntoGraf.

**Classes:** Classes are used to represent more general concepts or categories of entities. These classes define comprehensive characteristics, properties, and relationships that entities within those categories share. Below are the classes within the domain of ITSC:

- `Event`: Refers to any event that has the potential to impact traffic flow or road usage.
- `Participant`: Represents any entity actively involved in or interacting with traffic systems.
    - ▶ `Vehicle`: Represents a vehicle on the road network.
        - `WaitingTime`: Refers to the duration a vehicle is stationary or delayed.
        - `Route`: Signifies the specific path that a vehicle follows to reach its destination.
        - `Speed`: Refers to the rate at which a vehicle is moving.
        - `Destination`: Represents the specific location or endpoint to which a vehicle is intended to arrive or reach during its journey.
        - `Approach`: Represents the direction a vehicle takes as it nears and prepares to enter an intersection.

- `Priority`: Refers to the level of precedence that a vehicle has when navigating through an intersection. It determines which vehicle has the authority to proceed before others, often based on traffic rules and regulations.
- `Type`: Represents a general type or category of a vehicle.
  - ▶ `Big`: Represents a type of vehicle that is categorized as "big" in size.
  - ▶ `Small`: Represents a type of vehicle that is categorized as "small" in size.
  - ▶ `Normal`: Represents a normal or regular type of vehicle.
  - ▶ `Emergency`: Represents a type of vehicle used for emergency services.
- `Position`: Represents the position or location of a vehicle on a lane.
- `Length`: Represents the length attribute of a vehicle.

▶ `Cyclist`

▶ `Pedestrian`

- `Infrastructure`: Represents elements of the physical environment related to road networks.
  - ▶ `RoadNetwork`: Represents a network of interconnected roads.
    - `RoadNode`: Represents a node within the road network, often an intersection.
      - ▶ `Intersection`: Represents a road junction where multiple roads meet.
        - `NumberOfVehicles`: Represents the count or quantity of vehicles located within or near an intersection at a given point in time.
    - `RoadConnection`: Represents a connection between road segments.
      - ▶ `Lane`: Represents a lane on a road.
        - `Side`: Represents the side of a lane, often used for indicating left or right.
        - `Occupancy`: Refers to the percentage of time that a specific area or section of a lane is occupied by vehicles.
        - `Queue`: Represents the line or sequence of vehicles waiting in a specific lane.
        - `Density`: Refers to the number of vehicles occupying a specific section of a lane at a given moment.

- - **`SpeedLimit`**: Refers to the maximum legal speed at which vehicles are allowed to travel on a specific lane.
  - ▶ **`Road`**: Represents a road segment within a road network.
    - **`Direction`**: Represents a direction of movement on a road.
    - **`RoadStatus`**: Encompasses various conditions on a road network, including congestion, traffic jams, and accidents.
    - **`TrafficVolume`**: Refers to the number of vehicles passing through a specific road segment.
    - **`TrafficFlow`**: Represents the movement of vehicles through a specific road segment, often characterized by vehicle speed, density, and spacing.
    - **`RoadType`**: Classifies road segments based on their characteristics, distinguishing between various types such as rural roads, streets, and highways, often based on factors like speed limits, surrounding land use, and road design.
- ▶ **`TrafficSignalControl`**: Represents the control and regulation of traffic signals.
  - **`TrafficSignalAction`**: Represents an action related to traffic signal control.
  - **`TrafficSignalPhase`**
    - ▶ **`PhaseElapsedTime`**: Refers to the duration of a specific phase within a traffic signal cycle.
    - ▶ **`PhaseIndicator`**: Represents a phase or state of a traffic signal.

**Individuals:** Each ontology can contain various individuals representing different instances of concepts. There are instances of road status such as "Accident", "Free" and "Jam" denoting different traffic conditions. Emergency vehicles are represented by individuals like "Ambulance", "PoliceCar", and "FireTruck" belonging to the "Emergency" category. "Trailer" and "FuelTruck" represent the "Big" category of vehicles. Different directions are captured through instances like "East", "West", "North", and "South", while lane sides are portrayed with "Left" and "Right". Traffic signal phases like "Red", "Yellow", and "Green" and traffic signal actions like "Keep" and "Switch" are also represented. Moreover, vehicle positions like "Moving" and "Stationary" and lengths like "Long", "Short", and "Medium" are individual instances. These individuals collectively contribute to modeling and understanding various aspects of road networks, traffic conditions, and vehicle attributes within the ontology.

## 3.9.2 Ontology for Edge Computing Environment

To model the concepts in an EC environment, I propose using the ontology shown in Figure 3.4.

Classes and subclasses in the EC ontology are as follows:

- `EdgeDevice`: Represents a device within the EC environment.
    - ▶ `DeviceMobility`: Represents the mobility characteristics of the device.
        - `StationaryDevice`: A device that remains fixed in position.
        - `MobileDevice`: A device capable of movement.
    - ▶ `DeviceLocation`: Represents the spatial coordinates of the device.
        - `DeviceY`: The Y-coordinate of the device's location.
        - `DeviceX`: The X-coordinate of the device's location.
    - ▶ `DeviceSoftware`: Represents software-related information of the device.
        - `DeviceOperatingSystem`: The operating system of the device.
        - `DeviceDatabase`: The database software used by the device.
    - ▶ `DeviceType`: The type of the device, e.g., Laptop, Sensor, Smartphone, WearableGadget.
    - ▶ `DeviceHardware`: Represents hardware-related information about the device.
- `User`: Represents a user interacting with the EC environment.
    - ▶ `ContactInfo`: Contact information of the user, including Name, Address, and Telephone.
    - ▶ `FinancialInfo`: Financial information of the user, including CardNumber.
    - ▶ `UserGroup`: Represents the user's group affiliation.
        - `SimilarityUserGroup`: A group based on user similarity.
        - `GeographicalUserGroup`: A group based on user geographical location.
    - ▶ `UsageHistory`: Represents the historical usage behavior of the user.
- `Connection`: Represents a connection between devices or servers.
    - ▶ `Network`: The network type or protocol used for the connection.
    - ▶ `Bandwidth`: The available bandwidth of the connection.
    - ▶ `ConnectionCost`: The cost associated with using the connection.

- `Service`: Represents a service provided within the EC environment.
  - `ServiceType`: The type of service provided.
    - `TaskOffloading`: Service for offloading tasks to servers.
    - `LoadBalancing`: Service for load balancing among servers.
    - `Clustering`: Service for grouping related devices or servers.
  - `Quality`: Represents quality attributes of the service.
    - `Fairness`: The fairness of service allocation.
    - `ServiceCost`: The cost associated with using the service.
- `Task`: Represents a task that needs to be processed.
  - `TaskSize`: The size of the task, categorized as SmallTask, MediumTask, or BigTask.
  - `ApplicationType`: The type of application the task belongs to.
  - `QoS`: Quality of Service attributes for the task.
    - `Security/Privacy`: Security and privacy requirements of the task.
    - `ErrorRate`: Allowable error rate for the task.
    - `Delay`: Maximum acceptable delay for the task.
  - `TaskOwner`: The user or entity that owns the task.
  - `ExpectedCompletionTime`: The expected time required to complete the task.
  - `WaitingTime`: The time the task spends waiting in the queue.
  - `ArrivingTime`: The time when the task arrives for processing.
  - `State`: The current state of the task.
  - `Requirement`: The resource requirements of the task, including Data, Software, and Hardware.
  - `Priority`: The priority level of the task, categorized as Low, Medium, or High.
  - `Latency`: The latency requirements of the task, categorized as VertLowLatency, LowLatency, or HighLatency.
- `EdgeServer`: Represents an edge server within the environment.
  - `Buffer`: The buffer space available for storing tasks.
  - `Resource`: The computational resources available on the server.
    - `Access`: Access type to the server's resources.

- ▶ `Shared`: Resources are shared among tasks.
- ▶ `Physical`: Dedicated physical resources for tasks.
- ▶ `Exclusive`: Exclusive access to resources.
- ▶ `Virtualized`: Resources are virtualized for tasks.
- `ServerSoftware`: Software-related information of the server.
  - ▶ `ServerDatabase`: The database software used by the server.
  - ▶ `ServerOperatingSystem`: The operating system of the server.
- `ServerHardware`: Hardware-related information of the server.
  - ▶ `Storage`: Storage Technology, TransferRate, StorageSpace, RotationalSpeed, and DataStorageCompany information.
  - ▶ `RAM`: RAMType, AccessTime, RAMCapacity, and RAMManufacturer.
  - ▶ `CPU`: CPU details, including Core, CPUType, and CPUManufacturer.

▶ `ServerLocation`: The spatial location of the server.
- `ServerX`: The X-coordinate of the server's location.
- `ServerY`: The Y-coordinate of the server's location.

▶ `ServerCost`: Represents the cost-related information of the edge server.
- `ServerSubscription`: Indicates the subscription status of the server.
  - ▶ `FreeServer`: A server available for free usage.
  - ▶ `PaidServer`: A server that requires payment for usage.
- `CostType`: Represents the type of cost associated with the server.
  - ▶ `Energy`: Energy-related costs.
  - ▶ `Maintenance`: Costs related to server maintenance.

▶ `ServerSize`: Represents the size category of the edge server, e.g., SmallServer, MediumServer, and BigServer.

▶ `Workload`: Represents the current workload of the server.

▶ `Board`: Represents the board category of the server, e.g., LowBoard and HighBoard.

▶ `ServerGroup`: Represents a group of servers, possibly based on specific criteria.
- `SimilarityServerGroup`: A group of servers with similar attributes.
- `GeographicalServerGroup`: A group of servers based on geographical location.

- ▶ `ServerMobility`: Represents the mobility characteristics of the server.
  - • `MobileServer`: A server capable of movement.
  - • `StationaryServer`: A server that remains fixed in position.
- ▶ `ServerLimit`: Represents the limitation or capacity of the server.
- ▶ `ServerOwner`: Represents the user or entity that owns the server.
- ▶ `ServerCapacity`: Represents the overall processing capacity of the server.

**Individuals:** A range of instances exist within the categories of "DeviceOperatingSystem" and "DeviceDatabase". Additionally, various predefined user and server groups have their own instances. Various "ApplicationTypes", including "DataCollection", "Entertainment", "HealthCare", and "VoIP", are defined as instances. Moreover, different states of a Task, such as "Offloading", "Sending", "Processing", "Migrating", "SendingBack", and "Disconnecting", are also instantiated.

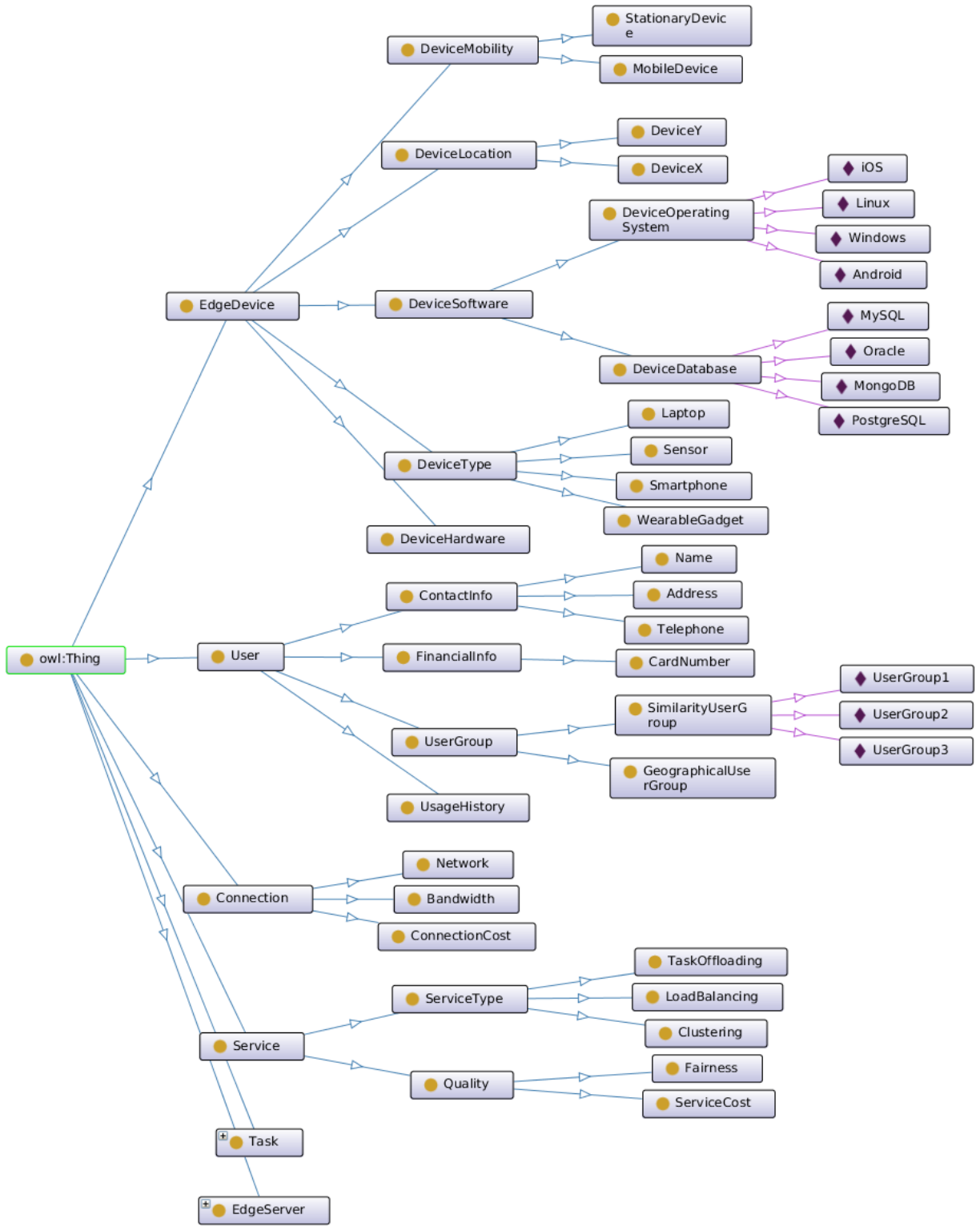

(a) Ontology for edge computing, as represented by OntoGraf.

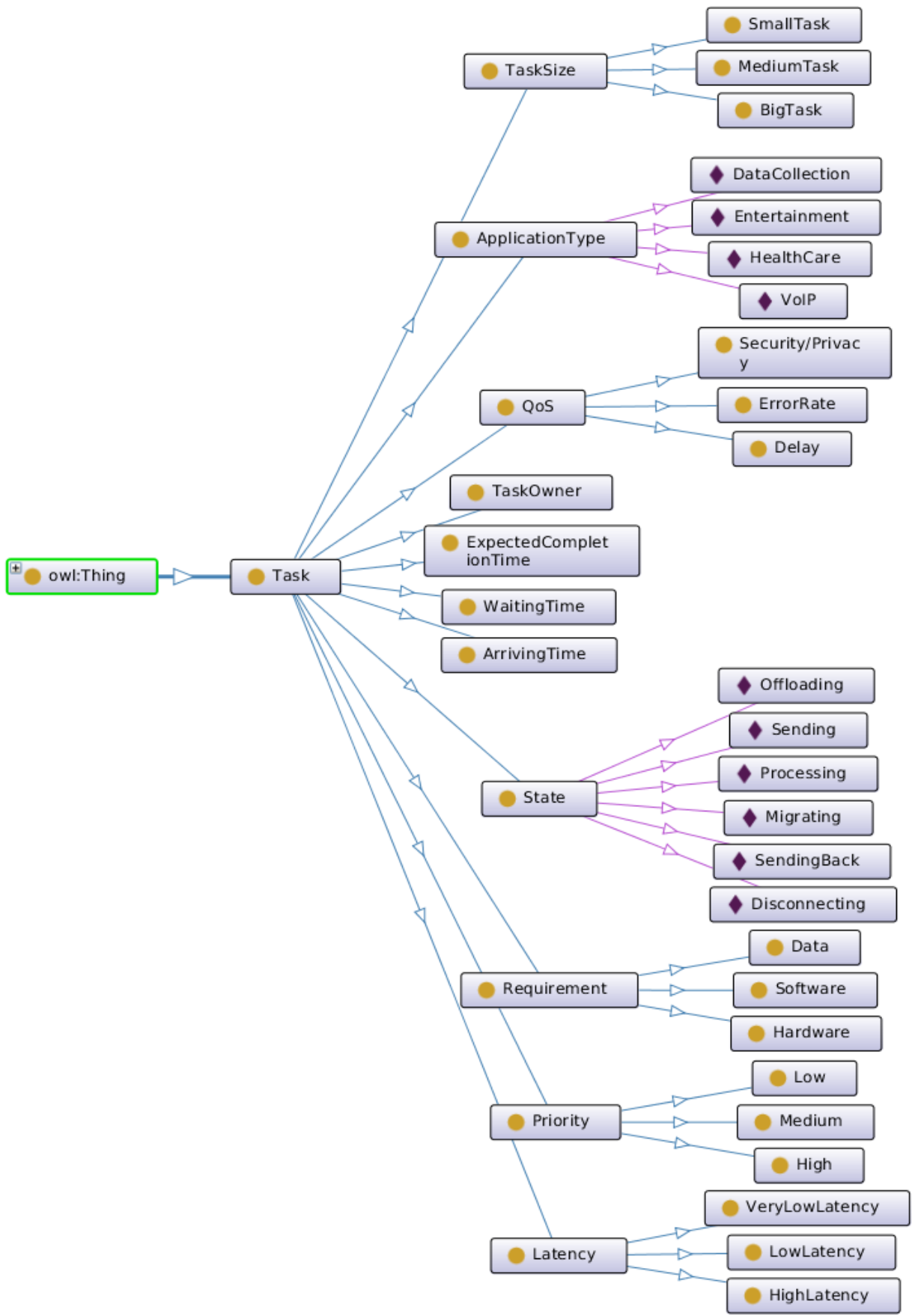

(b) (Continued): Ontology for edge computing, as represented by OntoGraf.

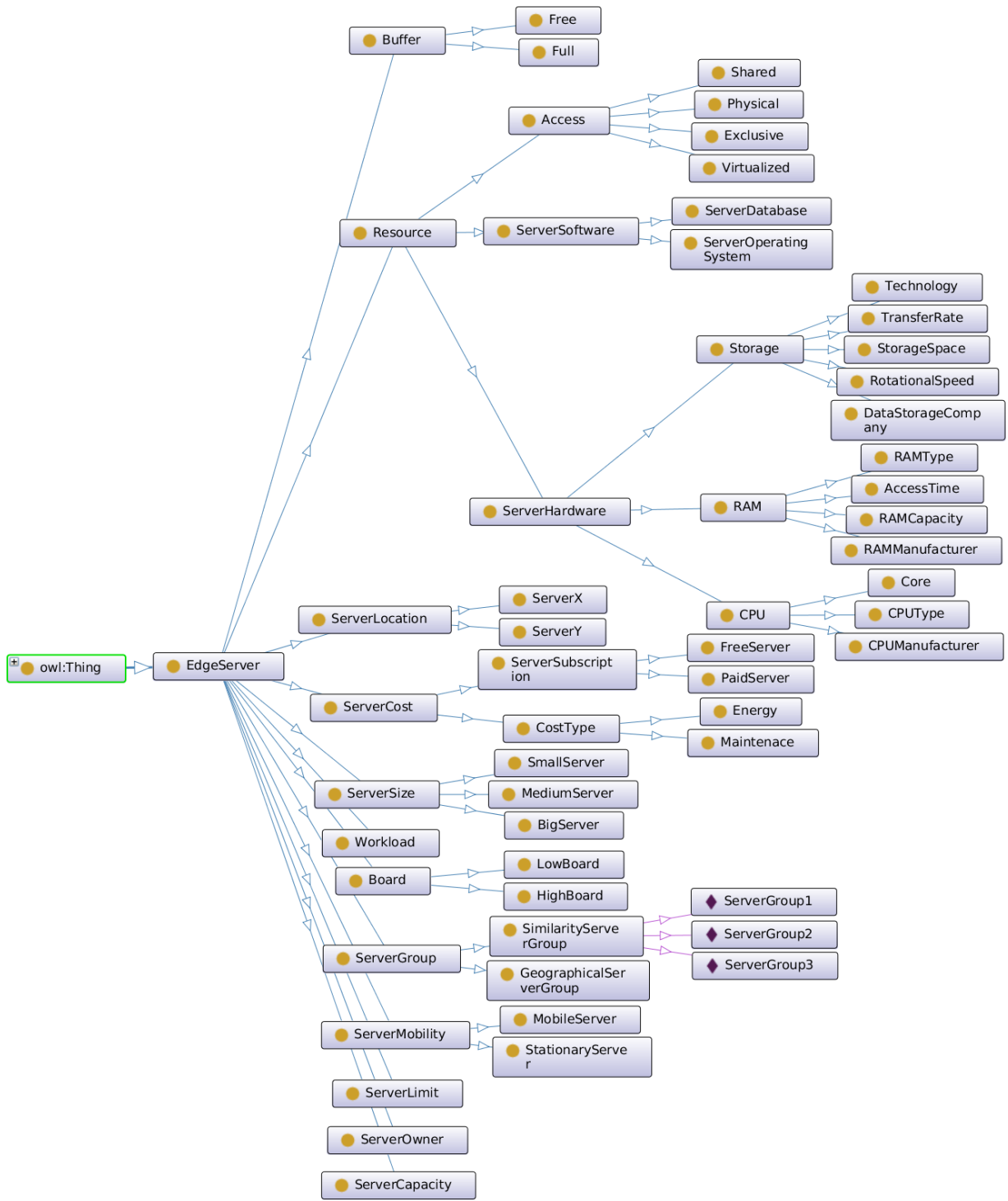

Figure 3.4 (Continued): Ontology for edge computing, as represented by OntoGraf.

### 3.9.3 Ontology for Job Shop Scheduling

To model the related concepts in the JSS environment, I use the ontology shown in Figure 3.5.

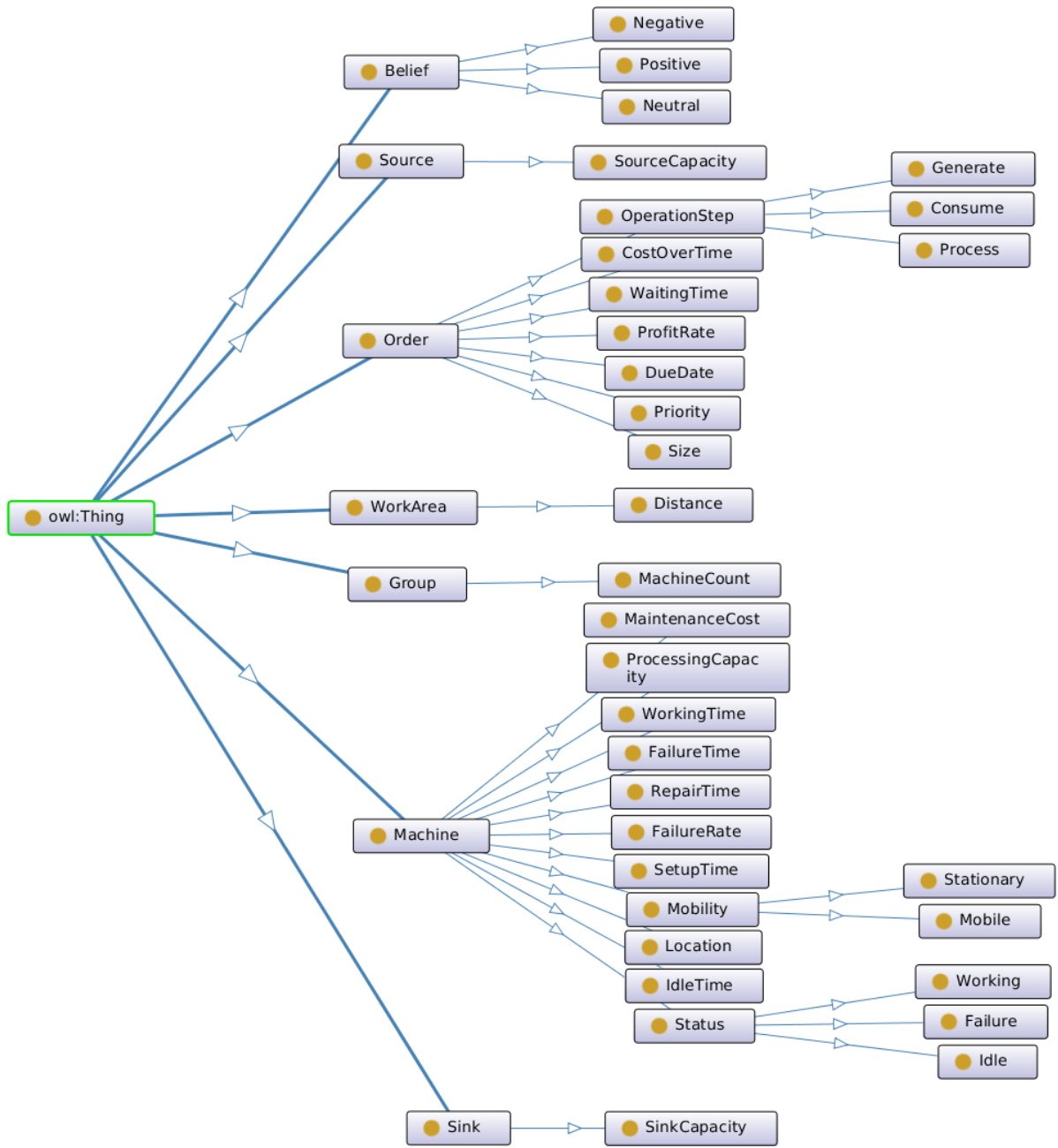

Figure 3.5: Ontology for job shop scheduling
, as represented by OntoGraf.

**Classes:** In the JSS ontology, the following classes represent different aspects or attributes within the domain:

- `Belief`: Represents a category of beliefs. Beliefs are shaped by human experiences with various concepts. For instance, WorkingTime is linked to Positive Belief, while WaitingTime is connected to Negative Belief.
- `Source`: Represents an entity of origin.
    - ▶ `SourceCapacity`: Signifies the capacity of a source.

- `Order`: Denotes a task or job.
  - ▶ `CostOverTime`: Depicts fluctuating costs during order processing.
  - ▶ `DueDate`: Specifies the deadline for an order.
  - ▶ `Priority`: Represents the level of importance of an order.
  - ▶ `Size`: Represents the size of an order.
  - ▶ `WaitingTime`: Quantifies the duration orders spend waiting.
  - ▶ `ProfitRate`: Represents the rate of financial gain achieved upon order completion.
  - ▶ `OperationStep`: Describes a distinct task within a broader operational process.
- `WorkArea`: Represents a designated working space for the machine.
  - ▶ `Distance`: Denotes spatial measurement between the job shop scheduler agent and work area.
- `Group`: Represents a collection of machines.
  - ▶ `MachineCount`: Indicates the number of machines within a group.
- `Machine`: Denotes a machine or equipment.
  - ▶ `FailureRate`: Signifies the frequency of machine failures.
  - ▶ `FailureTime`: Indicates the moment when a machine failure occurs.
  - ▶ `IdleTime`: Quantifies the duration of machine inactivity.
  - ▶ `Location`: Specifies the position or coordinates of a machine.
  - ▶ `MaintenanceCost`: Represents the expenses related to machine maintenance.
  - ▶ `Mobility`: Describes the ability of a machine to move.
  - ▶ `ProcessingCapacity`: Specifies the capability of a machine to process orders.
  - ▶ `WorkingTime`: Quantifies the time a machine is in an active state.
  - ▶ `RepairTime`: Denotes the time required for machine repair.
  - ▶ `SetupTime`: Specifies the time needed for machine setup.
  - ▶ `Status`: Represents the condition or state of a machine.
    - `Working`: Represents a machine in an active state.
    - `Failure`: Represents a machine in a failure state.
    - `Idle`: Represents a machine in an inactive state.

> ▶ `Mobility`: Describes the ability of a machine to move.
>> • `Mobile`: Represents a machine capable of movement.
>> • `Stationary`: Represents a machine that remains fixed.
> • `Sink`: Represents a destination entity.
>> ▶ `SinkCapacity`: Denotes the capacity of a sink or destination.

In addition to the classes, there are subclass relationships established between some of the classes. For example, "Generate", "Consume", and "Process" are subclasses of "OperationStep", indicating that these activities are various types of operation steps.

### 3.9.4 Ontology for Temporal Information

The Human Time Ontology (HuTO) [58] is used in this thesis to enable standardized representation and reasoning of temporal information related to different events (See Figure 3.6).

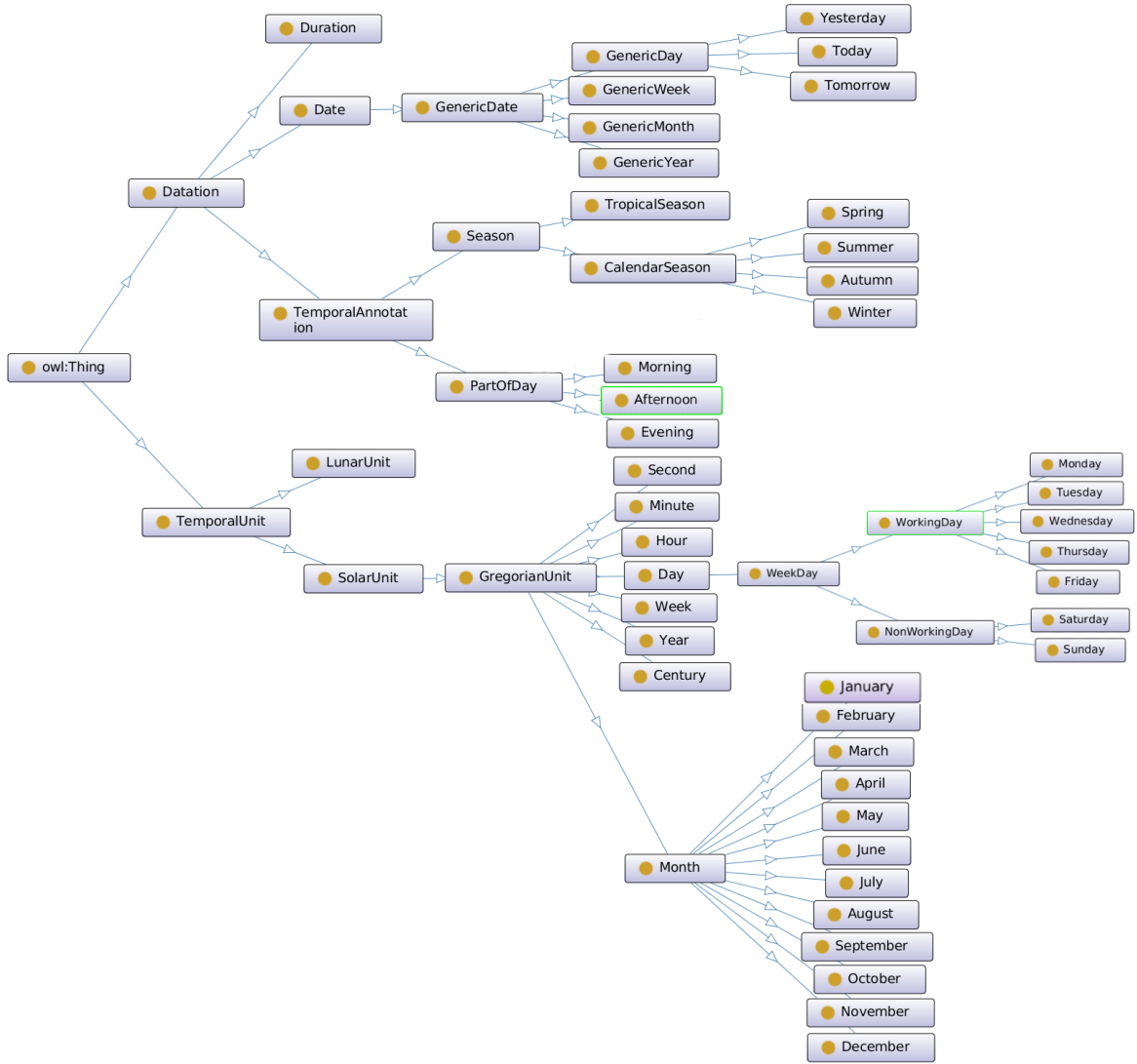

Figure 3.6: Human Time Ontology (HuTO)
, as represented by OntoGraf [58].

Temporal information classes and their subclasses are defined as follows within the HuTO ontology:

- `Datation`: Represents the concept of time in a general sense.
    - ▶ `Duration`: Represents a period of time.
    - ▶ `Date`: Represents a specific point in time.
        - `GenericDate`: Represents a date with a specific relationship to the current time, including days, weeks, months, and years.
            - ▶ `GenericDay`: Represents a day relative to the current time, including Yesterday, Today, and Tomorrow.
            - ▶ `GenericWeek`: Represents a week relative to the current time.

- ▶ `GenericMonth`: Represents a month relative to the current time.
- ▶ `GenericYear`: Represents a year relative to the current time.
- ▶ `TemporalAnnotation`: Provides a means to link temporal concepts with actual data resources.
  - `Season`: Represents a division of the year marked by changes in weather.
    - ▶ `TropicalSeason`: Divides the year into two distinct periods based on the tropical climate: the dry season and the wet season.
    - ▶ `CalendarSeason`: Encompasses Spring, Summer, Autumn, and Winter.
  - `PartOfDay`: Represents a division of the day, including Morning, Afternoon, and Evening.
- `TemporalUnit`: Represents different units of time.
  - ▶ `LunarUnit`: Represents units of time in a lunar calendar.
  - ▶ `SolarUnit`: Represents units of time in a solar calendar.
    - `GregorianUnit`: Represents various units of time in the Gregorian calendar system, such as Second, Minute, Hour, Day (including WorkingDay and NonWorkingDay), Week, Year, Century, and Month (including January to December).

## 3.10 Summary

The chapter commences by presenting ontology development through the NeOn methodology, which encompasses phases including requirement analysis, ontology design, ontology development, ontology integration, and ontology evaluation. It then highlights Protégé as a versatile platform for creating and managing ontologies, supporting various ontology languages. Once the domain-specific ontology is created, the procedure of mapping low-level sensor data with high-level concepts using the Semantic Sensor Network (SSN) ontology is introduced. This process leads to the development of an ontology-based schema for agents' observations. This schema incorporates calculated importance weights and similarities for concepts. Also, the Jaccard Distance measure is used to quantify differences between ontologies, particularly in assessing changes in the agent's schema over time. Furthermore, the differentiation between general concepts known as Subsumers and specific concepts termed subsumees effectively encompasses both general and specialized aspects within the specified domain.

An inference engine, with forward and backward chaining strategies, helps intelligent agents in deducing new insights from existing knowledge. Apart from the ontology tailored for agents' observations, separate ontologies are established for individual actions, encompassing pertinent concepts and properties. To conclude, the chapter delves into crafting application-specific ontologies designed for diverse domains. Collectively, these approaches constitute the foundation for ontology-based environment modeling within the proposed OntoDeM model, providing a robust foundation for reasoned decision-making and informed actions.

# CHAPTER
## FOUR

# ONTOLOGY-ENHANCED DECISION-MAKING

This chapter presents the problem statement, the key contributions of the thesis, and how these contributions form the novel Ontology-Enhanced Decision-Making model (OntoDeM).

First, in the *problem statement* section, we discuss the challenges that RL encounters in POMDPs when solving real-world problems. Specifically, these challenges include dealing with ambiguous observations and handling unforeseen events that may arise in an environment. In the subsequent *Ontology-enhanced decision-making* section, I elaborate on OntoDeM's contribution through three stages: Ontology-enhanced observation modeling, Ontology-driven goal selection/generation, and Ontology-driven action selection. These stages use several tools such as ontology-based schema, concept weighting, similarity measures, inference rules, and semantic constraints to improve the agent's decision-making process. Lastly, in the *OntoDeM guidelines* section, a comprehensive collection of structured recommendations is presented, offering a systematic approach for selecting the most suitable ontology-based method according to distinct situations and specific prerequisites.

## 4.1 Problem Statement

*Partially Observable Markov Decision Processes*, POMDPs pose several challenges for RL algorithms when applied to real-world problems.

**Ambiguous Observation:** The challenge pertains to a scenario wherein a sequence of observations, denoted as historical experience $h^t$, which consists of state-action pairs $((s^1, a^1), (s^2, a^2), \ldots, (s^t, a^t))$ is given over a historical time span $[1, t]$, where $t$ represents the previous time step. The primary complexity arises from the fact that current ob-servation $v^{t+1}$ can correspond to multiple latent states $s^i$, where $i$ indexes the potential underlying states. Mathematically, this can be represented as $P(s^i|v^{t+1})) > 0$ for multiple $i$ values. Consequently, the task involves understanding the true underlying state $s^{t+1}$ from the set of feasible states $\{s^i\}$ for each observation $v^{t+1}$, indicating the inherent ambiguity introduced by these observations and their associations with different states.

**Unforeseen Events:** Unforeseen events can introduce sudden changes in the environment which can significantly affect the dynamics and state transitions of the POMDP, making it difficult for RL algorithms to learn optimal policies. The limited visibility of the environment (i.e., partial observability) during unforeseen events makes it challenging for the agent to make accurate predictions and take appropriate actions. Additionally, the agent's knowledge gained from past experiences may not be relevant or reliable in the face of unforeseen events, further complicating the learning process.

Given an unforeseen event function $E : S \times S \times A \to \{0, 1\}$, where $E(s^{t+1}, s^t, a^t) = 1$ indicates that an unforeseen event occurred, resulting in a transition from state $s^t$ to state $s^{t+1}$ upon taking action $a^t$. Otherwise, $E(s^{t+1}, s^t, a^t) = 0$ indicates the absence of an unforeseen event. The challenge arises when unforeseen events $E(s^{t+1}, s^t, a^t) = 1$ introduce unexpected state transitions. These transitions might result in states $s^{t+1}$ that is not represented or contemplated within $h^t$. Consequently, the historical experience $h^t$ might not contain relevant or informative state-action pairs to guide decision-making in the new unforeseen state. This lack of relevant experience can hinder the agent's ability to accurately determine the best action $a^{t+1}$ to take given the unforeseen event-induced changes, leading to ambiguity in action selection due to the divergence between its past learned behaviors and the novel dynamics introduced by the unforeseen event.

In these situations, effective decision-making requires unforeseen event detection and action selection through exploration of the action space to uncover suitable actions, as the historical experience alone may lack the necessary guidance to address the novel circumstances arising from unforeseen events. In situations where the action space is extensive, the exploration process becomes arduous and leads to another challenge called the curse of dimensionality. So, the agent needs to evaluate numerous actions to identify those that lead to favorable outcomes, consequently, finding optimal policies within a reasonable timeframe becomes challenging. Therefore, adaptive exploration and learning strategies are needed to handle these challenges, allowing the RL algorithm to actively explore and adapt to the changing dynamics of the POMDP, even in the presence of unforeseen events.

To tackle these challenges, we have introduced a novel ontology-enhanced decision-making model (OntoDeM) that utilizes the potential of ontology. Ontology plays a pivotal role in overcoming these challenges through the following ways. Firstly, by employing ontology, we construct a structured representation (i.e., entities, relationships, and properties) of domain-specific knowledge pertinent to the POMDP problem. Consequently, our RL agents gain the ability to maintain a more accurate representation of their state, allowing them to adeptly reason about the ever-evolving dynamics of the environment. Furthermore, ontology allows the integration of contextual information, encompassing factors like environmental conditions, temporal aspects, and external events, into the POMDP framework. By representing and reasoning about this contextual data, ontology enhances the RL agent's ability to adapt its behavior and decision-making process in response to unforeseen events and evolving conditions. Lastly, ontology's capacity to facilitate the sharing of past experiences and learned models in a standardized format enhances knowledge transfer across different time steps. This capability proves instrumental in addressing the issues of partial observability and unforeseen events, as it allows the agents to draw upon insights derived from similar scenarios to make more informed decisions.

## 4.2 Ontology-Enhanced Decision-Making Model (OntoDeM)

This section describes the contribution of the thesis, **Onto**logy-enhanced **De**cision-**M**aking model (OntoDeM), in three stages of the decision-making process using the ontology-enhanced environment modeling process that was described in the previous chapter.

Using OntoDeM, agents use the ontology-enhanced environment model that allows them to access an ontology-based schema, a concept weighting method, a concept similarity measure, inference rules, forward and backward reasoning, and semantic constraints, which they use as a tool to augment their observations, reason about unforeseen events in the environment, and ultimately improve their performance. OntoDeM includes the following stages: (1) ontology-enhanced observation modeling, (2) ontology-driven goal selection/generation, and (3) ontology-driven action selection (see Figure 4.1). The details of these stages are discussed further in the following subsections.

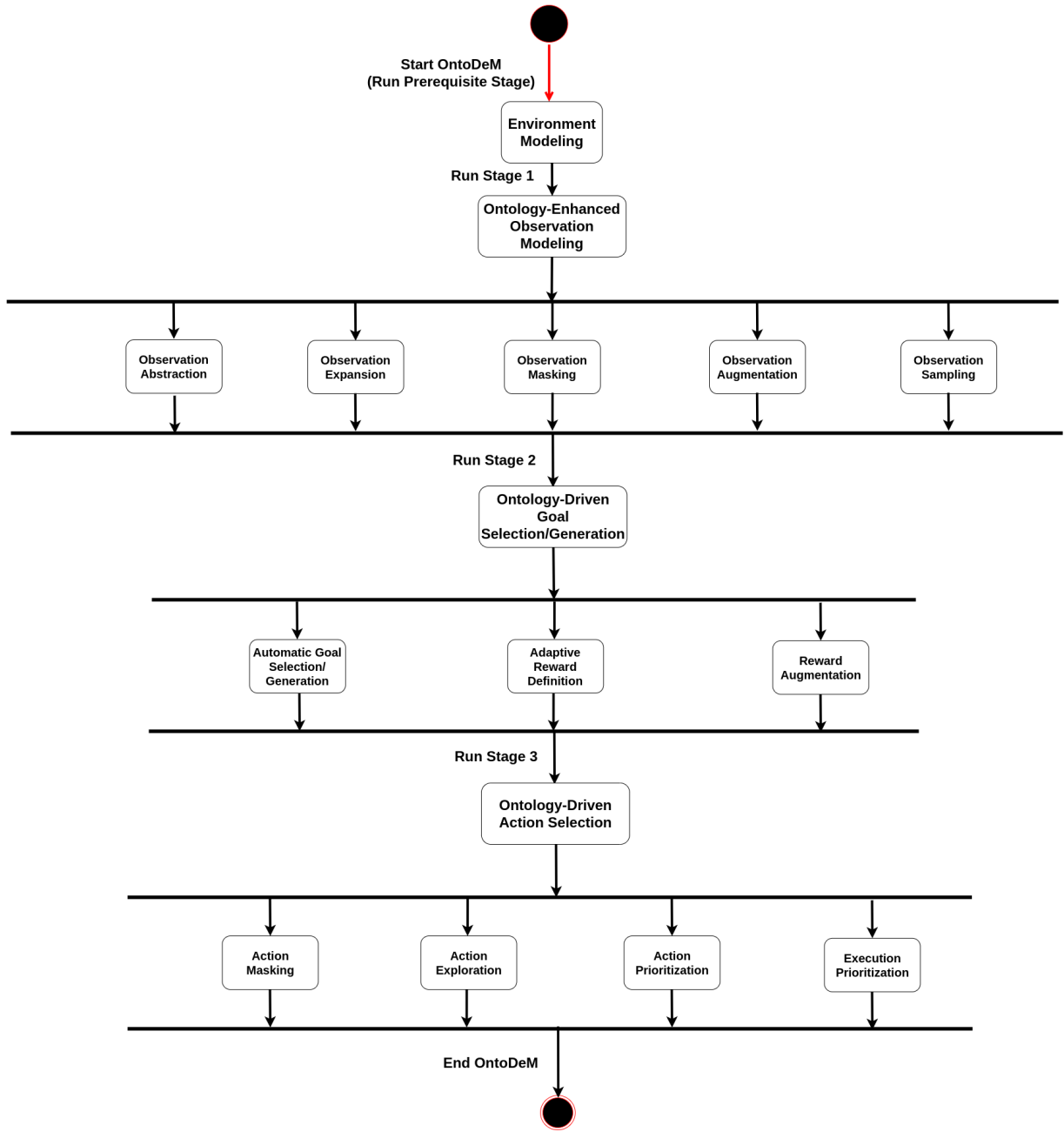

Figure 4.1: The state diagram of the ontology-enhanced decision-making model (OntoDeM).

## 4.2.1 Ontology-Enhanced Observation Modeling

The ontology-enhanced observation modeling stage includes five methods to improve the agent's observation and consequently its decision-making in partially observable and dynamic environments. These methods are as follows: *observation abstraction, observation expansion, observation masking, observation augmentation, and observation sampling*. The details of these methods are described in the following subsections.

#### 4.2.1.1 Observation Abstraction

In ontology, an abstract concept (superclass) refers to a higher-level categorization or general concept that captures common characteristics shared by multiple more specific instances or concrete concepts (subclass). Ontologies employ a hierarchical structure, known as a taxonomy or ontology tree, where concepts are organized in a parent-child relationship. Abstract concepts are represented by higher-level nodes, and concrete concepts are represented by lower-level nodes. This hierarchical arrangement facilitates the organization, classification, and understanding of concepts, enabling more efficient knowledge representation and reasoning within the ontology. This hierarchical structure enables abstracting and grouping of agent's observations at different levels of granularity. In the proposed observation abstraction method, when there is a significant difference between the observed values and the expected values for a property, the agent identifies that property as having a noisy value. Then the agent substitutes a noisy property value $f_x$ with the average of non-noisy values from previous observations in the same class. This classification is accomplished by employing the abstract concept $c_y$ within the agent's observation so that the value of property $f_x$ changes in response to variations in the concept $c_y$ (see Algorithm 1). For instance, within the context of ITSC, if the agent acquires a noisy queue length value for vehicles at a specific time on a particular day (e.g., Saturday at 7:00 a.m.), it can classify past observations based on the abstract concept "Time". Rather than specifying the exact time (e.g., 7:00 a.m.), the agent can employ a broader category such as morning. Subsequently, it deduces the noisy data value by averaging the queue length values from its prior observations during Saturday mornings.

#### 4.2.1.2 Observation Expansion

The observation expansion method [93] aims to enrich the ontology-based schema by leveraging domain-specific ontology knowledge. The method consists of two phases: the **conceptual expansion approach** and the **property addition approach**. In the former, newly deduced concepts from the domain-specific ontology are systematically added to the ontology-based schema, enhancing its capacity to encompass a broader range of observations. In the latter, new deduced properties are associated with the relevant concepts within the ontology-based schema (see Algorithm 2). Through this method, the ontology-based schema is iteratively expanded with both new concepts and properties, leading to a more comprehensive and informed representation of observed data, which is crucial for informed decision-making in dynamic environments. For example, in the ITSC domain, the ontology can incorporate contextual factors such

**Algorithm 1** Ontology-Enhanced-Observation-Modeling

**Require:** $O_{g_i}^t$ - Ontology-based schema
**Ensure:** $(v_{g_i}^t)'$ - Abstraction-enhanced observation
1: **function** OBSERVATION-ABSTRACTION($O_{g_i}^t$)
2:    **for** $f_x \in F_{g_i}^t$ **do** ▷ Iterate over properties in ontology-based schema
3:      **if** property $f_x$ is noisy **then**
4:        $c_y \leftarrow$ Find superclass concept related to $f_x$ ▷ Utilize ontology reasoning
5:        $G \leftarrow$ Group observations by $c_y$ ▷ Organize observations
6:        **for** $group \in G$ **do**
7:           $avgValue \leftarrow$ Calculate average non-noisy property value $f_x$ in $group$ ▷ Abstraction
8:           **for** $observation \in group$ **do**
9:              Replace noisy value of $f_x$ with $avgValue$ in $observation$ ▷ Substitution
10:               Update $(v_{g_i}^t)'$ based on observation ▷ Update observation
11:           **end for**
12:        **end for**
13:      **end if**
14:    **end for**
15:    **return** $(v_{g_i}^t)'$ ▷ Abstraction-enhanced observation
16: **end function**

as weather conditions, time of day, road infrastructure, and data about traffic flow, including the number of vehicles on each road segment and their respective speeds to further enhance agent's reasoning and enable dynamic adaptation of signal control strategies. To extract the values for newly deduced concepts and properties from available sensor data, the agent can map the sensor data to the newly deduced concepts/properties. For example, if the agent deduces a concept like "TrafficCongestion", it could correlate traffic flow and vehicle speed data to identify congestion.

### 4.2.1.3 Observation Masking

Ontology enables the agent to identify concepts/properties that are relevant to its decision-making process. By employing reasoning capabilities, the agent can infer the relevance or usefulness of observations that might initially appear unrelated. In our proposed method for observation masking [93], the agent compares the similarity of concepts in two ontologies: domain-specific ontology and action ontology. This helps the agent identify the relevant concepts or properties it needs to make decisions. Once identified, the agent can then filter out or disregard observations related to concepts that are not useful or relevant to its decisions. Filtering out irrelevant information enables the agent to identify similarities between its present and past observations, allowing it to leverage previous experiences in addressing the current situation effect-ively, a process known as generalization. During observation masking, similarities of the concept $c_x$ in the ontology-based schema $O_{g_i}^t$ and the concept $c_y$ in the action ontology $O_{a_j}$ are compared. If $\forall c_y \in C_{a_j} : Sim(c_x, c_y) < Similarity - Threshold$, then the agent does not use the observed data of the concept $c_x$ in its decision-making for the action $a_j$ (see Algorithm 3). In experiments, establishing threshold values like $Similarity - Threshold$ involves the following steps: analyze historical data, conduct experiments with various values, and find a balance between sensitivity and stability to achieve the desired agent behavior.

**Algorithm 2** Ontology-Enhanced-Observation-Modeling (continued)
___
**Require:** $O_{d_i}$ - Domain-specific ontology
**Require:** $O_{g_i}^t$ - Ontology-based schema
**Ensure:** $(v_{g_i}^t)'$ - Expanded observation data
 1: **function** OBSERVATION-EXPANSION($O_{d_i}, O_{g_i}^t$)
 2:     **for** newConcept in $O_{d_i}$ **do**         ▷ Conceptual expansion approach
 3:         Add $newConcept$ to $O_{g_i}^t$
 4:     **end for**
 5:     **for** newProperty in $O_{d_i}$ **do**         ▷ Property addition approach
 6:         Associate $newProperty$ with relevant concepts in $O_{g_i}^t$
 7:         Map sensor to the value of $newProperty$     ▷ Sensor data mapping
 8:         Update $(v_{g_i}^t)'$ based on value of $newProperty$     ▷ Update observation
 9:     **end for**
10:     **return** $(v_{g_i}^t)'$         ▷ Expanded observation data
11: **end function**
___

As an example in the OTSC domain, the agent's ontology-based schema $O_{g_i}^t$ includes concepts related to traffic flow, congestion levels, weather conditions, and vehicle types. The action ontology $O_{a_j}$ comprises concepts representing different traffic signal phases, such as "green phase for main street" or "green phase for side street". Suppose the concept $c_x$ represents a "CongestionLevel". It is important for the agent to determine whether or not the concept of "CongestionLevel" is relevant to its decision-making process for the "green phase for main street".

### 4.2.1.4 Observation Augmentation

Utilizing the existing ontological knowledge and reasoning strategies, the agent can derive new knowledge from observed data. This reasoning process enhances the augmentation of observations by inferring missing information or uncovering hidden relationships. In the observation augmentation method [90, 105, 95], the automatic inference mechanism can deduce the augmented observations for properties that have unknown information, i.e., either there is no sensor to observe them, or there is a

**Algorithm 3** Ontology-Enhanced-Observation-Modeling (continued)

**Require:** $O_{g_i}^t$ - Ontology-based schema
**Require:** $O_{a_j}$ - Action ontology
**Require:** Similarity-Threshold
**Ensure:** $(v_{g_i}^t)'$ - Masked observation data

1: **function** OBSERVATION-MASKING($O_{g_i}^t, O_{a_j}$, Similarity-Threshold)
2:    $(v_{g_i}^t)' \leftarrow \emptyset$    ▷ Initialize masked observation data
3:    **for** $c_x \in C_{g_i}^t$ **do**    ▷ Iterate over concepts in ontology-based schema
4:       useConcept $\leftarrow$ False    ▷ Flag to determine if concept is used
5:       **for** $c_y \in C_{a_j}$ **do**    ▷ Iterate over concepts in action ontology
6:          **if** $Sim(c_x, c_y) \geq$ Similarity-Threshold **then**
7:             useConcept $\leftarrow$ True    ▷ Concept is used
8:             **break**
9:          **end if**
10:       **end for**
11:       **if** useConcept **then**
12:          $(v_{g_i}^t)' \leftarrow (v_{g_i}^t)' \cup \{v_{g_i}^t[c_x]\}$    ▷ Add concept's observation to masked data
13:       **end if**
14:    **end for**
15:    **return** $(v_{g_i}^t)'$    ▷ Masked observation data
16: **end function**

fault in the sensors (i.e., partial observation), so the agent has not observed them while they exist in the environment. To do so, using forward chaining over ontology logical rules (i.e., relationship) between $c_x$ and $c_y$ and the explicitly observed information $f_y$ of concept $c_y$, the agent extracts implicit observation data $f_x$ of concept $c_x$ (see Algorithm 4).

**Algorithm 4** Ontology-Enhanced-Observation-Modeling (continued)

**Require:** $L_{g_i}^t$ - Logical rules between concepts
**Require:** $f_y$ - Explicitly observed information of concept $c_y$
**Ensure:** $(v_{g_i}^t)'$ - Augmented observation data

1: **function** OBSERVATION-AUGMENTATION($L_{g_i}^t, f_y$)
2:    **for** each logical rule $l$ in $L_{g_i}^t$ **do**
3:       **if** $l$ expresses relationship between $c_x$ and $c_y$ **then**
4:          **if** $f_y$ is known and satisfies conditions of $l$ **then**
5:             Infer $f_x$ using forward chaining and $f_y$ based on $l$
6:             Update $(v_{g_i}^t)'$ based on observation    ▷ Update observation
7:          **end if**
8:       **end if**
9:    **end for**
10:    **return** $(v_{g_i}^t)'$    ▷ Augmented observation data
11: **end function**

As an example in the ITSC domain, the signal controller agent has sensors that can directly measure the number of vehicles passing through the intersection ($f_y$). However, it lacks a direct sensor for measuring traffic congestion ($f_x$). Using ontological reasoning and logical rules, the agent can infer a relationship between the number of vehicles and traffic congestion. The logical rule states that a higher vehicle count ($f_y$) typically indicates higher traffic congestion ($f_x$). So, even though the agent has not directly observed the congestion levels, it can augment its observations by inferring congestion levels based on vehicle count data.

### 4.2.1.5 Observation Sampling

Samples are typically representative subsets (capturing the essential characteristics) of an agent's observation. They are used when the agent cannot observe an environment completely (i.e., due to complexity or noise) in order to reduce the influence of random fluctuations in data [263, 271]. The size of a sample influences the agent's action selection precision. Assuming that the sample size is constant, the agent needs to allocate the sample size proportional to the importance of the observed data to draw precise conclusions [90, 105]. By incorporating concepts and properties that capture the contextual factors influencing the sampling process (e.g., time, location, environmental conditions), the ontology provides a holistic representation of the context ensuring that the sampling process takes into account relevant contextual factors for generating representative samples. In the proposed observation sampling method, agent $g_i$ samples observation data $v_{g_i}^t$ at different sampling rates proportional to the weight of concepts $c_x \in C_{g_i}^t$ (see Algorithm 5). So, the observation data of concept $c_x$ with higher iweighting indicator $iw(c_x)$ is to be sampled at a higher rate than the observation data of concept $c_y$ with lower importance weight $iw(c_y)$. The sampled observation $v_{g_i}^{t\prime}$ is formed as follows:

$$(v_{g_i}^t)' = \{x_1, \ldots, x_{\lfloor iw(c_x) \times u/n \rfloor}, y_1, \ldots, y_{\lfloor iw(c_y) \times u/n \rfloor}\}$$
$$\text{if } iw(c_x) + iw(c_y) = n \text{ and sample size} = u \quad (4.1)$$

For instance, in the ITSC domain, during rush hours, when congestion becomes a significant concern (time-based importance weight), the observation sampling mechanism focuses on collecting more data related to congestion, including information or measurements that indicate the extent of traffic congestion at specific locations or times. This data could include metrics such as traffic flow rates, vehicle densities, average speeds, or the size of queues at intersections. At other times of the day, when congestion is less of a concern, the weight assigned to congestion-related observations could

**Algorithm 5** Ontology-Enhanced-Observation-Modeling (continued)

**Require:** $v_{g_i}^t$ - Current state observation
**Require:** $C_{g_i}^t$ - Concepts of the agent's observation
**Require:** $\{iw_{g_i}^t\}$ - Importance weights of concepts
**Ensure:** $(v_{g_i}^t)'$ - Sampled observation data
1: **function** OBSERVATION-SAMPLING($v_{g_i}^t, C_{g_i}^t, \{iw_{g_i}^t\}$)
2:     $(v_{g_i}^t)' \leftarrow \{\}$                                        ▷ Initialize sampled observation
3:     $iw_{sum} \leftarrow \sum_{c_x \in C_{g_i}^t} iw(c_x)$       ▷ Sum of importance weights for all concepts
4:     **for** $c_x \in C_{g_i}^t$ **do**                                    ▷ Iterate through concepts
5:        $n_{sample} \leftarrow$ Round(Sample-Size $\times (iw(c_x)/iw_{sum})$)
6:        $\overline{v_{g_i}^t} \leftarrow \overline{v_{g_i}^t} \cup$ Random-Sample($v_{g_i}^t, n_{sample}$)
7:     **end for**
8:     **return** $(v_{g_i}^t)'$                           ▷ Sampled observation based on importance
9: **end function**

be lower, allowing the sampling to be more balanced and reflective of the overall traffic conditions.

### 4.2.2 Ontology-Driven Goal Selection/Generation

The goal is what the agent is trying to achieve, and the reward is the feedback signal that tells the agent how well it is doing with respect to that goal. The reward function is typically defined according to the agent's goals, enabling the computation of the reward signal. The design of a proper reward function can be challenging in partially observable and dynamic environments [268]. The challenge of goal selection/generation in such environments lies in the complexity of identifying and formulating new goals that are relevant to the dynamic conditions of the environment. In such scenarios, the agent's limited observability of the environment and the constant evolution of its surroundings make it difficult to define a comprehensive set of predefined goals.

To improve the goal selection/generation process, the following three methods are proposed. *Automatic goal selection/generation* to generate new goals in unforeseen situations, *adaptive reward definition* to generate various reward functions based on changes in the environment, and *reward augmentation* to augment reward functions with additional parameters extracted from an ontology.

#### 4.2.2.1 Automatic Goal Selection/Generation

Ontology can play a crucial role in incorporating contextual information for goal selection and generation. It achieves this by defining concepts and properties that en-compass contextual factors including user preferences, environmental conditions, and available resources. This contextual framework enables the automatic generation of tailored goals based on the situation. Utilizing logical rules and inference mechanisms within the ontology enables goal reasoning and linking specific conditions or states to goals.

In the proposed automatic goal selection/generation method [91], when unforeseen situations occur, the agent continuously observes its environment, identifies the changes, and has access to a mechanism that enables it to decide whether to continue with its RL process, change its current goal to a pre-defined goal, or generate a new goal. The automatic goal selection/generation method enables the agent to make such decisions in the following steps (see state diagram of automatic goal selection/generation method in Figure 4.2 and algorithmic procedure in Algorithm 6):

---

**Algorithm 6** Ontology-Driven-Goal-Selection/Generation

---
**Require:** $v_{g_i}^t$ - Current state observation
**Require:** $v_{g_i}^{t-1}$ - Previous state observation
**Require:** $P_{g_i}$ - Agent's preferences and constraints
**Require:** $\{iw_{g_i}^t\}$ - Importance weights of concepts
**Require:** $R_{g_i}^t$ - Problem-specific reward function
**Require:** $O_{g_i}^t$ - Ontology-based schema
**Ensure:** Selected action based on new goal selection/generation
1: **function** AUTOMATIC-GOAL-SELECTION/GENERATION($v_{g_i}^{t-1}, v_{g_i}^t, P_{g_i}, \{iw_{g_i}^t\}, R_{g_i}^t, O_{g_i}^t$)
2:　　Compute state distance $D_{g_i}^t$ using Equation 4.2
3:　　Evaluate observation $v_{g_i}^t$ using reward, state distance, and importance weight
4:　　**if** $r_{g_i}^t$ is out of range Discrepancy-Low-Threshold and Discrepancy-High-Threshold OR $D_{g_i}^t$ > State-Distance-Threshold OR Concept $c_x$ with high $iw(c_x)$ appears OR $iw_{g_i}^t$ > Importance-Weight-Threshold **then**
5:　　　　Change current goal to a predefined goal OR Generate a new goal based on contextual information
6:　　**end if**
7:　　**if** Change in goal is required **then**
8:　　　　Reason about goals using logical rules and forward chaining
9:　　　　Choose a predefined goal based on preferences and importance weight of concepts
10:　　　　Update problem-specific reward function $R_{g_i}^t$
11:　　**else if** New goal needs to be generated **then**
12:　　　　Define state similarity reward function $(R_{g_i}^t)'$ using Equation 4.3
13:　　　　Use backward chaining to maximize $(R_{g_i}^t)'$
14:　　　　Select action based on the combination of $R_{g_i}^t$ and $(R_{g_i}^t)'$
15:　　**end if**
16:　　**return** Selected action
17: **end function**

---

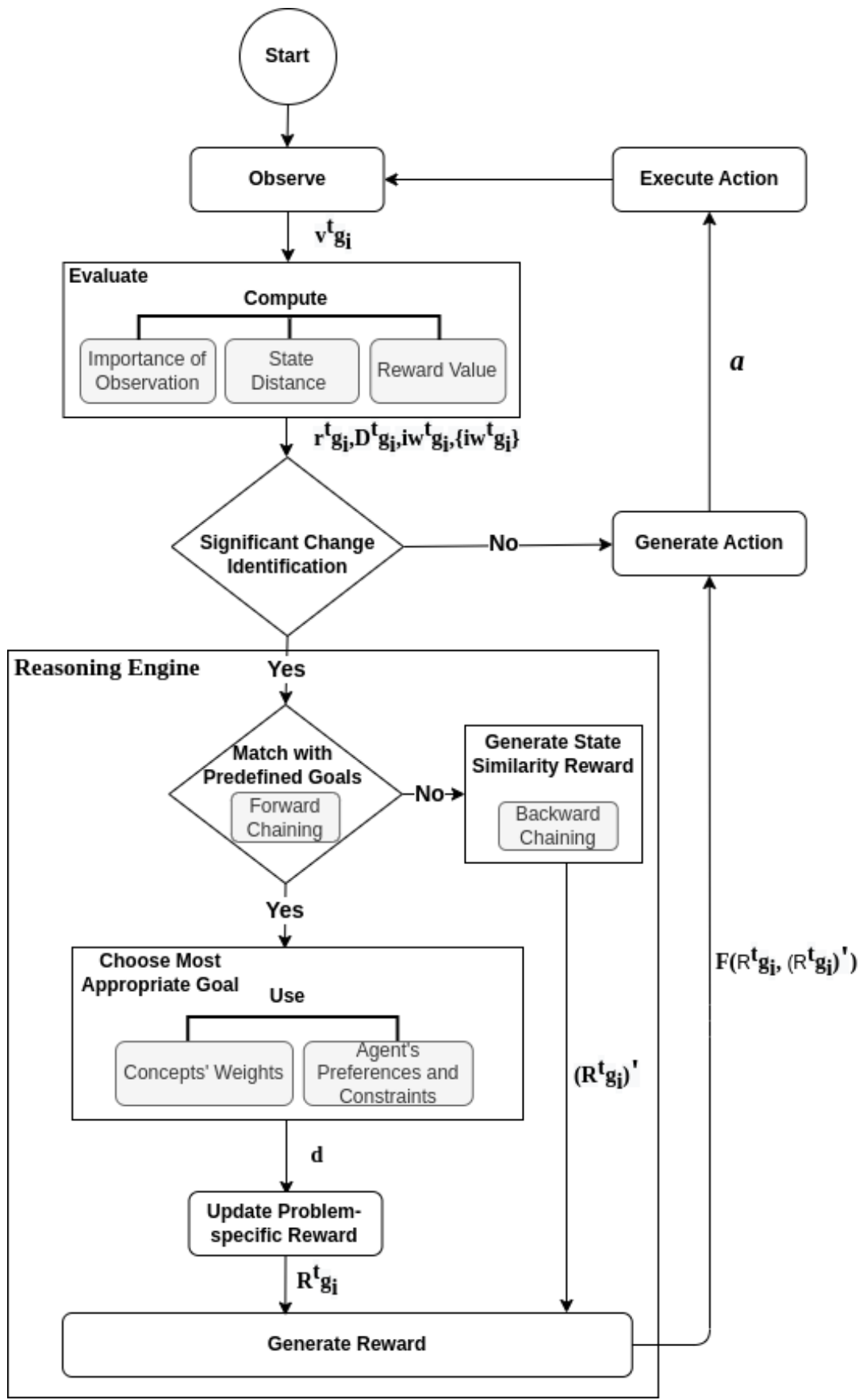

Figure 4.2: The state diagram of the automatic goal selection/generation method.

In the **evaluate** step, agent $g_i$ evaluates its observation using the received reward, the state distance, and the importance of the observation. The agent computes the state distance $D_{g_i}^t$ as the absolute difference between the current state $v_{g_i}^t$ and the previous state $v_{g_i}^{t-1}$ (see Equation 4.2). To do so, I define $V_{v_{g_i}^t}$ to be a quantifying value that describes $v_{g_i}^t$. The importance of the observation $w_{g_i}^t$ is determined by summation of the importance weight of the associated concepts $C_{g_i}^t$. In the **significant change identification** step, based on the output from the evaluate step, the agent decides whether there is a need to change its current goal or create a new one, following the below cases:

- **Case 1.** When $r_{g_i}^t$ is out of the predefined range ($r_{g_i}^t < Discrepancy - Low - Threshold$ or $r_{g_i}^t > Discrepancy - High - Threshold$). $Discrepancy - Low - Threshold$ and $Discrepancy - High - Threshold$ define the minimum and maximum values of the reward that can be received.

- **Case 2.** When $D_{g_i}^t$ is bigger than the predefined $State - Distance - Threshold$ which is the maximum difference that can exist between two successive environment states.

- **Case 3.** When a concept $c_x$ with a high importance weight $iw(c_x)$ appears in an environment or the importance of observation $iw_{g_i}^t$ becomes higher than a predefined $Importance - Weight - Threshold$ which is the maximum importance weight that has been experienced for the current goal.

$$D_{g_i}^t = |V_{v_{g_i}^{t-1}} - V_{v_{g_i}^t}| \qquad (4.2)$$

In the **reasoning** step, using the reasoning engine, agent $g_i$ continuously reasons about the goals it is pursuing, when it is required to change or generate a new goal, two cases are possible:

- **Choosing a predefined goal.** An agent uses logical rules to deduce a predefined goal from a goal-set through forward chaining. The goal set specifies tuples of $(v_{g_i}^t, d)$ where $d$ is a goal/desire that can be adopted when observation $v_{g_i}^t$ is observed (*IF ($v_{g_i}^t$) THEN (d)*). When more than one goal in the goal-set is matched with $v_{g_i}^t$, the agent's preferences and constraints $P_{g_i}$ and/or the concepts' weights $\{iw_{g_i}^t\}$ (i.e., the weights associated with all concepts within the agent's ontology-based schema) will be used as decision criteria. Finally, the problem-specific reward function $R_{g_i}^t$ is updated with the selected goal $d$. For example, consider the two following alternative rules in the goal set:

    ▶ **rule 1:** $c_1 = k_1, c_2 = k_2, ... c_5 = k_5, ..., c_n = k_n -> d_1$

▶ rule 2: $c_1 = k_1, c_2 = k_2, ...c_6 = k_6, ..., c_n = k_n -> d_2$

$c_i$ is a concept and $k_i$ is its value. The agent's state $v_{g_i}^t$ is matched with both rules. If iweighting indicator $iw(c_6)$ is higher than iweighting indicator $iw(c_5)$ then $d_2$ will be selected.

- **Creating a new goal.** When agent $g_i$ cannot find a suitable goal (meaning no goal within the goal-set aligns with the agent's current state observation), the agent formulates a state similarity reward function denoted as $(R_{g_i}^t)'$. This function is designed as the inverse difference between $v_{g_i}^t$ and $v_{g_i}^{t-1}$. Reducing the difference between $v_{g_i}^t$ and $v_{g_i}^{t-1}$ leads to an increase in the state similarity reward (see Equation 4.3). Agent $g_i$ uses backward chaining to maximize $(R_{g_i}^t)'$.

$$(R_{g_i}^t)' = 1/D_{g_i}^t \tag{4.3}$$

Finally, agent $g_i$ selects an action based on the recommendation of the function that combines the two reward functions $R_{g_i}^t$ and $(R_{g_i}^t)'$. Depending on the problem, various combinations can be defined for these two reward functions by domain experts. For instance, the agent $g_i$ can prioritize maximizing the state similarity reward $(R_{g_i}^t)'$ over the problem-specific reward $R_{g_i}^t$ optimization and take actions that can contribute to experiencing its previous known state (see Figure 4.3). To do so, agent $g_i$ selects actions that minimize the difference between $v_{g_i}^t$ and $v_{g_i}^{t-1}$.

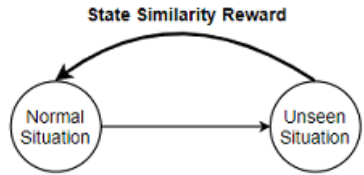

Figure 4.3: Handling an unforeseen situation in the automatic goal selection/generation method, the functionality of the state similarity reward.

In the ITSC domain, when unexpected changes in traffic conditions occur, the agent identifies disparities in the current state compared to the previous one. In response, the agent employs a state similarity reward function that incentivizes actions leading to a traffic state akin to the earlier condition. By minimizing the divergence between the two states, the agent dynamically adjusts signal timings to mitigate congestion, swiftly adapting to unforeseen events and enhancing overall traffic flow efficiency at the intersection.

### 4.2.2.2 Adaptive Reward Definition

Ontology provides the necessary context for adaptive reward definition by defining concepts and properties that capture the contextual factors influencing reward definition, such as environmental conditions, user preferences, or task-specific constraints. This allows for rewards to be dynamically adjusted based on the specific situation or requirements.

Reward Machines (RMs) are defined in terms of propositional symbols $E$ that denote features or events of the concrete state of the environment. Intuitively, an RM represents what reward function should currently be used to provide the reward signal, given the sequence of propositional symbols that the agent has seen so far. The utilization of ontology empowers agents to extract new propositional symbols and novel reward functions [94], enabling the dynamic creation or modification of RMs. This adaptability is suitable for agents' partial observations operating in a continuously evolving environment. Adaptive reward definition method includes four steps: (1) propositional symbol extraction, (2) belief and constraints deducing, (3) reward function extraction, and (4) learning. The details of the steps for this process are explained as follows (see state diagram of adaptive reward definition method in Figure 4.4 and the algorithmic procedure in Algorithm 7).

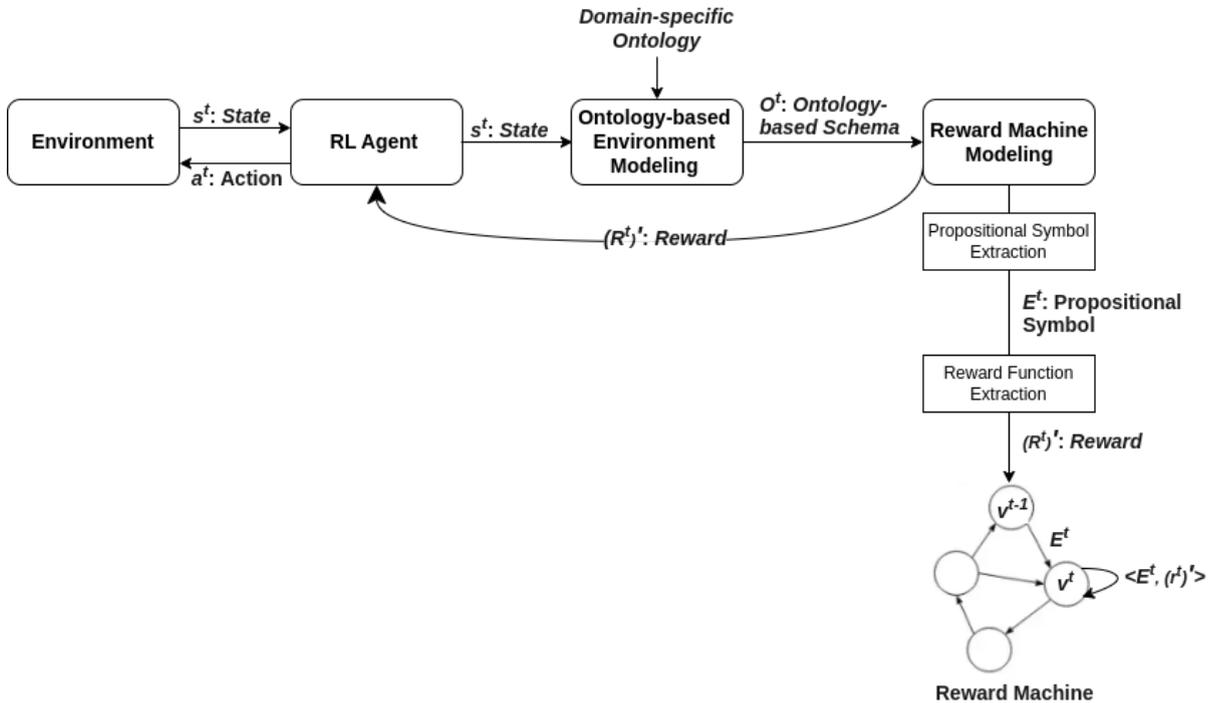

Figure 4.4: The state diagram of the adaptive reward definition method.

In **propositional symbol extraction** step, based on the concepts and properties of the agent's state at time step $t$, the agent $g_i$ extracts a relevant subsumer, then labels its observation by this subsumer as a new propositional symbol/event $E_{g_i}^t$. Subsequently, the agent creates a new transition to a new state in its RM based on the new propositional symbol. In the **belief and constraints deducing** step, through logical reasoning, the agent can derive belief values and allowable values for a specific property $f_x$ based on the relationships it has with other concepts or properties in the ontology. The result is a set of beliefs values (maximize/minimize) shown as $B^t_{g_i} = \{b_{f_x}, b_{f_y}, \ldots, b_{f_n}\}$ for different properties of the concepts observed by the agent at time step $t$ and a set of equations that define the constraints $H(f_x)$ for each property. In the **reward function extraction** step, a set of reward functions $(R^t_{g_i})' = \{R_{f_x}, R_{f_y}, \ldots, R_{f_n}\}$ is defined based on the properties with negative or positive beliefs and their constraints in the domain-specific ontology. To do so, the agent aims to maximize/minimize the value of properties associated with positive/negative beliefs subjecting to the relevant constraints. The constraints are written in the form of inequality constraints, where each constraint should be less than or equal to zero (see Equation 4.4).

**Algorithm 7** Ontology-Driven-Goal-Selection/Generation (continued)
___
**Require:** $O_{g_i}^t$ - Ontology-based schema
**Require:** $O_{d_i}$ - Domain-specific ontology
**Ensure:** $(R_{g_i}^t)'$ - Adaptive reward function
**Ensure:** $RM'$ - New reward machine
 1: **function** ADAPTIVE-REWARD-DEFINITION($O_{g_i}^t$)
 2:     $E_{g_i}^t \leftarrow$ NEW-PROPOSITIONAL-SYMBOL-EXTRACTION($O_{g_i}^t$)     ▷ Extract new propositional symbol
 3:     $B_{g_i}^t \leftarrow$ BELIEF-DEDUCING-STEP($O_{d_i}$)     ▷ Deduce belief
 4:     $H(f) \leftarrow$ CONSTRAINTS-DEDUCING-STEP($O_{d_i}$)     ▷ Deduce constraints
 5:     $(R_{g_i}^t)' \leftarrow$ NEW-REWARD-FUNCTION-EXTRACTION($B_{g_i}^t, H(f)$)     ▷ Extract new reward function
 6:     **return** $(R_{g_i}^t)'$     ▷ Adaptive reward function
 7: **end function**
 8: **function** NEW-PROPOSITIONAL-SYMBOL-EXTRACTION($O_{g_i}^t$)
 9:     Extract relevant subsumer from concepts and properties of the agent's state at time step $t$
10:     Label observation with extracted subsumer as new propositional symbol/event $E_{g_i}^t$
11:     Create a new transition in RM based on the new propositional symbol $E_{g_i}^t$ and the new reward function $(R_{g_i}^t)'$
12:     **return** $RM'$
13: **end function**
14: **function** BELIEF-DEDUCING-STEP($O_{d_i}$)
15:     Deduce belief values for properties $f \in F_{g_i}^t$ based on relationships with Belief property
16:     **return** $B_{g_i}^t$
17: **end function**
18: **function** CONSTRAINTS-DEDUCING-STEP($O_{d_i}$)
19:     Deduce allowable values for property $f$ based on relationships with other concepts/properties
20:     Formulate constraints $H(f)$ for property $f$ using derived allowable values
21:     **return** $H(f)$
22: **end function**
23: **function** NEW-REWARD-FUNCTION-EXTRACTION($F_{g_i}^t, B_{g_i}^t, H(f) \forall f \in F$)
24:     Define a set of reward functions $(R_{g_i}^t)'$ based on properties $F_{g_i}^t$ and their belief values $B_{g_i}^t$
25:     Optimize reward functions $(R_{g_i}^t)'$ subject to constraints $H(f)$
26:     **return** $(R_{g_i}^t)'$
27: **end function**
___

$$\text{if Belief}(f) = \text{Positive: } a \leftarrow \arg\max_{\forall a \in A} f, H(f) \leq 0 \tag{4.4}$$

$$\text{else if Belief}(f) = \text{Negative: } a \leftarrow \arg\min_{\forall a \in A} f, H(f) \leq 0$$

$$\text{else if Belief}(f) = \text{Neutral: } f \text{ not considered for rewards}$$

The reward function's value becomes meaningful for learning only when the specific actions chosen by the RL agent directly influence this value (i.e., exposing action-reward bindings). During the **learning** step, a separate learner can operate within each state of the RM, while the agent's action selection relies on the policy of the current learner. Also, the agent can use a multi-advisor RL [145] in which there are multi-learners solving the same task using a different focus or reward function. Multi-learners' advice, taking the form of action values, is then communicated to an aggregator responsible for merging the advisors' recommendations into a global policy by voting.

In the context of ITSC, by integrating contextual factors such as traffic flow, environmental conditions, and user preferences into the ontology, the system will be capable of dynamically adjusting reward definitions for traffic signal control. For instance, the system can incorporate real-time traffic data to define rewards that prioritize reducing congestion during peak hours. In the propositional symbol extraction step, the agent could identify unique traffic patterns as new propositional symbols, allowing the agent to respond to specific traffic situations. The belief and constraints deducing step could involve deriving belief values that represent the importance of different traffic parameters, such as minimizing waiting times or maximizing traffic flow. Consequently, the signal controller agent can dynamically adapt to varying traffic conditions by adjusting rewards to achieve specific traffic management goals, like minimizing delays for emergency vehicles during certain hours or maximizing traffic throughput during off-peak times.

### 4.2.2.3 Reward Augmentation

Reward shaping is a technique used in RL that allows a reward function to be modified to provide more frequent feedback on appropriate behaviors and improve the RL agents' action selection [134]. In the proposed reward augmentation method [90], the augmented reward signal $r' = r + \phi$ is proposed to provide more frequent feedback to the system where $r$ is the output of the original reward function $R$, $\phi$ represents the additional reward signal resulting in a shaping function $\Phi$, and $r'$ represents the signal generated by the augmented reward function $R'$ [134]. The proposed reward signal $r_{g_i}^t$ is augmented using the actions' efficiency rate $e$ (see Equation 8). Efficient actions are those that with equal or fewer observations, they can make more changes in the environment when compared to other valid actions in the current state.

---
**Algorithm 8** Ontology-Driven-Goal-Selection/Generation (continued)
---
**Require:** $r_{g_i}^t$ - Original reward signal
**Require:** $X$ - Observation matrix
**Require:** $Y$ - Change matrix
**Require:** $\{iw_{g_i}^t\}$ - Concepts' weights
**Ensure:** $(r_{g_i}^t)'$ - Augmented reward signal
 1: **function** REWARD-AUGMENTATION($r_{g_i}^t, X, Y, \{iw_{g_i}^t\}$)
 2:      $e \leftarrow \frac{\sum_{u=1}^{n} iw(c_u) \times y_{uq}}{\sum_{u=1}^{m} iw(c_u) \times x_{uq}}$      ▷ Calculate efficiency rate
 3:      $(r_{g_i}^t)' \leftarrow r_{g_i}^t + e$      ▷ Augment reward signal
 4:      **return** $(r_{g_i}^t)'$      ▷ Augmented reward signal and efficiency rate
 5: **end function**
---

Suppose we have a set of $k$ actions $a_1, a_2, \ldots, a_k$. Each action changes $n$ items of state while observing $m$ items of state. For example, by observing the waiting time of $m$ vehicles, the signal controller agent decides to switch to a green phase which causes a decrease in the waiting time of $n$ vehicles. Let us take into account an observation matrix $X = [x_{uv}, u = 1, 2, \ldots, m, v = 1, 2, \ldots, k]$ and a change matrix $Y = [y_{uv}, u = 1, 2, \ldots, n, v = 1, 2, \ldots, k]$. The $q^{\text{th}}$ line (i.e., $X_q$ and $Y_q$) of these matrices shows quantified values of the observed concepts before applying action $a_q$ and quantified values of the changed concepts after applying the action, respectively. The concept weights are used to compute the weighted sum of the quantified values, and the efficiency rate ($e$) of the action is expressed as follows:

$$e = \frac{\text{weighted sum of changed concepts' quantified values}}{\text{weighted sum of observed concepts' quantified values}}$$

$$= \frac{\sum_{u=1}^{n} iw(c_u) \times y_{uq}}{\sum_{u=1}^{m} iw(c_u) \times x_{uq}} \quad (4.5)$$

For instance, in the ITSC domain, the signal controller agent observes how many vehicles are waiting in each lane before selecting an action. Once it adjusts the green phase duration, it looks at how the number of waiting vehicles changes in each lane. Some lanes might affect congestion more than others, for instance, a lane with heavy traffic. The efficiency rate ($e$) of the action is calculated based on the extent of reduction in the number of waiting vehicles, particularly in critical lanes when compared to alternative actions available in the current state.

### 4.2.3 Ontology-Driven Action Selection

In dynamic and unpredictable environments, agents occasionally encounter the unavailability of specific actions. So, masking the irrelevant actions is important because it can limit the number of actions to those that are feasible and improve the agent's exploration performance. Moreover, prioritizing actions is important in partially observable and dynamic environments due to the complexities posed by limited information and constantly changing circumstances. In such scenarios, making well-informed decisions becomes challenging, and not all actions carry equal weight. Prioritization ensures that the agent focuses its attention on actions that are more likely to yield favorable outcomes. In the ontology-driven action selection stage, I have proposed four methods including action masking, action exploration, action prioritization, and execution prioritization.

#### 4.2.3.1 Action Masking

The agent's possible or feasible actions $A'$ are a small percentage of the available actions $A$ at each time step. The primary function of the action masking in RL is to filter out impossible actions, which improves its exploration performance by only considering the feasible action set and outputting a better policy [93]. By encoding relevant information about the environment, capabilities, resources, and contextual factors, the ontology establishes a foundation for assessing the feasibility of taking action. This knowledge can include physical limitations, legal restrictions, operational requirements, or any other relevant factors that determine the action's feasibility at a certain point in time. Also, certain actions may have prerequisites, dependencies, or mutually exclusive conditions that can impact their feasibility. By representing these relationships in the ontology, actions that are not compatible with the associated constraints can be filtered. Therefore, the agent's action selection in any state might be constrained by semantic constraints $H_{d_i} = (H^+, H^-)$ imposed by domain-specific ontology. An ontology language (e.g., SWRL) expresses them, specifying both the consequences $H^+$ that must hold in each state and the consequences $H^-$ that cannot hold. Any action that complies with these constraints is considered feasible and any action that violates these restrictions is considered infeasible (see the algorithmic procedure in Algorithm 9). For instance in the ITSC domain, when deciding on potential signal phase change, the agent ensures that it never suggests executing a left turn signal phase and a through movement signal phase for the conflicting directions at the same time. These infeasible combinations are masked, guiding the agent to choose actions that comply with safety regulations and prevent risky vehicle interactions.

---
**Algorithm 9** Ontology-Driven-Action-Selection
---
**Require:** $A$ - Set of all available actions
**Require:** $O_{d_i}$ - Domain-specific ontology
**Ensure:** $A'$ - Set of valid actions
1: **function** ACTION-MASKING($A, O_{d_i}$)
2:     $A' \leftarrow \emptyset$     ▷ Valid action set
3:     **for** $a_i$ in $A$ **do**
4:         **if** $a_i$ satisfies semantic constraints $H_{d_i}$ **then**
5:             Add $a_i$ to $A'$
6:         **end if**
7:     **end for**
8:     **return** $A'$     ▷ Filtered set of valid actions
9: **end function**
---

### 4.2.3.2 Action Exploration

The primary function of action exploration in a large action space is to systematically search for and discover potentially effective actions that can lead to better outcomes [90, 105]. In the **concepts importance ranking** step of the proposed action exploration method, ontology enables the ranking or prioritization of concepts $C_{g_i}^t$ based on their importance $\{w_{g_i}^t\}$ within the domain. By assigning weights, scores, or other measures of importance to concepts, ontology provides a structured representation of their significance. This ranking helps in identifying the most important concepts that should be favored during action exploration. In the **action-concept-reward mapping** step, ontology includes mappings between actions and the associated concept rewards or outcomes. This information helps prioritize action exploration based on the expected rewards for important concepts. In the **adaptive hybrid policy** step, the RL agent dedicates a certain percentage of exploration time to the actions that reward them for the most important concepts of the environment. The automatic inference mechanism deduces the appropriate action based on the logical rules. Agent $g_i$ employs a hybrid policy that combines the proposed ontology-based policy $\pi_{\text{OntoDeM}}$ and an RL policy $\pi_{\text{RL}}$, and will have the flexibility to switch between these two policies (refer to Equation 4.6). The parameter $\alpha(t)$ acts as a percentage function, ensuring a dynamic balance between them, and is computed according to the occurrence pattern of unforeseen events until time step $t$. Suppose unforeseen events occurred every 2 minutes, then $\alpha(t)$ will equal 0.5 (see the algorithmic procedure in Algorithm 10).

$$\pi(a \mid s) = \alpha(t) \times \pi_{\text{OntoDeM}}(a \mid s) + (1 - \alpha(t)) \times \pi_{\text{RL}}(a \mid s) \qquad (4.6)$$

**Example in the Intelligent Traffic Signal Control Domain:** In the context of ITSC, in the concepts importance ranking step, the ontology is used to prioritize concepts such as traffic flow in different lanes, emergency vehicle movement, and congestion levels based on their importance scores. These scores are assigned through ontology-based reasoning, providing a structured representation of their significance. For instance, lanes with high traffic volume might be given higher importance. In the action-concept-reward mapping step, the ontology establishes connections between specific actions and their corresponding outcomes, considering important concepts such as reduced congestion or improved emergency vehicle response times. This mapping helps in guiding the agent toward exploring actions that lead to positive outcomes for critical concepts. In the adaptive hybrid policy step, during periods with frequent congestion events, a higher percentage of exploration time might be allocated to actions that alleviate congestion.

---

**Algorithm 10** Ontology-Driven-Action-Selection (continued)

---

**Require:** $C_{g_i}^t$ - Concepts in the ontology-based schema
**Require:** $\{w_{g_i}^t\}$ - Concepts' weights
**Require:** $A$ - Set of all available actions
**Require:** $\pi_{\text{RL}}$ - RL policy
**Require:** $\pi_{\text{OntoDeM}}$ - OntoDeM policy
**Require:** $\alpha(t)$ - Hybrid policy parameter
**Ensure:** Selected action $a_{\text{selected}}$
1: **function** ACTION-EXPLORATION($C_{g_i}^t, \{w_{g_i}^t\}, \alpha(t), A$)
2: $\quad C_{\text{important}} \leftarrow$ RANK-CONCEPTS($C_{g_i}^t, \{w_{g_i}^t\}$) ▷ Rank concepts by importance
3: $\quad A_{\text{concept\_rewards}} \leftarrow$ MAP-ACTION-CONCEPT-REWARDS($A, C_{g_i}^t$) ▷ Map action-concept rewards
4: $\quad A_{\text{exploration}} \leftarrow$ FILTER-ACTIONS-FOR-EXPLORATION($A, C_{\text{important}}, A_{\text{concept\_rewards}}$) ▷ Filter actions for exploration
5: $\quad a_{\text{selected}} \leftarrow$ SELECT-ACTION($\pi_{\text{OntoDeM}}, \pi_{\text{RL}}, A_{\text{exploration}}, \alpha(t)$) ▷ Select action using hybrid policy
6: $\quad$ **return** $a_{\text{selected}}$ ▷ Selected action for execution
7: **end function**

### 4.2.3.3 Action Prioritization

In action prioritization, the emphasis is on prioritizing a single action over others, using predefined rules (defined by ontology engineers) that specify certain requirements, and ontology to infer whether or not an action satisfies them [93, 105]. To do so, firstly, guided by the RL policy, a set of actions is chosen. Then the agent prioritises the actions that meet the predefined requirements. Ultimately, it proceeds by taking the action with the highest priority (see the algorithmic procedure in Algorithm 11). For instance, during heavy pedestrian hours, the traffic signal controller agent prioritizes actions that ensure safe pedestrian crossings over actions that could lead to increased vehicular flow.

---
**Algorithm 11** Ontology-Driven-Action-Selection (continued)
---
**Require:** $A$ - Set of all available actions
**Require:** $O_{d_i}$ - Domain-specific ontology
**Require:** $\pi_{\text{RL}}$ - RL policy
**Ensure:** Prioritized action $a_{\text{prioritized}}$
1: **function** ACTION-PRIORITIZATION($A, O_{g_i}^t, \pi_{\text{RL}}$)
2:     $A_{\text{selected}} \leftarrow$ ACTION-SELECTION($A, \pi_{\text{RL}}$)           ▷ Action selection by RL policy
3:     $A_{\text{requirements}} \leftarrow$ ASSESS-REQUIREMENTS($A_{\text{selected}}, O_{d_i}$)           ▷ Assess action requirements
4:     $A_{\text{ranked}} \leftarrow$ COMPUTE-ACTION-RANKINGS($A_{\text{requirements}}$)           ▷ Compute action rankings
5:     $a_{\text{prioritized}} \leftarrow$ SELECT-PRIORITY-ACTION($A_{\text{ranked}}$)           ▷ Select priority action
6:     **return** $a_{\text{prioritized}}$           ▷ Prioritized action for execution
7: **end function**
---

### 4.2.3.4 Execution Prioritization

Execution prioritization deals with the scenario where multiple actions are simultaneously scheduled for execution and the most optimal sequence of action execution should be defined. The proposed execution prioritization method [93] involves handling temporal dependencies, adapting to contextual factors, and optimizing the action sequence. In the **temporal dependency handling** step, ontology logical rules can capture precedence relationships between actions by defining rules that specify the order in which certain actions should be executed based on their temporal constraints (e.g., action $a_i$ must be completed before action $a_j$). For instance, if the green phase for a pedestrian crossing ($a_i$) must occur before a left turn signal ($a_j$) to ensure pedestrian safety, the ontology rules ensure that $a_i$ is executed prior to $a_j$. Also, in the **contextual adaptation** step, logical rules take contextual factors that may influence execution prioritization into consideration. Contextual factors refer to the specific conditions, user preferences, constraints, or requirements (e.g., resource availability or time constraints) that may impact the order and the importance of executing certain actions. By incorporating these factors into the execution prioritization process, the agent computes the optimized sequence for actions execution (see the algorithmic procedure in Algorithm 12). For instance, during rush hours, when the main road experiences heavy traffic, the logical rules prioritize extending the green phase duration for the main road ($a_i$) to alleviate congestion and ensure efficient traffic flow on the main route. However, when a major event causes a sudden surge in traffic on the side road, the logical rules may temporarily adjust the execution prioritization to extend the green phase duration for the side road ($a_j$).

---

**Algorithm 12** Ontology-Driven-Action-Selection (continued)
---
**Require:** $A$ - Set of scheduled actions
**Require:** $O_{d_i}$ - Domain-specific ontology
**Ensure:** $A_{\text{optimized}}$ - Optimized action sequence
1: **function** EXECUTION-PRIORITIZATION($A, O_{d_i}$)
2:     $A_{\text{temporal}} \leftarrow$ HANDLE-TEMPORAL-DEPENDENCIES($A, O_{d_i}$)    ▷ Handle temporal dependencies
3:     $A_{\text{contextual}} \leftarrow$ ADAPT-TO-CONTEXT($A, O_{d_i}$)    ▷ Adapt to contextual factors
4:     $A_{\text{optimized}} \leftarrow$ OPTIMIZE-ACTION-SEQUENCE($A_{\text{temporal}}, A_{\text{contextual}}$)    ▷ Optimize action sequence
5:     **return** $A_{\text{optimized}}$    ▷ Optimized action sequence for execution
6: **end function**

---

### 4.2.4 OntoDeM Guidelines

The following flowchart (see Figures 4.5) summarizes the guidelines for incorporating the proposed ontology-based methods to enhance different stages of the RL agents' decision-making process in dynamic and partially observable environments.

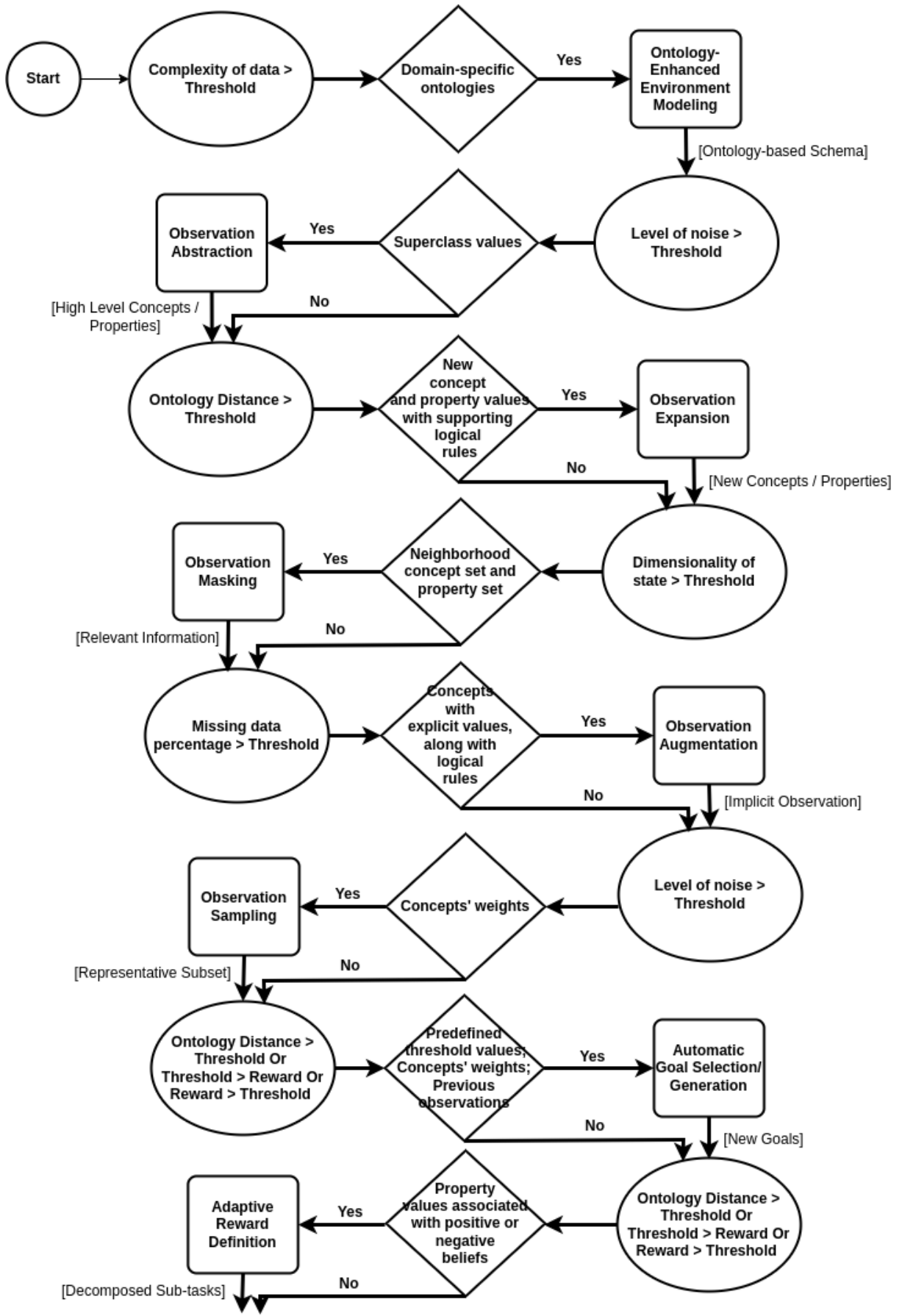

Figure 4.5: Flowchart guiding the use of OntoDeM's methods.

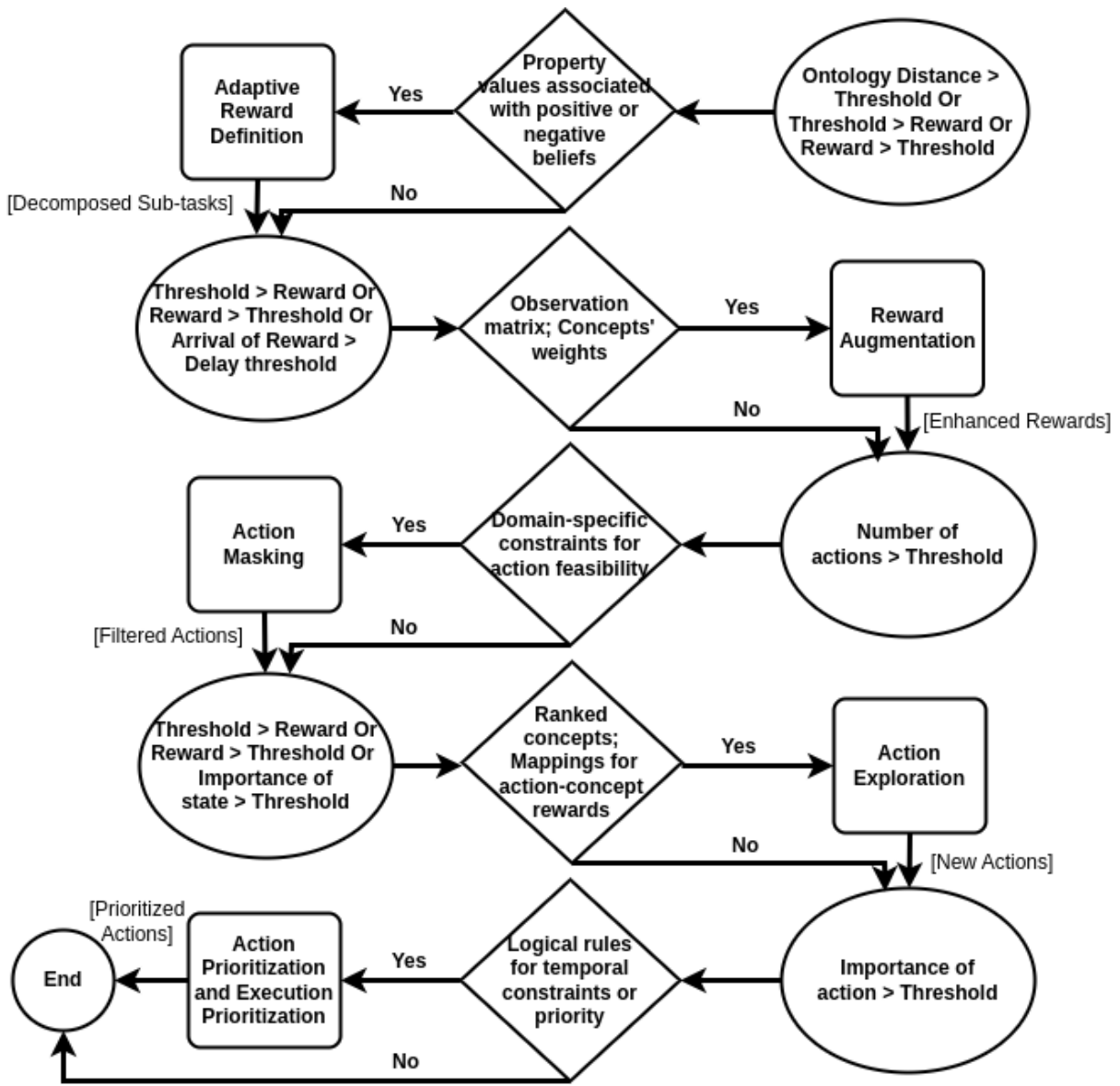

Figure 4.5 (Continued): Flowchart guiding the use of OntoDeM's methods.

The following guidelines are defined based on the triggers and preconditions that must be present for specific methods to be applicable. A trigger outlines a scenario that necessitates certain considerations or the proposed methods, while A precondition specifies the conditions that must be met before those methods can be carried out. A detailed description of each trigger and precondition are provided in the following subsections:

#### 4.2.4.1  Ontology-based Environment Modeling

- Trigger: When the complexity of observed data surpasses a predetermined $Threshold$. The complexity of observed data can be assessed by considering factors such as Dimensionality (count of features), Entropy (randomness within the data), Variability (fluctuations within the data), Interactions (dependencies among variables), and Hierarchy (levels of nesting within the data).

- Precondition: Availability of domain-specific ontologies to map low-level sensor data to high-level concepts, relations, and properties.

#### 4.2.4.2 Ontology-Enhanced Observation Modeling

- *Observation Abstraction*:

    ▶ Trigger: When the level of noise (difference between observed and expected values) in the agent's observations reaches a predetermined $Threshold$.

    ▶ Precondition: Availability of superclass values for concepts with noisy information.

- *Observation Expansion*:

    ▶ Trigger: When the Ontology Distance between two consecutive ontology-based schemas surpasses a predefined $Threshold$.

    ▶ Precondition: Availability of values for properties intended to be added to the observation, along with logical rules that enable the deduction of relationships between concepts of the ontology-based schema and new properties and concepts in the domain ontology.

- *Observation Masking*:

    ▶ Trigger: When the dimensionality of state observations exceeds a predefined $Threshold$.

    ▶ Precondition: Availability of the neighborhood concept set and property set associated with concepts in both the domain-specific ontology and action ontology.

- *Observation Augmentation*:

    ▶ Trigger: When missing data percentage exceeds a predefined $Threshold$.

    ▶ Precondition: Availability of ontology logical rules that allow for the identification of existing relationships between concepts with unknown values and concepts with explicit values.

- *Observation Sampling*:

- ▶ Trigger: When the level of noise in the agent's observations reaches a predetermined $Threshold$.

- ▶ Precondition: Availability of importance weights associated with each concept.

### 4.2.4.3 Ontology-Driven Goal Selection/Generation

- *Automatic Goal Selection/Generation*:

  - ▶ Trigger: When the Ontology Distance between two consecutive ontology-based schemas surpasses a predefined $Threshold$. When the reward signal is out of the predefined range $Discrepancy - Low - Threshold$ and $Discrepancy - High - Threshold$.

  - ▶ Precondition: Availability of predefined threshold values ($Discrepancy - Low - Threshold$ and $Discrepancy - High - Threshold$, $State - Distance - Threshold$, $Importance - Weight - Threshold$), access to concepts' weights, and the ability to compare the current observation with the previous one.

- *Adaptive Reward Definition*:

  - ▶ Trigger: When the Ontology Distance between two consecutive ontology-based schemas surpasses a predefined $Threshold$.

  - ▶ Precondition: Availability of property values associated with positive or negative beliefs.

- *Reward Augmentation*:

  - ▶ Trigger: When reward signal is out of the predefined range $Discrepancy - Low - Threshold$ and $Discrepancy - High - Threshold$. When the arrival of the reward signal exceeds a predefined allowable delay value.

  - ▶ Precondition: Availability of an observation matrix that represents quantified values of observed concepts before and after applying each action.

### 4.2.4.4 Ontology-Driven Action Selection

- *Action Masking*:

  - ▶ Trigger: When the total number of available actions exceeds a predefined $Threshold$, trigger action masking to exclude the actions that have the least compatibility with the predefined constraints encoded in the ontology.

▶ Precondition: Availability of domain-specific constraints (i.e., semantic constraints) that define the feasibility of actions, such as physical limitations, legal restrictions, operational requirements, prerequisites, dependencies, or mutually exclusive conditions.

- *Action Exploration*:

   ▶ Trigger: When reward signal is out of the predefined range $Discrepancy - Low - Threshold$ and $Discrepancy - High - Threshold$. When the importance of the state observation becomes higher than a predefined $Importance - Weight - Threshold$.

   ▶ Precondition: Availability of rankings or prioritization of concepts based on their importance within the domain and access to the mappings between actions and the associated concept rewards through logical rules.

- *Action Prioritization and Execution Prioritization*:

   ▶ Trigger: When the importance of actions exceeds a predefined $Threshold$. The importance of actions is determined by summation of the weights of importance associated with the concepts involved.

   ▶ Precondition: Availability of logical rules to capture precedence relationships between actions based on their temporal constraints or priority.

### 4.2.5 Computational Complexity Analysis

In this section, I analyze the time complexity of OntoDeM considering the operations performed at each stage of the decision-making process.

- Ontology-based Environment Modeling:

  **Ontology Development:** Designing the ontology involves identifying crucial concepts, relationships, and properties representing the domain knowledge. The complexity here depends on the size and complexity of the domain, the number of concepts and relationships to be modeled, and the structuring of the ontology into modular components.

  1. Hierarchy Creation: In this step, the focus is on organizing concepts into a structured taxonomy or hierarchy. The hierarchy defines the parent-child relationships between concepts, indicating which concepts are more general (parent) and which are more specific (child). The complexity of hierarchy creation depends on the depth ($h$) and breadth ($b$) of the hierarchy.

The depth represents the maximum number of levels in the hierarchy, while the breadth represents the maximum number of child concepts for any given parent concept. The complexity of hierarchy creation is expressed as $O(h \times b)$.

2. Defining Relationships: This step involves specifying how concepts are related to each other beyond the hierarchical structure. While hierarchy creation establishes parent-child relationships, defining relationships focuses on capturing other types of connections between concepts, such as association, composition, or dependency. The complexity of defining relationships ($r$) depends on the number of such relationships to be defined.

3. Assigning Properties: Assigning properties to concepts involves specifying attributes or characteristics associated with each concept. The complexity increases with the number of properties ($p$) to be assigned.

Overall, the computational complexity of ontology design can be approximated as the sum of the complexities related to the hierarchy creation, defining relationships, and assigning properties:

$$O(h \times b) + O(r) + O(p)$$

**Action Ontology Modeling:** The computational complexity of action ontology modeling depends on several factors, including the number of action categories, the complexity of creating the hierarchy of concepts, and the process of defining relationships and assigning properties across all action categories.

Let's denote:

- $k$ as the total number of action categories in the action ontology.

- $h$ as the maximum depth of the hierarchy for organizing concepts.

- $b$ as the maximum branching factor, representing the average number of child concepts for any given parent concept.

- $r$ as the total number of relationships to be defined across all action categories.

- $p$ as the total number of properties to be assigned across all action categories.

Then, the computational complexity of action ontology modeling can be approximated as:

$$O(k \times (h \times b + r + p))$$

This expression considers the overhead associated with creating and managing the hierarchy of concepts, defining relationships, and assigning properties across all action categories in the ontology.

**Sensor Data Mapping:** The computational complexity of the sensor data mapping method depends on the size of the sensor data streams and the complexity of the mapping process.

Let's denote the following:

- $k$ as the total number of sensor data streams.

- $m$ as the average size of each sensor data stream.

- $n$ as the number of high-level concepts to which the sensor data is mapped.

- $p$ as the average number of properties associated with each high-level concept.

Then, the computational complexity of the sensor data mapping method can be approximated as:

$$O(k \times (m + n \times p))$$

So, $m + n \times p$ represents the total operations required to map each sensor data stream to high-level concepts. The $m$ term represents the operations involved in processing each sensor data stream, and $n \times p$ represents the operations involved in mapping each data stream to $n$ high-level concepts, each with $p$ associated properties. Multiplying this by $k$, the total number of sensor data streams, gives the overall computational complexity of the sensor data mapping method.

**Measuring Concept Similarity in Different Ontologies:** By representing the number of concepts in the ontology-based schema $O_{g_i}^t$ as $n_{g_i}$ and the number of concepts in the action ontology $O_{a_j}$ as $n_{a_j}$, calculating the similarity between each concept in $O_{g_i}^t$ and each concept in $O_{a_j}$ has a time complexity of $O(n_{g_i} \times n_{a_j})$.

**Measuring Ontology Distance:** The computational complexity of computing the Jaccard Similarity (JS) or Jaccard Distance (JD) between two sets $A$ and $B$ depends on the sizes of the sets and the operations required to calculate the intersection and union of the sets.

Let:

- $|A|$ be the size of set $A$,

- $|B|$ be the size of set $B$.

Computing the intersection $|A \cap B|$ typically requires iterating over the elements of one set and checking for membership in the other set. Similarly, computing the union $|A \cup B|$ involves combining the elements of both sets while ensuring that each element appears only once.

The computational complexity of calculating the intersection and union of two sets is often linear with respect to the size of the sets. Therefore, the computational complexity of computing the Jaccard Similarity (JS) or Jaccard Distance (JD) can be expressed as:

$$O(|A| + |B|)$$

**Subsumer Extraction:** The computational complexity of subsumer extraction depends on the size and structure of the ontology, as well as the algorithm or method used for extraction.

Let's denote:

- $n_t$ as the total number of concepts in the ontology,

- $r$ as the average number of relationships (properties) associated with each concept,

- $n_s$ as the average number of concepts in the subsumee set for each subsumer concept.

If we use an algorithm that traverses the ontology and performs comparisons for each concept and its associated relationships, the computational complexity can be approximated as:

$$O(n_t \times (r + n_s))$$

This expression accounts for the overhead associated with processing each concept and its relationships, as well as the complexity of comparing or analyzing the subsumer-subsumee relationships.

**Rule-Based Inference Strategies:** The time complexity of forward chaining is linear, denoted as $O(l)$, where $l$ represents the number of logical rules in the knowledge base. Backward chaining can be more efficient than forward chaining, especially when dealing with complex knowledge bases. Since backward chaining starts from the goal and progresses in reverse, it can skip over rules that are unrelated to the goal, thus potentially reducing the overall computational overhead.

**Assessment of Observation Significance:** The computational complexity of assessing the significance of observations based on the iweighting indicator involves several steps, including calculating importance weights and aggregating them. I break down the complexity as follows:

1. Calculation of Importance Weights: For each concept $c_x \in C_{g_i}^t$, the importance weight $iw(c_x)$ is calculated based on the average importance weights of its relationships. Let $(n)$ represent the number of concepts and $(r)$ represent the average number of relationships per concept. Computing the importance weight for each concept involves iterating over its relationships and calculating the average, resulting in a time complexity of $O(r \times n)$.

2. Aggregation of Importance Weights: After calculating the importance weights for each concept, they need to be aggregated to determine the overall significance of the observation. This step typically involves simple arithmetic operations like addition, which can be done in constant time. Therefore, the complexity of this step is negligible.

Considering these steps, the overall computational complexity of assessing observation significance based on the iweighting indicator can be approximated as $O(n \times r)$, where $n$ is the number of concepts in the ontology-based schema and $r$ is the average number of relationships per concept.

- Ontology-Enhanced Observation Modeling:

  **Observation Abstraction:** The computational complexity of the observation abstraction method can be analyzed based on the operations performed in the following:

  1. Identifying Noisy Property Values: This step involves comparing observed values with expected values for each property. The computational complexity of this step depends on the number of properties and the size of the observed data. If there are $(p)$ properties and $(k_o)$ observed data points, this step would have a time complexity of $O(p \times k_o)$.

  2. Classifying Past Observations: Classifying past observations based on abstract concepts involves grouping observations based on higher-level categories. The computational complexity of this step depends on the number of abstract concepts and the number of past observations. If there are $(n)$ abstract concepts and $(k_p)$ past observations, this step would have a time complexity of $O(n \times k_p)$.

  3. Substituting Noisy Property Values: After identifying the noisy property values, the agent substitutes them with the average of non-noisy values from previous observations in the same class. To perform this operation, the agent needs to: Identify the class to which the noisy property value belongs. This involves associating the property value with a specific abstract concept category, as explained earlier. Access and average the non-noisy values from previous observations within the same class. The time complexity of averaging the non-noisy values would depend on the number of previous observations ($k_p$) for each property within the same class and the number of properties ($p$) involved in the averaging process. Therefore, the time complexity of this step would be $O(k_p \times p)$.

Overall, the observation abstraction method involves multiple steps, each with its own computational complexity. The total complexity would be the sum of the complexities of these individual steps. Assuming that the number of properties, observed data points, previous observations, and abstract concepts are all within reasonable bounds, the overall computational complexity of the observation abstraction method would be polynomial, likely $O(p \times k_o + n \times k_p + k_p \times p)$, where $p$, $n$, $k_p$, and $k_o$ are as defined above.

**Observation Expansion:** The computational complexity of the observation expansion method can be analyzed as follows:

1. Conceptual Expansion Approach: The computational complexity of this phase depends on the number of newly deduced concepts ($n_c$) and the process of adding these concepts to the ontology-based schema. If the ontology hierarchy is large and finding the appropriate place for a new concept involves traversing through the ontology tree to identify the suitable parent concept or superclass, the time complexity would depend on the depth and breadth of the ontology hierarchy.

    Let us take into account the following:

    - $h$ as the depth of the ontology hierarchy (i.e., the maximum distance from the root concept to any leaf concept).

    - $b$ as the average branching factor (i.e., the average number of child concepts for each parent concept).

    In this scenario, the time complexity of finding the appropriate place for a new concept would depend on how efficiently the search algorithm can navigate through the ontology hierarchy to locate the suitable parent concept. Assuming a simple traversal algorithm such as Depth First Search (DFS) or Breadth First Search (BFS) is used, the time complexity would be determined by the worst-case scenario:

    - In BFS, each node is visited once, and the algorithm explores nodes level by level. The maximum number of nodes expanded at each level is equal to the average branching factor $b$. Therefore, the time complexity of BFS is

$O(b^h)$, where $b$ is the average branching factor and $h$ is the maximum depth of the ontology hierarchy.

- In DFS, the algorithm explores as far as possible along each branch before backtracking. The time complexity of DFS depends on the size of the search tree. In the worst-case scenario, DFS may visit all nodes before finding the solution, resulting in a time complexity of $O(b^h)$, similar to BFS.

If the ontology hierarchy is large, with considerable depth and breadth, finding the appropriate place for a new concept could potentially have a significant time complexity, especially if the search algorithm needs to traverse through many levels of the hierarchy. However, if the ontology hierarchy is well-structured with efficient search mechanisms (e.g., indexing, caching), the time complexity may be reduced.

2. Property Addition Approach: In this approach, newly deduced properties are associated with the relevant concepts within the ontology-based schema. Similarly, the computational complexity of this phase depends on the number of newly deduced properties ($p$) and the process of associating these properties with concepts. Assuming that associating a single property with a concept has a constant time complexity, the time complexity of associating $p$ properties would be $O(p)$.

Combining both approaches, the overall computational complexity of the observation expansion method would be the sum of the complexities of the conceptual expansion approach $O(b^h)$ and the property addition approach $O(p)$:

Total Complexity = $O(b^h) + O(p)$

However, since both phases involve iterating through the newly deduced concepts and properties and performing operations such as adding them to the ontology-based schema or associating them with concepts, we can simplify the total complexity to the maximum of the two complexities:

Total Complexity = $max(O(b^h), O(p))$

**Observation Masking:** The computational complexity of the observation masking method primarily depends on the process of comparing the similarity of concepts in two ontologies: the domain-specific ontology and the action ontology that has a time complexity of $O(n_{g_i} \times n_{a_j})$ (i.e., by representing the number of concepts in the ontology-based schema as $n_{g_i}$ and the number of concepts in the action ontology as $n_{a_j}$).

**Observation Augmentation:** The computational complexity of the observation augmentation method can be analyzed based on the steps involved in the following algorithm:

1. Forward Chaining over Ontology Logical Rules: The method involves applying forward chaining over ontology logical rules to deduce augmented observations for properties with unknown information.

   - Let $p$ denote the number of properties with unknown information in the ontology-based schema.

   - Let $l$ denote the average number of applicable logical rules for each property.

   Applying forward chaining over logical rules for each property $f_x$ has a time complexity of $O(l)$ per property.

2. Applying Forward Chaining to Multiple Properties: Since there are $p$ properties with unknown information, the method applies forward chaining to each of these properties.

Therefore, the overall time complexity for the observation augmentation method is $O(p \times l)$.

**Observation Sampling:** The computational complexity of the observation sampling method can be analyzed based on the steps involved in the algorithm provided:

1. Determining Sampling Rates Proportional to Concept Weights: The method involves determining sampling rates for each concept proportional to its importance weight indicator.

   - Let $n_t$ denote the number of concepts in the ontology-based schema.

   - Let $k$ denote the number of observations to be sampled.

   - Let $n_o$ denote the average number of concepts associated with each observation.

   Determining the sampling rates for each concept has a time complexity of $O(n_t)$ since it requires iterating over all concepts in the ontology-based schema.

2. Sampling Observations Based on Sampling Rates: Once the sampling rates are determined, the method samples observations based on these rates. Sampling $k$ observations based on the determined rates has a time complexity of $O(k)$.

Therefore, the overall time complexity for this method is $O(n_t + k)$.

**Automatic Goal Selection/Generation:** The computational complexity of the AGGM method involves several steps, each with its own complexity considerations:

1. Evaluate Step: Calculating the state distance $D_{g_i}^t$ involves comparing the current state $v_{g_i}^t$ with the previous state $v_{g_i}^{t-1}$. Let $d$ represent the dimensionality of the state vector. The time complexity for computing the state distance is $O(d)$. Computing the importance of the observation $w_{g_i}^t$ involves summing the importance weights of associated concepts. Let $n$ represent the number of concepts. The time complexity for this step is $O(n)$.

2. Significant Change Identification Step: This step involves checking conditions such as reward range, state distance, and concept importance weights. The time complexity for this step is $O(1)$.

3. Reasoning Step: For choosing a predefined goal, the agent needs to apply logical rules to deduce the appropriate goal. Let $l$ represent the number of rules. The time complexity for this step is $O(l)$. For creating a new goal, the agent uses backward chaining to maximize the state similarity reward. Let $k$ represent the number of steps in the backward chaining process. The time complexity for this step is $O(k)$.

4. Action Selection Step: In the action selection step, the agent evaluates the available actions based on the combined reward functions $R_{g_i}^t$ and $(R_{g_i}^t)'$ to select the most suitable action. The complexity of this step depends on the number of actions $a$ in the action space and the complexity of the function that combines the two reward functions $s$. If the decision-making process involves evaluating each action individually and comparing them based on their associated rewards, the time complexity is $O(s \times a)$.

Summing up the complexities of each step, the overall computational complexity of the AGGM method can be expressed as:

$$O(d + n + 1 + l + k + s \times a)$$

This represents the combined complexity of all steps involved in the AGGM method. Each term in the sum contributes to the overall computational complexity, with $s \times a$ being the most significant factor in the action selection step, as it depends on the number of actions $a$ and the complexity $s$ of the decision-making process.

**Adaptive Reward Definition:** To determine the computational complexity of the Adaptive Reward Definition method, I will analyze each step and consider the complexity of the operations involved:

1. Propositional Symbol Extraction Step: The complexity of subsumer extraction step depends on the number of concepts and relationships that need to be compared. The complexity could be $O(n^2)$ in the worst case, where $n$ is the number of concepts.

2. Belief and Constraints Deducing: This step involves logical reasoning to derive belief values and constraints for each property, and if there are $p$ properties for which reward functions are defined, then the overall time complexity should indeed be $O(p \times l)$, where $l$ represents the complexity of the logical reasoning process for each property.

Overall, the computational complexity of the Adaptive Reward Definition method can be expressed as the sum of the complexities of each step:

$$O(n^2 + p \times l)$$

**Reward Augmentation:** To analyze the computational complexity of the reward augmentation method, I break down the steps involved:

1. Observation Matrix Construction: Constructing the observation matrix involves recording the observed state before taking each action. Let $k_o$ be the number of observed items of state, and $a$ be the number of actions. Constructing the observation matrix has a time complexity of $O(k_o \times a)$.

2. Change Matrix Construction: Constructing the change matrix involves recording the changes in the state after taking each action. Let $k_c$ be the number of items of state changed by each action. Constructing the change matrix has a time complexity of $O(k_c \times a)$.

3. Efficiency Rate Computation: Computing the efficiency rate for each action involves calculating the weighted sum of the quantified values in the observation and change matrices. The time complexity of computing the efficiency rate for each action is $O(k_o + k_c)$. Since this computation is performed for each of the $a$ actions, the total complexity for this step is $O(a \times (k_o + k_c))$.

4. Reward Augmentation: Augmenting the reward signal involves adding the efficiency rate to the original reward signal for each action. Since this operation is performed for each of the $a$ actions, the time complexity for this step is $O(a)$.

Overall, the computational complexity of the reward augmentation method can be approximated as the sum of the complexities of the individual steps:

$$O(k_o \times a) + O(k_c \times a) + O(a \times (k_o + k_c)) + O(a)$$

If the number of actions is large, then the efficiency rate computation step could dominate the overall complexity. Conversely, if the number of observed and changed items is large, then the construction of the observation and change matrices could dominate.

**Action Masking:** The computational complexity of the action masking method depends on the number of actions and the complexity of evaluating the semantic constraints imposed by the domain-specific ontology. I break down the complexity as follows:

1. Semantic Constraint Evaluation: The complexity of evaluating the semantic constraints $H_{d_i} = (H^+, H^-)$ imposed by the domain-specific ontology depends on the number of constraints and the complexity of each constraint. Let $c$ represent the number of semantic constraints. The time complexity for evaluating these constraints is $O(c)$.

2. Feasibility Check for Each Action: For each action, the agent needs to check whether it complies with the semantic constraints. Let $a$ represent the number of actions. The time complexity for checking the feasibility of each action is $O(1)$ if the number of constraints is constant for each action.

Considering these factors, the total computational complexity of the action masking method can be expressed as:

$$O(a \times c)$$

**Action Exploration:** The computational complexity of the action exploration method involves several steps, each contributing to the overall complexity:

1. Concepts Importance Ranking Step: The computational complexity of the concepts importance method depends on the number of concepts $n$ and the complexity of computing the importance weights for each concept. Let $r$ represent the average number of relationships per concept. Computing the importance weight for each concept has a time complexity of $O(r \times n)$. The complexity of ranking or prioritizing concepts based on their importance depends on the number of concepts and the method used for ranking. If the ranking process involves sorting that scale with the number of concepts, the complexity is $O(n \log n)$ or $O(n)$ if a linear-time ranking algorithm is used.

2. Action-Concept-Reward Mapping Step: Establishing mappings between actions and associated concept rewards or outcomes has a complexity that depends on the number of actions and the number of concepts. Let $a$ represent the number of actions. If each action is mapped to rewards for multiple concepts, and vice versa, the complexity is $O(a \times n)$.

Considering these factors, the overall computational complexity of the action exploration method can be expressed as:

$$O(r \times n + n \log n + a \times n)$$

**Action Prioritization:** The computational complexity of the action prioritization method depends on the number of actions and the complexity of evaluating each action against the predefined rules.

Let $a$ represent the total number of candidate actions selected based on RL policy. Evaluating each candidate action against the predefined rules to determine its priority involves checking whether the action satisfies the requirements specified by the ontology. Let $l$ represent the number of predefined rules. The time complexity for evaluating all candidate actions against the rules is:

$$O(a \times l)$$

**Execution Prioritization:** The computational complexity of the execution prioritization method depends on the number of actions and the complexity of evaluating temporal dependencies and contextual factors.

Let $k$ represent the total number of actions available for execution.

1. Temporal Dependency Handling: Evaluating temporal dependencies involves checking the logical rules that define precedence relationships between actions. Let $l_t$ represent the number of logical rules related to temporal dependencies. The time complexity for handling temporal dependencies is $O(l_t)$.

2. Contextual Adaptation: Considering contextual factors for execution prioritization also involves evaluating logical rules. Let $l_c$ represent the number of logical rules related to contextual adaptation. The time complexity for handling contextual factors is $O(l_c)$.

So, the overall computational complexity of the execution prioritization method should be:

$$O(a \times (l_t + l_c))$$

This accounts for evaluating both temporal dependencies and contextual factors for each of the $a$ actions.

## 4.3 Summary

This chapter describes the contribution of the thesis, OntoDeM in the following stages of the agent's decision-making process: In the *ontology-enhanced observation modeling* stage, five methods are introduced to enhance an agent's observations for improved decision-making in partially observable and dynamic environments: observation abstraction, expansion, masking, augmentation, and sampling. These methods involve filtering irrelevant observations, inferring missing/excess data, and generating weighted samples. The ontology-driven goal selection/generation stage involves three methods for improving goal selection and reward definition: Automatic goal selection/generation, which uses ontological reasoning to adapt goals; adaptive reward definition, which dynamically adjusts rewards based on contextual factors; and reward augmentation, which enhances rewards with an efficiency rate to prioritize effective actions. The ontology-driven action selection stage introduces four methods for enhancing action selection in dynamic environments: Action masking, which filters out infeasible actions based on semantic constraints defined by the ontology; action exploration, which employs concepts weight to explore actions favoring more important concepts and contextual adapt exploration strategies; action prioritization, which uses ontology-defined constraints to rank and select actions; and execution prioritization, which leverages ontology logical rules to optimize the sequence of actions based on temporal dependencies and contextual factors. Finally, I delineate the triggers and prerequisites for employing ontology-driven methods in the various stages of RL agent decision-making.

# CHAPTER

# FIVE

# CASE STUDIES

This chapter describes several case studies in four different application areas that are used to evaluate OntoDeM in real-world environments, and are as follows: (1) Intelligent traffic signal control system, (2) Edge computing, (3) Job shop scheduling, and (4) Heating control system.

## 5.1 Intelligent Traffic Signal Control System

An Intelligent Traffic Signal Control (ITSC) uses AI and real-time data to optimize the operation of traffic signals at intersections [295], with the aim to improve the efficiency of traffic flow, reduce congestion, and travel time, decrease emissions, improve safety, and enhance overall transportation system performance. Traditional traffic signal systems operate on fixed timing plans regardless of the actual traffic conditions. In contrast, an ITSC system dynamically adjusts signal timings based on real-time traffic data, such as vehicle volumes, speeds, and queues (see Figure 5.1). Below, the details of why particularly ITSC is of interest and is selected as an application area in this thesis are listed.

**ITSC's Environment is Dynamic and Partially Observable.** Vehicle-To-Infrastructure (V2I) communication can support real-time traffic information sharing between vehicles and traffic signal controllers. With V2I communication, the existing literature mostly assumes access to accurate and complete information retrieved from all vehicles on the road [110], [30]. However, the assumptions are inherently conflicted with real-world traffic scenarios. Firstly, not all vehicles are connected to the vehicular network, which is undetected without communication devices [294]. In this case, ITSC cannot fully observe the traffic information. Secondly, traffic data transmission is not always reliable and might experience packet loss and delay. In some cases, the observed traffic information is inaccurate due to noisy sensors. Moreover, in ITSC systems, it is not pos-sible to predict all the events, and consequently encountering unforeseen situations is inevitable (i.e., dynamic and stochastic traffic flow) [101]. Due to unexpected events such as traffic accidents, disabled vehicles, adverse weather conditions, spilled loads, and hazardous materials, the traffic demand varies from time to time which increases travel times. So, the controller agents should be able to handle emerging requirements and adapt to their dynamic environment on a real-time basis to decrease congestion, and secondary crashes, improve roadway safety, and decrease traffic delays [92].

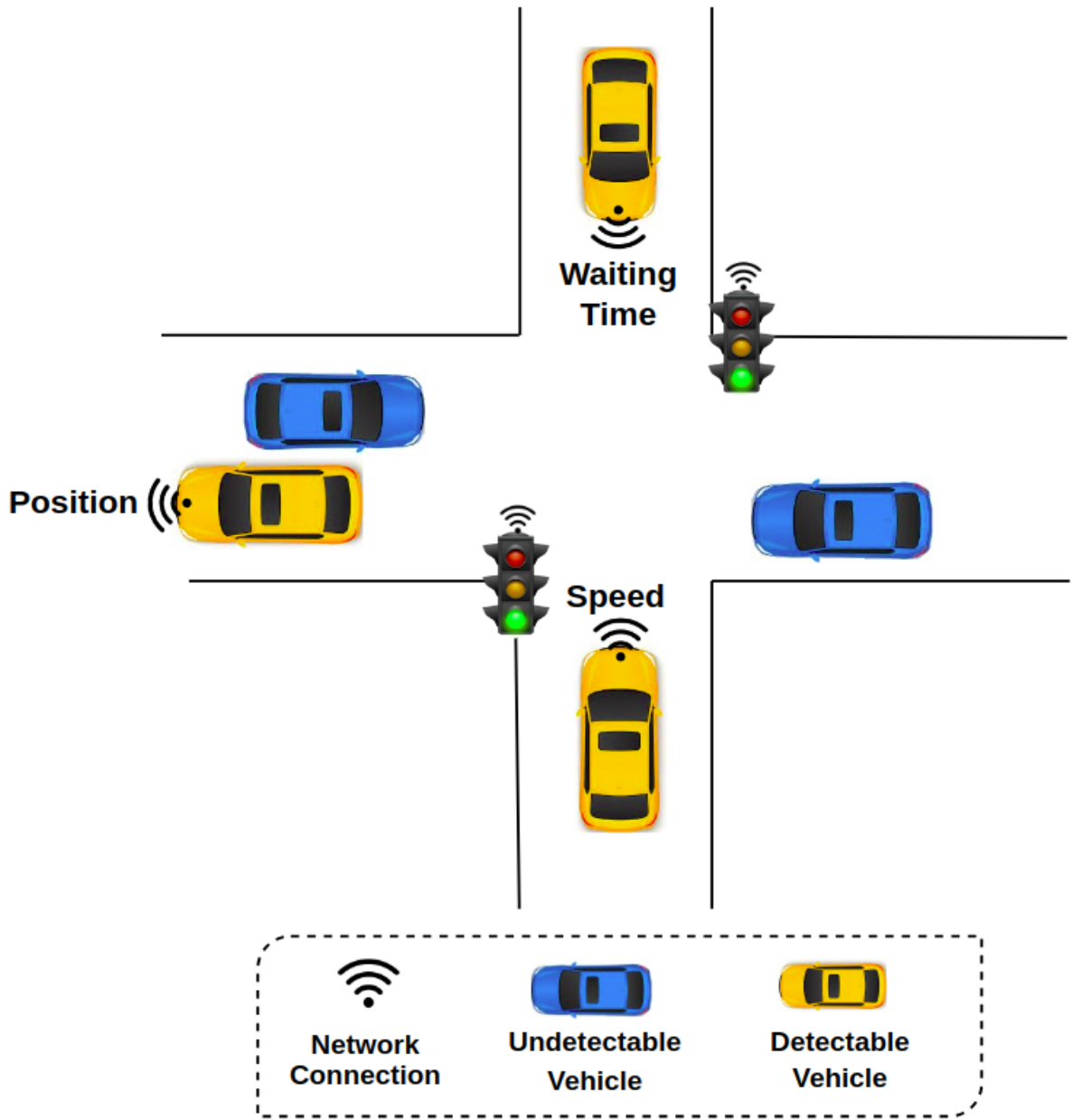

Figure 5.1: Representation of intelligent traffic signal control environment.

**ITSC can be managed through Autonomous agents.** ITSC can be managed by agents with learning capabilities controlling traffic signals [226, 19, 102]. RL agent observes and analyzes the traffic collected data and selects the appropriate traffic signal phase accordingly. RL techniques are useful for partial detection of vehicles [295], [294], [89] and dynamic traffic network [291], [201] as they do not require comprehensive modeling of the underlying environment dynamics. However, when the real environment is significantly different from the simulated one, the controllers trained in the simulator are not robust and cannot adapt to such environments [294]. Hence, in the early stages of the deployment, when the detection rate of vehicles is low, RL algorithms may not perform well and execute a wrong update because of a high degree of unpredictability of the environment and detected vehicle patterns [295].

**Stakeholders.** In the context of ITSC, various stakeholders are expected to utilize the proposed ontology-driven methods to enhance traffic signal control systems and optimize traffic flow. City transportation authorities, traffic engineers, urban planners, and transportation management agencies are among the key stakeholders who would deploy these solutions. These methods would operate by continuously collecting real-time traffic data from sensors and vehicles, processing the data using ontological reasoning to extract relevant information, and dynamically adjusting traffic signal timings and control strategies to optimize traffic flow and reduce congestion. Through this deployment, stakeholders aim to improve overall transportation system performance, enhance road safety, and reduce travel times for road users within urban traffic control systems.

### 5.1.1 Modeling Intelligent Traffic Signal Control as an RL Process

The RL agents' state, actions, and reward can be defined as follows:

- *State*: I model information of each state $s_{g_i}^t$ for traffic signal controller $g_i$ at time step $t$ as follows:

$$s_{g_i}^t = \{\text{Phase indicator}, \text{Phase elapsed time}, \text{Lane queue}, \text{Lane density}, \\ \text{Vehicle type}, \text{Vehicle position}, \text{Vehicle waiting time}\} \quad (5.1)$$

  Where the *phase indicator* represents the yellow, red, and green light, the *lane queue* symbolizes the number of vehicles awaiting in each lane divided by the lane capacity, *lane density* signifies the number of vehicles in each lane divided by the lane capacity, and *vehicle waiting time* indicates the number of seconds a vehicle maintains a speed lower than $0.1\ m/s$.

- *Actions*: Taking action $a$ means selecting an appropriate traffic signal phase for the next time step, which results in "Keep" or "Switch" the traffic signal phase accordingly:

$$A = \{\text{Keep}, \text{Switch}\} \quad (5.2)$$

- *Reward*: The reward $R_{g_i}^t$ is computed based on the waiting times of individual vehicles in two consecutive time steps $t$ and $t+1$:

$$R_{g_i}^t = \left( \sum_{v_i \in \text{Vehicles}} w_{v_i}^t - \sum_{v_j \in \text{Vehicles}} w_{v_j}^{t+1} \right) \quad (5.3)$$

Where $w_{v_i}^t$ represents the waiting time of vehicle $v_i$ at time step $t$, and $w_{v_j}^{t+1}$ represents the waiting time of vehicle $v_j$ at time step $t+1$. The objective of the ITSC system is to minimize the total cumulative waiting time of all vehicles over two consecutive time steps, and the reward function encourages actions that lead to a reduction in the cumulative waiting times of individual vehicles.

### 5.1.2 Case Studies in Intelligent Traffic Signal Control

I have formulated the following two case studies to assess the effectiveness of OntoDeM in the ITSC application area:

- **ITSC-CS1:** This case study aims to minimize waiting times for all vehicles by addressing congestion issues and reducing waiting times for emergency vehicles on a simulated synthetic traffic grid through improvements in traffic signal control.

- **ITSC-CS2:** In this case study, the aim is to enhance traffic signal control within a realistic urban traffic setting. The study takes place in Dublin and seeks to minimize vehicles' waiting time and travel delays and maximize the number of finished trips.

## 5.2 Edge Computing Environment

Edge Computing (EC) is a computing paradigm that moves the processing of the data and computations closer to the source of data (e.g., edge devices), to reduce latency and enhance system performance [209]. This paradigm involves offloading the computational tasks generated at resource-constrained edge devices to nearby edge servers or centralized cloud resources to optimize resource utilization, reduce latency, and improve system performance [230]. Currently, the literature includes a vast body of AI-based task offloading techniques that have proposed intelligent decision-making algorithms to determine whether to process a task locally on an edge device or offload it to a centralized cloud or nearby edge server (see Figure 5.2). Many of such methods [169, 230, 267, 232, 233, 231] include a learning process that learns from past experiences to optimize offloading strategies based on factors like latency, resource availability, and user preferences.

**EC's Environment is Dynamic and Partially Observable.** Edge computing aims to process data and make decisions closer to the source to reduce latency, however, in certain scenarios, there might be a trade-off between latency and data availability. Limited observability can impact real-time decision-making, as edge devices may not have access to up-to-date or complete information. Moreover, unforeseen events can include sudden changes in workload, connectivity disruptions, hardware failures, and environmental factors. Edge computing environments often experience fluctuations in workload due to varying data generation rates, user demands, or changing application requirements. Unforeseen events can cause sudden spikes or drops in workload, which can strain the resources and impact the performance of edge devices. Also, edge devices typically have constrained computational power, storage capacity, and energy resources, and Unforeseen offloading requests that require additional processing, storage, or network bandwidth that may exceed the capabilities of edge devices, leading to performance degradation or system failures. Moreover, edge devices rely on network connectivity to communicate with each other or with centralized systems. Unforeseen events such as power failures, natural disasters, physical damage, network outages, interference, or intermittent connectivity can disrupt communication and coordination between edge devices, impacting the overall functioning of the edge computing system.

**Task offloading can be managed through Autonomous agents.** RL techniques [230, 162, 252] can be employed to enable edge devices to learn optimal offloading strategies through trial and error. RL agents can learn to maximize reward functions (e.g., minimizing latency or energy consumption) by observing the current state, selecting an offloading action, and receiving feedback on the success of the offloading decision. This allows edge devices to adapt their offloading decisions based on dynamic conditions.

**Stakeholders.** In the context of EC applications, stakeholders such as EC service providers, IoT device manufacturers, software developers, and end-users can leverage ontology-based methods to enhance decision-making processes and optimize system performance. EC service providers can integrate ontology-driven techniques into their platforms to offer intelligent data processing and decision-making capabilities at the network edge. IoT device manufacturers can embed ontology-based reasoning modules into their devices to enable context-aware data analysis and interpretation. Software developers can utilize ontology-based methods to build edge applications that incorporate semantic reasoning for more intelligent and adaptive behavior. End-users, such as organizations deploying EC solutions, can benefit from ontology-based meth-ods by achieving better insights, improved decision-making, and enhanced operational efficiency in their EC environments.

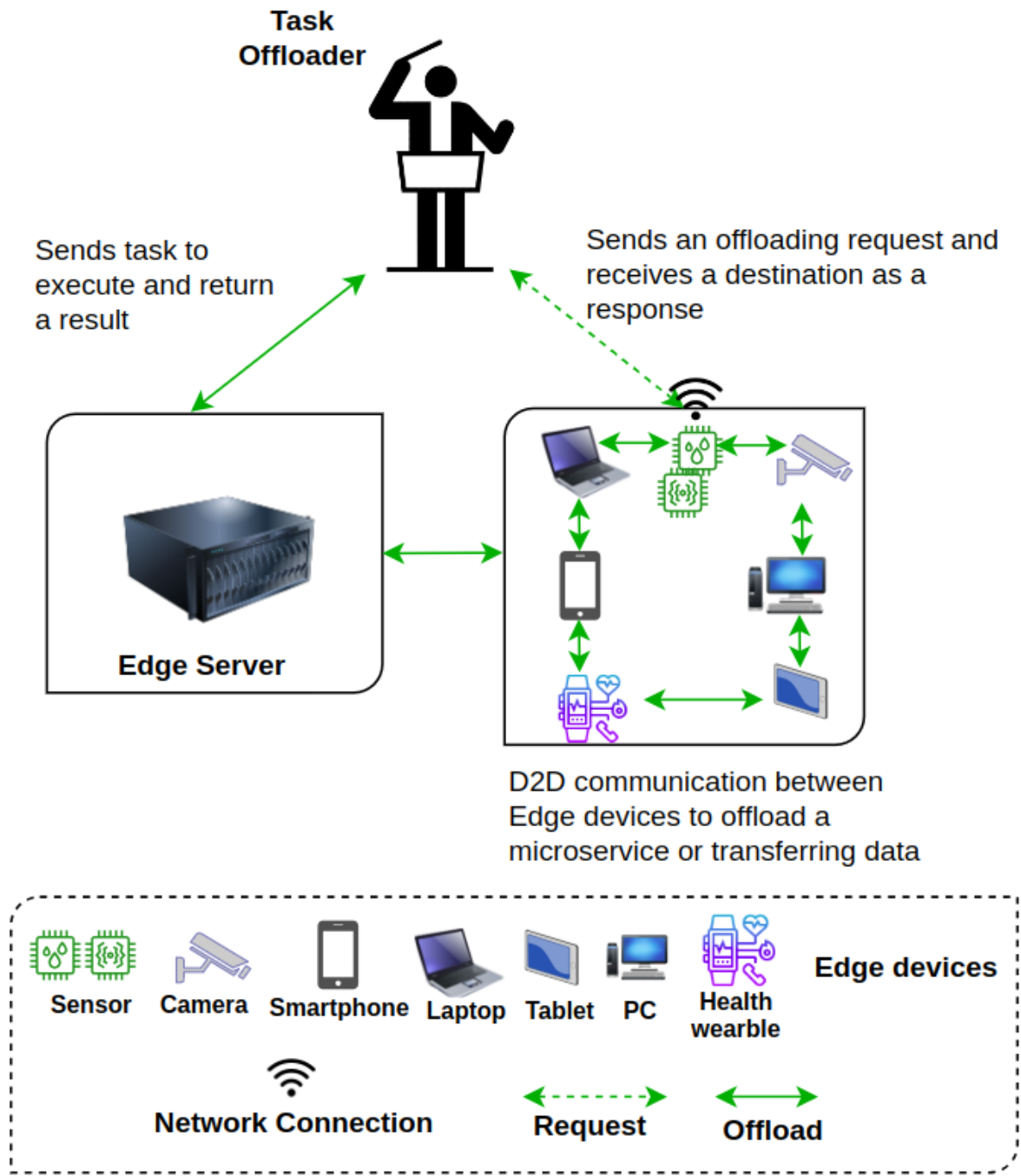

Figure 5.2: Representation of edge computing environment.

### 5.2.1 Modeling Task Offloading as an RL Process

In order to apply RL to the task offloading problem, it is essential to establish clear definitions for the state, actions, and reward. These fundamental components are detailed as follows:

- *State*: I model information of each state $s_{g_i}^t$ for task offloading agent $g_i$ at time step $t$ as following parameters:

$$s_{g_i}^t = \{\text{Server computing resource}, \text{Server workload}, \text{Server board}, \text{Server group},$$
$$\text{Server cost}, \text{Bandwidth}, \text{Task size}, \text{Task latency}, \text{Task priority},$$
$$\text{Application type}, \text{Device location}, \text{Device type}, \text{Device operating system},$$
$$\text{Device database}, \text{User financial information}, \text{User Group}, \text{User usage}\} \tag{5.4}$$

- *Actions*: Let $A$ represent the set of all possible actions available to each mobile user. Each action, denoted by $a \in A$, is defined as a tuple of three components:

$$a = (\text{Computing resources}, \text{Migration bandwidth}, \text{Offloading target}) \tag{5.5}$$

Where *computing resources* represents the amount of computing resources required by the mobile user's task, *migration bandwidth* represents the required bandwidth for the task during the offloading process, and *offloading target* indicates the destination or target for offloading the task to an edge server.

- *Reward*: The reward function is defined as the total number of processed tasks in each time step:

$$R_{g_i}^t = \sum_{i=1}^{N} P_i^t \tag{5.6}$$

Where $N$ is the total number of mobile users (or tasks) in the system, and $P_i^t$ represents the number of tasks processed for mobile user $i$ at time step $t$.

### 5.2.2 The Case Study in Edge Computing

The edge computing case study, **EC-CS** aims to assess the efficacy of OntoDeM in enhancing task offloading performance within a mobile user environment. The study takes place in a dynamic setting, specifically at the Seoul subway station, with the primary goals of maximizing task success rate and minimizing task failure rate while optimizing task distribution between mobile devices and edge resources.

## 5.3 Job Shop Scheduling

Job Shop Scheduling (JSS) refers to the problem of processing multiple orders on multiple machines, with several operational steps required in each order, completed in a specific sequence [2]. For instance, the order may involve manufacturing consumer products like automobiles. As orders are generated by sources, they are placed in the queue. A job shop scheduler agent then selects an order from the queue of orders in the source. Then, it moves to a work area and a group and selects a machine in the desired group to process the order. Finally, it selects an order from the queue of orders in the machine and places the processed order in a sink for consumption (see Figure 5.3). The job shop scheduler agent is responsible for scheduling orders so that all of them can be completed in the shortest amount of time. AI approaches are applied to tackle the complex optimization problem of JSS [155], [293], [41], which involve determining the most efficient sequence and timing for executing a set of orders on a set of machines. RL techniques can be employed to learn scheduling policies [205].

**JSS's Environment is Dynamic and Partially Observable.** Partial observability in JSS refers to the limited or incomplete knowledge of the current system state and future events. It means that the scheduler does not have complete information about the status of orders, machines, resources, and their interactions at any given time. Unforeseen events in JSS are unexpected disruptions or changes that can impact the scheduling process and the overall performance of the system. Unforeseen events can include machine breakdowns, order arrivals or cancellations, priority changes, resource unavailability, and other unexpected circumstances.

**JSS can be managed through Autonomous agents.** JSS as a complex task in manufacturing and production environments, can be effectively managed through autonomous agents. By autonomously coordinating and optimizing the scheduling of orders, resources, and machines, they can significantly reduce production delays and improve overall operational efficiency. Through trial and error, the RL agent learns to maximize rewards (e.g., minimizing makespan or total tardiness) and improves its schedul-ing decisions over time. Deep RL [155, 219, 288] and multi-agent [279, 298, 208] algorithms allow agents to explore different scheduling strategies while exploiting previously learned knowledge. By continuously improving their decisions based on rewards obtained from completed orders, RL agents can efficiently tackle the challenges of scheduling in dynamic job shop environments, ultimately leading to enhanced production efficiency and resource utilization.

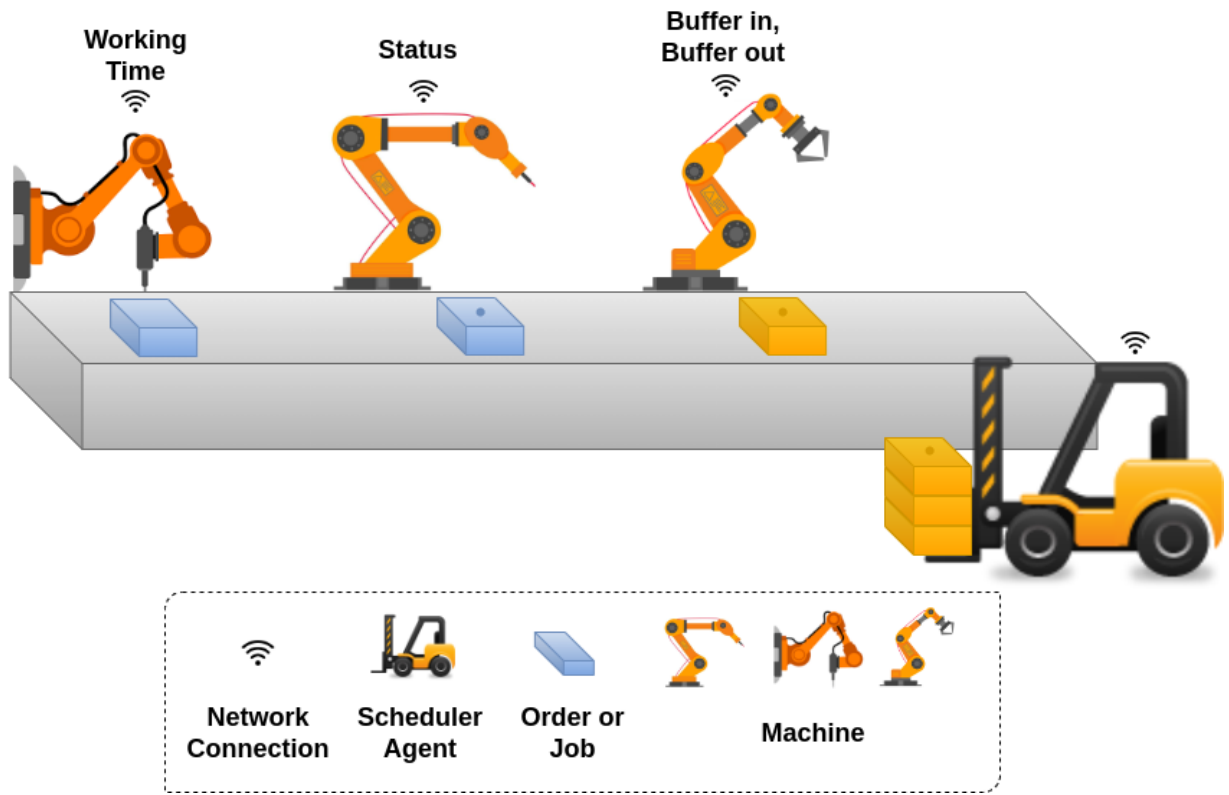

Figure 5.3: Representation of job shop scheduling environment.

**Stakeholders.** In the JSS environment, stakeholders such as manufacturing companies, production managers, scheduling software developers, and industrial engineers can leverage ontology-based methods to improve scheduling efficiency and optimize production processes. Manufacturing companies can integrate ontology-driven scheduling modules into their existing systems to automate scheduling decisions and optimize resource utilization. Production managers can utilize ontology-based scheduling tools to generate optimized production schedules that consider various constraints and objectives, such as minimizing makespan or maximizing throughput. Scheduling software developers can incorporate ontology-driven reasoning engines into their scheduling software to enable more intelligent decision-making and adaptive scheduling algorithms. Industrial engineers can use ontology-based methods to model and analyze complex production environments, identifying critical factors and dependencies that impact scheduling decisions.

### 5.3.1 Modeling Job Shop Scheduling as an RL Process

The job shop scheduler is modeled as an RL agent that receives reward and state observation from the environment and takes action accordingly. Agent's state, actions, and reward are modeled as follows:

- *State*: The information of each state $s_{g_i}^t$ for the job shop scheduler $g_i$ at time step $t$ is modelled as following parameters:

$$\begin{aligned}
s_{g_i}^t = \{&\text{Order waiting time}, \\
&\text{Order due date}, \text{Order actual processing time}, \\
&\text{Order current processing time}, \text{Order priority}, \text{Machine working time}, \\
&\text{Machine idle time}, \text{Machine failure time}, \text{Machine failure rate}, \\
&\text{Machine last broken start time}, \text{Machine last process start time}, \\
&\text{Machine status}, \text{Machine input buffer}, \text{Machine output buffer}, \\
&\text{Machine processing buffer}, \text{Machine capacity}, \text{Work area distance}, \\
&\text{Sink capacity}\}
\end{aligned}$$
(5.7)

Where *order waiting time* signifies the number of seconds the order has been waiting for its turn to begin processing by a machine, *order due date* represents the required or preferable time for order completion, categorized as low (1000 seconds), medium (3000 seconds), and high (5000 seconds), *order priority* is categorized as low, medium, or high, *order current processing time* quantifies the duration since the order entered the machine's processing buffer until the present moment, *machine working time* quantifies the total processing duration of the machine, *machine idle time* defines intervals when the machine remains available but unproductive, *machine failure rate* pertains to the likelihood of machine failure, classified as low, medium, or high, *last broken start time* refers to the most recent occurrence of machine failure, *last process start time* indicates the most recent initiation time of a process within the machine, *machine status* encompasses failure, working, or idle states, *machine input buffer* is the count of orders within the machine's input buffer, *machine output buffer* indicates the count of orders within the output buffer, *machine processing buffer* signifies the count of orders currently under processing, *machine capacity* corresponds to the total order handling capacity of the machine, and *work area distance* refers to the distance between the work area and the scheduler agent.

- *Actions*: Let $A_{\text{order}}$ be the set of actions for selecting an order to be processed from multiple available orders:

$$A_{\text{order}} = \{\text{Order 1}, \text{Order 2}, \ldots, \text{Order n}\} \tag{5.8}$$

Similarly, let $A_{\text{machine}}$ be the set of actions for selecting a machine for processing an order from multiple available machines:

$$A_{\text{machine}} = \{\text{Machine 1}, \text{Machine 2}, \ldots, \text{Machine m}\} \tag{5.9}$$

And let $A_{\text{sink}}$ be the set of actions for selecting a sink to consume the processed order from multiple available sinks:

$$A_{\text{sink}} = \{\text{Sink 1}, \text{Sink 2}, \ldots, \text{Sink k}\} \tag{5.10}$$

The scheduler agent needs to make decisions by selecting an appropriate order, machine, or sink from the respective action sets.

- *Reward*: The reward function maximizes the average utilization rate of all the machines. The percentage utilization of an individual machine $U_{m_i}^t$ is calculated based on its working time as follows:

$$U_{m_i}^t = W_{m_i}^t / F_{m_i}^t \tag{5.11}$$

Where $F_{m_i}$ represents the failure time of machine $m_i$ ($F_{m_i}$) can be mathematically described as the summation of failure occurrences ($f_{m_i}^t$) over all time steps until time step $t$:

$$F_{m_i}^t = \sum_{j=1}^{t} f_{m_i}^j \tag{5.12}$$

The working time of machine $m_i$ ($W_{m_i}$) can be mathematically described as the accumulation of processing times ($w_{m_i}^t$) across all time steps up to time step $t$:

$$W_{m_i}^t = \sum_{j=1}^{t} w_{m_i}^j \tag{5.13}$$

Then the reward is computed by the average of the utilization rate as follows:

$$R_{g_i}^t = \frac{1}{|M|} \sum_{i=0}^{|M|-1} U_{m_i}^t \tag{5.14}$$

### 5.3.2 Case Studies in Job Shop Scheduling

I have developed the following two case studies to assess the effectiveness of OntoDeM in the application of JSS:

- **JSS-CS1:** This case study focuses on improving JSS efficiency in a simple setup that includes eight machines. It primarily addresses the task of selecting the optimal machine for order processing, with a focus on maximizing machine utilization rate and maximizing the total number of processed orders.

- **JSS-CS2:** In this case study, the objective is to optimize the performance of the JSS algorithm in a more complex setup, involving sixteen machines. The study encompasses a wider range of tasks, including machine selection, order selection, work area selection, and group selection, all of which are distributed across various locations within the factory. The study aims to maximize machine utilization rate, and total processed orders, and minimize order waiting time and the total number of failed orders.

## 5.4 Heating System Control

A Heating System Control (HSC) is an automated system designed to regulate the indoor temperature of buildings by adjusting the temperature and other parameters (e.g., humidity levels, ventilation rates). These systems aim to maintain a comfortable and consistent temperature while optimizing energy consumption and minimizing wastage. These controllers utilize various sensors and algorithms to monitor indoor and outdoor conditions, such as temperature, humidity, occupancy, and weather forecasts (see Figure 5.4). AI can significantly enhance HSC by intelligently analyzing data (e.g., historical temperature patterns) and making informed decisions to optimize heating efficiency (i.e., reducing energy consumption and costs) [198].

**HSC's Environment is Dynamic and Partially Observable.** The environment of an HSC involves constantly changing factors such as external weather conditions, varying heating demands, and user preferences. Additionally, the system may not have complete information about the entire environment at all times, making it partially observable. One common challenge is dealing with noisy temperature data, which can be affected by measurement errors or environmental fluctuations.

**HSC can be managed through Autonomous agents.** In the domain of HSC, autonomous agents can achieve optimal comfort and energy conservation by continuously adjusting heating settings in response to real-time temperature readings. RL agents learn from interactions with the HSC environment, observing the state (e.g., temperature), selecting actions (e.g., thermostat adjustments), and receiving rewards for actions that lead to comfortable indoor temperatures and reduced energy consumption [296, 125, 124].

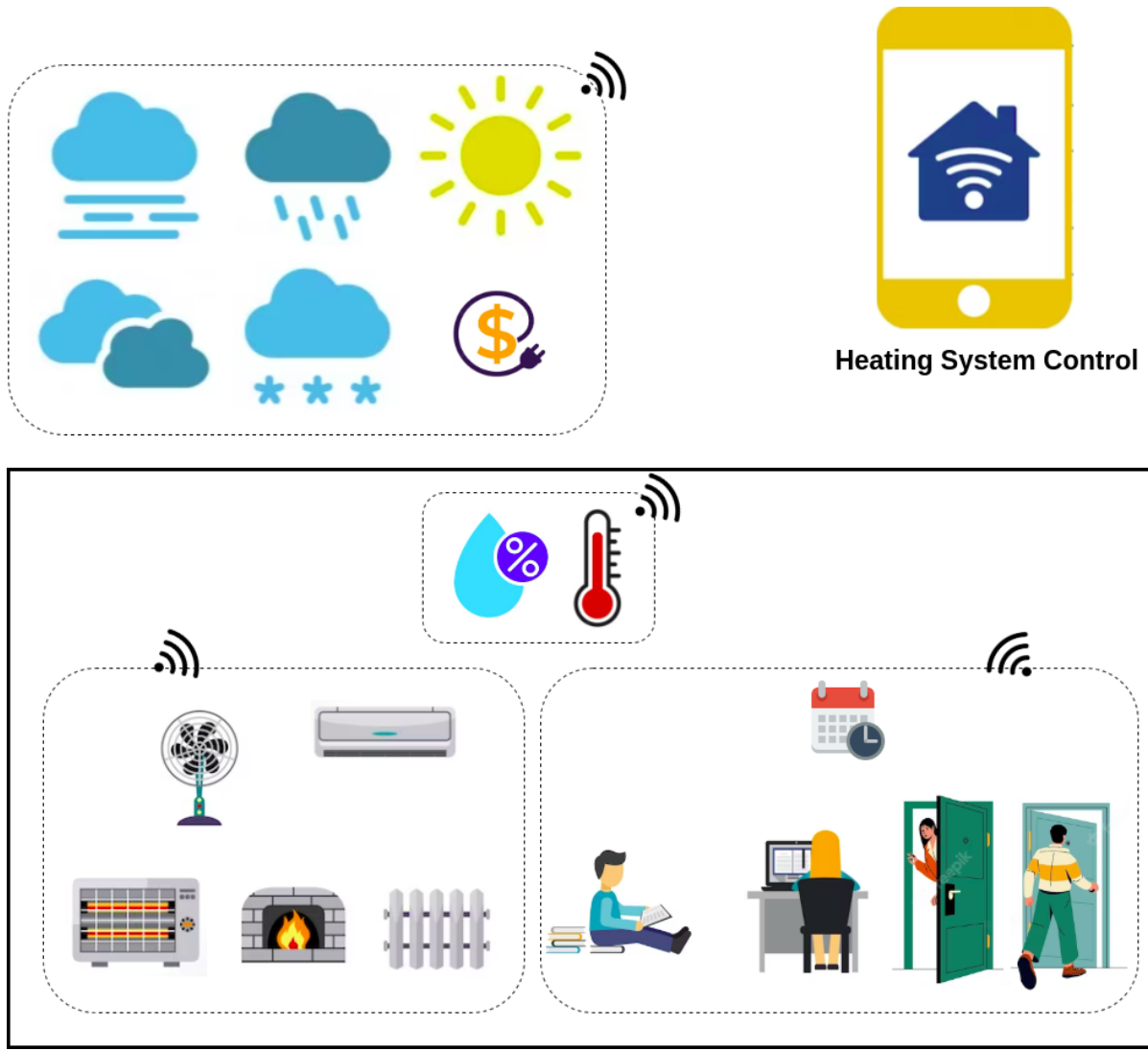

Figure 5.4: Representation of heating system control environment.

**Stakeholders.** In the HSC environment, stakeholders such as building owners, facility managers, HVAC system integrators (professionals or companies specializing in the design, installation, and integration of Heating, Ventilation, and Air Conditioning (HVAC) systems), and energy efficiency consultants can utilize ontology-based methods to enhance heating system performance and energy efficiency. Building owners and facility managers can deploy ontology-driven control systems to optimize heating schedules and setpoints based on occupancy patterns, weather forecasts, and energy pricing signals. HVAC system integrators can incorporate ontology-based reasoning engines into building automation systems to enable adaptive heating control strategies that respond dynamically to changing environmental conditions and occupant preferences. Energy efficiency consultants can leverage ontology-based modeling tools to analyze heating system data and identify opportunities for energy savings through system optimization and retrofitting initiatives.

### 5.4.1 Modeling Heating System Control as an RL Process

To model HSC as an RL problem, I outline the definitions for state, actions, and the reward function as presented below:

- *State*: I model information of each state $s_{g_i}^t$ for heating system controller $g_i$ at time step $t$ as follows:

$$s_{g_i}^t = \{\text{Previous inside temperature}, \text{Previous outside temperature}, \\ \text{Expected outside temperature}, \text{Expected energy price}\} \quad (5.15)$$

Where the previous inside/outside temperature is calculated using a random normal distribution that relies on the initial temperature, time of day, the building's heat mass capacity, and heat transmission to the outside. The expected outside temperature and the expected energy price are calculated using a random normal distribution that relies on the specific time frame in the future.

- *Actions*: The action represents the heating power level (i.e., how much energy is being used to produce heat) that the system will use in the next time step to regulate the indoor temperature, and it should be chosen from the subsequent range:

$$A = [0, 11000] \quad (5.16)$$

Where 0 Watt represents no heating power applied and 11000 Watt signifies the maximum heating power achievable.

- *Reward*: The reward function can be composed of two components:
  - ▶ The penalty is applied for deviating from the desired inside temperature interval.
  - ▶ The agent is penalized according to the cost of the energy consumed.

$$R_{g_i}^t = -\text{MSE}(T_{\text{in}}^t) - E_{\text{cost}}^t \quad (5.17)$$

Where MSE($T_{\text{in}}^t$) is the Mean Squared Error penalizing deviation from the desired inside temperature interval, and $E_{\text{cost}}^t$ is the cost of the energy used (computed based on heating power and time).

The MSE($T_{\text{in}}^t$) is calculated as follows:

$$\text{MSE}(T_{\text{in}}^t) = \frac{1}{n} \sum_{i=1}^{n} (T_{\text{in},i}^t - T_{\text{mid}})^2 \tag{5.18}$$

Where $n$ is the number of observations at time step $t$, and $T_{\text{in},i}^t$ represents the observed temperature value for the $i$-th observation at time $t$. Additionally, $T_{\text{mid}}$ denotes the midpoint of the desired temperature interval. The summation iterates over all observations at time step $t$, calculating the squared difference between each observed temperature value and the midpoint of the desired temperature interval. The sum of squared differences is divided by the number of observations $n$ to compute the average squared difference, which is the MSE for time step $t$.

### 5.4.2 Case Studies in Heating System Control

I have devised the subsequent case studies to evaluate the efficacy of OntoDeM within the domain of HSC:

- **HSC-CS1:** In this case study, the primary objective is to maintain desired temperature levels while minimizing energy consumption. This case study involves training an RL model with a pre-trained base without adding noisy data.

- **HSC-CS2:** This case study focuses on enhancing the performance of a heating system controller. The approach involves an RL model that utilizes the same pre-trained base and includes additional runs with noisy data for training. The key measures for this case study include minimizing the deviation from the desired temperature interval and minimizing energy costs.

## 5.5 Summary

The presented case studies encompass diverse domains, each facing challenges arising from dynamic and partially observable environments. In the Intelligent Traffic Signal Control System (ITSC), Vehicle-To-Infrastructure (V2I) communication assumes all vehicles provide accurate data, neglecting unconnected vehicles. Unreliable data transmission and sensor inaccuracies further disrupt real-time data. Unforeseen events, such as accidents, and weather changes, result in dynamic traffic flows. Edge Computing (EC) involves processing data and performing computations at or near the data source, reducing the need for centralized cloud resources and minimizing latency. Similarly, in the context of EC, limited observability can impact real-time decision-making, especially when edge devices lack access to up-to-date or comprehensive information. Unforeseen events, including sudden workload fluctuations, connectivity disruptions, hardware failures, and environmental factors, further complicate the situation. Job Shop Scheduling (JSS) is a critical domain in production planning, aiming to optimize the arrangement of jobs or orders on machines to enhance resource utilization, minimize production time, and meet delivery deadlines effectively. JSS operates in a dynamic environment where unexpected events like machine failures and urgent order introductions disrupt schedules, and it's partially observable due to delayed, corrupted, or missing observations for machines and orders. Heating System Control (HSC) aims to regulate indoor temperatures efficiently while minimizing energy consumption. The environment's dynamics stem from changing external conditions and heating demands. Unforeseen events like component malfunctions, power outages, occupant activities, and weather fluctuations further contribute to the dynamic and challenging environment. Also, its partial observability arises from incomplete information about the entire environment due to noisy temperature data and measurement errors.

**Considerations of Agent Architecture in Application Areas.** In the ITSC application, we have a decentralized architecture with one decision-maker agent at each intersection. However, in the other application areas (EC, JSS, and HSC), a single centralized agent is responsible for decision-making and control across the entire system. This centralized approach offers several advantages, including simplified coordination, unified decision-making, and centralized knowledge representation. However, it also introduces certain implications for system performance and applicability:

- Scalability: In scenarios where the system grows in size or complexity, such as increasing the number of devices in EC, machines in JSS, or heating zones in HSC, a centralized agent may encounter scalability challenges. As the number of system components increases, the centralized agent may struggle to efficiently manage and coordinate all elements.

- Responsiveness: While a centralized agent can make globally optimal decisions, it may face challenges in responding quickly to local changes or events. In dynamic environments like EC, JSS, and HSC, rapid decision-making is crucial to adapt to changing conditions and ensure system performance.

- Fault Tolerance: The centralized nature of the agent architecture may pose vulnerabilities in terms of fault tolerance. A single point of failure in the centralized agent could potentially disrupt the entire system's operation, leading to downtime or degraded performance.

- Visibility: The visibility challenge stems from the decentralized and dynamic nature of the system's elements. For example, in EC, the centralized agent responsible for task offloading and resource allocation may lack comprehensive visibility over individual edge devices' status and capabilities. This limited visibility can result from factors such as network latency and communication delays.

However, the choice of a centralized agent architecture across the application scenarios in the thesis was deliberate and aligned with the primary research focus, which is centered on addressing the challenges of decision-making in partially observable and dynamic environments. While decentralized decision-making approaches, such as multi-agent systems, offer advantages in certain contexts, the thesis aims to demonstrate the effectiveness of proposed OntoDeM methods in facilitating decision-making processes within real-world scenarios characterized by partial observability and unforeseen events.

In Table 5.1, I present the mapping of OntoDeM methods to specific case studies, showing the addressed research questions and associated evaluation metrics for each case study.

Table 5.1: Mapping of case studies to the OntoDeM's methods, research questions, and related evaluation metrics.

| Case study | Method | Research question | Evaluation metric |
|---|---|---|---|
| ITSC-CS1 | Observation Augmentation<br>Observation Sampling<br>Automatic Goal Selection/ Generation<br>Reward Augmentation<br>Action Exploration | RQ1<br>RQ2 | **Minimize**<br>Vehicle waiting time |
| ITSC-CS2 | Observation Augmentation<br>Observation Sampling<br>Action Exploration<br>Action Prioritization | RQ1<br>RQ3 | **Minimize**<br>Vehicle waiting time<br>Travel time<br>Travel delay<br>**Maximize**<br>Number of finished trips |
| EC-CS | Observation Masking<br>Action Masking<br>Action Prioritization<br>Execution Prioritization | RQ1<br>RQ3 | **Maximize**<br>Task success rate<br>**Minimize**<br>Task failure rate |
| JSS-CS1 | Observation Augmentation | RQ1 | **Maximize**<br>Machine utilization rate<br>Total processed orders |
| JSS-CS2 | Adaptive Reward Definition | RQ2 | **Maximize**<br>Machine utilization rate<br>Total processed orders<br>**Minimize**<br>Order waiting time<br>Total failed orders |
| HSC-CS1<br>HSC-CS2 | Observation Abstraction | RQ1 | **Minimize**<br>Average time spent out of range<br>Deviation from desired temperature interval<br>**Maximize**<br>Total reward |

CHAPTER

SIX

# EVALUATION

This chapter evaluates OntoDeM's performance in different application domains including intelligent traffic signal control, edge computing, job shop scheduling, and heating system control. In separate sections, I describe the simulation environments, their settings, scenarios, evaluation criteria, and baseline algorithms that have been used to evaluate OntoDeM's performance, then discuss the results in the results analysis section.

## 6.1 Intelligent Traffic Signal Control Experiment Design

Simulation of Urban MObility (SUMO) which provides a microscopic real-time traffic simulation [168] is employed to evaluate the performance of OntoDeM in the ITSC case study. I used the interface to instantiate RL environments with SUMO for ITSC provided by [9] to interact with the traffic signal-controlled intersections. The proposed OntoDeM model is used by the RL agent to adjust the traffic signal phase to reduce accumulated vehicles' waiting time.

**Hardware and Operating System Configuration.** Simulations are conducted on a machine equipped with an Intel(R) Core(TM) i7-5500U CPU @ 2.40GHz (1 core, 1 thread, 2.39 GHz base) and VMware SVGA II Adapter, running on an Ubuntu 18.04.5 LTS operating system with 4.8GB of RAM.

### 6.1.1 Evaluation Scenarios, Metrics and Baseline

**Evaluation scenarios:** Four Evaluation scenarios are designed for this case study and summarised in Table 6.1. In these scenarios, the OntoDeM-enabled RL controllers and the baseline RL controllers' performance in addressing different traffic issues are evaluated and compared. These issues include handling traffic congestion, managing un-foreseen situations, dealing with partial observation and noisy data, and prioritizing special vehicles.

Table 6.1: Evaluation scenarios for intelligent traffic signal control environment.

| Scenario | Description |
| --- | --- |
| Scenario 1 - Partial observation | The controller agents cannot observe the waiting time of 50% of vehicles. This scenario will test OntoDeM's performance in handling the partial detection of vehicles problem. |
| Scenario 2 - Noisy observation | The waiting time information of 50% of vehicles will not be accurate to simulate observation. Vehicles can be detected to get their real-time speed and position, however, the data received is not always real-time due to packet loss and latency during the transmission. This scenario will test OntoDeM's performance in handling the noisy observation data. |
| Scenario 3 - Vehicle prioritization | Special vehicle types (i.e., ambulance, fuel truck, and trailer) that have higher priorities over default vehicles are generated which are considered unexpected traffic situations. This scenario will test OntoDeM's performance to adapt to such special vehicles as unforeseen entities. |
| Scenario 4 - Event prioritization | Road/lane closure caused by accidents burden the surrounding traffic, possibly lasting a few hours. This scenario will test OntoDeM's performance to solve the unexpected congestion problem. |

**Evaluation metrics:** The metrics employed to evaluate the enhancement at the system level in each scenario are as follows:

- Average vehicle waiting time as the average of the cumulative duration when the vehicle speed is lower than $0.1\ m/s$.

- Average vehicle travel time as the average duration it takes for the vehicle to travel from its origin to its destination within the simulated road network.

- Average travel delay as the average of the vehicle's time loss due to traveling below the maximum allowed speed (e.g., slowing down near an intersection).

- Average number of finished trips.

**Baselines:** I compared the results obtained from the baseline RL algorithms, including Q-learning, SARSA, and DQN [9], to the same algorithms when they use OntoDeM along with their usual functionality. Q-learning's simplicity and SARSA's on-policy learning complement DQN's deep neural network architecture and off-policy learning, providing a comprehensive assessment of OntoDeM's performance in the ITSC application area.

### 6.1.2 Case studies and Employed OntoDeM Methods

**ITSC-CS1.** The whole simulated traffic network is a $750m \times 750m$ area. Each intersection is a $300m \times 300m$ area, with 16 intersections, each including two incoming roads and two exit roads (see Figure 6.1). Each road is marked with a name such as 0to1 and includes two lanes, for example, road 0to1 includes two lanes 0to1_0 and 0to1_1. So, there are eight lanes with a length of 120 meters. The vehicles on incoming roads from west to east are allowed to take right-turn and pass-through traffic. The vehicles on incoming roads from north to south are allowed to take a left turn and pass-through traffic. The minimal gap between the two vehicles is 2.5 meters. Four types of vehicles are defined in the simulation: default, ambulance, fuel truck, and trailer. The length of default and ambulance vehicles is 5 meters and the length of the fuel truck and trailer is 10 meters. The default vehicles arrive in the environment following a random process. All the vehicles are assigned a predefined route with varying lengths. Vehicles are discarded if they cannot be inserted. For all types of vehicles, the max speed is $55.55\ m/s$, which is equal to $200\ km/h$ [24]. However, depending on the usage, different roads of the simulated network have varying speed limits, most of them are lower than this max speed. Also, the max accelerating acceleration is $2.6\ m/s^2$ and the decelerating acceleration is $4.5\ m/s^2$. SUMO uses the Krauss Following Model [141], which guarantees safe driving on the road. The duration of the yellow phase is set to 2 seconds. The minimum duration of the green phase is set to 5 seconds and the maximum one is set to 100 seconds. The number of simulated seconds on SUMO is set to 1000 seconds. The number of simulation seconds run before learning begins is set to 300 seconds. The simulation seconds between actions are set to 5 seconds [1].

**OntoDeM's Methods Employed in ITSC-CS1.**

*Observation augmentation (related to S1, CH1, RO1, RQ1):* When the OntoDeM-enabled RL signal controller is unable to collect data on undetectable vehicles in the environment, it replaces the waiting time of "Stationary" vehicles $a$ on lane $l$ with the current traffic signal phase elapsed time $e$, following the inference rule shown in Table 6.2.

*Observation sampling (related to S1, CH1, RO1, RQ1):* When the observation is noisy, the controller realizes that "Stationary" vehicles are more important than "Moving" vehicles according to the inference rules in Table 6.3, and their waiting times are sampled more frequently.

---

[1]The code of OntoDeM for synthetic traffic grid is publicly available: `https://github.com/saeedehghanadbashi/Ontology-based-Intelligent-Traffic-Signal-Control-Model`

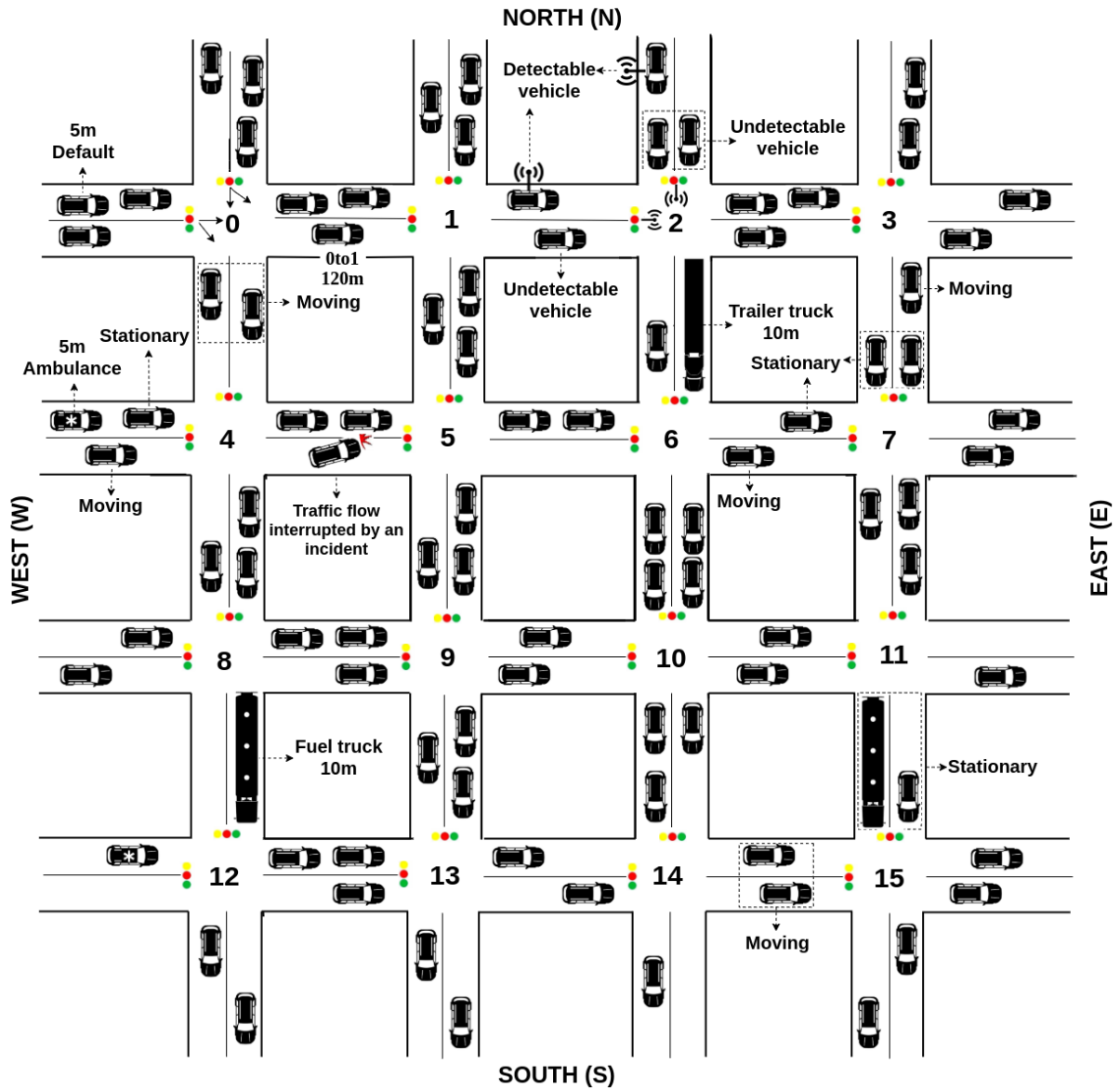

Figure 6.1: Simulated synthetic traffic grid in SUMO.

Table 6.2: Augmenting waiting time of undetectable vehicles, an example of inference rules.

| Inference rules |
| --- |
| TrafficSignalControl(?i), Intersection(?s), Road(?r), Lane(?l), consistOf(?r, ?l), Vehicle(?a), isOn(?a, ?l), atIntersection(?i, ?s), isRegulatedBy(?r, ?i), hasPosition(?a, Stationary), hasElapsedTime(?i, ?e) |
| − > hasWaitingTime(?a, ?e) |
| [a]In Semantic Web Rule Language (SWRL), variables are indicated using the standard convention of prefixing them with a question mark. |

Table 6.3: Sampling observation proportional to the vehicles' position, an example of inference rules.

| Inference rules |
|---|
| TrafficSignalControl(?i), Intersection(?s), Road(?r), Lane(?l), consistOf(?r, ?l), Vehicle(?a), Vehicle(?b), isOn(?a, ?l), isOn(?b, ?l), atIntersection(?i, ?s), isRegulatedBy(?r, ?i), hasPosition(?a, Stationary), hasPosition(?b, Moving), hasImportanceWeight(?a, Low), hasImportanceWeight(?b, Lowest) |
| $->$ hasHigherSamplingRate(?a, ?b) |

*Reward augmentation (related to S2, CH2, RO2, RQ2):* When special vehicles enter intersections, the controller assigns the "Highest" importance weight to the relationship "hasType" and its domain "Vehicle" and range "Emergency", thus giving priority to the road that has the most important vehicles (see the related inference rules in Table 6.4). In this case, it has a higher efficiency rate to switch to a green phase for the road containing emergency vehicles than to keep a green phase on another road because the former action reduces the waiting time for special vehicles with the "Highest" importance weight.

Table 6.4: Selecting action based on the vehicles' type, an example of inference rules.

| Inference rules |
|---|
| TrafficSignalControl(?$i^a$), Intersection(?s), Road(?$r_1$), Road(?$r_2$), Vehicle(?a), Vehicle(?b), isOn(?a, ?$r_1$), isOn(?b, ?$r_2$), atIntersection(?i, ?s), DifferentFrom(?$r_1$, ?$r_2$), isRegulatedBy(?$r_1$, ?i), isRegulatedBy(?$r_2$, ?i), hasType(?a, Emergency), hasSubType(?a, Ambulance), hasType(?b, Default), hasPosition(?a, Stationary), hasPosition(?b, Stationary), hasImportanceWeight(?a, Highest), hasImportanceWeight(?b, Low), hasTrafficSignalPhase(?$r_1$, Red), hasTrafficSignalPhase(?$r_2$, Green) |
| $->$ hasHigherImportance(?$r_1$, ?$r_2$), isNextTrafficSignalAction(?i, Switch) |

*Automatic goal selection/generation (related to S2, CH2, RO2, RQ2):* When congestion happens as an unforeseen situation according to the inference rules shown in Table 6.5, OntoDeM creates a new goal to revert the road to a familiar previous state by defining the state similarity reward function. So, the agent maximizes the position coordinates of all vehicles $b$ on the road $r1$ (i.e., hasPosition(?b, Moving)), thereby, the environment will be reverted to a known previous state.

**ITSC-CS2.** The testing traffic scenario is extracted from the open data in [104] to simulate realistic urban traffic in Dublin. The simulated road network covers approximately 1 square kilometer as shown in Figure 6.2, located at the Dublin city center in Dublin 2. There are 29 signalized intersections of 119 intersections in total. These traffic signal plans vary according to the intersection shape, so the number of phases ranges from 3 to 6. The green phase lasts from 27 seconds to 82 seconds. The biggest intersection has 5 controlled roads, including 4 incoming and 6 outgoing lanes. In the experiments, the simulated traffic lasts 2000 seconds from 7:30 AM, while the learning process starts after a 500-second warm-up period. There would be nearly 1500 vehicles departing during the simulations[2].

Table 6.5: Inferring maximizing the position coordinates of all vehicles involved in congestion, an example of inference rules.

| Inference rules |
| --- |
| TrafficSignalControl(?i), Intersection(?s), Road(?r1), Lane(?l1), Vehicle(?b), isOn(?b, ?l1), consistOf(?r1, ?l1), atIntersection(?i, ?s), isRegulatedBy(?r1, ?i), hasPosition(?b, Stationary) |
| $->$ hasCongestion(?r1) |

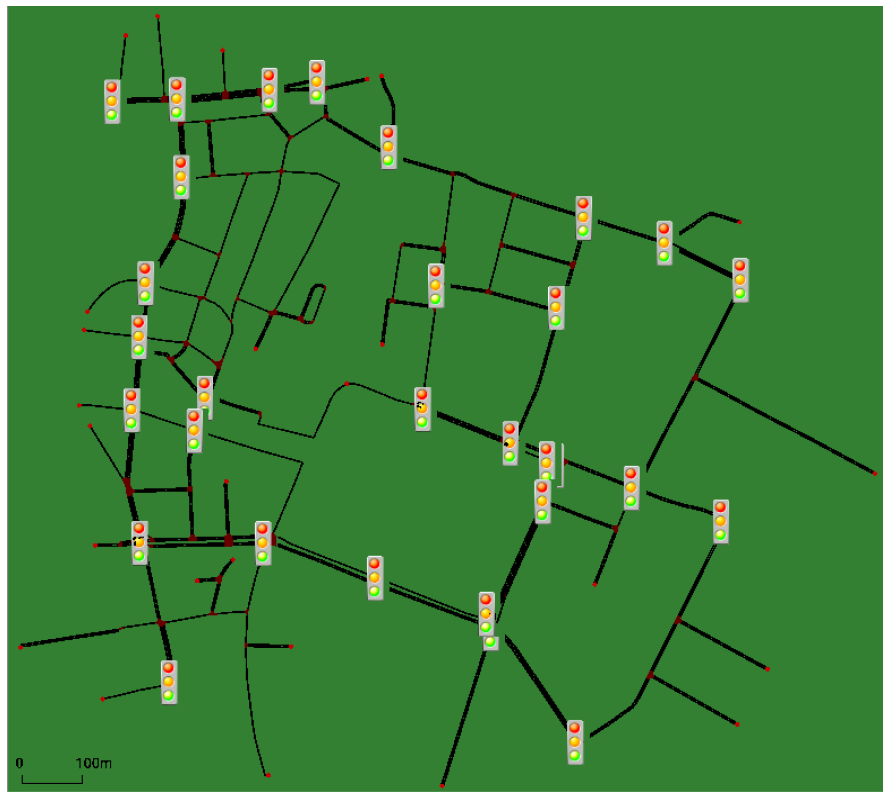

Figure 6.2: Simulated urban traffic network at Dublin city center.

**OntoDeM's Methods Employed in ITSC-CS2.**

In Scenario 1, the OntoDeM model utilizes *observation augmentation (related to S1, CH1, RO1, RQ1)* to deduce waiting time information for undetected vehicles. This logical inference results in accessing more accurate data, and consequently enabling precise traffic signal decisions. In Scenario 2, the OntoDeM model employs *observation*

---

[2]The code of OntoDeM for urban traffic network is publicly available: https://github.com/Guojyjy/OITSC_Urban

*sampling (related to S1, CH1, RO1, RQ1)* to filter noise and address data inconsistencies, thereby improving traffic signal control performance in the presence of noise. Scenario 3 demonstrates the OntoDeM *action prioritization (related to S3, CH2, RO3, RQ3)* method in adjusting signal timings based on the importance of different vehicle types, illustrating its adaptability to dynamic traffic scenarios. Finally, in Scenario 4, the OntoDeM model's adeptness in interpreting road closures and traffic impacts is highlighted, as its *action exploration (related to S3, CH2, RO3, RQ3)* method strategically modifies signal phases to alleviate congestion and minimize delays, showcasing its ability to respond effectively to real-time events.

### 6.1.3 Results Analysis

**Results in the ITSC-CS1.** The results report the average waiting time for all types of vehicles in 10 runs in each scenario. For scenario 3, special vehicles are generated in (A) four *parallel* roads: 0to1, 4to5, 9to10, 14to15, (B) three pairs of *intersecting* roads: (0to4, 21to4), (5to9, 8to9), (10to14, 13to14) or (C) random roads in *Low* and *High* frequencies. In *Low* frequency setting, 3 special vehicles are generated per 2 minutes, and in *High* frequency, 5 special vehicles are generated per 2 minutes. For scenario 4, traffic congestion is created at even-numbered intersections (e.g., 0, 2) with a line of more than 10 vehicles at an intersection. When OntoDeM is employed, the results show that all algorithms outperform their basic implementation (see Figure 6.3).

As shown in Figure 6.3a, OntoDeM performs almost as well as the baseline algorithms when the average waiting time of default vehicles is compared in scenarios 3.A, 3.B, and 3.C, and outperforms the baseline algorithms in scenario 4. As shown in Figures 6.3b, 6.3c, and 6.3d, OntoDeM significantly decreases the average waiting time of ambulances, fuel trucks, and trailers compared to the baseline algorithms.

As shown in Table 6.6 (p-value results computed by the Mann-Whitney U test to justify that the observed differences between the two methods (i.e., baseline and OntoDeM) are shown in Table 6.7), the average waiting decreases significantly when OntoDeM is applied, compared to the baseline algorithms, however, OntoDeM is more effective in low-frequency settings in a more complex scenario.

Also, I observed the improvement of OntoDeM compared to the baseline algorithms in scenario 3.A where ambulances, fuel trucks, and trailers are generated in parallel roads is more than scenario 3.B where they are generated in intersecting roads. This is because the special vehicles entering intersecting roads create a more complex situation. Also, my approach shows a better performance in more complex scenarios where unforeseen situations can happen simultaneously in multiple parallel or inter-

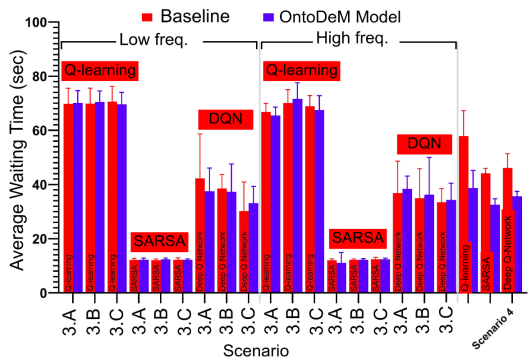

(a) The average waiting time for default vehicles in Scenarios 3 and 4: baseline and OntoDeM.

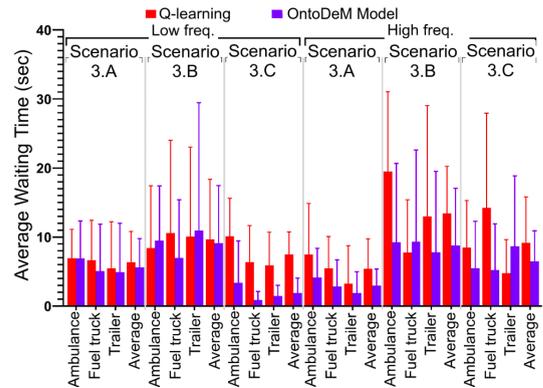

(b) The average waiting time for special vehicles in scenario 3: Q-learning and OntoDeM.

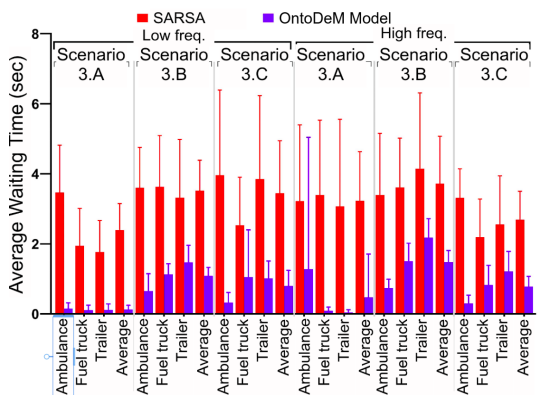

(c) The average waiting time for special vehicles in scenario 3: SARSA and OntoDeM.

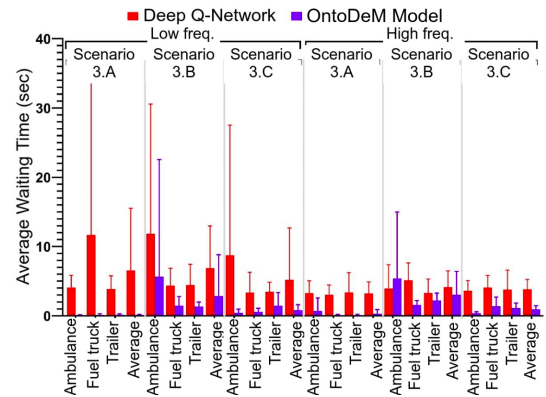

(d) The average waiting time for special vehicles in scenario 3: Deep Q-Network and OntoDeM.

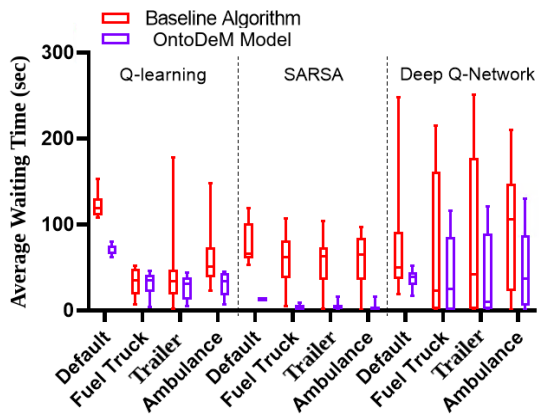

(e) The average waiting time for all vehicle types in the combination of two scenarios 1 and 3.C: baseline and OntoDeM.

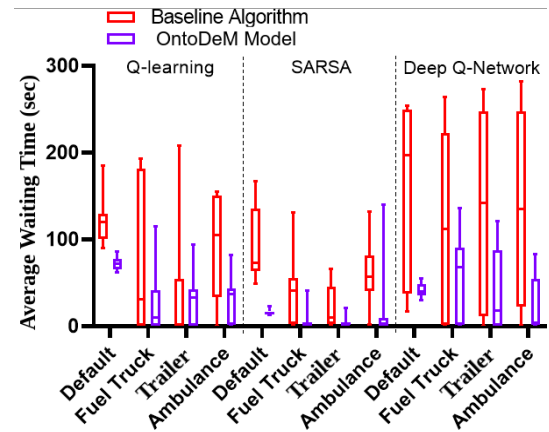

(f) The average waiting time for all vehicle types in the combination of two scenarios 2 and 3.C: baseline and OntoDeM.

Figure 6.3: The average waiting time in the ITSC-CS1, the performance comparison between the baseline algorithms and OntoDeM.

Table 6.6: Decrease in average waiting time in the ITSC-CS1, the baseline algorithms and OntoDeM.

| Type of vehicle | Scenario | Freq. | Q-learning | SARSA | DQN | AVG |
|---|---|---|---|---|---|---|
| **Ambulance** | 1 & 3.C | High | 50% | 97% | 64% | 70% |
| | 2 & 3.C | High | 31% | 90% | 66% | 62% |
| | 3.A | Low | 1% | 96% | 98% | 65% |
| | | High | 45% | 60% | 79% | 61% |
| | 3.B | Low | -12% | 82% | 52% | 41% |
| | | High | 53% | 78% | -37% | 31% |
| | 3.C | Low | 67% | 92% | 95% | 85% |
| | | High | 35% | 91% | 91% | 72% |
| **Fuel truck** | 1 & 3.C | High | 6% | 96% | 53% | 52% |
| | 2 & 3.C | High | 43% | 91% | 63% | 66% |
| | 3.A | Low | 24% | 95% | 99% | 73% |
| | | High | 49% | 98% | 97% | 81% |
| | 3.B | Low | 34% | 69% | 66% | 56% |
| | | High | -20% | 58% | 70% | 36% |
| | 3.C | Low | 87% | 58% | 84% | 76% |
| | | High | 63% | 63% | 67% | 64% |
| **Trailer** | 1 & 3.C | High | 39% | 97% | 48% | 61% |
| | 2 & 3.C | High | 21% | 94% | 79% | 65% |
| | 3.A | Low | 11% | 94% | 96% | 67% |
| | | High | 42% | 99% | 97% | 79% |
| | 3.B | Low | -9% | 56% | 69% | 39% |
| | | High | 40% | 48% | 33% | 40% |
| | 3.C | Low | 76% | 74% | 59% | 70% |
| | | High | -82% | 53% | 70% | 14% |
| **AVG Ambulance, Fuel truck, and Trailer** | 1 & 3.C | High | 32% | 97% | 55% | 61% |
| | 2 & 3.C | High | 32% | 92% | 69% | 64% |
| | 3.A | Low | 12% | 95% | 98% | 68% |
| | | High | 46% | 85% | 91% | 74% |
| | 3.B | Low | 6% | 69% | 59% | 45% |
| | | High | 35% | 60% | 26% | 40% |
| | 3.C | Low | 75% | 77% | 85% | 79% |
| | | High | 30% | 71% | 75% | 59% |
| **Default** | 1 & 3.C | High | 42% | 91% | 51% | 61% |
| | 2 & 3.C | High | 39% | 86% | 71% | 65% |
| | 4 | - | 33% | 26% | 23% | 27% |

secting roads (i.e., scenario 3.C) and it does not perform as well in simple scenarios where congestion happens (i.e., scenario 4).

Moreover, I observed that OntoDeM decreases the waiting time of ambulances more than the other types of vehicles in scenario 3.C. Also, in scenario 3.B, the improvement of waiting time for fuel trucks is higher than the other types of vehicles. Since in scenarios 3.B and 3.C, special vehicles can enter the environment in intersecting roads, higher priority is given to the road that has the most important vehicles. Therefore, when an ambulance or a fuel truck is observed on one road and a trailer on another road, the road with the ambulance or fuel truck will have a higher priority.

Table 6.7: p-value results in average waiting time in the ITSC-CS1, the baseline algorithms and OntoDeM.

| Scenario | Baseline | Vehicle type | p-value | Statistically significant |
|---|---|---|---|---|
| 1 & 3.C | Q-learning | Default | 0.000183 | ✓ |
| | | Fuel Truck | 0.623176 | ✗ |
| | | Trailer Truck | 0.472676 | ✗ |
| | | Ambulance | 0.021134 | ✓ |
| | SARSA | Default | 0.000183 | ✓ |
| | | Fuel Truck | 0.000583 | ✓ |
| | | Trailer Truck | 0.003611 | ✓ |
| | | Ambulance | 0.001706 | ✓ |
| | DQN | Default | 0.053903 | ✗ |
| | | Fuel Truck | 0.520523 | ✗ |
| | | Trailer Truck | 0.241322 | ✗ |
| | | Ambulance | 0.212294 | ✗ |
| 2 & 3.C | Q-learning | Default | 0.000183 | ✓ |
| | | Fuel Truck | 0.301041 | ✗ |
| | | Trailer Truck | 0.338267 | ✗ |
| | | Ambulance | 0.025748 | ✓ |
| | SARSA | Default | 0.000183 | ✓ |
| | | Fuel Truck | 0.009082 | ✓ |
| | | Trailer Truck | 0.058687 | ✗ |
| | | Ambulance | 0.007285 | ✓ |
| | DQN | Default | 0.088973 | ✗ |
| | | Fuel Truck | 0.241322 | ✗ |
| | | Trailer Truck | 0.053903 | ✗ |
| | | Ambulance | 0.01133 | ✓ |

Finally, from waiting time results for all types of vehicles in the system, I can conclude that the performance of the default vehicles is not compromised to increase the performance of special vehicles in scenario 3. This is particularly an important result as OntoDeM can handle unforeseen situations while keeping the performance of the system for seen situations.

**Results in the ITSC-CS2.** I trained the RL algorithm for over 300 runs and obtained the average results across all runs. For scenario 3, 15 special vehicles are generated per 300 seconds and assigned random routes. The accident is set to randomly occur around signalized intersections in the middle of the simulation. Table 6.8 (p-value results shown in Table 6.9) shows the traffic improvements of OntoDeM in the simulated urban envir-onment. Compared to the baseline RL algorithm, my OntoDeM-enabled RL algorithm can achieve up to a 12% and 5% decrease in the average vehicle waiting time and travel time, about 7% reduction in the average travel delay, and nearly 26 more vehicles completing the trips. In scenario 1, the average waiting time of OntoDeM-enabled DQN is reduced by 13% compared to the baseline DQN algorithm. Correspondingly, travel delays are reduced by 8%. Moreover, it indicates that the vehicles can travel at relatively high speeds throughout the trip. These results show that OntoDeM can address partial observation of the environment. Similarly, OntoDeM is also able to reduce the negative impact of noisy data.

Table 6.8: Traffic statistics in the ITSC-CS2.

| **Scenario** | | 1 | 2 | 3 | 4 |
|---|---|---|---|---|---|
| Average waiting time (sec) | DQN | 135.23 | 130.80 | 129.95 | 140.70 |
| | OntoDeM | 117.30 | 111.94 | 117.97 | 122.77 |
| % Decrease | | 13% | 14% | 9% | 13% |
| Average travel time (sec) | DQN | 264.28 | 262.28 | 261.33 | 269.10 |
| | OntoDeM | 252.14 | 250.81 | 250.75 | 256.81 |
| % Decrease | | 5% | 4% | 4% | 5% |
| Average travel delay (sec) | DQN | 184.09 | 181.65 | 180.66 | 189.39 |
| | OntoDeM | 170.11 | 168.60 | 168.74 | 175.53 |
| % Decrease | | 8% | 7% | 7% | 7% |
| Average number of finished trip | DQN | 1157.14 | 1163.67 | 1169.42 | 1143.58 |
| | OntoDeM | 1186.06 | 1186.98 | 1194.42 | 1171.77 |
| # Increase | | 28.92 | 23.31 | 25.00 | 28.19 |

When a vehicle with high priority, i.e., an ambulance, requests to pass through the intersection as quickly as possible, the OntoDeM-enabled DQN algorithm can adjust the traffic signal considering different vehicle types in scenario 3. Among the four types of vehicles as shown in Figure 6.4, ambulances get the most significant decrease in average waiting time (i.e., approximately 11%), fuel truck keeps the same improvement as the default vehicle, and trailers do not have any improvement due to low importance weight in urban areas compared other types of vehicles. Among the four scenarios, traffic congestion caused by unexpected road closures (i.e., scenario 4) is more serious, and as shown in Table 6.8, OntoDeM can significantly recover the traffic.

## 6.2 Edge Computing Experiment Design

To evaluate the performance of OntoDeM in an EC environment, I simulated a neighborhood consisting of a varied number of users moving according to the mobility data collected from the users' devices at the subway station in Seoul, Korea [111].

Table 6.9: p-values results in the ITSC-CS2, the DQN algorithm and OntoDeM.

| Scenario | Evaluation metric | p-value | Statistically significant |
|---|---|---|---|
| 1 | Average waiting time | 0.000498 | ✓ |
|   | Average travel delay | 0.000038 | ✓ |
|   | Average travel time | 0.000038 | ✓ |
|   | Average number of finished trip | 0.000008 | ✓ |
| 2 | Average waiting time | <0.000001 | ✓ |
|   | Average travel delay | <0.000001 | ✓ |
|   | Average travel time | <0.000001 | ✓ |
|   | Average number of finished trip | <0.000001 | ✓ |
| 3 | Average waiting time | <0.000001 | ✓ |
|   | Average travel delay | 0.000006 | ✓ |
|   | Average travel time | 0.000004 | ✓ |
|   | Average number of finished trip | 0.000003 | ✓ |
| 4 | Average waiting time | 0.000498 | ✓ |
|   | Average travel delay | 0.000038 | ✓ |
|   | Average travel time | 0.000038 | ✓ |
|   | Average number of finished trip | 0.000008 | ✓ |

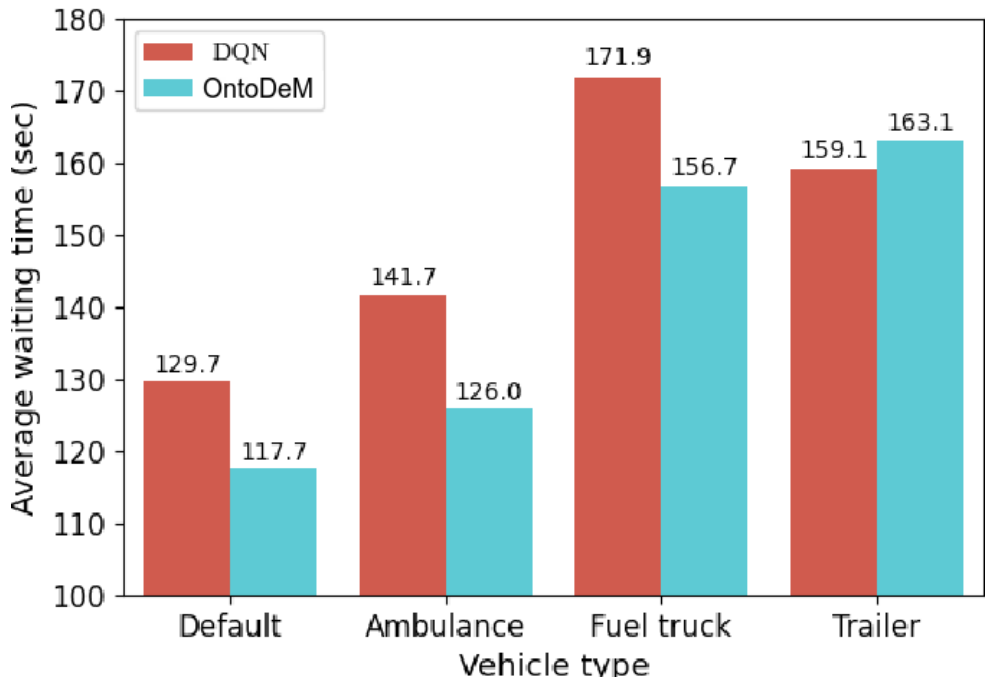

Figure 6.4: The average waiting time for different vehicle types in scenario 3 of the ITSC-CS2, the performance comparison between the DQN and OntoDeM.

These users, equipped with smartphones, wearable gadgets, and laptops, engaged in resource-intensive applications such as online games or VoIP. By implementing task offloading techniques, computationally intensive tasks from these applications were intelligently distributed and executed in the EC environment. This approach allowed for a comprehensive assessment of OntoDeM's performance in handling the complexities of diverse applications while optimizing resource utilization.

**Hardware and Operating System Configuration.** Simulations are conducted on a machine equipped with an Intel(R) Core(TM) i7-5500U CPU @ 2.40GHz (4 cores, 16 threads, 2.40 GHz base, 0.6 GHz turbo) and Intel Corporation HD Graphics 5500, running on an Ubuntu 20.04.6 LTS operating system with 7.7GB of RAM.

### 6.2.1 Evaluation Scenarios, Metrics and Baseline

**Evaluation scenarios.** To evaluate OntoDeM in the EC environment, four scenarios are defined to cover different order loads (i.e., number of orders per time step) and noise levels (See Table 6.10).

Table 6.10: Evaluation scenarios for the edge computing environment.

| Scenario | Workload | Latency-sensitive tasks |
|---|---|---|
| **10-simple** | 10 users | 5% of tasks are generated as health care, |
| **25-simple** | 25 users | 15% of them are VoIP, 55% are data collection, |
| **50-simple** | 50 users | and 25% are entertainment. |
| **10-medium** | 10 users | 10% of tasks are generated as health care, |
| **25-medium** | 25 users | 30% of them are VoIP, 35% are data collection, |
| **50-medium** | 50 users | and 25% are entertainment. |
| **10-hard** | 10 users | 20% of tasks are generated as health care, |
| **25-hard** | 25 users | 40% of them are VoIP, 10% are data collection, |
| **50-hard** | 50 users | and 30% are entertainment. |

In the base scenario, I defined that 80% of the servers can be used free of charge for task offloading, and the rest can be used at a cost. Also, 70% of the servers have a high board, and 30% of servers have a low board. 20% of servers can process four tasks simultaneously, 30% of servers process three, 30% process two, and 20% of servers can only process one task simultaneously. Additionally, users and servers are organized into three groups based on user access type. In this setting, 30% of users are in Group 1, 30% are in Group 2, and 40% are in Group 3. Also, 30% of servers are in Group 1, 30% in Group 2, and 40% in Group 3. I defined four heterogeneous tasks: (1) Healthcare tasks are: very low latency, small size, and high priority (priority 3). (2) VoIP tasks are: very low latency, big size, and middle priority (priority 2). (3) Entertainment tasks are: low latency, medium size, and low priority (priority 1). (4) Data collection tasks are: high latency, very big size, and low priority (priority 1). Also, I created a fluctuating workload, where the number of users generating tasks in the environment and the rate of the latency-sensitive tasks are not constant throughout the simulation time.

**Evaluation metrics:** The following metrics have been used to evaluate the performance of the methods:

- Task success rate (i.e., average total processed tasks) is calculated by the number of processed tasks divided by the total number of generated tasks.

- Failures in EC can arise due to network issues, incomplete processing, and due to delays. Task failure rate (i.e., average total failed tasks) is calculated by dividing the number of tasks that failed due to the delay by the number of generated tasks.

**Baseline:** To evaluate the performance of OntoDeM, I compared different task offloading strategies obtained from the baseline Deep Deterministic Policy Gradient (DDPG) algorithm [38] to the OntoDeM-enabled DDPG algorithm. Among DRL algorithms, the DDPG algorithm is a popular choice for task offloading in EC [82].

### 6.2.2 Case studies and Employed OntoDeM Methods

**EC-CS.** Users' devices offload tasks to one of 10 edge servers to obtain computation service. Edge servers are responsible for offering computational resources and processing tasks for mobile users. After a requested task has been processed, users need to receive the processed task from the edge server and offload a new task to an edge server again.

Edge servers are connected with a mesh topology which represents a fully connected network, where each edge server has direct connectivity to every other edge server. In this topology, every edge server maintains incoming and outgoing links to all other edge servers in the network. The realism of the network lies in its simulation of a dense and interconnected edge infrastructure, facilitating low-latency communication and efficient resource allocation for mobile users. The assumption is that tasks can be migrated from one edge server to another, with the constraint of limited bandwidth (1e9 byte/sec). Information flow can only exist from one side to the other (i.e., the bandwidth is shared). All simulations are run during 25 steps; each of them takes 1500 seconds[3].

**OntoDeM's Methods Employed in EC-CS.**

*Observation masking (related to S1, CH1, RO1, RQ1):* In the problem of task offloading based on RL, the user device information, including its location, type, operating system, and database, is part of the agent's observation (see Section 5.2.1). However, according to the EC ontology's inference rules shown in Table 6.11, only the device location must be kept, and the rest can be masked.

---

[3]The code of OntoDeM for edge computing environment is publicly available: https://github.com/saeedehghanadbashi/ontology-based-RL

Table 6.11: Masking irrelevant user device information, an example of inference rules.

| Inference rules |
| --- |
| EdgeServer(?s), User(?u), Task(?t), EdgeDevice(?d), isService(TaskOffloading), DeviceLocation(?l), DeviceOperatingSystem(?o), DeviceDatabase(?b), DeviceType(?v), hasEdgeDevice(?u, ?d), hasLocation(?d, ?l), hasType(?d, ?v), hasOperatingSystem(?d, ?o), hasDatabase(?d, ?b), hasTask(?u, ?t), requestFrom(?u, ?s) |
| $->$ isRelevant(?l), isIrrelevant(?o), isIrrelevant(?b), isIrrelevant(?v) |

According to the scenario setting discussed earlier, some edge servers can be accessed at a cost. In this case, users' payment information is required during the task offloading process. However, when using the free-of-charge servers, the users' payment information can be masked, according to the EC ontology's inference rules, as shown in Table 6.12.

Table 6.12: Masking user card number for free-of-charge servers, an example of inference rules.

| Inference rules |
| --- |
| Server(?s), User(?u), Task(?t), CardNumber(?c), isService(TaskOffloading), hasCardNumber(?u, ?c), hasTask(?u, ?t), hasServerCost(?s, FreeServer), requestFrom(?u, ?s) |
| $->$ isIrrelevant(?c) |

*Action masking (related to S3, CH2, RO3, RQ3):* Latency can be significantly reduced when the edge servers are closer to the place where data or task is generated. Using the inference rules of the proposed EC ontology (see Table 6.13), those servers far from the users are masked by giving the zero selection probability to the inaccessible servers, and the OntoDeM-enabled DDPG agent will not choose them.

Table 6.13: Masking inaccessible servers, an example of inference rules.

| Inference rules |
| --- |
| Server(?s), User(?u), Task(?t), ServerBoard(?b), Distance(?d), isService(TaskOffloading), hasServerBoard(?s, ?b), hasTask(?u, ?t), requestFrom(?u, ?s), hasDistance(?u, ?s, ?d), isGreaterThan(?d, ?b) |
| $->$ isInaccessibleFor(?s, ?u) |

Furthermore, considering the servers' maximum capacity (server limit) is essential to balance the servers' workload. Using the predefined rules shown in Table 6.14, servers with computationally intensive workloads (i.e., unavailable servers) can be masked.

This prevents servers from becoming overworked, which could cause them to slow down, drop requests, and even crash.

Table 6.14: Masking unavailable servers, an example of inference rules.

| **Inference rules** |
| --- |
| Server(?s), User(?u), Task(?t), ServerLimit(?l), ServerWorkload(?w), isService(TaskOffloading), hasServerLimit(?s, ?l), hasServerWorkload(?s, ?w), hasTask(?u, ?t), requestFrom(?u, ?s), isEqualTo(?l, ?w) |
| $->$ isUnavailableFor(?s, ?u) |

*Action prioritization (related to S3, CH2, RO3, RQ3):* Edge devices can be grouped based on user access type before getting access to the servers of the EC environment to satisfy security requirements. According to the inference rules shown in Table 6.15, OntoDeM prioritizes users' access to the servers in the same group.

Table 6.15: Prioritizing users' access to the servers in the same group, an example of inference rules.

| **Inference rules** |
| --- |
| Server(?$s_1$), Server(?$s_2$), User(?u), Task(?t), Group(?$g_1$), Group(?$g_2$), isService(TaskOffloading), hasTask(?u, ?t), inGroup(?u, ?$g_1$), inGroup(?$s_1$, ?$g_1$), inGroup(?$s_2$, ?$g_2$), requestFrom(?u, ?$s_1$), requestFrom(?u, ?$s_2$) |
| $->$ hasPriorityOver(?$s_1$, ?$s_2$) |

*Execution prioritization (related to S3, CH2, RO3, RQ3):* Moreover, in the EC environment, since not all task requests from edge servers can be scheduled on time, thus, guaranteeing fairness among the users (i.e., edge devices offloading tasks) while considering the priorities of the tasks becomes a critical issue. Based on the inference rules, OntoDeM proposes four execution prioritization strategies, namely the user fair, priority fair, application type fair, and latency fair, that account for the user usage history (i.e., the number of user tasks that have been processed), task priorities, application types, and task latency requirements respectively (see Tables 6.16-6.19). So, the tasks are first prioritized, and then the servers are assigned to them.

Table 6.16: Prioritizing tasks based on usage history, an example of inference rules.

| **Inference rules** |
| --- |
| Server(?s), User(?$u_1$), User(?$u_2$), Task(?$t_1$), Task(?$t_2$), UsageHistory(?$h_1$), UsageHistory(?$h_2$), isService(TaskOffloading), hasTask(?$u_1$, ?$t_1$), hasTask(?$u_2$, ?$t_2$), hasUsageHistory(?$u_1$, ?$h_1$), hasUsageHistory(?$u_2$, ?$h_2$), requestFrom(?$u_1$, ?s), requestFrom(?$u_2$, ?s), isLessThan(?$h_1$, ?$h_2$) |
| $->$ hasPriorityOver(?$u_1$, ?$u_2$) |

Table 6.17: Prioritizing tasks based on task priority, an example of inference rules.

| Inference rules |
|---|
| Server(?s), User($?u_1$), User($?u_2$), Task($?t_1$), Task($?t_2$), Priority($?p_1$), Priority($?p_2$), isService(TaskOffloading), hasTask($?u_1$, $?t_1$), hasTask($?u_2$, $?t_2$), hasPriority($?t_1$, $?p_1$), hasPriority($?t_2$, $?p_2$), requestFrom($?u_1$, $?s$), requestFrom($?u_2$, $?s$), isGreaterThan($?p_1$, $?p_2$) $->$ hasPriorityOver($?u_1$, $?u_2$) |

Table 6.18: Prioritizing tasks based on application type, an example of inference rules.

| Inference rules |
|---|
| Server(?s), User($?u_1$), User($?u_2$), Task($?t_1$), Task($?t_2$), isService(TaskOffloading), hasTask($?u_1$, $?t_1$), hasTask($?u_2$, $?t_2$), hasApplicationType($?t_1$, HealthCare), hasApplicationType($?t_2$, Entertainment), requestFrom($?u_1$, $?s$), requestFrom($?u_2$, $?s$) $->$ hasPriorityOver($?u_1$, $?u_2$) |

Table 6.19: Prioritizing tasks based on task latency requirement, an example of inference rules.

| Inference rules |
|---|
| Server(?s), User($?u_1$), User($?u_2$), Task($?t_1$), Task($?t_2$), Latency($?l_1$), Latency($?l_2$), isService(TaskOffloading), hasTask($?u_1$, $?t_1$), hasTask($?u_2$, $?t_2$), hasLatency($?t_1$, $?l_1$), hasLatency($?t_2$, $?l_2$), requestFrom($?u_1$, $?s$), requestFrom($?u_2$, $?s$), isLessThan($?l_1$, $?l_2$) $->$ hasPriorityOver($?u_1$, $?u_2$) |

### 6.2.3 Results Analysis

**Results in the EC-CS.** The average total processed and failed tasks in 10 runs for the proposed methods in each scenario are reported in the following. The results show that the OntoDeM-enabled DDPG increases the average number of total processed tasks and decreases the average number of total failed tasks compared to the DDPG algorithm.

I observed that masking users' device information as irrelevant information increases the average total processed tasks by 7%, 8%, and 7% in simple, medium, and hard scenarios, respectively (see Figure 6.5, p-value results shown in Table 6.20). This improvement is 4%, 8%, and 9% for scenarios with 10, 25, and 50 users. The percentage increase is more significant when the number of users increases. The average total failed tasks decreased by 6%, 3%, and 8% in simple, medium, and hard scenarios and 8%, 6%, and 3% in the scenarios with 10, 25, and 50 users, respectively. This indicates that there is a higher percentage decrease in the scenarios with less number of users.

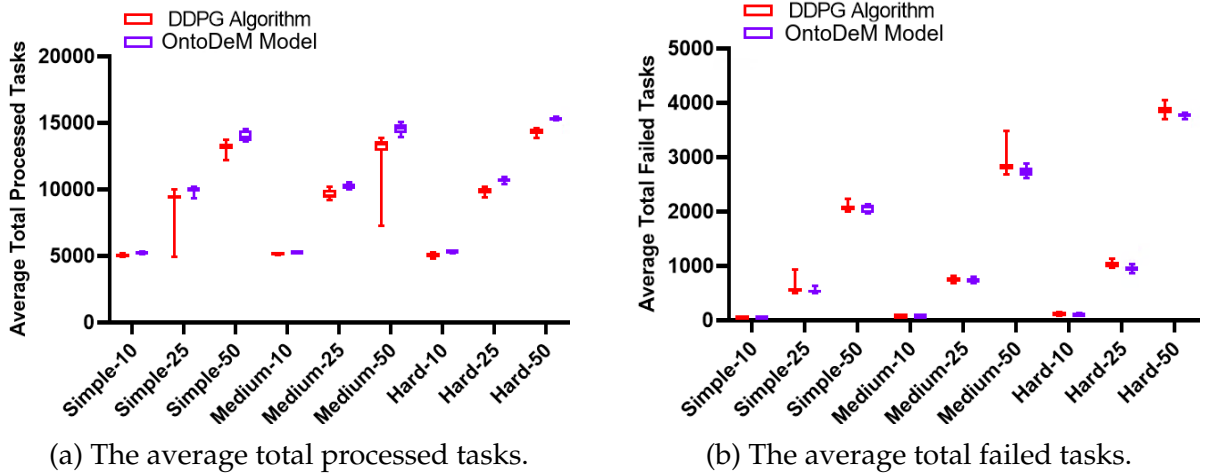

(a) The average total processed tasks.    (b) The average total failed tasks.

Figure 6.5: Masking irrelevant user device information, the performance comparison between the DDPG algorithm and OntoDeM.

Table 6.20: Masking irrelevant user device information, p-value results, the DDPG algorithm and OntoDeM.

| Evaluation metric | Scenario | p-value | Statistically significant |
|---|---|---|---|
| Total processed tasks | Simple-10 | 0.000076 | ✓ |
| | Simple-25 | 0.001505 | ✓ |
| | Simple-50 | 0.00013 | ✓ |
| | Medium-10 | 0.000011 | ✓ |
| | Medium-25 | 0.00105 | ✓ |
| | Medium-50 | 0.000011 | ✓ |
| | Hard-10 | 0.000022 | ✓ |
| | Hard-25 | 0.000011 | ✓ |
| | Hard-50 | 0.000011 | ✓ |
| Total failed tasks | Simple-10 | 0.060058 | ✗ |
| | Simple-25 | 0.247451 | ✗ |
| | Simple-50 | 0.911797 | ✗ |
| | Medium-10 | 0.435872 | ✗ |
| | Medium-25 | 0.528849 | ✗ |
| | Medium-50 | 0.035463 | ✓ |
| | Hard-10 | 0.063013 | ✗ |
| | Hard-25 | 0.011496 | ✓ |
| | Hard-50 | 0.011496 | ✓ |

I observed that masking users' card numbers as irrelevant information increases the average total processed tasks by 8%, 13%, and 4% in simple, medium, and hard scenarios, respectively. This improvement is 13%, 5%, and 7% for scenarios with 10, 25, and 50 users. The improvement is lower when the number of low-sensitive tasks increases. Also, the percentage increase is lower when the number of users increases. This is because the number of paid servers for which the user card number should be included in the observation increases, and the observation in both methods (i.e., baseline and OntoDeM) will be similar. The same results are achieved in the comparison of the average total failed tasks so that it decreases by 8%, 9%, and 3% in simple, medium, and hard scenarios and 11%, 5%, and 5% in the scenarios with 10, 25, and 50 users respectively (see Figure 6.6, p-value results shown in Table 6.21).

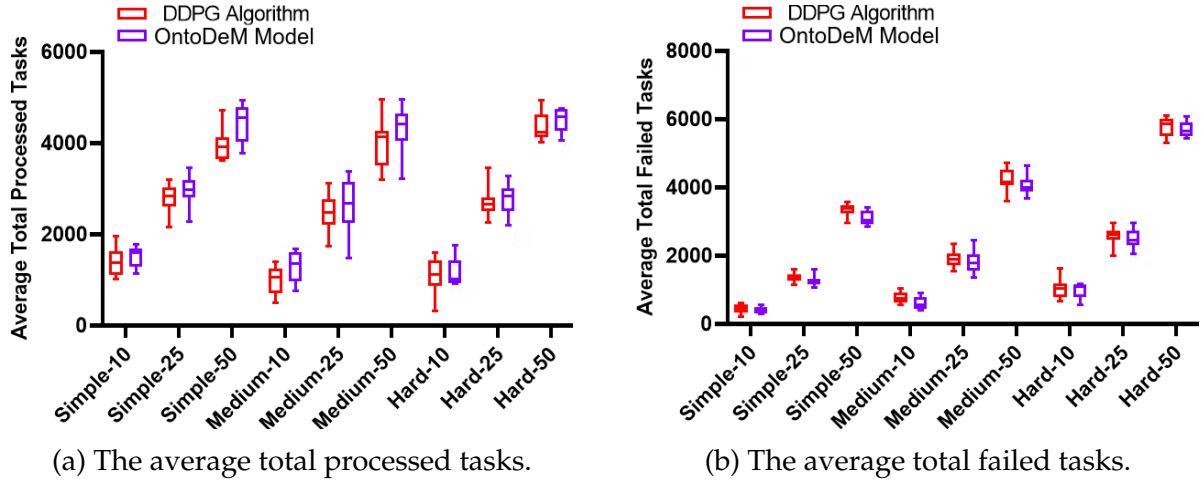

(a) The average total processed tasks.  (b) The average total failed tasks.

Figure 6.6: Masking user card number for free-of-charge servers, the performance comparison between the DDPG algorithm and OntoDeM.

Table 6.21: Masking user card number for free-of-charge servers, p-value results, the DDPG algorithm and OntoDeM.

| Evaluation metric | Scenario | p-value | Statistically significant |
|---|---|---|---|
| | Simple-10 | 0.352681 | ✗ |
| | Simple-25 | 0.217563 | ✗ |
| | Simple-50 | 0.008931 | ✓ |
| | Medium-10 | 0.063013 | ✗ |
| Total processed tasks | Medium-25 | 0.435872 | ✗ |
| | Medium-50 | 0.123005 | ✗ |
| | Hard-10 | 0.970512 | ✗ |
| | Hard-25 | 0.435872 | ✗ |
| | Hard-50 | 0.279861 | ✗ |
| | Simple-10 | 0.393048 | ✗ |
| | Simple-25 | 0.123005 | ✗ |
| | Simple-50 | 0.011496 | ✓ |
| | Medium-10 | 0.063013 | ✗ |
| Total failed tasks | Medium-25 | 0.435872 | ✗ |
| | Medium-50 | 0.165494 | ✗ |
| | Hard-10 | 0.970512 | ✗ |
| | Hard-25 | 0.435872 | ✗ |
| | Hard-50 | 0.435872 | ✗ |

The results show restricting the number of servers that can be assigned to each user based on the server board and user request latency requirement significantly increases the average total processed tasks and decreases the average total failed tasks compared to the baseline algorithm. The average total processed tasks increased by 18%, 16%, and 14% in simple, medium, and hard scenarios, respectively. This improvement is 23%, 16%, and 9% for scenarios with 10, 25, and 50 users. When the number of users increases, the number of available servers that can be assigned to each user based on the appropriate distance is limited, so less improvement can be the result. The same effects are observed in the percentage of decreases in the average total failed tasks by 32%, 29%, and 29% in simple, medium, and hard scenarios and 50%, 29%, and 11% in the scenarios with 10, 25, and 50 users respectively (see Figure 6.7, p-value results shown in Table 6.22).

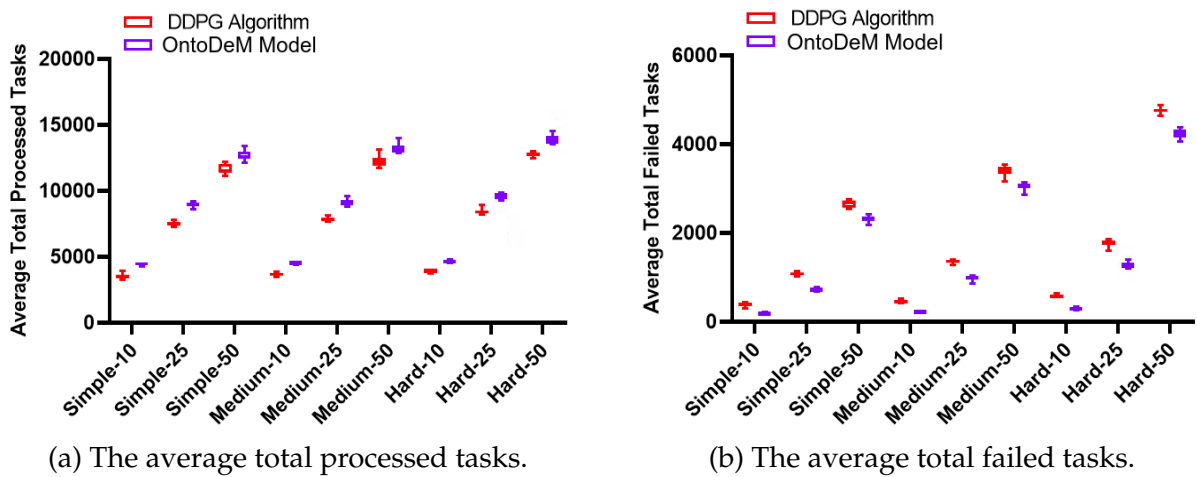

(a) The average total processed tasks.   (b) The average total failed tasks.

Figure 6.7: Masking inaccessible servers, the performance comparison between the DDPG algorithm and OntoDeM.

In Figure 6.8 (p-value results shown in Table 6.23), it is shown that masking the action set based on the server limit increases the average total processed tasks by 9%, 8%, and 7% in simple, medium, and hard scenarios and 8%, 8%, and 7% for scenarios with 10, 25, and 50 users respectively. Also, the proposed model significantly decreases the average total failed tasks by 77%, 69%, and 72% in simple, medium, and hard scenarios and 88%, 83%, and 47% in the scenarios with 10, 25, and 50 users, respectively. The reason for the significant improvement in the average total failed tasks is that when load balancing is considered, it will prevent a server from saturating more than its limit, thus failing the tasks.

The results indicate that prioritizing the servers for each user based on server group significantly increases the average total processed tasks satisfying security requirements compared to the baseline algorithm. The percentage increase is 82%, 78%, and

Table 6.22: Masking inaccessible servers, p-value results, the DDPG algorithm and OntoDeM.

| Evaluation metric | Scenario | p-value | Statistically significant |
|---|---|---|---|
| Total processed tasks | Simple-10 | 0.000011 | ✓ |
| | Simple-25 | 0.000011 | ✓ |
| | Simple-50 | 0.000022 | ✓ |
| | Medium-10 | 0.000011 | ✓ |
| | Medium-25 | 0.000011 | ✓ |
| | Medium-50 | 0.00013 | ✓ |
| | Hard-10 | 0.000011 | ✓ |
| | Hard-25 | 0.000011 | ✓ |
| | Hard-50 | 0.000011 | ✓ |
| Total failed tasks | Simple-10 | 0.000011 | ✓ |
| | Simple-25 | 0.000011 | ✓ |
| | Simple-50 | 0.000011 | ✓ |
| | Medium-10 | 0.000011 | ✓ |
| | Medium-25 | 0.000011 | ✓ |
| | Medium-50 | 0.000011 | ✓ |
| | Hard-10 | 0.000011 | ✓ |
| | Hard-25 | 0.000011 | ✓ |
| | Hard-50 | 0.000011 | ✓ |

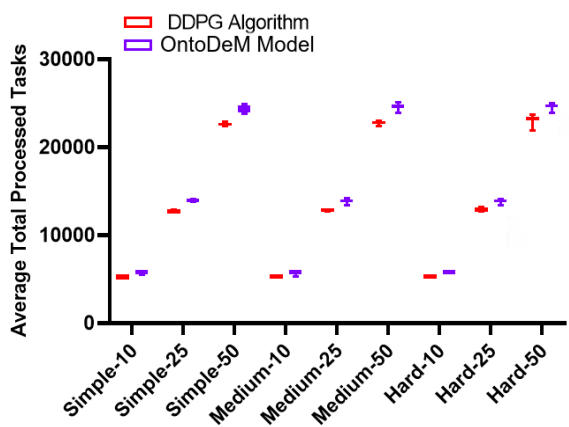
(a) The average total processed tasks.

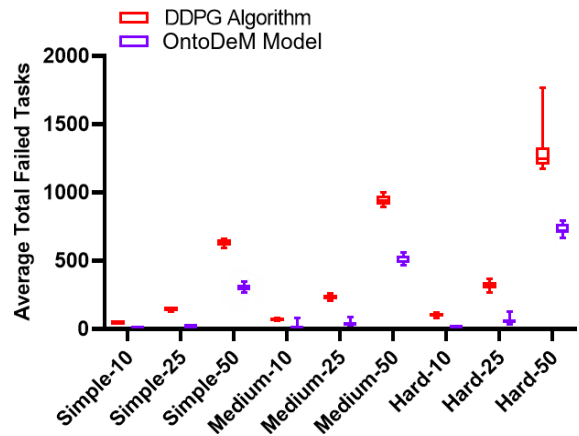
(b) The average total failed tasks.

Figure 6.8: Masking unavailable servers, the performance comparison between the DDPG algorithm and OntoDeM.

84% in simple, medium, and hard scenarios and 78%, 83%, and 83% for scenarios with 10, 25, and 50 users respectively (see Figure 6.9a, p-value results shown in Table 6.24). The average total processed tasks and average total failed tasks are not compromised to increase the satisfying security requirements (see Figures 6.9b and 6.9c). This is a significant result that OntoDeM improves satisfying security requirements while maintaining overall system performance.

Table 6.23: Masking unavailable servers, p-value results, the DDPG algorithm and OntoDeM.

| Evaluation metric | Scenario | p-value | Statistically significant |
|---|---|---|---|
| Total processed tasks | Simple-10 | 0.000011 | ✓ |
| | Simple-25 | 0.000011 | ✓ |
| | Simple-50 | 0.000011 | ✓ |
| | Medium-10 | 0.000725 | ✓ |
| | Medium-25 | 0.000011 | ✓ |
| | Medium-50 | 0.000011 | ✓ |
| | Hard-10 | 0.000011 | ✓ |
| | Hard-25 | 0.000011 | ✓ |
| | Hard-50 | 0.000011 | ✓ |
| Total failed tasks | Simple-10 | 0.000011 | ✓ |
| | Simple-25 | 0.000011 | ✓ |
| | Simple-50 | 0.000011 | ✓ |
| | Medium-10 | 0.000465 | ✓ |
| | Medium-25 | 0.000011 | ✓ |
| | Medium-50 | 0.000011 | ✓ |
| | Hard-10 | 0.000011 | ✓ |
| | Hard-25 | 0.000011 | ✓ |
| | Hard-50 | 0.000011 | ✓ |

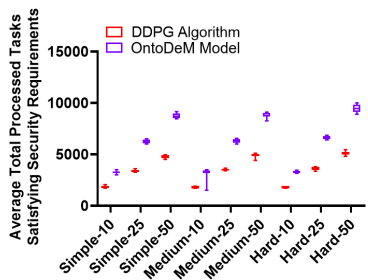
(a) The average total processed tasks satisfying security requirements.

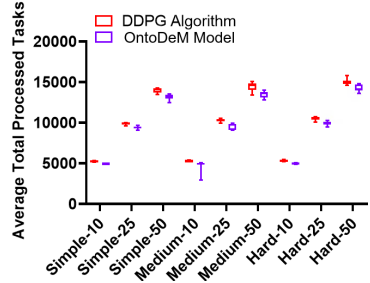
(b) The average total processed tasks.

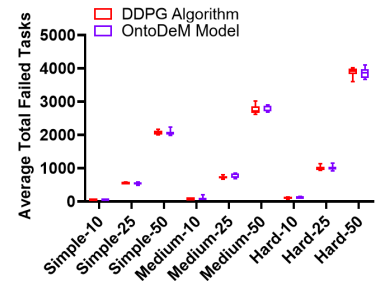
(c) The average total failed tasks.

Figure 6.9: Prioritizing users' access to the servers in the same group, the performance comparison between the DDPG algorithm and OntoDeM.

The results indicate that prioritizing the execution of the actions based on usage history significantly decreases the average standard deviation of usage history compared to the baseline algorithm. The percentage decrease is 64%, 62%, and 64% in simple, medium, and hard scenarios, respectively. This improvement is 44%, 67%, and 79% for scenarios with 10, 25, and 50 users respectively (see Figure 6.10a, p-value results shown in Table 6.25). So, the proposed model shows more improvement when the number of users increases. Also, I saw that the average total processed tasks and average total failed tasks are not compromised to decrease the standard deviation of usage history

Table 6.24: Prioritizing users' access to the servers in the same group, p-value results, the DDPG algorithm and OntoDeM.

| Evaluation metric | Scenario | p-value | Statistically significant |
|---|---|---|---|
| | Simple-10 | 0.000011 | ✓ |
| | Simple-25 | 0.000011 | ✓ |
| | Simple-50 | 0.000011 | ✓ |
| | Medium-10 | 0.001505 | ✓ |
| Total processed tasks | Medium-25 | 0.000011 | ✓ |
| satisfying security requirements | Medium-50 | 0.000011 | ✓ |
| | Hard-10 | 0.000011 | ✓ |
| | Hard-25 | 0.000011 | ✓ |
| | Hard-50 | 0.000011 | ✓ |

(see Figures 6.10b and 6.10c). This is particularly important as OntoDeM can guarantee fairness among the users while keeping the whole system's performance.

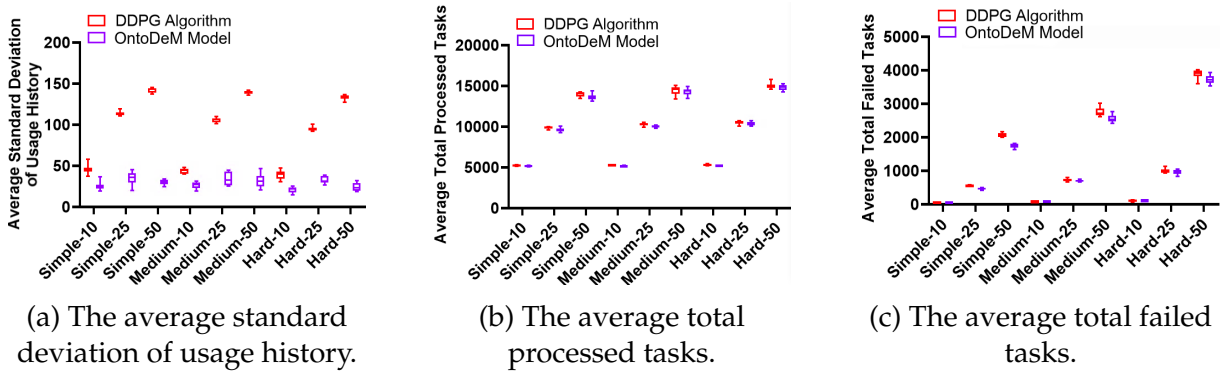

(a) The average standard deviation of usage history.

(b) The average total processed tasks.

(c) The average total failed tasks.

Figure 6.10: Prioritizing tasks based on usage history, the performance comparison between the DDPG algorithm and OntoDeM.

Table 6.25: Prioritizing tasks based on usage history, p-value results, the DDPG algorithm and OntoDeM.

| Evaluation metric | Scenario | p-value | Statistically significant |
|---|---|---|---|
| | Simple-10 | 0.000011 | ✓ |
| | Simple-25 | 0.000011 | ✓ |
| | Simple-50 | 0.000011 | ✓ |
| | Medium-10 | 0.000011 | ✓ |
| Standard deviation of | Medium-25 | 0.000011 | ✓ |
| usage history | Medium-50 | 0.000011 | ✓ |
| | Hard-10 | 0.000011 | ✓ |
| | Hard-25 | 0.000011 | ✓ |
| | Hard-50 | 0.000011 | ✓ |

According to the results shown in Figure 6.11 (p-value results shown in Table 6.26), prioritizing execution of the actions based on the task priority decreases the average total failed tasks by 25%. The percentage decrease for tasks with priority 2 is 69%, 58%, and 31% in simple, medium, and hard scenarios, respectively. This improvement is 43%, 56%, and 59% for scenarios with 10, 25, and 50 users. In the tasks with the highest priority (i.e., priority 3), the average total failed tasks decreased by 81%, 83%, and 73% in simple, medium, and hard scenarios and 67%, 82%, and 88% for scenarios with 10, 25, and 50 users respectively. OntoDeM decreases the average failure rate for the tasks with priority 3 more than those with priority 2. This is because when important tasks with the highest priority are generated in the EC environment, the execution prioritization step prioritizes them over other tasks to assign the requested server to them.

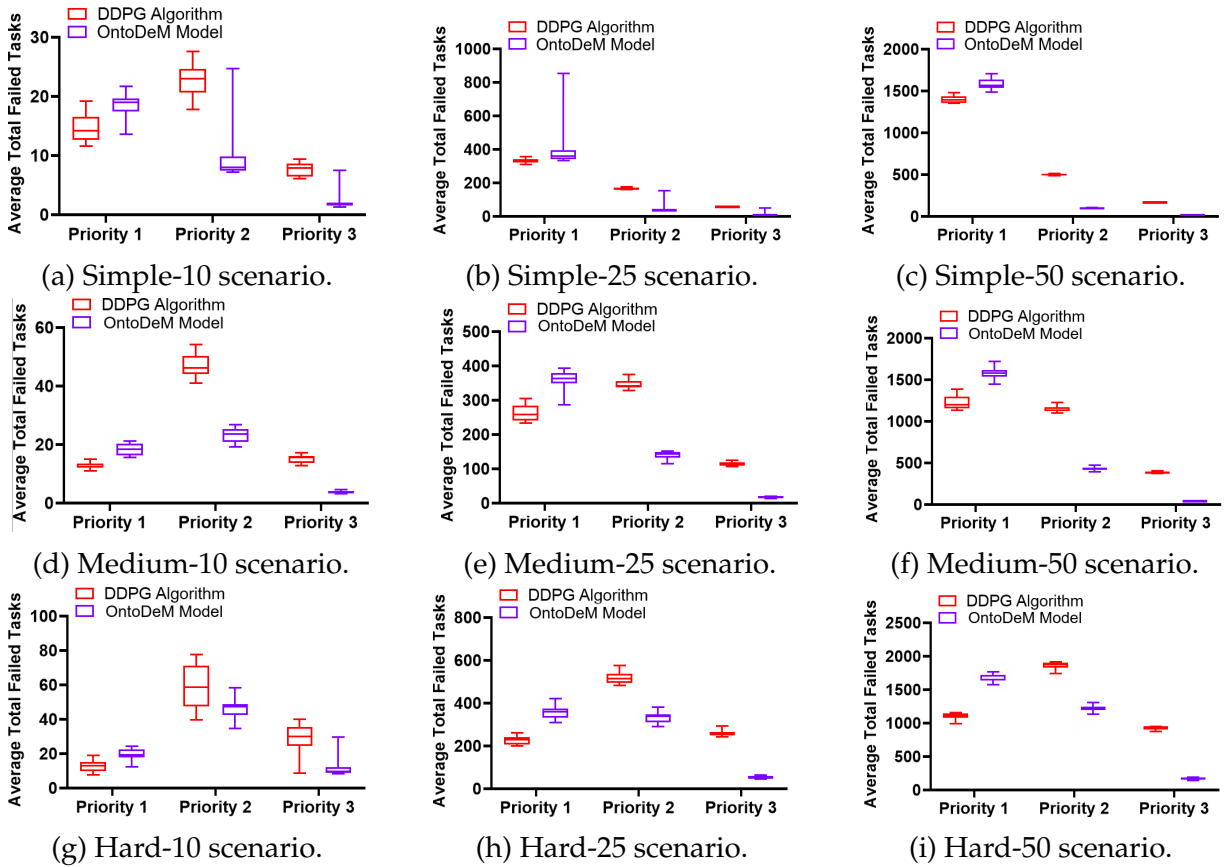

Figure 6.11: Prioritizing tasks based on task priority, the performance comparison between the DDPG algorithm and OntoDeM
(Priority 3: Remote health care, Priority 2: VoIP, Priority 1: Data collection, Priority 1: Entertainment).

The results show that the execution prioritization based on the application type decreases the average total failed tasks overall by 31% (see Figure 6.12, p-value results shown in Table 6.27). The percentage decrease for remote healthcare tasks (i.e., App Type 1) is 79%, 80%, and 67% in simple, medium, and hard scenarios, respectively. This improvement is 67%, 80%, and 79% for scenarios with 10, 25, and 50 users, re-spectively. In the VOIP tasks (i.e., App Type 2), the average total failed tasks decreased by 69%, 56%, and 28% in simple, medium, and hard scenarios and 44%, 54%, and 54% for scenarios with 10, 25, and 50 users respectively. OntoDeM decreases the average failure rate for remote healthcare tasks more than VOIP tasks. The reason is that the real-time requirement for remote healthcare applications is essential, and any delay is unacceptable. Hence, my model gives high priority to the remote healthcare applications in the task offloading service.

Table 6.26: Prioritizing tasks based on task priority, p-value results, the DDPG algorithm and OntoDeM.

| Evaluation metric | Scenario | p-value | Statistically significant |
|---|---|---|---|
| | Simple-10 | 0.000206 | ✓ |
| | Simple-25 | 0.001505 | ✓ |
| | Simple-50 | 0.000011 | ✓ |
| | Medium-10 | 0.000011 | ✓ |
| Total failed tasks | Medium-25 | 0.000011 | ✓ |
| | Medium-50 | 0.000011 | ✓ |
| | Hard-10 | 0.023231 | ✓ |
| | Hard-25 | 0.000011 | ✓ |
| | Hard-50 | 0.000011 | ✓ |

Table 6.27: Prioritizing tasks based on application type, p-value results, the DDPG algorithm and OntoDeM.

| Evaluation metric | Scenario | p-value | Statistically significant |
|---|---|---|---|
| | Simple-10 | 0.000011 | ✓ |
| | Simple-25 | 0.000011 | ✓ |
| | Simple-50 | 0.01469 | ✓ |
| | Medium-10 | 0.000011 | ✓ |
| Total failed tasks | Medium-25 | 0.001505 | ✓ |
| | Medium-50 | 0.000011 | ✓ |
| | Hard-10 | 0.018543 | ✓ |
| | Hard-25 | 0.000011 | ✓ |
| | Hard-50 | 0.14314 | ✗ |

Figure 6.13 (p-value results shown in Table 6.28) shows that the execution prioritization based on task latency decreases the average total failed tasks by 33%. The percentage decrease for very low latency tasks is 75%, 63%, and 44% in simple, medium, and hard scenarios and 53%, 63%, and 66% for scenarios with 10, 25, and 50 users, respectively. In the low latency tasks, the percentage decrease is 70%, 36%, and -33% in simple, medium, and hard scenarios and 25%, 23%, and 25% for scenarios with 10, 25, and 50 users, respectively. I observed that OntoDeM decreases the average failure rate for very low latency tasks more than others. Also, the improvement in the hard scenario is lower than in medium and simple scenarios. This is because the number of very low latency tasks increases, and my proposed model shows less improvement due to the limited number of servers.

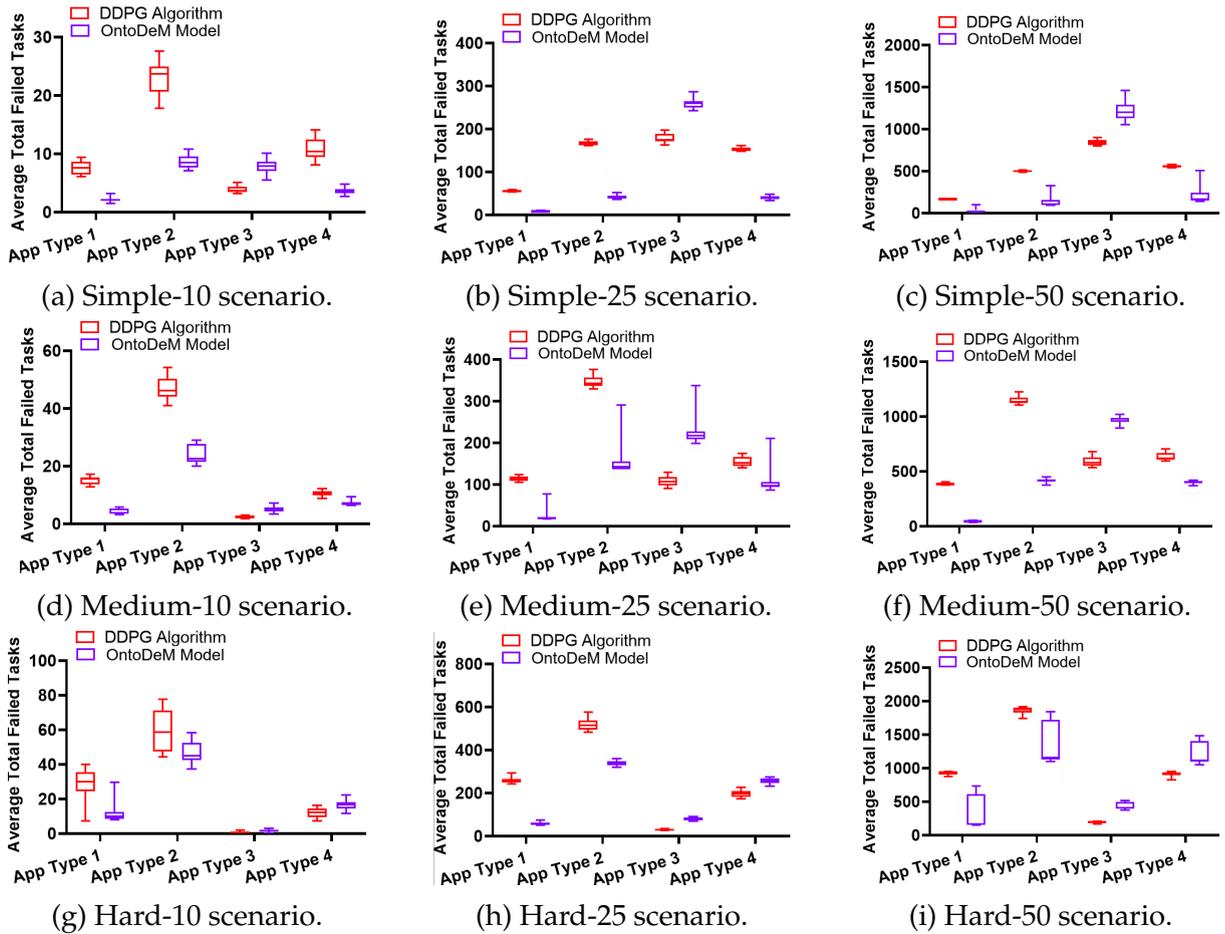

Figure 6.12: Prioritizing tasks based on application type, the performance comparison between the DDPG algorithm and OntoDeM
(App Type 1: Remote health care, App Type 2: VoIP, App Type 3: Data collection, App Type 4: Entertainment).

## 6.3  Job Shop Scheduling Experiment Design

I have simulated a JSS environment consisting of three sources, $I_1$, $I_2$, and $I_3$ to generate orders and a number of machines processing orders based on the specified sequence of operations. Each machine has one processing capacity, so only one order can be processed at a time. Machines are categorized into three groups, $n_1$, $n_2$, and $n_3$, placed at three work areas, $z_1$, $z_2$, and $z_3$. The processed orders are consumed by three sinks, $X_1$, $X_2$, and $X_3$. The simulation model of production planning and control of complex job shop systems implemented by [142] is used to compare the performance of the baseline algorithm and OntoDeM-enabled baseline algorithm. Simulated episodes are set to 1000, each with 100 simulation steps.

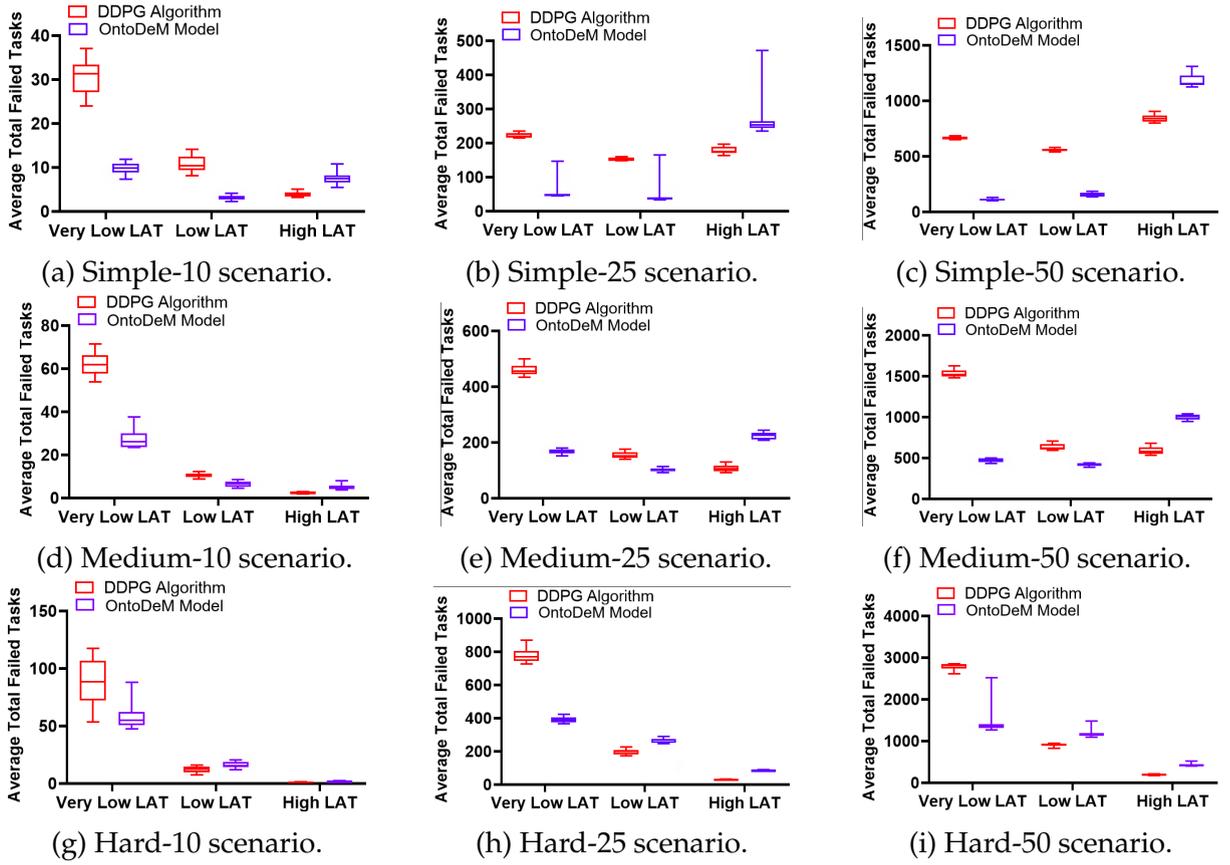

Figure 6.13: Prioritizing tasks based on task latency requirement, the performance comparison between the DDPG algorithm and OntoDeM.

Table 6.28: Prioritizing tasks based on task latency requirement, p-value results, the DDPG algorithm and OntoDeM.

| Evaluation metric | Scenario | p-value | Statistically significant |
|---|---|---|---|
|  | Simple-10 | 0.000011 | ✓ |
|  | Simple-25 | 0.001505 | ✓ |
|  | Simple-50 | 0.000011 | ✓ |
|  | Medium-10 | 0.000011 | ✓ |
| Total failed tasks | Medium-25 | 0.000011 | ✓ |
|  | Medium-50 | 0.000011 | ✓ |
|  | Hard-10 | 0.006722 | ✓ |
|  | Hard-25 | 0.000011 | ✓ |
|  | Hard-50 | 0.001505 | ✓ |

**Hardware and Operating System Configuration.** Simulations are conducted on a machine equipped with an Intel Core i5-4200U CPU (2 cores, 4 threads, 1.60 GHz base) and an NVIDIA GeForce GT 740M, running on a Windows 10 operating system with 4GB of RAM.

### 6.3.1 Evaluation Scenarios, Metrics and Baseline

**Evaluation scenarios:** To evaluate OntoDeM in the JSS environment, six scenarios are defined to cover different order loads (i.e., number of orders per time step), with different due date requirements and different noise levels (see Table 6.29). Also, I defined 30%, 50%, and 20% of orders with low, medium, and high priority, respectively. The failure rate for 50% of machines is low, for 30% is medium, and for 20% is high.

Table 6.29: Evaluation scenarios for job shop scheduling environment.

|  |  | Description |
|---|---|---|
| **Order load** | Light | Three orders are generated at each time step. |
|  | Heavy | Six orders are generated at each time step. |
| **Due date level** | High | 5% of orders have a low due date, 80% of them have a medium due date, and 15% have a high due date. |
|  | Low | 25% of orders have a low due date, 70% of them have a medium due date, and 5% have a high due date. |
| **Noise level** | Low | Noisy capacity data: In 0.5% of cases where a machine's capacity is full, it is observed as free. Noisy working time data: In two of eight randomly selected machines, the working time is randomly noised with a number between zero and four. |
|  | High | Noisy capacity data: In 1% of cases where a machine's capacity is full, it is observed as free. Noisy status data: The status of four of eight randomly selected machines has noise, i.e., if the machine is a failure, it will be observed as no failure. |

**Evaluation metrics:** The following metrics are used to evaluate the performance of OntoDeM:

- Average utilization rate of machines (see Equation 5.14).

- Average waiting time $\overline{w}$: The average waiting time of orders is computed as follows:

$$\overline{w} = \frac{1}{|D|} \sum_{i=0}^{|D|-1} w_{d_i}^t \qquad (6.1)$$

Where The notation $w_{d_i}^t$ represents the waiting time for order $d_i$ at time step $t$, while $|D|$ represents the total number of orders.

- Total failed orders: The number of orders that failed due to delay (i.e., due date requirement).

- Total processed orders: The number of orders successfully processed.

**Baselines:** The baseline method is based on the framework for job shop manufacturing systems proposed by [142, 143]. In the *JSS-CS1*, the TRPO algorithm is selected as the baseline algorithm. In the JSS domain, small changes can have significant consequences on production efficiency and resource utilization, for example, delaying the processing of one order may lead to cascading delays for subsequent orders, impacting overall throughput and efficiency. So, stability and incremental improvements are essential for maintaining smooth operations, adapting to dynamic conditions, and optimizing overall system performance. TRPO ensures stable updates to the policy by constraining the size of policy updates within a trust region. This stability ensures that policy improvements are monotonic, which is crucial for scheduling tasks where small changes can have significant consequences.

In *JSS-CS2*, different strategies are used for different tasks of the job shop scheduler agent as follows:

- Random strategy for tasks $\tau_{u_1}$, $\tau_{u_2}$, and $\tau_{u_5}$ which selects a work area, group, or sink randomly.

- LIFO strategy for task $\tau_{u_0}$, which selects the last order entered into the queue for resource allocation.

- FIFO strategy for task $\tau_{u_4}$, which selects the first order entered into the queue for processing.

- RL strategy for task $\tau_{u_3}$, which selects the machine with the highest reward to process an order.

### 6.3.2 Case studies and Employed OntoDeM Methods

**JSS-CS1.** This case study consists of eight machines, $\{m_1, m_2, \ldots, m_8\}$ (see Figure 6.14). In this case study, OntoDeM's performance is evaluated across different order loads and noise levels.

**OntoDeM's Methods Employed in JSS-CS1.**

*Observation augmentation (related to S1, CH1, RO1, RQ1):* In the JSS environment, several inference rules can be used by the job shop scheduler agent. Table 6.30 describes the inference rules that can infer the implicit observation related to the capacity of machine $a$. Suppose that the sum of orders in the input buffer $x$, the output buffer $y$, and the processing buffer $z$ of the machine is less than the machine's capacity $c$. In that case, the machine's remaining capacity is free.

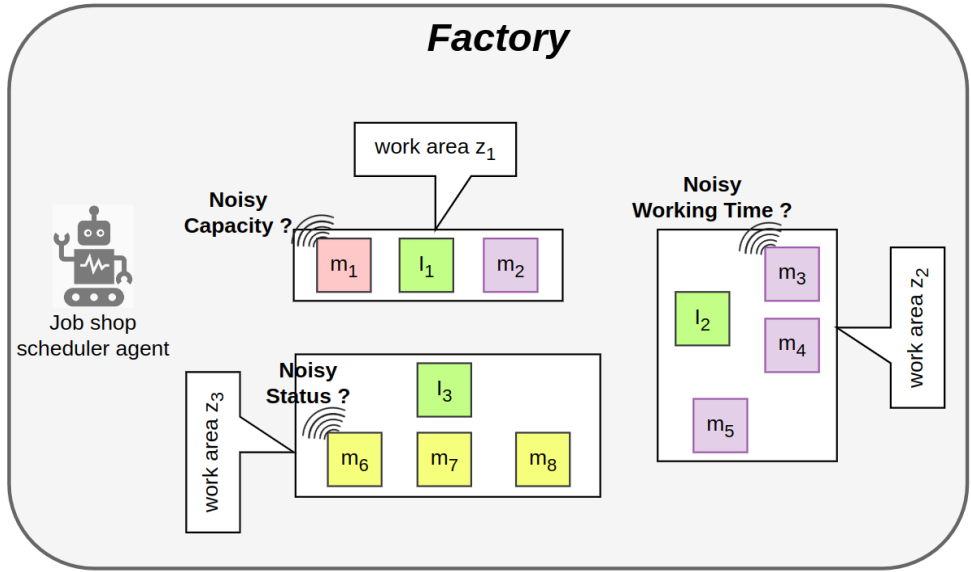

Figure 6.14: Simulated job shop scheduling environment, JSS-CS1. The machines of group $n_1$ are marked with purple color, group $n_2$ with pink color, and group $n_3$ with yellow color. There is noise in the machines' capacity, status, and working time data.

Table 6.30: Augmenting implicit observation related to the capacity of the machine, an example of inference rules

| Inference rules |
|---|
| JobShopScheduler(?i), Machine(?a), hasInputBuffer(?a, ?x), hasOutputBuffer(?a, ?y), hasProcessingBuffer(?a, ?z), hasNumber(?x, ?$n_1$), hasNumber(?y, ?$n_2$), hasNumber(?z, ?$n_3$), hasInitialCapacity(?a, ?c), hasSum(?$n_1$, ?$n_2$, ?$n_3$, ?N), isLess(?N, ?c) |
| $->$ hasRemainingCapacity(?a, $Free$) |

Table 6.31 describes the inference rules that can infer the implicit observation related to the status of machine $a$. The machine is currently failing if the last time it failed $m$ was greater than the last time it started a process $n$.

Table 6.32 describes the inference rules that can infer the implicit observation related to the working time of the machine $a$. Based on the actual processing time of orders in the machine's input $p_2$ and output buffers $p_3$, as well as the order processing time at the moment $p_1$, the scheduler agent can estimate the machine's working time.

Table 6.31: Augmenting implicit observation related to the status of the machine, an example of inference rules.

| Inference rules |
|---|
| JobShopScheduler(?i), Machine(?a), hasLastBrokenStart(?a, ?m), hasLastProcessStart(?a, ?n), isGreater(?m, ?n) |
| $->$ hasStatus(?a, $Failure$) |

Table 6.32: Augmenting implicit observation related to the working time of the machine, an example of inference rules

| Inference rules |
|---|
| JobShopScheduler(?i), Machine(?a), hasInputBuffer(?a, ?x), hasOutputBuffer(?a, ?y), Order(?$d_1$, ?$d_2$, ?$d_3$), inProcess(?$d_1$, ?a), inInputBuffer(?$d_2$, ?x), inOutputBuffer(?$d_3$, ?y), hasCurrentProcessingTime(?$d_1$, ?$p_1$), hasActualProcessingTime(?$d_2$, ?$p_2$), hasActualProcessingTime(?$d_3$, ?$p_3$), hasSum(?$p_1$, ?$p_2$, ?$p_3$, ?P) |
| $->$ hasWorkingTime(?a, ?P) |

**JSS-CS2.** This case study consists of sixteen machines, $\{m_0, m_2, \ldots, m_{15}\}$. To illustrate the JSS environment, see Figure 6.16. In this case study, the scenarios are defined based on order load and due date levels. The factory is divided into three areas including sink, source, and work areas. There are different tasks associated with every location of the factory that must be done by the JSS agent sequentially. I assumed the factory has six different locations $\{u_0, \ldots, u_5\}$ and the tasks associated with each location are as follows:

- $\tau_{u_0}$ selects an order from the source order queue.

- $\tau_{u_1}$ selects a work area for moving the order there.

- $\tau_{u_2}$ selects a group for selecting a machine from it.

- $\tau_{u_3}$ selects a machine for processing the order.

- $\tau_{u_4}$ selects an order from the machine order queue to be processed.

- $\tau_{u_5}$ selects a sink to consume the processed order.

When the job shop scheduler agent is located in each area it can only observe the entities that are specifically located in that area. For example, in location $u_0$, only $I_1$, $I_2$, and $I_3$ and their order queue are visible (i.e., the green dotted rectangle area). When the agent reaches the location $u_1$, it can only observe work areas $z_1$, $z_2$, and $z_3$ (i.e., the purple dotted rectangle area), while at the location $u_2$ it observes groups $n_1$ and $n_2$ (i.e., the red rectangle area). Location $u_3$ is where the agent can observe machines $m_0$ and $m_1$ (the pink dotted rectangle), whereas location $u_4$ is where it can observe only machine $m_1$ and its order queue (the yellow dotted rectangle). Finally, in location $u_5$, the agent observes sinks $X_1$, $X_2$, and $X_3$ (i.e., the blue dotted rectangle area)[4].

**OntoDeM's Methods Employed in JSS-CS2.**

*Adaptive reward definition (related to S2, CH2, RO2, RQ2):* Based on OntoDeM, I proposed the following method: in state $u_0$, the agent defines three separate TRPO learners $\mathcal{M}^1_{\tau_{u_0}}, \mathcal{M}^2_{\tau_{u_0}}, \mathcal{M}^3_{\tau_{u_0}}$ for task $\tau_{u_0}$ with different reward functions $r^1_{\tau_{u_0}}$ (maximizing the due date of orders), $r^2_{\tau_{u_0}}$ (minimizing the waiting time of orders), $r^3_{\tau_{u_0}}$ (minimizing the priority of orders), extracted based on its ontology-based schema. For the sake of simplicity, for tasks $\tau_{u_1}$ and $\tau_{u_2}$ the agent uses a random strategy same as the baseline method. The agent uses one learner $\mathcal{M}_{\tau_{u_3}}$ for task $\tau_{u_3}$ with reward function $r_{\tau_{u_3}}$ (maximizing the working time of machines). For task $\tau_{u_4}$, in 4 out of 16 machines the agent uses one learner $\mathcal{M}_{\tau_{u_4}}$ with reward function $r_{\tau_{u_4}}$ (minimizing the waiting time of orders). For 12 machines left, the agent uses the baseline FIFO strategy. For task $\tau_{u_5}$ the agent uses the random strategy same as the baseline method (see Table 6.33).

Table 6.33: Settings for the baseline algorithm and OntoDeM, JSS-CS2.

|  |  | Baseline method | OntoDeM-enabled proposed method | |
|---|---|---|---|---|
|  |  |  | Without multi-advisor | With multi-advisor |
| **State** | $s_{\tau_{u_0}}$ | LIFO | TRPO ($r^1_{\tau_{u_0}}$) | TRPO ($r^1_{\tau_{u_0}}, r^2_{\tau_{u_0}}, r^3_{\tau_{u_0}}$) |
|  | $s_{\tau_{u_1}}$ | Random | Random | Random |
|  | $s_{\tau_{u_2}}$ | Random | Random | Random |
|  | $s_{\tau_{u_3}}$ | TRPO | TRPO ($r_{\tau_{u_3}}$) | TRPO ($r_{\tau_{u_3}}$) |
|  | $s_{\tau_{u_4}}$ | FIFO | TRPO ($r_{\tau_{u_4}}$), FIFO | TRPO ($r_{\tau_{u_4}}$), FIFO |
|  | $s_{\tau_{u_5}}$ | Random | Random | Random |

*Modify reward machine:* In the JSS environment, when the agent is in state $s_{\tau_{u_0}}$, it observes subsumer $E_{\tau_{u_0}}$ as follows:

$E_{\tau_{u_0}}$ = {Thing, Time, Number, Belief, ($\exists$has.DueDate, $\forall$has.PositiveBelief), ($\exists$has.WaitingTime, $\forall$has.NegativeBelief), ($\exists$has.Priority, $\forall$has.NegativeBelief)}.

Also, for task $\tau_{u_0}$, one of the reward functions could be defined as maximizing the due date of orders because it is associated with a positive belief in the JSS ontology. So, the agent tries to select orders with a low due date to be processed first (i.e., keep orders with a high due date to be processed later):

---

[4]The code of OntoDeM for job shop scheduling environment is publicly available: https://github.com/akram0618/ontology-based-adaptive-reward-machine-RL

$r_{\tau_{u_0}}$ = Maximizing the due date of orders (i.e., select an order with a low due date to be processed first).

When the agent observes the new subsumer $E_{\tau_{u_1}}$ as:

$E_{\tau_{u_1}}$ = {Thing, Time, Number, Belief, ($\forall$has.Distance, $\forall$has.NegativeBelief), $\exists$has.Group, ($\forall$has.MachineCount, $\forall$has.PositiveBelief), $\forall$has.Machine, ($\exists$has.WorkingTime, $\forall$has.PositiveBelief), ($\exists$has.IdleTime, $\forall$has.NegativeBelief), ($\exists$has.FailureTime, $\forall$has.NegativeBelief), ($\forall$has.FailureRate, $\forall$has.NegativeBelief)}.

It recognizes the need to create a new transition to the new state $s_{\tau_{u_1}}$ (see Figure 6.15).

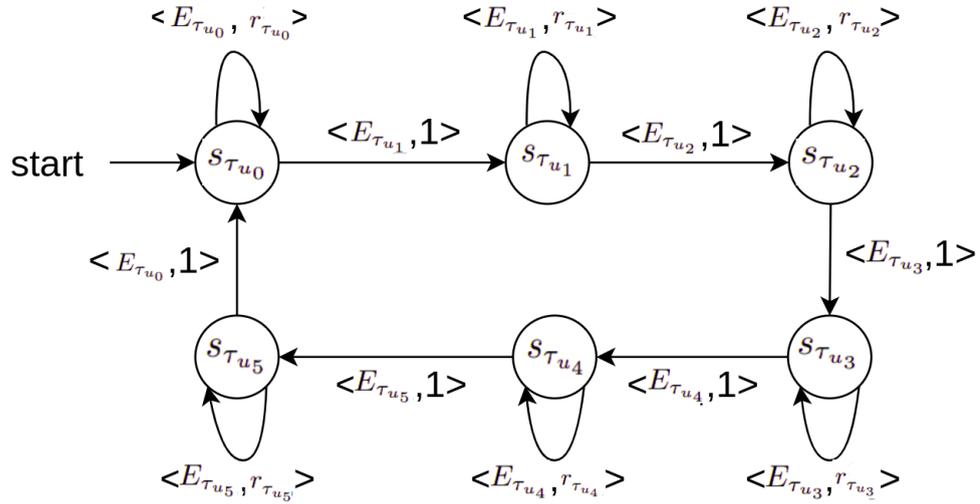

Figure 6.15: $s_{\tau_{u_0}}, \ldots, s_{\tau_{u_5}}$ indicates the agent's different states, $E_{\tau_{u_0}}, \ldots, E_{\tau_{u_5}}$ are subsumers extracted from the agent's states as new propositional symbols/events, and $r_{\tau_{u_0}}, \ldots, r_{\tau_{u_5}}$ are the ontology-based reward functions. I assume that the agent receives a fixed reward value equal to 1 to go from one state to the next.

Consider that some reward functions can not be used for learning of the agent's different action selections (task), for example, selecting a machine to process an order $\tau_{u_3}$ does not impact maximizing the machine count of a group, so the group's property (i.e., machine count) cannot be used as a reward function for the task $\tau_{u_3}$.

### 6.3.3 Results Analysis

**Results in the JSS-CS1.** The average utilization rate and total processed orders in 10 runs in each scenario are reported in Figure 6.17 for the TRPO algorithm and the OntoDeM-enabled TRPO algorithm. The results show that the OntoDeM-enabled

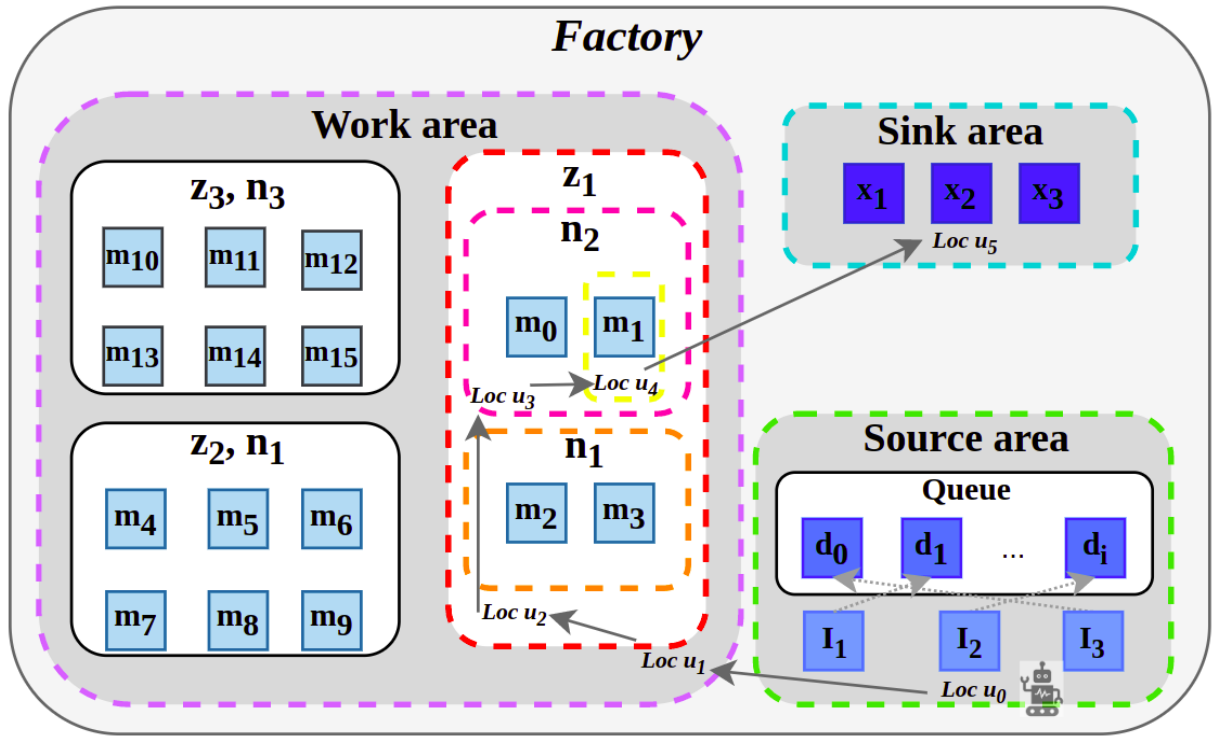

Figure 6.16: Simulated job shop scheduling environment, JSS-CS2.

TRPO algorithm increases the average utilization rate and total processed orders compared to the TRPO algorithm. I observed that the average utilization rate increased by 19%, 8%, and 24%, and the total processed orders increased by 22%, 15%, and 38% in augmenting partial observable data of machines' capacity, status, and working time, respectively (see Tables 6.34, 6.35, and 6.36, p-value results shown in Table 6.37). Thus, when the job shop scheduler agent augments noisy machines' working time and capacity, the percentage change is more significant than augmenting noisy machines' status. Also, the average utilization rate and total processed orders improve better in Heavy scenarios than in Light scenarios. The reason for this is that the number of orders generated has increased, resulting in increased times the agent has to choose machines to process orders and an increase in the impact of improving the machines' noisy data on the agent's performance. With noisy capacity data, where there is both low and high noise, I saw that the improvement for High scenarios is more significant than for Low scenarios. Since the amount of noise is more remarkable, an improvement in noise will significantly impact the agent's decision-making.

**Results in the JSS-CS2.** The average utilization rate, waiting time, total processed orders, and failed orders in 10 runs in each scenario are reported in Figure 6.18 for the baseline method and the OntoDeM-enabled proposed method (without multi-advisor RL). The results show that the OntoDeM-enabled proposed method increases the average utilization rate and the total processed orders and decreases the average waiting time and the total failed orders compared to the baseline method. I observed that the total processed orders increased by 6%, 14%, 14%, and 15%, and the average utilization rate increased by 5%, 11%, 12%, and 16% in Light-High, Light-Low, Heavy-High, and Heavy-Low scenarios, respectively (see Table 6.38, p-value results shown in Table 6.39). So, the percentage change is more significant when the number of orders increases. Also, the total failed orders decreased by 68%, 73%, 41%, and 32%, and the average waiting time decreased by 13%, 8%, 9%, and 6% in Light-High, Light-Low, Heavy-High, and Heavy-Low scenarios, respectively. The improvement in the Low scenarios is lower than in the High scenarios. This is because the number of low due date orders increases, and my proposed method shows less improvement due to the limited number of resources.

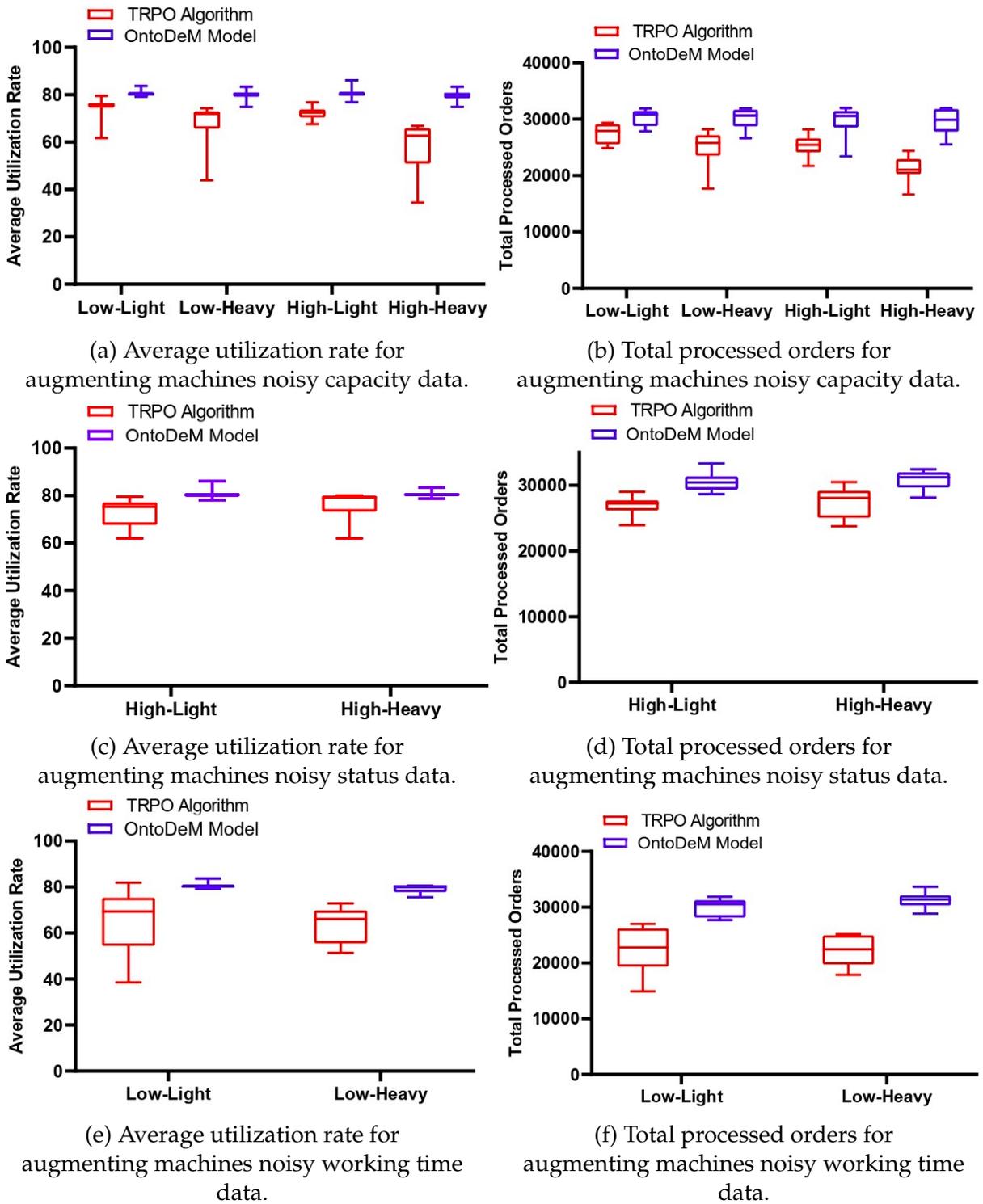

Figure 6.17: Job shop scheduling environment, JSS-CS1, the performance comparison between the TRPO algorithm and OntoDeM.

Table 6.34: The percentage change in evaluation metrics for augmenting machines' noisy capacity data, TRPO algorithm and OntoDeM.

| Scenario | Evaluation metric | |
| --- | --- | --- |
| | Utilization Rate | Processed Orders |
| Low-Light | 8% | 11% |
| Low-Heavy | 19% | 21% |
| High-Light | 12% | 17% |
| High-Heavy | 36% | 40% |
| AVG | 19% | 22% |

Table 6.35: The percentage change in evaluation metrics for augmenting machines' noisy status data, TRPO algorithm and OntoDeM.

| Scenario | Evaluation metric | |
| --- | --- | --- |
| | Utilization Rate | Processed Orders |
| High-Light | 10% | 17% |
| High-Heavy | 6% | 12% |
| AVG | 8% | 15% |

Table 6.36: The percentage change in evaluation metrics for augmenting machines' noisy working time data, TRPO algorithm and OntoDeM.

| Scenario | Evaluation metric | |
| --- | --- | --- |
| | Utilization Rate | Processed Orders |
| Low-Light | 23% | 35% |
| Low-Heavy | 24% | 41% |
| AVG | 24% | 38% |

Table 6.37: Job shop scheduling environment, JSS-CS1, p-value results, the TRPO algorithm and OntoDeM.

| Experiment | Evaluation metric | Scenario | p-value | Statistically significant |
| --- | --- | --- | --- | --- |
| Augmenting machines noisy capacity data | Total processed orders | Low-Light | 0.002879 | ✓ |
| | | Low-Heavy | 0.00013 | ✓ |
| | | High-Light | 0.000725 | ✓ |
| | | High-Heavy | 0.000011 | ✓ |
| | Average utilization rate | Low-Light | 0.000022 | ✓ |
| | | Low-Heavy | 0.000011 | ✓ |
| | | High-Light | 0.000011 | ✓ |
| | | High-Heavy | 0.000011 | ✓ |
| Augmenting machines noisy working time data | Total processed orders | Low-Light | 0.000011 | ✓ |
| | | Low-Heavy | 0.000011 | ✓ |
| | Average utilization rate | Low-Light | 0.000639 | ✓ |
| | | Low-Heavy | 0.000011 | ✓ |

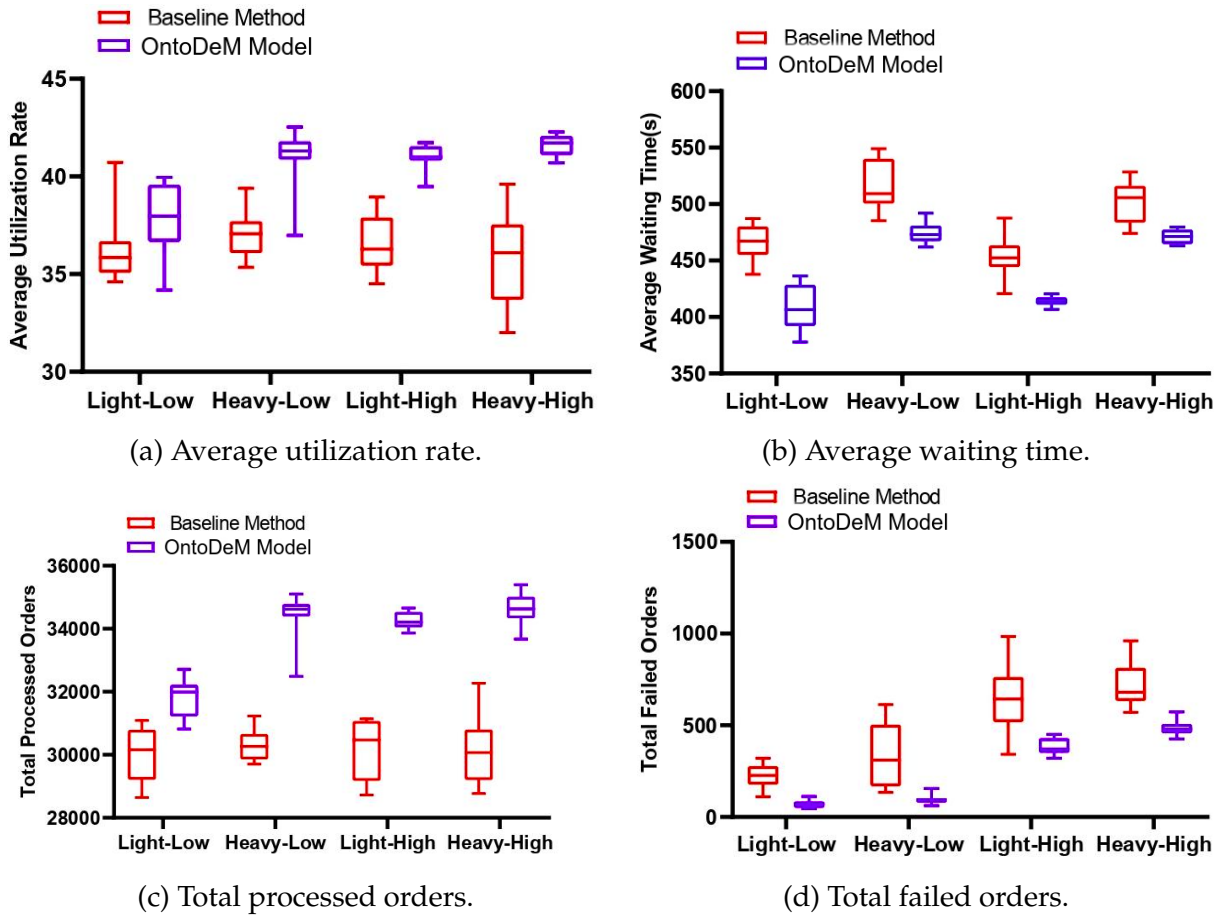

(a) Average utilization rate.

(b) Average waiting time.

(c) Total processed orders.

(d) Total failed orders.

Figure 6.18: Job shop scheduling environment, JSS-CS2, the performance comparison between the baseline algorithm and OntoDeM without multi-advisor RL.

Similarly, compared to the baseline method, the OntoDeM-enabled proposed method (with multi-advisor RL) increases the average utilization rate and the total processed orders while decreasing the average waiting time and the total failed orders (see Figure 6.19, p-value results shown in Table 6.41). I observed that the total processed orders increased by 16%, 16%, 16%, and 18% and the average utilization rate increased by 15%, 14%, 16%, and 19% in Light-High, Light-Low, Heavy-High, and Heavy-Low scenarios, respectively (see Table 6.40). So, the percentage change is more significant with a high number of orders. Also, the total failed orders decreased by 51%, 53%, 18%, and 4%, and the average waiting time decreased by 3%, 0%, 1%, and -1% in Light-High, Light-Low, Heavy-High, and Heavy-Low scenarios, respectively. In Low scenarios, limited resources result in less improvement due to the increased number of low due date orders. When I compared the results of the OntoDeM-enabled proposed method with and without multi-advisor RL, I observed that the improvement without multi-advisor RL is more significant than the other one. The reason can be multiple conflicting reward functions/objectives (e.g., reward functions $r^1_{\tau_{u_0}}, r^2_{\tau_{u_0}}, r^3_{\tau_{u_0}}$ for task $\tau_{u_0}$), which must be balanced based on their relative importance, so the simple voting algorithm does not work well.

Table 6.38: The percentage change in evaluation metrics, the baseline method and the OntoDeM-enabled proposed method (without multi-advisor RL).

|  | Evaluation metrics | | | | |
| --- | --- | --- | --- | --- | --- |
| Scenario | Utilization Rate (increase) | Waiting Time (decrease) | Processed Orders (increase) | Failed Orders (decrease) | AVG |
| **Light-High** | 5% | 13% | 6% | 68% | 23% |
| **Light-Low** | 11% (0.03) | 8% | 14% | 73% | 27% |
| **Heavy-High** | 12% | 9% | 14% | 41% | 19% |
| **Heavy-Low** | 16% | 6% | 15% | 32% | 17% |
| **AVG** | 11% | 9% | 12% | 54% | - |

Table 6.39: p-value results, the baseline method and the OntoDeM-enabled proposed method (without multi-advisor RL).

| Evaluation metric | Scenario | p-value | Statistically significant |
| --- | --- | --- | --- |
| Utilization Rate | Light-Low | 0.037635 | ✓ |
|  | Heavy-Low | 0.000768 | ✓ |
|  | Light-High | 0.000182 | ✓ |
|  | Heavy-High | 0.000182 | ✓ |
| Waiting Time | Light-Low | 0.000182 | ✓ |
|  | Heavy-Low | 0.000329 | ✓ |
|  | Light-High | 0.000182 | ✓ |
|  | Heavy-High | 0.001314 | ✓ |
| Processed Orders | Light-Low | 0.000329 | ✓ |
|  | Heavy-Low | 0.000182 | ✓ |
|  | Light-High | 0.000182 | ✓ |
|  | Heavy-High | 0.000182 | ✓ |
| Failed Orders | Light-Low | 0.000246 | ✓ |
|  | Heavy-Low | 0.000244 | ✓ |
|  | Light-High | 0.002193 | ✓ |
|  | Heavy-High | 0.000244 | ✓ |

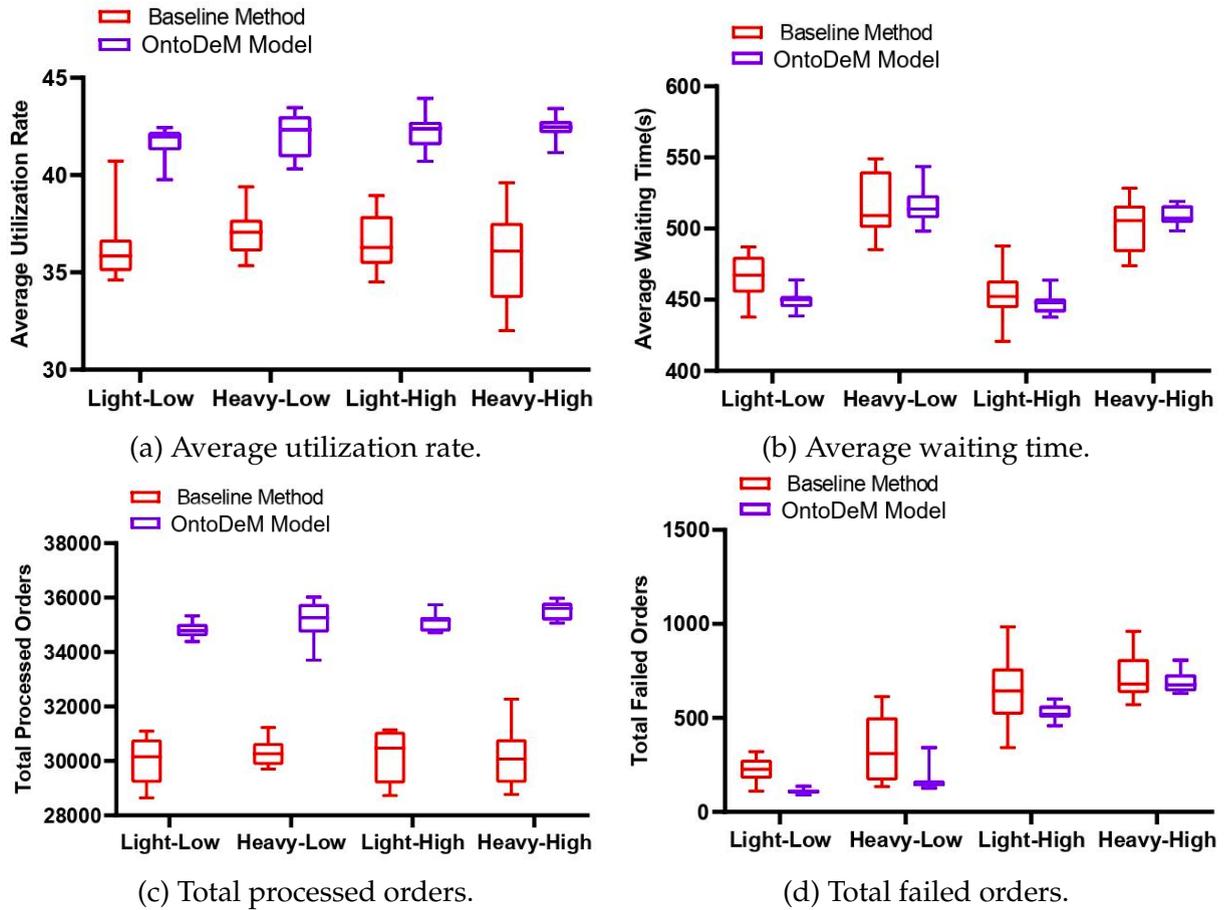

Figure 6.19: Job shop scheduling environment, JSS-CS2, the performance comparison between the baseline algorithm and OntoDeM with multi-advisor RL.

Table 6.40: The percentage change in evaluation metrics, the baseline method and the OntoDeM-enabled proposed method (with multi-advisor RL).

| Scenario | Evaluation metrics | | | | |
| --- | --- | --- | --- | --- | --- |
| | Utilization Rate (increase) | Waiting Time (decrease) | Processed Orders (increase) | Failed Orders (decrease) | AVG |
| **Light-High** | 15% | 3% | 16% | 51% | 21% |
| **Light-Low** | 14% | 0% | 16% | 53% | 21% |
| **Heavy-High** | 16% | 1% | 16% | 18% | 13% |
| **Heavy-Low** | 19% | -1% | 18% | 4% | 10% |
| **AVG** | 16% | 1% | 17% | 32% | - |

## 6.4 Heating System Control Experiment Design

In HSC experiments, the RL algorithm's performance is assessed by testing it on a dataset that represents the environment. The dataset collects temperature measurements taken over time, and the RL algorithm's behavior and decision-making are analyzed based on its interactions with this dataset. This approach is referred to as "offline RL". In offline RL, the algorithm learns from a fixed dataset of previously collected experiences, as opposed to online RL where the agent interacts with the environment in real-time.

Table 6.41: p-value results, the baseline method and the OntoDeM-enabled proposed method (with multi-advisor RL).

| Evaluation metric | Scenario | p-value | Statistically significant |
|---|---|---|---|
| Utilization Rate | Light-Low | 0.000244 | ✓ |
| | Heavy-Low | 0.000182 | ✓ |
| | Light-High | 0.000182 | ✓ |
| | Heavy-High | 0.000182 | ✓ |
| Waiting Time | Light-Low | 0.021133 | ✓ |
| | Heavy-Low | 0.623176 | ✗ |
| | Light-High | 0.384673 | ✗ |
| | Heavy-High | 0.496129 | ✗ |
| Processed Orders | Light-Low | 0.000182 | ✓ |
| | Heavy-Low | 0.000182 | ✓ |
| | Light-High | 0.000182 | ✓ |
| | Heavy-High | 0.000182 | ✓ |
| Failed Orders | Light-Low | 0.000666 | ✓ |
| | Heavy-Low | 0.006461 | ✓ |
| | Light-High | 0.053812 | ✗ |
| | Heavy-High | 0.879784 | ✗ |

**Hardware and Operating System Configuration.** Simulations are conducted on a machine equipped with an Intel Core i5-4200U CPU (2 cores, 4 threads, 1.60 GHz base) and an NVIDIA GeForce GT 740M, running on a Windows 10 operating system with 4GB of RAM.

### 6.4.1 Evaluation Scenarios, Metrics and Baseline

**Evaluation scenarios:** To assess OntoDeM within the HSC environment, four scenarios are established, characterized by varying noise levels of Low and High, along with the classification of data based on time and day (see Table 6.42).

**Evaluation metrics:** The subsequent metrics are employed for assessing the performance of the OntoDeM:

- Average Time Spent Out of Range (ATSOR) is calculated by measuring how much time the temperature remains outside the desired range over a certain period, providing a measure of how well the HSC maintains the desired temperature range. Let $T_{\text{in}}^t$ denote the temperature at time $t$. Let $L$ be the lower bound and $U$ be the upper bound of the desired temperature range. The ATSOR is mathematically defined as follows:

1. Define the deviation function $D(t)$ as:

$$D(t) = \begin{cases} T_{\text{in}}^t - U, & \text{if } T_{\text{in}}^t > U \\ L - T_{\text{in}}^t, & \text{if } T_{\text{in}}^t < L \\ 0, & \text{if } L \leq T_{\text{in}}^t \leq U \end{cases} \quad (6.2)$$

Table 6.42: Evaluation scenarios for heating system control.

| | | Description |
|---|---|---|
| **Noise level** | Low | Noise is present in 0.02% of the temperature data. |
| | High | 0.05% of the temperature data contains noise. |
| **Classification** | Day | Guided by the "WeekDay" concept, I define two classes: "WorkingDay" (Monday-Friday) and "Non-WorkingDay" (Saturday and Sunday). |
| | Time | In accordance with the "PartOfDay" concept, I establish the following classes: "Morning" (5am-12pm), "Afternoon" (12pm-5pm), "Evening" (5pm-8pm), and "Night" (8pm-5am). |

2. Define the set $I = \{t \mid D(t) \neq 0\}$, which represents the time intervals during which the temperature $T_{\text{in}}^t$ is outside the desired range.

3. For each interval $I_k = [t_{k,\text{start}}, t_{k,\text{end}}]$ in $I$, calculate its duration $\Delta t_k = t_{k,\text{end}} - t_{k,\text{start}}$.

4. Calculate the total time spent outside the desired range by summing the durations of all intervals in $I$:

$$\text{Total time spent outside range} = \sum_k \Delta t_k \quad (6.3)$$

5. Define the observation period $\Delta t_{\text{obs}}$ as the total time span of the temperature measurements.

6. Finally, compute the average time spent out of range as the ratio of the total time spent outside the range to the observation period:

$$\text{ATSOR} = \frac{\text{Total time spent outside range}}{\Delta t_{\text{obs}}} \quad (6.4)$$

- The MSE provides a measure of how well the observed temperatures align with the desired interval midpoint, with lower values indicating better alignment (see Equation 5.18).

- Total reward is the sum of the rewards (see Equation 5.17) earned by the heating system controller in each run. This total reward is indicative of the agent's overall performance across multiple scenarios.

**Baselines:** I compared the outcomes achieved by the foundational RL algorithms, encompassing PPO and DDPG [254], with those of the same algorithms augmented by OntoDeM while retaining their conventional operations.

HSC often involves dealing with noisy temperature data. PPO's trust region optimization and DDPG's off-policy learning with experience replay make them suitable choices for handling noisy data. PPO's stability stems from its constraint on policy updates within a trust region, preventing large updates in response to noisy data points. DDPG, although off-policy, utilizes a replay buffer to store past experiences, which can help mitigate the impact of noisy data on its learning process. In the Thesis, the performance of PPO and DDPG, both as baseline algorithms and when augmented with OntoDeM, is compared. This allows for an empirical evaluation of how these algorithms perform in handling noisy temperature data and the effectiveness of OntoDeM in enhancing their performance.

### 6.4.2 Case studies and Employed OntoDeM Methods

**HSC-CS1.** The dataset, covering the years 2018 to 2019, presents a collection of temperature measurements taken over time[5]. The recorded temperatures range between -3.75°C and 19.96°C. The average temperature is about 6.47°C, with a standard deviation of 4.67°C, indicating a high level of variability around the average. The data is symmetrically distributed, evident from both the mean and median temperatures aligning closely. Temperature fluctuations are captured within the interquartile range of approximately 2.77°C to 9.86°C, showcasing the central 50% of the observations.

In this case study, there is a pre-trained model that has not been exposed to noisy data previously. I then initiated an evaluation of the performance of both the baseline algorithms and the OntoDeM model when dealing with noisy data as they begin to operate with the pre-trained model as a foundation. The pre-trained model is gener-ated over the course of 1000 episodes. The performance of the learning algorithm is assessed through a single testing round involving the entire dataset[6].

---

[5]Dataset is publicly available: `https://github.com/tilkb/thermoAI/blob/master/simulator/weather_data/temp_2018_2019_basel.csv`

**HSC-CS2.** The dataset, covering the years 2016 to 2017, presents a collection of temperature measurements taken over time[7]. The recorded temperatures range between -6.28°C and 21.62°C. The average temperature is about 5.8°C, with a standard deviation of 4.84°C, indicating a high level of variability around the average. Temperature fluctuations are captured within the interquartile range of approximately 2.2°C to 9.09°C, showcasing the central 50% of the observations.

In this case study, I utilized the same pre-trained RL model (as presented in HSC-CS1), however, I proceeded to train it through five episodes, each comprising 64 steps and involving data with varying levels of noise. Subsequently, I assessed the performance of both the proposed and baseline algorithms when confronted with the challenges posed by noisy data.

**OntoDeM's Methods Employed in HSC.**

*Observation abstraction (related to S1, CH1, RO1, RQ1):* Using the OntoDeM model, the agent classifies the previous observations (historical experiences) based on concepts "PartOfDay" or "WeekDay" (i.e., ontology-based observation abstraction). Then it calculates the average temperature for noise-free data in each class. Subsequently, the agent assigns the current observation to an appropriate class. Ultimately, the average temperature within this class is employed to replace the noisy temperature in the current observation.

### 6.4.3 Results Analysis

**Results in the HSC-CS1.** The average time spent out of range, MSE, and total reward in 10 runs in each scenario are reported in Figure 6.20 for the PPO and DDPG baseline algorithms and the OntoDeM-enabled algorithm. The results show that the OntoDeM-enabled algorithm decreases the ATSOR and MSE and increases total reward compared to the baseline algorithms.

Observing the PPO algorithm's performance, we note reductions of 6% and 31% in ATSOR and MSE, respectively, alongside a 15% increase in total reward (refer to Table 6.43, p-value results shown in Table 6.44). In contrast, the DDPG algorithm saw improvements of 3% in MSE and 13% in total reward (see Table 6.45, p-value results shown in Table 6.46). Evidently, PPO exhibits greater enhancement compared to DDPG. PPO tends to be more robust to noisy data compared to DDPG. This is because PPO uses a form of trust region optimization that helps stabilize training and prevents the policy from making large updates in response to noisy data points. However DDPG is an off-policy algorithm that uses a replay buffer to store past experiences, noisy data can have a more significant impact on its learning process. Furthermore, notable improvements become evident when confronted with "High" scenarios, highlighting OntoDeM's adeptness at effectively managing high noise levels.

---

[6]The code of OntoDeM for heating system control is publicly available: https://github.com/akram0618/ontology-based-observation-abstraction-RL

[7]Dataset is publicly available: https://github.com/tilkb/thermoAI/blob/master/simulator/weather_data/temp_2016_2017_basel.csv

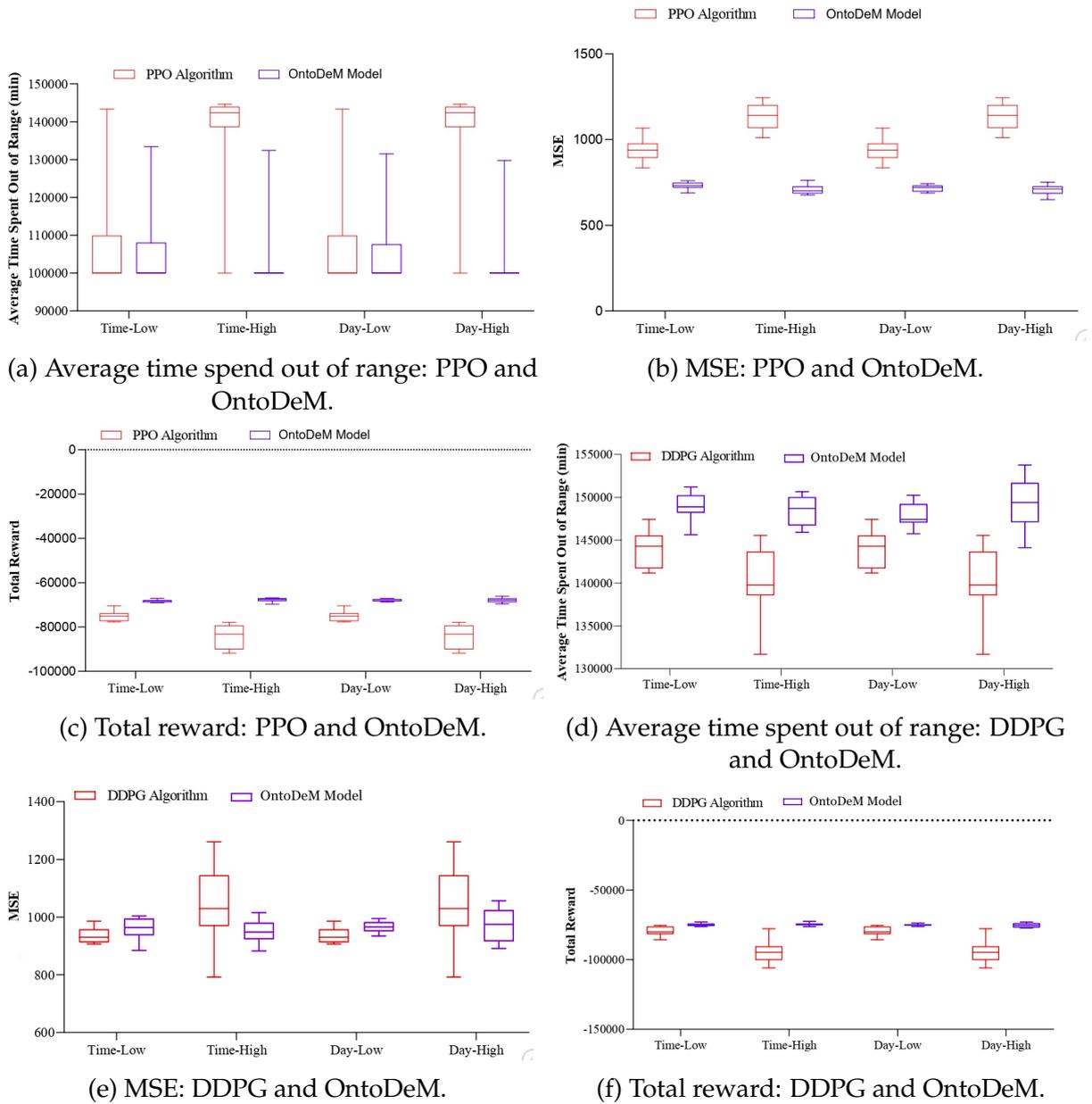

Figure 6.20: Heating system control, HSC-CS1, the performance comparison between the baseline algorithms and OntoDeM.

Table 6.43: The percentage change in evaluation metrics, HSC-CS1, the baseline PPO algorithm and the OntoDeM-enabled proposed method.

| Scenario | Evaluation metrics | | | |
| --- | --- | --- | --- | --- |
| | Average Time Spent Out of Range (decrease) | MSE (decrease) | Total Reward (increase) | AVG |
| **Time-Low** | 5% | 22% | 9% | 12% |
| **Time-High** | 6% | 38% | 20% | 21% |
| **Day-Low** | 6% | 24% | 9% | 13% |
| **Day-High** | 7% | 38% | 20% | 22% |
| **AVG** | 6% | 31% | 15% | - |

Table 6.44: p-value results, HSC-CS1, the baseline PPO algorithm and the OntoDeM-enabled proposed method.

| **Evaluation metric** | **Scenario** | **p-value** | **Statistically significant** |
| --- | --- | --- | --- |
| Average Time Spent Out of Range | Time-Low | 0.871151 | × |
| | Time-High | 0.000379 | ✓ |
| | Day-Low | 0.871151 | × |
| | Day-High | 0.000379 | ✓ |
| MSE | Time-Low | 0.000181 | ✓ |
| | Time-High | 0.000182 | ✓ |
| | Day-Low | 0.000182 | ✓ |
| | Day-High | 0.000182 | ✓ |
| Total Reward | Time-Low | 0.000182 | ✓ |
| | Time-High | 0.000182 | ✓ |
| | Day-Low | 0.000182 | ✓ |
| | Day-High | 0.000182 | ✓ |

Table 6.45: The percentage change in evaluation metrics, HSC-CS1, the baseline DDPG algorithm and the OntoDeM-enabled proposed method.

| Scenario | Evaluation metrics | | | |
| --- | --- | --- | --- | --- |
| | Average Time Spent Out of Range (decrease) | MSE (decrease) | Total Reward (increase) | AVG |
| **Time-Low** | -4% | -3% | 6% | 0% |
| **Time-High** | -6% | 9% | 21% | 8% |
| **Day-Low** | -3% | -3% | 6% | 0% |
| **Day-High** | -6% | 7% | 20% | 7% |
| **AVG** | -5% | 3% | 13% | - |

**Results in the HSC-CS2.** Figure 6.21 presents the ATSOR, MSE, and total reward across 10 runs in each scenario. This data encompasses the performance of the PPO and DDPG baseline algorithms alongside the OntoDeM-enhanced algorithm. The outcomes demonstrate that the employment of the OntoDeM-enabled algorithm leads to a reduction in the ATSOR, as well as a decrease in MSE, while simultaneously enhancing the total reward when contrasted with the baseline algorithms.

Table 6.46: p-value results, HSC-CS1, the baseline DDPG algorithm and the OntoDeM-enabled proposed method.

| Evaluation metric | Scenario | p-value | Statistically significant |
|---|---|---|---|
| Average Time Spent Out of Range | Time-Low | 0.000329 | ✓ |
|  | Time-High | 0.000181 | ✓ |
|  | Day-Low | 0.001007 | ✓ |
|  | Day-High | 0.000328 | ✓ |
| MSE | Time-Low | 0.053812 | × |
|  | Time-High | 0.031082 | ✓ |
|  | Day-Low | 0.021037 | ✓ |
|  | Day-High | 0.161814 | × |
| Total Reward | Time-Low | 0.001007 | ✓ |
|  | Time-High | 0.000182 | ✓ |
|  | Day-Low | 0.000582 | ✓ |
|  | Day-High | 0.000182 | ✓ |

Using the OntoDeM-enabled PPO algorithm, the ATSOR and MSE decrease by 32% and 64%, and the total reward increases by 23% (see Table 6.47, p-value results shown in Table 6.48). In the OntoDeM-enabled DDPG algorithm, the percentage decrease is 10% and 52% and the percentage increase is 27% for evaluation metrics ATSOR, MSE, and total reward, respectively (see Table 6.49, p-value results shown in Table 6.50). We can see that the amount of improvement in the HSC-CS2 is more than the HSC-CS1. The reason is that when the model is trained with noisy data, it can act as a form of regularization. Regularization helps prevent overfitting by introducing noise or perturbations into the training process. This can make the model more robust and able to generalize better to unseen data. Also, when training with noisy data, the RL model might have explored more diverse actions due to the introduction of noise. This increased exploration could lead to a better policy that performs well on the dataset during testing.

## 6.5 Discussion

In this section, I provide an overview of the strengths and weaknesses associated with the utilization of OntoDeM's methods throughout various stages of an agent's decision-making process:

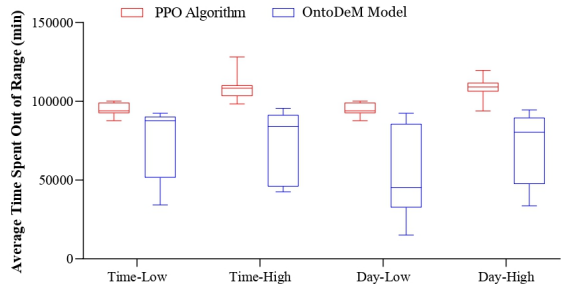
(a) Average time spend out of range: PPO and OntoDeM.

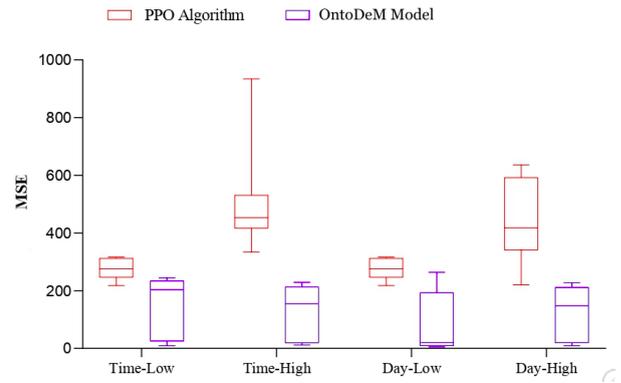
(b) MSE: PPO and OntoDeM.

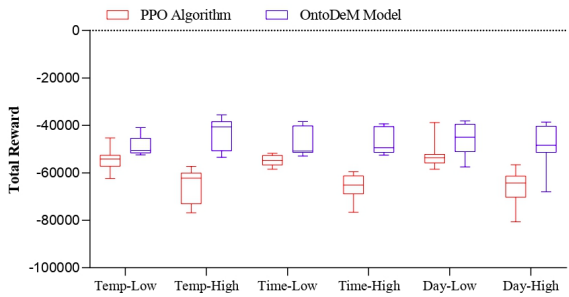
(c) Total reward: PPO and OntoDeM.

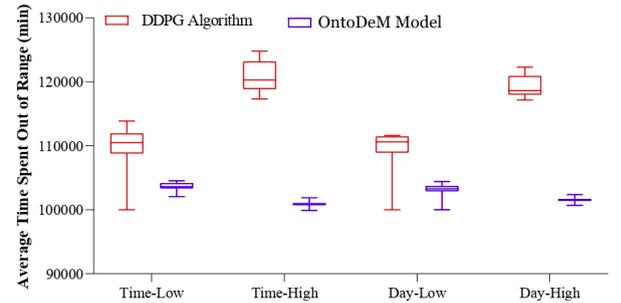
(d) Average time spent out of range: DDPG and OntoDeM.

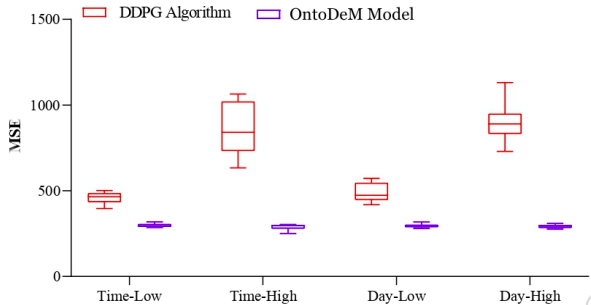
(e) MSE: DDPG and OntoDeM.

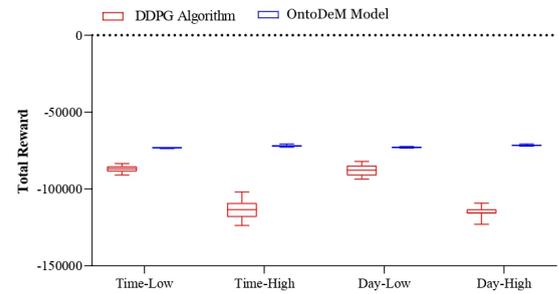
(f) Total reward: DDPG and OntoDeM.

Figure 6.21: Heating system control, HSC-CS2, the performance comparison between the baseline algorithms and OntoDeM.

### 6.5.1 Discussion on Ontology-based Environment Modeling Process

**Strengths.** "Sensor Data Mapping" leverages the Semantic Sensor Network (SSN) ontology to provide a semantic representation of low-level sensor data. By using established ontologies like SSN, the approach enhances interoperability. It ensures that sensor data can be easily integrated with other systems and applications that also understand these ontologies. The use of domain-specific ontologies allows for flexibility in defining and adapting the concepts and properties relevant to a specific application or domain. This flexibility is valuable when working with diverse sensor data sources.

Table 6.47: The percentage change in evaluation metrics, HSC-CS2, the baseline PPO algorithm and the OntoDeM-enabled proposed method.

| Scenario | Evaluation metrics | | | |
| --- | --- | --- | --- | --- |
| | Average Time Spent Out of Range (decrease) | MSE (decrease) | Total Reward (increase) | AVG |
| **Time-Low** | 20% | 43% | 13% | 25% |
| **Time-High** | 33% | 75% | 29% | 46% |
| **Day-Low** | 42% | 66% | 20% | 43% |
| **Day-High** | 34% | 73% | 30% | 46% |
| **AVG** | 32% | 64% | 23% | - |

Table 6.48: p-value results, HSC-CS2, the baseline PPO algorithm and the OntoDeM-enabled proposed method.

| Evaluation metric | Scenario | p-value | Statistically significant |
| --- | --- | --- | --- |
| Average Time Spent Out of Range | Time-Low | 0.001314 | ✓ |
| | Time-High | 0.000181 | ✓ |
| | Day-Low | 0.000439 | ✓ |
| | Day-High | 0.000246 | ✓ |
| MSE | Time-Low | 0.001314 | ✓ |
| | Time-High | 0.000182 | ✓ |
| | Day-Low | 0.000504 | ✓ |
| | Day-High | 0.000246 | ✓ |
| Total Reward | Time-Low | 0.000439 | ✓ |
| | Time-High | 0.000182 | ✓ |
| | Day-Low | 0.021133 | ✓ |
| | Day-High | 0.001314 | ✓ |

Table 6.49: The percentage change in evaluation metrics, HSC-CS2, the baseline DDPG algorithm and the OntoDeM-enabled proposed method.

| Scenario | Evaluation metrics | | | |
| --- | --- | --- | --- | --- |
| | Average Time Spent Out of Range (decrease) | MSE (decrease) | Total Reward (increase) | AVG |
| **Time-Low** | 6% | 35% | 16% | 19% |
| **Time-High** | 15% | 66% | 37% | 39% |
| **Day-Low** | 6% | 40% | 17% | 21% |
| **Day-High** | 14% | 67% | 38% | 40% |
| **AVG** | 10% | 52% | 27% | - |

Table 6.50: p-value results, HSC-CS2, the baseline DDPG algorithm and the OntoDeM-enabled proposed method.

| Evaluation metric | Scenario | p-value | Statistically significant |
|---|---|---|---|
| Average Time Spent Out of Range | Time-Low | 0.002827 | ✓ |
| | Time-High | 0.030303 | ✓ |
| | Day-Low | 0.002459 | ✓ |
| | Day-High | 0.030303 | ✓ |
| MSE | Time-Low | 0.000181 | ✓ |
| | Time-High | 0.000179 | ✓ |
| | Day-Low | 0.000180 | ✓ |
| | Day-High | 0.000181 | ✓ |
| Total Reward | Time-Low | 0.000182 | ✓ |
| | Time-High | 0.000182 | ✓ |
| | Day-Low | 0.000182 | ✓ |
| | Day-High | 0.000182 | ✓ |

**Weaknesses.** "Ontology Development" can be complex, particularly when dealing with large and intricate domains. Managing the complexity and ensuring that ontologies remain maintainable can be challenging. Developing ontologies, especially from scratch, can be resource-intensive in terms of time, expertise, and computational resources. Also, effective ontology development often requires a deep understanding of the domain, and it may be challenging to find individuals with the necessary expertise. Moreover, ontologies are not static; they evolve over time. Ensuring that ontologies remain up-to-date and reflective of changing domain knowledge can be a continuous effort. Furthermore, managing large-scale ontologies can still be a significant challenge. Lastly, ontology evaluation can be subjective, and there may not always be clear criteria for assessing ontology quality.

### 6.5.2 Discussion on OntoDeM Methods Employed in Ontology-enhanced Observation Modeling Stage

**Strengths.** "Observation Abstraction" enables the agent to handle noisy sensor information by abstracting observations into higher-level concepts. "Observation Expansion" enriches the agent's ontology-based schema with domain-specific knowledge, allowing for a more comprehensive representation of observed data, thus enhancing decision-making in dynamic environments. "Observation Masking" filters out irrelevant observations, enabling the agent to focus on essential information and generalize from past experiences. "Observation Augmentation" empowers the agent to infer missing information, increasing the completeness of observations, especially when dealing with partially observable environments. Lastly, "Observation Sampling" ensures representative sampling based on contextual factors. These methods collectively strengthen the agent's observation modeling capabilities, enabling it to adapt to partially observable and dynamic environments.

**Weaknesses.** In the "Observation Abstraction" method, challenges may arise in determining the appropriate level of abstraction and managing noisy observations. In "Observation Expansion", challenges may involve determining which concepts and properties to add, ensuring the ontology remains manageable and adaptable, and handling the integration of newly deduced concepts and properties into the agent's reasoning process. In "Observation Masking" challenges include setting suitable similarity thresholds, and ensuring that masked observations do not lead to loss of critical information. In the "Observation Augmentation" method challenges encompass defining appropriate logical rules, handling complex relationships, and ensuring that the agent's inferences align with domain-specific knowledge. Finally, in "Observation Sampling" challenges include determining concept weighting indicators.

### 6.5.3 Discussion on OntoDeM Methods Employed in Ontology-driven Goal Selection/Generation Stage

**Strengths.** The "Automatic Goal Selection/Generation" method leverages ontology to dynamically generate tailored goals based on the agent's observations, allowing it to adapt to unforeseen situations. The "Adaptive Reward Definition" method utilizes ontology to dynamically adjust reward functions based on contextual factors in changing environments. Finally, the "Reward Augmentation" method augments reward signals with efficiency rates, providing more frequent feedback and encouraging actions that have a greater impact on the environment. These methods collectively enable agents to make adaptive decisions in partially observable and dynamic environments.

**Weaknesses.** In "Automatic Goal Selection/Generation", one of the main challenges is the identification of unforeseen situations and the need to generate or modify goals in response. This requires continuous monitoring of the environment, evaluation of observations, and decision-making about whether to change the current goal, create a new one, or continue with the existing goal. The challenge lies in defining thresholds for such decision-making processes. In "Adaptive Reward Definition", the challenge lies in dynamically adjusting reward functions to align with the specific situation or requirements at hand. Ontology plays a crucial role in providing context for reward function extraction, but it requires careful belief values definition. Determining how to formulate and adapt reward functions based on the agent's beliefs and constraints can be intricate, especially in domains with numerous interrelated concepts/properties. In "Reward Augmentation", ensuring that the augmented rewards provide meaningful feedback to the agent without introducing distortions is a crucial challenge.

### 6.5.4 Discussion on OntoDeM Methods Employed in Ontology-driven Action Selection Stage

**Strengths.** "Action Masking" filters out irrelevant actions based on semantic constraints, improving exploration performance. "Action Exploration" leverages ontology-driven concept prioritization and reward mapping to systematically explore actions, enhancing the agent's ability to discover new actions for unforeseen situations. "Action Prioritization" uses ontology-defined logical rules to focus on actions meeting specific requirements. "Execution Prioritization" handles temporal dependencies and contextual factors to optimize action sequences, enabling adaptation to changing circumstances. These methods collectively strengthen the agent's decision-making capabilities, particularly in scenarios with limited information and dynamic conditions.

**Weaknesses.** The effectiveness of "Action Masking" heavily relies on the completeness and accuracy of the domain-specific ontology. If the ontology is missing certain critical constraints or prerequisites for actions, infeasible actions might not be correctly identified, leading to suboptimal/incorrect decision-making. The success of "Action Exploration" is dependent on the quality of the concept weighting method and action-concept-reward mappings. If the ontology fails to properly rank important concepts or establish accurate action-reward associations, the exploration process may not prioritize actions that genuinely lead to desirable outcomes. Moreover, striking the right balance between exploration and exploitation can be challenging. Overemphasis on exploration may lead to slow learning and delayed exploitation of optimal actions, while overemphasis on exploitation may cause the agent to miss out on discovering new actions to handle unforeseen situations. The effectiveness of "Action Prioritization" is subject to the accuracy of the predefined logical rules. If the rules become overly complex or are not reflective of the domain requirements, the agent may prioritize actions incorrectly, potentially leading to suboptimal/incorrect decisions. Keeping predefined rules up-to-date with evolving domain conditions can be challenging. As the environment changes, the rules may need frequent updates, requiring domain experts to maintain the ontology and logical rule set. Also, the correct handling of temporal dependencies in "Execution Prioritization" relies on accurately defined ontology-based logical rules. If the rules do not accurately capture the temporal relationships between actions, the execution prioritization may result in suboptimal/incorrect action sequences.

Furthermore, adapting to contextual factors is dependent on the ontology's ability to model and capture these factors accurately.

Domain experts provide domain-specific knowledge, and valuable contextual information for decision-making through ontologies. These experts are essential for creating and maintaining the ontological knowledge base, defining semantic constraints, and fine-tuning the logical rules in dynamic environments. The success of OntoDeM's methods depends on the quality, completeness, and accuracy of the ontology, as well as the effectiveness of predefined logical rules and concepts' weight. So, continuous collaboration with domain experts and ongoing ontology maintenance is essential for addressing these challenges and improving the methods in the future.

### 6.5.5 Scalability Assessment

- In the ITSC application area, considering the difference in the number of intersections between the two case studies (16 intersections in ITSC-CS1 and 29 intersections in ITSC-CS2), OntoDeM has effectively managed traffic signal control in larger and more complex networks, as evidenced by its improved performance in ITSC-CS2 compared to ITSC-CS1. Despite the increased number of intersections, OntoDeM maintains efficient traffic flow and reduces vehicle waiting times, showcasing its scalability in handling larger traffic networks with numerous agents.

- In the EC domain, I defined multiple scenarios representing different levels of user demand and task complexity. I analyzed the performance of the OntoDeM model by measuring its ability to efficiently manage latency-sensitive tasks under increasing user loads. By conducting these evaluations, I aimed to assess how the OntoDeM model scales with the growing number of agents (users) and actions (tasks) in the EC environment.

- To address the scalability concerns in the JSS domain, I have defined various evaluation scenarios, that encompass different levels of order load (number of actions), due date levels, and noise levels. By varying the order load, I aimed to evaluate how our scheduling approach scales with increasing workload demands. Also, my evaluation included scenarios with different levels of due date urgency: high and low. This allows me to assess how well the OntoDeM model handles varying due date constraints. Additionally, I considered two levels of noise in the job shop scheduling environment: low and high. Through experi-mentation and analysis, I provided insights into how the OntoDeM scales with increasing complexity and workload demands.

Additionally, in assessing the scalability of OntoDeM in JSS environments, we investigate its efficacy across two distinct case studies: JSS-CS1 and JSS-CS2. Initially focusing on a simpler setup with eight machines in JSS-CS1, OntoDeM demonstrates its scalability by efficiently selecting optimal actions (machines) to maximize the machine utilization rate and the total number of processed orders. Transitioning to the more complex environment of JSS-CS2, which encompasses sixteen machines, OntoDeM demonstrated its scalability by effectively managing a larger number of actions.

- In HSC, I have designed evaluation scenarios to encompass varying levels of noise in the temperature data: low and high. By evaluating OntoDeM's performance under these different noise conditions, I provided insights into how it scales with increasing complexity and variability in environmental conditions, addressing concerns about scalability in HSC.

## 6.6 Summary

The chapter encompasses various aspects of applying the OntoDeM model in different domains. In the ITSC domain, utilizing SUMO (Simulation of Urban MObility) software, two traffic scenarios, synthetic and urban Dublin networks, are evaluated. Key metrics like average vehicle waiting time, travel time, travel delay, and finished trips are considered. Baseline RL algorithms (Q-learning, SARSA, DQN) are compared against OntoDeM-enhanced versions, yielding improved traffic flow by reducing waiting times and travel delays. The evaluation of OntoDeM across various traffic scenarios led to reductions in average waiting time for default vehicles by 9% to 33%. Furthermore, there are improvements in average waiting time for emergency vehicles from 31% to 72%. These reductions in waiting time are crucial in real-world traffic management as they directly translate to decreased congestion, shorter travel times, and improved overall traffic flow efficiency, thereby enhancing road safety and reducing environmental impacts. The chapter also presents an experiment assessing OntoDeM's performance in an EC context. Various scenarios are tested, and metrics such as task success and failure rates are used. Results show that the OntoDeM model enhances task processing and reduces failures compared to the DDPG baseline. OntoDeM led to an increase in the average total processed tasks by 6% to 26% and decreases in failed tasks by 15% to 55% across different scenarios. These enhancements directly contribute to optimized resource utilization, increased task processing efficiency, and improved system reliability. Moreover, the agent aims to enhance JSS using OntoDeM's augmentation of noisy machine data and define adaptive reward functions. Scenarios with different order loads and noise levels are analyzed. The OntoDeM-enabled TRPO algorithm surpasses the baseline in terms of machines' utilization rate and processed orders, demonstrating its efficacy in diverse scenarios. Then the total processed orders increased by 12% to 38%, and the average utilization rate increased by 8% to 24%. Meanwhile, the total failed orders decreased by 54%, and the average waiting time decreased by 9%. In scenarios with heavier workloads, the impact of these improvements becomes even more evident, ensuring smoother operations and better response to dynamic demands. Lastly, the evaluation of OntoDeM in the HSC environment is explored by abstracting observations based on different scenarios to augment various levels of noisy data. The results show improvements achieved by the OntoDeM-enhanced PPO and DDPG algorithms in terms of average time spent out of range, MSE, and total reward. Using OntoDeM, the average time spent out of range and MSE decreased by 6% to 32% and 3% to 64%, respectively. These enhancements leads to improved energy efficiency by optimizing heating system operations, resulting in reduced energy consumption and associated costs. Also, these improvements ensure better temperature regulation, enhancing comfort and safety for occupants of buildings or facilities. Moreover, by minimizing energy waste and greenhouse gas emissions, the optimized control strategy contributes to environmental sustainability.

# CHAPTER SEVEN

# CONCLUSION

This chapter summarises the challenges addressed, and contributions, and provides a discussion on avenues for future works.

## 7.1 Addressed Challenges in the Thesis

Many real-world problems are characterized by being partially observable and dynamic, meaning that the decision-making process becomes challenging due to the incomplete and dynamic nature of the environment. Partial observability refers to situations where decision-makers lack complete information about the current state of the system or have limited access to relevant data. This can be due to various factors such as sensor limitations, uncertain or hidden variables, latency or delays, and environments complexities. Additionally, dynamic environments are those that exhibit constant changes over time, making it difficult to predict future states accurately, and unforeseen events are more likely to occur. Therefore, decision-making in such environments becomes challenging because it requires adaptability, the ability to handle uncertainty and make timely decisions based on incomplete or rapidly changing information.

This Ph.D. research focuses on intelligent, adaptive, and autonomous agents operating in dynamic and partially observable environments. The agents are capable of making decisions, learning, and adapting based on their perception of the environment. The challenges arising from partial observations and unforeseen situations are discussed in connection with various stages of the agent's decision-making process. In the *observation modeling* stage, the agent's observation is accurately modeled despite the available incomplete and noisy information. In the *goal selection* stage, the challenges are related to efficiently searching for relevant goals, updating preferences over time, generating new goals in unforeseen situations, and ensuring efficient and effective goal achieve-ment. The *action selection* stage discusses the difficulties in choosing optimal actions due to partial observation and uncertainty, as well as the trade-off between exploration and exploitation, particularly in large action spaces and when quick decisions are required.

The literature presents a variety of strategies aimed at addressing challenges related to partial observability and unforeseen situations within an agent's decision-making process. The observation modeling approaches encompass diverse techniques including Hidden Markov Models (HMMs), Partially Observable Markov Decision Processes (POMDPs), scene graphs, Monte Carlo methods, Bayesian learning, and neural network-based models. However, these approaches exhibit limitations such as struggling to capture complex dependencies, computational challenges with larger state spaces, limited expressive power, sensitivity to noisy data, insufficient training data, and struggles with long-term dependencies present in past observations. The goal selection approaches include methods like goal formation, generation, reasoning, learning, heuristic-based strategies, and adaptive reward signals. However, they face challenges such as reliance on historical data, lack of predefined strategies for unforeseen events, and sub-optimal solutions stemming from heuristic-based methods, and the intricate task of designing suitable reward functions, particularly in dynamic environments. In the action selection approaches, techniques like Reinforcement Learning (RL), online planning, self-adaptive systems, exploration algorithms, and semi-stochastic action selection are discussed. However, the existing approaches encounter several limitations. RL can struggle with high-dimensional state spaces and require extensive exploration and learning time. Online planning techniques face difficulties in generating plans within time constraints and adapting to large state spaces. Self-adaptive systems may grapple with handling rare or unknown events and need to strike a balance between exploration and real-time decision-making. Exploration algorithms, although crucial for uncertainty reduction, can be resource-intensive and hindered by exponential growth in the number of possible actions. Semi-stochastic action selection's effectiveness is impacted by limited or noisy observations, while the broader challenge lies in swift adaptation to dynamic changes over time.

## 7.2 Overview of Thesis Achievements

I developed a comprehensive model called OntoDeM (ontology-enhanced decision-making model) for an agent's decision-making in partially observable and dynamic environments. The model aims to improve agents' performance by enabling them to augment their observations and reason about them. The research presents various methods within the RL agent's decision-making process, where agents observe the environment, choose actions, execute them, and update their experience.

The thesis leverages ontology to empower agents in structuring and comprehending their observations through ontological frameworks. Utilizing ontology facilitates the modeling of partial observability by explicitly deducing connections between observable and unobservable factors, and employing reasoning mechanisms. Additionally, ontology incorporation allows for contextual integration such as time, location, weather conditions, and user preferences into the environment model. This contextual representation enhances agents' grasp of environmental dynamics, supporting adaptive decision-making in response to varying contextual conditions.

In the **ontology-based environment modeling** stage, the agent annotates raw environment data streams using the semantic description defined by the domain-specific ontology and creates an ontology-based schema for its observation.

The **ontology-enhanced observation modeling** stage presents five methods designed to enhance agents' observations and subsequently improve their decision-making in dynamic and partially observable environments. These techniques include *observation abstraction, observation expansion, observation masking, observation augmentation, and observation sampling*. In *observation abstraction*, ontology's hierarchical structure enables the agent to categorize observations at different levels of granularity, allowing for the substitution of noisy values with averages of non-noisy values from similar past observations. *Observation expansion* enriches the ontology-based schema by systematically incorporating newly deduced concepts and properties from domain-specific ontology knowledge, thus providing a more comprehensive representation of observed data. *Observation masking* leverages ontology's reasoning capabilities to filter out or disregard irrelevant observations, enabling the agent to identify similarities between past and present observations, particularly useful in scenarios involving unforeseen events, thereby enhancing its decision-making process. *Observation augmentation* utilizes ontology-based reasoning to deduce missing information or hidden relationships from observed data, enhancing the augmentation of noisy observations. *Observation sampling*, which uses subsets of observations, is influenced by contextual factors captured in the ontology, ensuring precise and context-aware sampling processes based on the importance weights of concepts.

In the **ontology-driven goal selection/generation** stage, the challenge of selecting and generating meaningful goals in dynamic and partially observable environments is addressed. This challenge arises due to the complexities of adapting goals to changing conditions and limited environment observability. OntoDeM proposes three methods for enhancing goal selection/generation: *automatic goal selection/generation*, *adaptive re-* *ward definition*, and *reward augmentation*. These methods leverage ontology's capabilities to incorporate contextual information, dynamically adjust rewards, and augment reward functions. The *automatic goal selection/generation* method addresses the challenge of dynamically selecting and generating goals in complex and evolving environments using ontology. By incorporating contextual information through ontology, the method enables the automatic creation of tailored goals that adapt to changing conditions. Logical rules and inference mechanisms within the ontology allow the agent to reason about the goals and link them to specific environmental states. In unforeseen situations, the agent evaluates its observations, identifies significant changes, and decides whether to maintain its current goal, switch to a predefined goal, or generate a new goal based on contextual factors and logical reasoning. The *adaptive reward definition* method employs ontology to adapt rewards based on domain-specific constraints and beliefs. Lastly, the *reward augmentation* method enhances the reward signal by incorporating the efficiency rate of actions in shaping the reward function.

In the **ontology-driven action selection** stage, four distinct methods are proposed within OntoDeM: *action masking*, *action exploration*, *action prioritization*, and *execution prioritization*. In the *action masking* method, ontology helps filter out impossible or invalid actions by encoding contextual information about the environment, operational constraints, and dependencies between actions. By representing these constraints, the agent's action selection is guided by semantic restrictions imposed by the ontology. The *action exploration* method utilizes ontology to rank the importance of concepts within the domain, mapping actions to concept rewards, and implementing a hybrid policy that blends ontology-based and RL-based decision-making, allowing the agent to dynamically explore actions that align with important concepts. *Action prioritization* involves using ontology to define rules that determine which actions should be given precedence based on specific requirements or conditions, enabling the agent to prioritize actions that meet these criteria. Lastly, *execution prioritization* uses ontology to handle temporal dependencies and priorities, optimizing the sequence in which multiple actions are executed to achieve optimal outcomes.

In summary, this thesis highlights how ontology supports the agent's decision-making in partially observable and dynamic environments. It explains how ontology facilitates observation modeling, goal selection, and action selection by providing structured representations, capturing constraints and dependencies, and incorporating contextual information.

### 7.2.1 Real World Applications

To illustrate and evaluate the applicability of the proposed method in real world applications, this research explored applications in Intelligent Traffic Signal Control (ITSC) systems, Edge Computing (EC) environments, Job Shop Scheduling (JSS), and Heating System Control (HSC).

The Intelligent Traffic Signal Control (ITSC) system is an advanced AI-powered system that optimizes traffic signal operation at intersections to enhance traffic flow efficiency, reduce congestion, travel times, fuel consumption, emissions, and improve overall transportation system performance. Unlike traditional fixed-timing traffic signal systems, ITSC dynamically adjusts signal timings based on real-time traffic data. The ITSC environment is dynamic and partially observable, with challenges like incomplete data from undetected vehicles, noisy information, and unforeseen events such as traffic accidents, disabled vehicles, and adverse weather conditions. The case studies conducted evaluate OntoDeM in synthetic and real urban traffic scenarios. OntoDeM-enabled RL controllers are compared to baseline RL controllers in various scenarios, including addressing congestion, handling unforeseen situations, dealing with partial observation, and prioritizing special vehicles. OntoDeM utilizes methods such as observation augmentation, observation sampling, automatic goal selection/generation, reward augmentation, action exploration, and action prioritization to improve decision-making. Results indicate that OntoDeM improves average waiting times, reduces travel delays, and enhances the number of completed trips, demonstrating its effectiveness in addressing partial observation, noisy data, and handling dynamic traffic scenarios.

Edge Computing (EC) involves processing data and computations closer to the source to reduce latency and improve system performance. Task offloading is a key concept in EC, where computational tasks are transferred from edge devices to edge servers or centralized cloud resources for optimization. The challenges of partial observability in EC arise from limited access to up-to-date or complete information. Additionally, the dynamic nature of EC environments, marked by workload fluctuations and unforeseen events include sudden spikes in data generation rates, network outages, hardware failures, and abrupt changes in user demands. A case study is defined to evaluate OntoDeM's effectiveness in enhancing task offloading performance within a mobile user environment. OntoDeM employs methods like observation masking, action masking, action prioritization, and execution prioritization to improve task offloading within the context of partially observable and dynamically changing EC environments. The results of the experiments show that OntoDeM significantly improves task processing and reduces task failures compared to the baseline DDPG algorithm.

Job Shop Scheduling (JSS) focuses on the problem of scheduling multiple orders on multiple machines with specific operational steps in a sequence. The objective is to optimize the utilization of resources and minimize completion times. Partial observability hinders accurate decision-making due to incomplete knowledge of orders, machines, and resources. Unforeseen events like machine breakdowns and priority changes disrupt the scheduling process. Autonomous agents are used to manage the complex task of JSS, with RL techniques being applied to learn effective scheduling policies. The proposed method, OntoDeM, employs observation augmentation to enhance the agent's knowledge of machine capacity, status, and working time. In more complex setups, OntoDeM leverages an adaptive reward definition method based on ontological information to guide the agent's decision-making. Comparative experiments reveal that OntoDeM significantly enhances performance in terms of utilization rate, waiting time, processed orders, and failed orders compared to the baseline TRPO method. Different scenarios encompass diverse order loads, due date constraints, noise levels, and failure rates, aiming to test the model's adaptability and performance across varying challenges.

A Heating System Control (HSC) is described as an automated system for regulating indoor temperatures in buildings while optimizing energy consumption. The environment of HSC is dynamic and partially observable due to factors like changing weather conditions and varying heating demands. Also, the HSC gathers indoor temperature data through sensors but might lack full information due to sensor issues or missing data, resulting in a partially observable environment. The use of autonomous agents, particularly RL agents, is proposed for optimizing HSC by adjusting heating settings in response to real-time temperature readings. Two case studies are designed to evaluate the OntoDeM in HSC. The OntoDeM employs ontology-based observation abstraction to mitigate the impact of noisy temperature data. Performance metrics like Average Time Spent Out of Range (ATSOR), Mean Squared Error (MSE), and total reward are used for assessment. Baseline algorithms PPO and DDPG are compared to OntoDeM-enabled versions of these algorithms. Results analysis shows that OntoDeM enhances the performance of the baseline algorithms in managing noisy data, resulting in reduced ATSOR and MSE and increased total reward, especially in scenarios with high noise levels.

Overall, OntoDeM proves to be effective in improving the performance of RL algorithms in various domains. It addresses challenges related to unforeseen events, noisy data, and incomplete information. The integration of OntoDeM enhances the agent's decision-making in the ITSC, EC, JSS, and HSC environments, leading to improved agent performance.

## 7.3 Future Research Directions and Limitations

It is worth mentioning that the OntoDeM model proposed in this thesis is the first attempt to show that using ontology has merit and significant benefits in agents' decision-making process, however further research is needed to make this model more practical.

**Use of ontologies:** OntoDeM uses domain-specific ontologies to formulate states and their distinguishing parameters, so, its performance depends on the accuracy and completeness of the used ontology. Additionally, operating in dynamic environments requires the ontology to evolve and update frequently. Ontology evolution techniques [285] can be used to address this issue. Also, ontology learning can happen in collaboration with knowledge experts, further enriching the knowledge base. More than one ontology can be used to model some problems, allowing for flexibility and precision in representing complex domains. Moreover, ontologies can relate to each other, enabling the integration of diverse knowledge sources.

Furthermore, a study about the impact of the goodness of ontology with respect to the performance of the agents could be useful. Moreover, concepts, their properties, importance weights, and semantic constraints can change depending upon the context, so the agent can have different ontologies for different "modes of operation", for example, during a hurricane, flood, or power outage.

Investigating how multiple agents (i.e., heterogeneous agents) with different ontologies, inferred knowledge and understandings of the environment can interact, communicate, and collaborate in dynamic and distributed environments can be challenging. Research can focus on techniques for aligning and integrating ontologies to enable effective knowledge-sharing and collaboration between agents. This can involve ontology mapping [118], ontology merging [37], or ontology mediation [56] approaches to reconcile differences and inconsistencies between ontologies. Furthermore, comprehending, exploring, and exploiting ontology models can be complex, and summarization techniques [215] can be used to provide an overview of such models which only covers important concepts, properties, and relationships for a better understanding. Furthermore, partial observability often leads to uncertainty and incomplete information in dynamic environments. Future research can explore how ontologies can be utilized to model and reason about uncertainty, probabilistic knowledge, and missing information. Techniques such as probabilistic ontologies [48] or extensions to handle incomplete information can enhance an agent's decision-making capabilities in uncertain and dynamic contexts.

**Online guidelines:** Currently, OntoDeM's guidelines serve as an offline tool, wherein system designers determine the integration of OntoDeM methods into the steps of a typical RL algorithm based on application area properties. However, in future work, there is potential for these guidelines to transition into an online framework. In such a setup, the selection of OntoDeM's methods could dynamically adapt to changes in the environment, ensuring the system remains responsive and effective over time. This shift towards online adaptability would enhance the practicality of the guidelines, allowing for real-time adjustments based on evolving conditions.

In this thesis, I have evaluated each method in various application areas to gain a comprehensive understanding of its effectiveness and applicability. As part of my future work, I plan to consolidate these findings by applying all methods within a single scenario, facilitating a more in-depth study of their impact, interactions, and scalability.

**Threshold sensitivity:** Regarding the potential impact of static thresholds on the dynamic nature of OntoDeM, I intend to address this issue through a comprehensive sensitivity analysis as part of future work. This analysis would involve systematically varying threshold values across different scenarios to understand their influence on performance gains and trade-offs. By doing so, I aim to uncover optimal threshold settings and gain deeper insights into the relationship between thresholds and method effectiveness.

**Adaptive reward settings:** Future work should explore methods to identify and deal with multiple types of unforeseen events in the environment. These events are not easily predictable and can include a diverse set of scenarios such as accidents, sudden changes in conditions (e.g., weather), equipment failures, emergencies, and other unanticipated circumstances. One approach can involve defining flexible reward functions that consider different types of events as separate objectives and developing mechanisms to dynamically assign weights or prioritize objectives based on the current situation [8]. Generative policies (e.g., [51], [4]) refer to policy models that are capable of generating policies that can adapt to different situations, and make appropriate decisions based on the observed environment. Generative policies can be integrated into OntoDeM to enable agents to observe, learn, and adapt high-level policy models. Also, by incorporating ontological concepts, relationships, and semantic constraints, generative policies can leverage structured knowledge to gain a deeper semantic understanding and make context-aware decisions adhering to specific rules or conditions.

**Environment types:** In this thesis, I have only tested OntoDeM in partially observable and dynamic environments. However, this model can be generalized to other types of environments through modeling and using the relevant ontologies, and applying the inference mechanism accordingly. Further exploring the application of OntoDeM in various real-world domains can also provide valuable insights and practical solutions. Examples include healthcare systems, smart cities, and autonomous driving, where dynamic and partially observable environments pose challenges to decision-making processes. Conducting empirical studies and evaluating the performance of ontology-based decision-making methods in these domains can validate their effectiveness and identify specific challenges and requirements.

By addressing these research directions, the utilization of ontology in improving agents' decision-making can be further advanced, leading to more intelligent and adaptive systems capable of handling complex real-world scenarios.

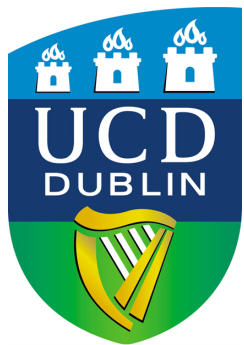
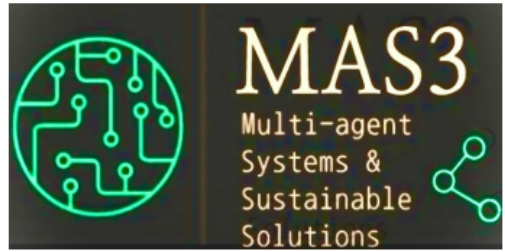